\documentclass[10pt,twocolumn,letterpaper]{article}

\usepackage{wacv}              %

\definecolor{wacvblue}{rgb}{0.21,0.49,0.74}
\usepackage[pagebackref,breaklinks,colorlinks,allcolors=wacvblue]{hyperref}
\usepackage{amsmath}
\usepackage[export]{adjustbox}
\usepackage{multirow}
\usepackage{placeins}
\usepackage{wrapfig}
\usepackage[T1]{fontenc}

\usepackage{float}
\usepackage[hypcap=false]{caption}
\usepackage{stfloats}

\usepackage[table]{xcolor}
\newcommand{\best}{\cellcolor{best}}
\newcommand{\sbest}{\cellcolor{sbest}}
\newcommand{\tbest}{\cellcolor{tbest}}

\definecolor{best}{rgb}{0.65, 0.92, 0.65}
\definecolor{sbest}{rgb}{0.82, 0.95, 0.70}
\definecolor{tbest}{rgb}{0.95, 0.98, 0.82}

\def\our{OmniStyle-INR}

\def\wacvPaperID{160} %
\def\confName{WACV}
\def\confYear{2027}

\title{\our{}: Universal and Multimodal Style Transfer for INRs}

\author{
    \begin{tabular}{c}
        Rafał Kajca\textsuperscript{1*} \\
        {\tt\small rafal.kajca@student.uj.edu.pl}
    \end{tabular}
    \qquad
    \begin{tabular}{c}
        Michał Miziołek\textsuperscript{1*} \\
        {\tt\small michal.miziolek@student.uj.edu.pl}
    \end{tabular} \\[1.5em]
    Kornel Howil\textsuperscript{1,3} \qquad Rafał Tobiasz\textsuperscript{1,2,3} \qquad Przemysław Spurek\textsuperscript{1,3} \\[1.5em]
    {\small \textsuperscript{1}Jagiellonian University, Faculty of Mathematics and Computer Science}\\
    {\small \textsuperscript{2}Jagiellonian University, Doctoral School of Exact and Natural Sciences}\\
    {\small \textsuperscript{3}IDEAS Research Institute}\\
    {\small \textsuperscript{*}Equal contribution}
}

\begin{document}

\twocolumn[{
    \maketitle
    \vspace{-1.3cm}
    \begin{center}
        \includegraphics[trim={0 0 10 0},clip,width=0.76\textwidth]{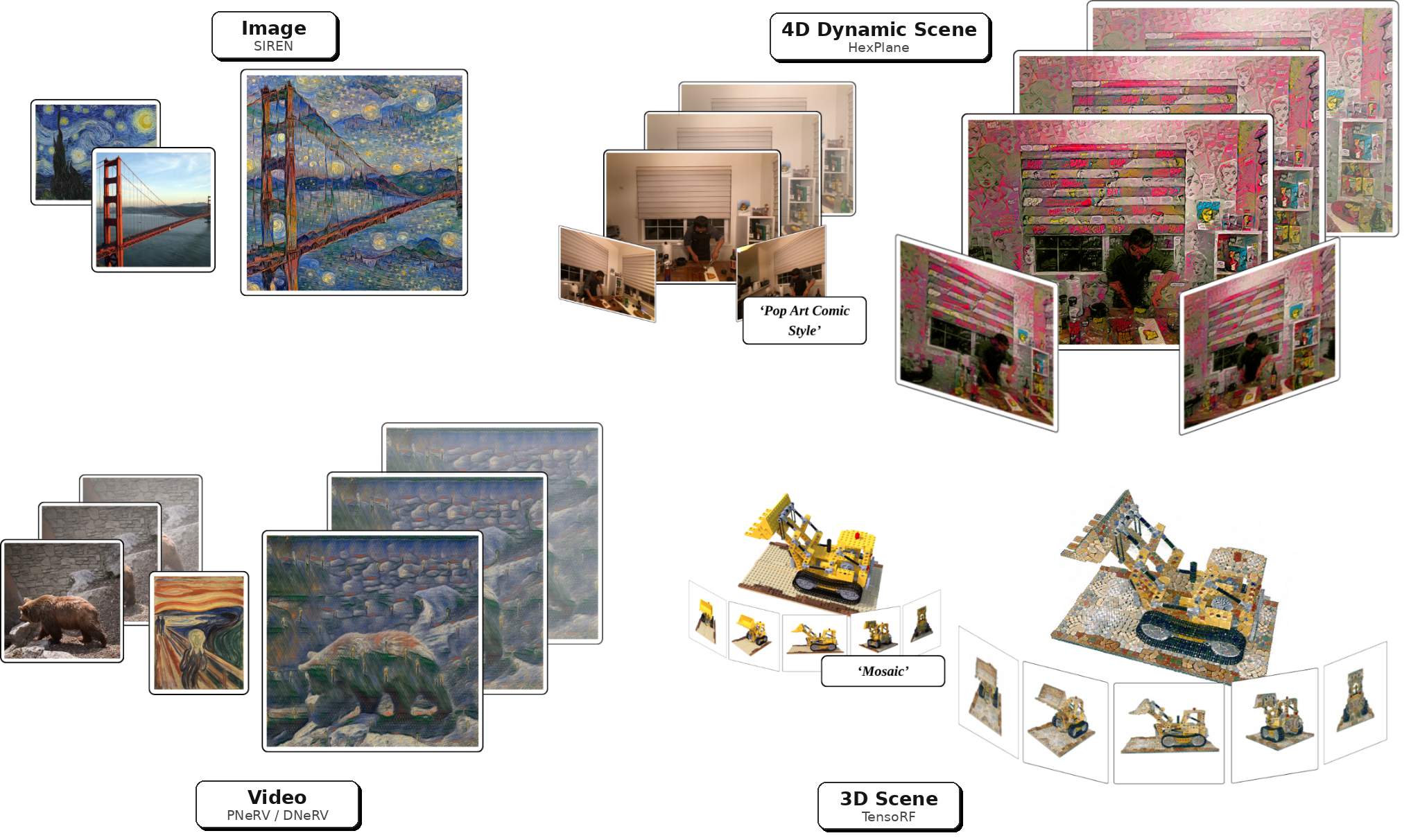}
        \captionof{figure}{\our{}, a universal style transfer framework that supports a broad spectrum of data modalities, including images, videos, 3D objects, and 4D dynamic scenes. Stylization in \our{} can be driven by either an image example or a text prompt. At the core of our method lies an Implicit Neural Representation (INR), which we leverage to model both the color and geometric components of the transferred style.}
        \label{fig:teaser}
    \end{center}
    \vspace{-0.2cm}
}]

\begin{abstract}
Style transfer remains a fundamental and highly important task across various data modalities, enabling creative manipulation conditioned by both reference images and textual descriptions. Recently, methods utilizing Gaussian Splatting have emerged as a unified representation for 2D images, video, 3D scenes, and 4D dynamics. However, representing videos and 2D images with Gaussian Splatting is structurally sub-optimal for dense continuous domains. The number of required Gaussians often approaches the total number of pixels, raising questions about the actual utility of such a representation for these specific modalities. In contrast, Implicit Neural Representations have established themselves as a much more popular and natural choice across all these data domains. Implicit Neural Representations naturally provide significant advantages, including data compression, inherent capabilities for super resolution, and seamless integration with deep generative models. To this end, we introduce \our{}, a novel framework that leverages network-based continuous representations as a truly universal domain. Our approach successfully performs high-quality style transfer across all visual modalities, guided seamlessly by both text prompts and visual exemplars.
\end{abstract}

\section{Introduction}

Visual style transfer has long been a fundamental task in computer vision and computer graphics~\cite{gatys2016image,jing2019neural}. It empowers users to creatively manipulate the visual appearance of data guided by either textual prompts or reference images. Recently, methods utilizing Gaussian Splatting have emerged as a unified representation for 2D images, video, 3D scenes, and 4D dynamics~\cite{howil2026clipgaussian}. For instance, recent frameworks such as MiraGe~\cite{waczynskamirage} and GaINeR~\cite{jakubowska2025gainer} have adapted this explicit formulation to represent 2D images, while VeGaS~\cite{smolak2025vegas} extended it to video processing. However, modeling 2D images and standard videos with Gaussian Splatting presents certain structural limitations. In such explicit setups, the number of required Gaussians often approaches the raw pixel count. While visually effective, this dense parameterization can limit the natural compression and continuous interpolation capabilities that are often desired in these domains.

In stark contrast, Implicit Neural Representations~\cite{sitzmann2020implicit} provide a fundamentally continuous and unified domain that is well-suited to modeling signals of arbitrary dimensionality. Implicit Neural Representations have proven exceptionally effective for 2D images~\cite{sitzmann2020implicit, chen2021learning, liu2024finer, kania2025fresh} and video compression~\cite{chen2021nerv, chen2023hnerv}. Furthermore, they serve as the foundational architecture for static spatial environments and complex 4D spatiotemporal scenes~\cite{pumarola2021d, cao2023hexplane}. By mapping input coordinates to visual properties via a neural network, these representations inherently provide powerful data compression and continuous interpolation capabilities, such as super-resolution. They also seamlessly integrate with deep generative models~\cite{spurek2020hypernetwork}.

\begin{figure}[t]
\centering
\small
\setlength{\tabcolsep}{1pt}
\resizebox{\columnwidth}{!}{%
    \begin{tabular}{ccccc}
    GT \& Style &  StyleGaussian~\cite{liu2024stylegaussian} &  G-Style~\cite{kovacs2024G}  &  CLIPGaussian~\cite{howil2026clipgaussian} &  \textbf{\our} \\
      \includegraphics[trim={50 130 0 0},clip, width=0.12\textwidth,valign=c]{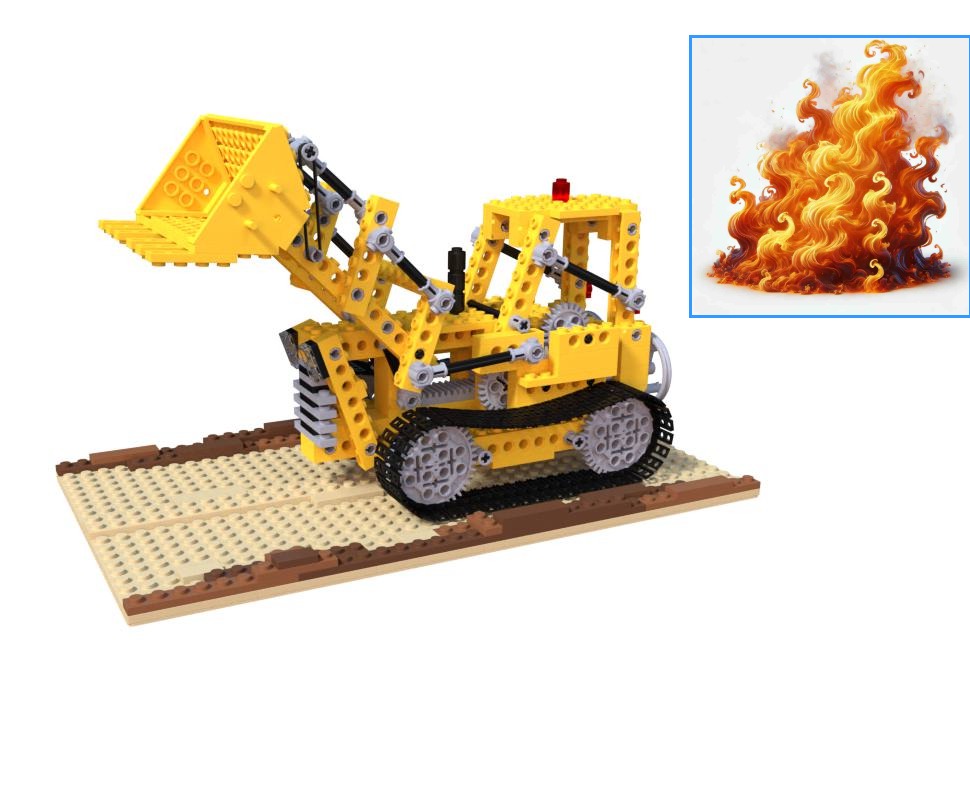}   & 
    \includegraphics[trim={50 130 50 100},clip, width=0.12\textwidth,valign=c]{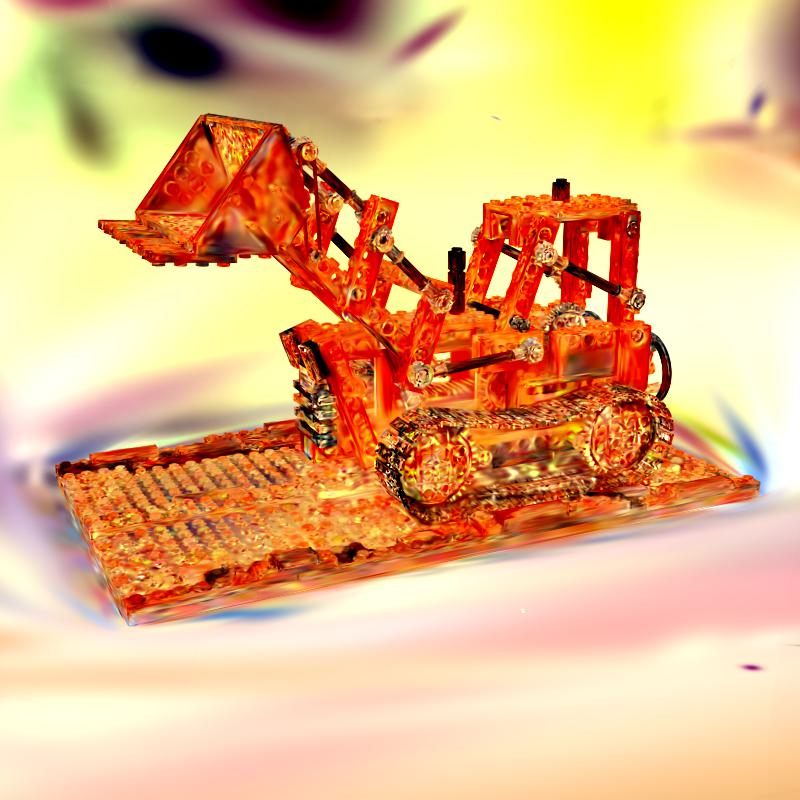}    & 
    \includegraphics[trim={50 130 50 100},clip, width=0.12\textwidth,valign=c]{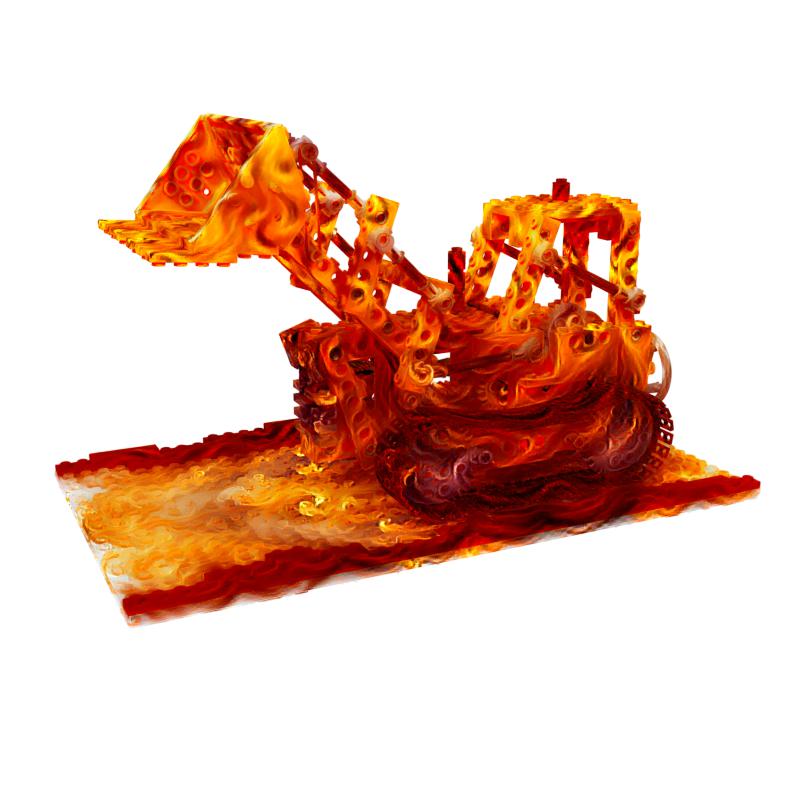}   & 
    \includegraphics[trim={50 130 50 100},clip, width=0.12\textwidth,valign=c]{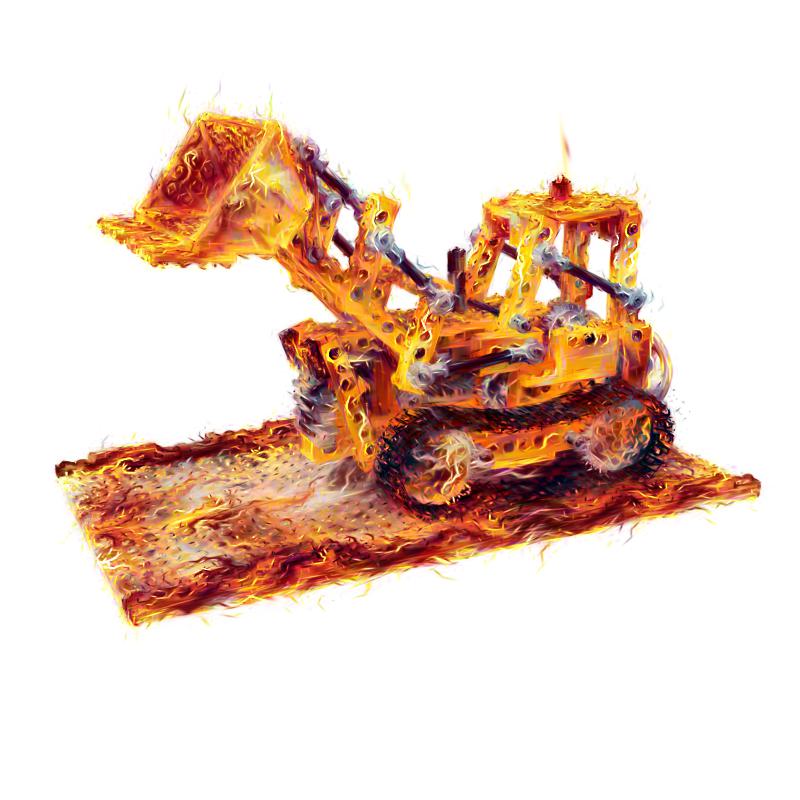}  &
    \includegraphics[trim={50 130 50 100},clip, width=0.12\textwidth,valign=c]{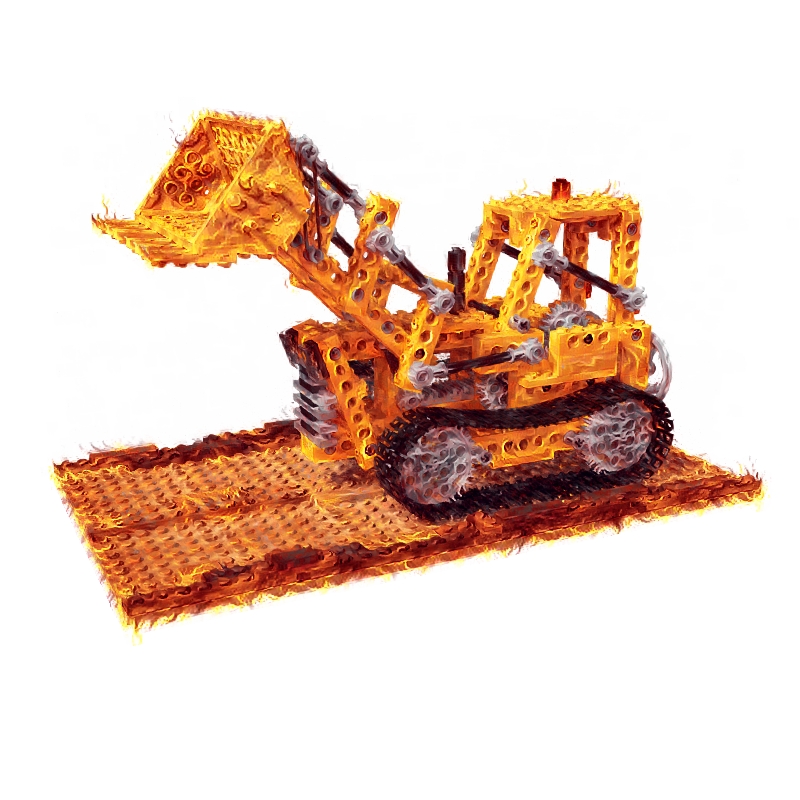}  \\ 
       \includegraphics[trim={50 100 0 0},clip, width=0.12\textwidth,valign=c]{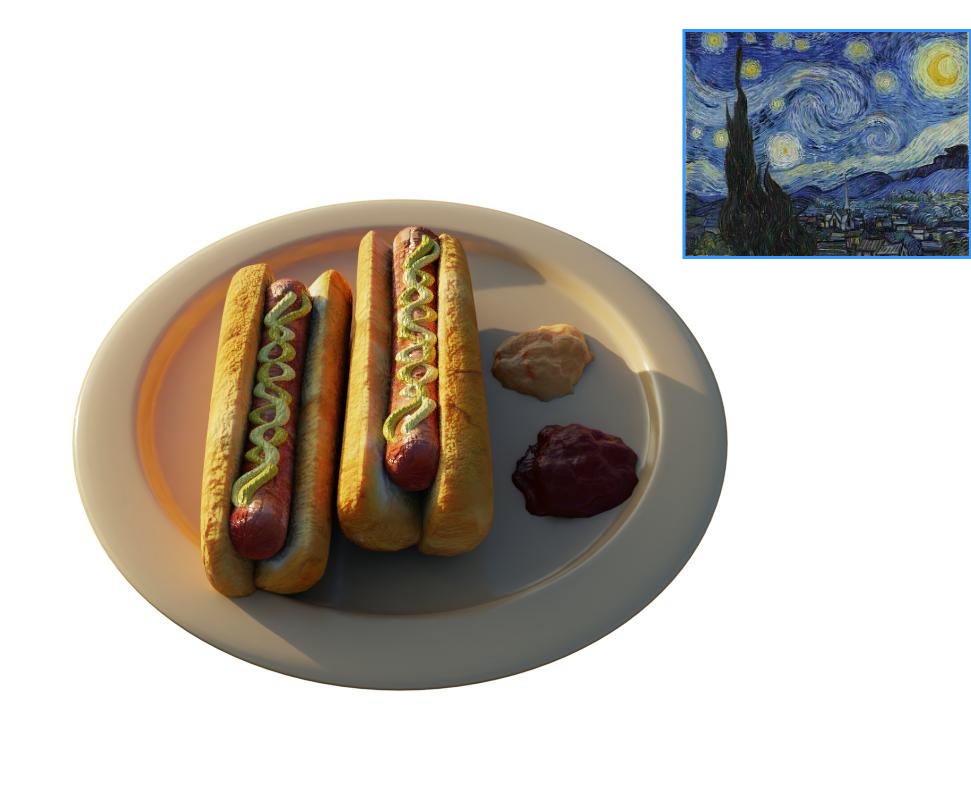}  &  
     \includegraphics[trim={50 100 50 190},clip, width=0.12\textwidth,valign=c]{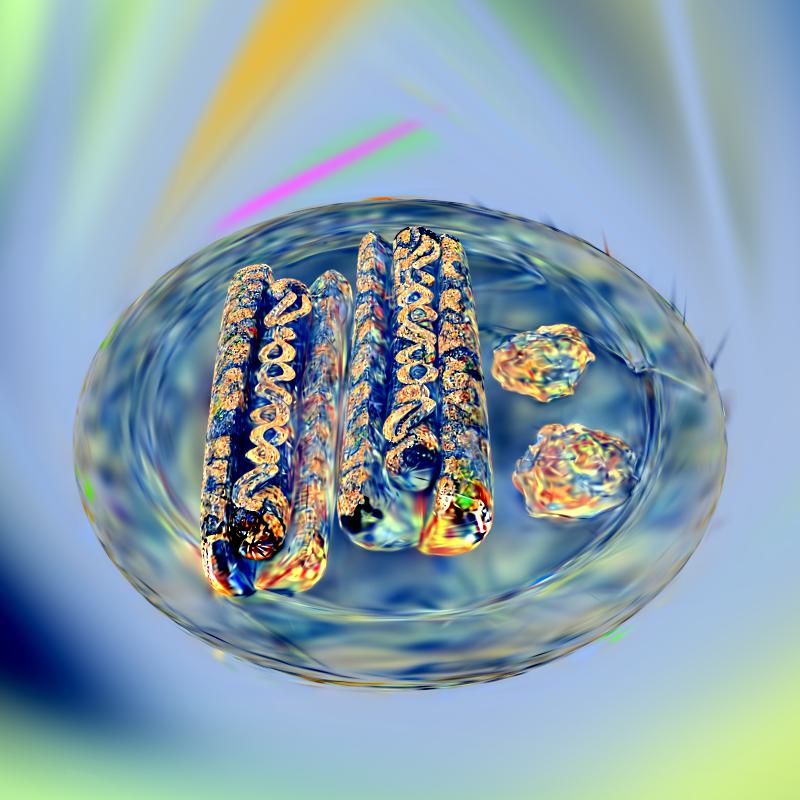}    &  
     \includegraphics[trim={50 100 50 190},clip, width=0.12\textwidth,valign=c]{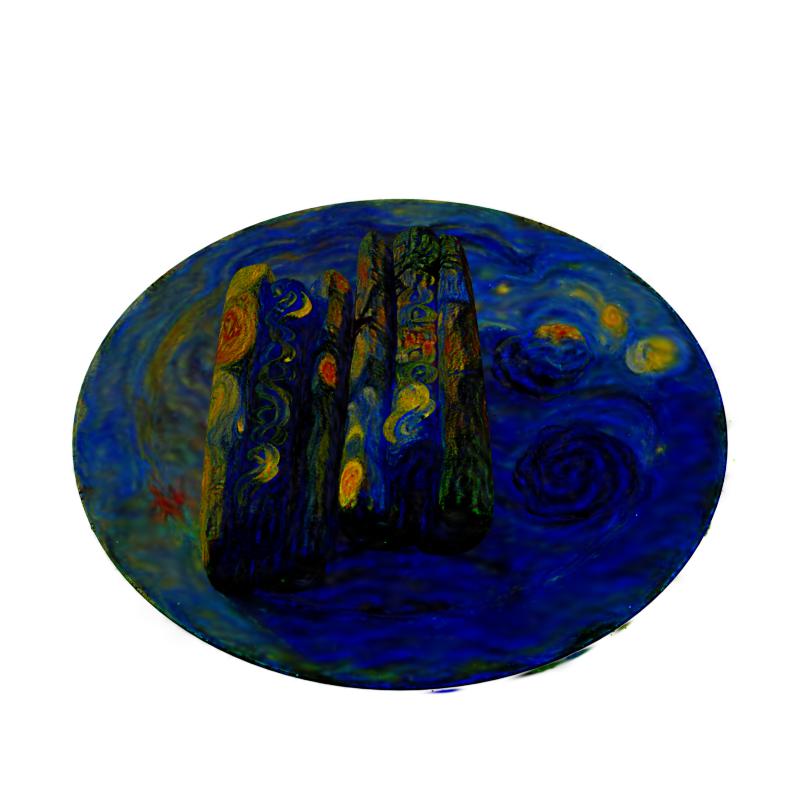}   &  
     \includegraphics[trim={50 100 50 190},clip, width=0.12\textwidth,valign=c]{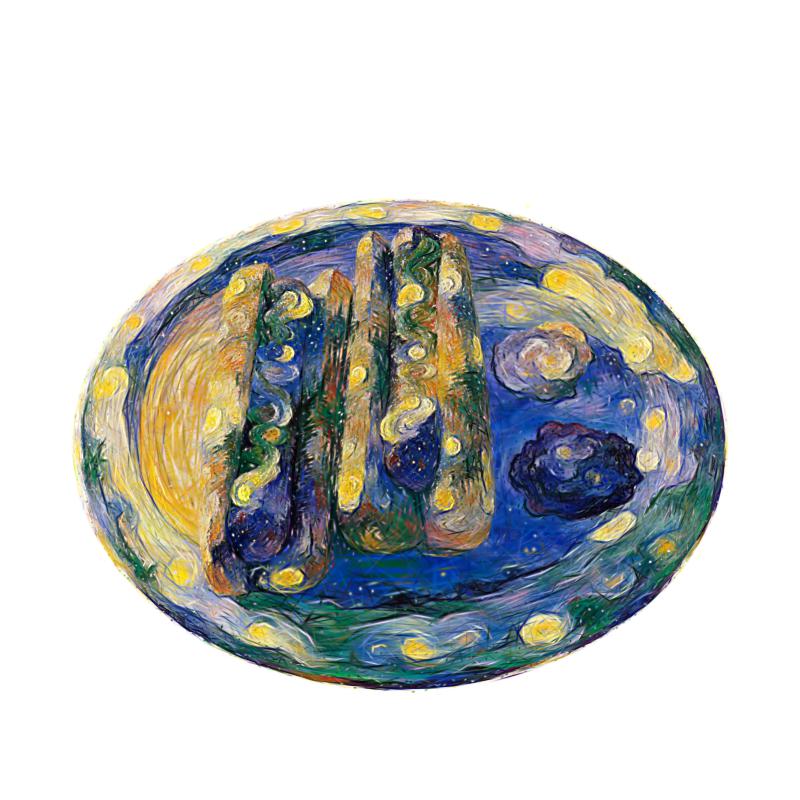} &
     \includegraphics[trim={50 100 50 190},clip, width=0.12\textwidth,valign=c]{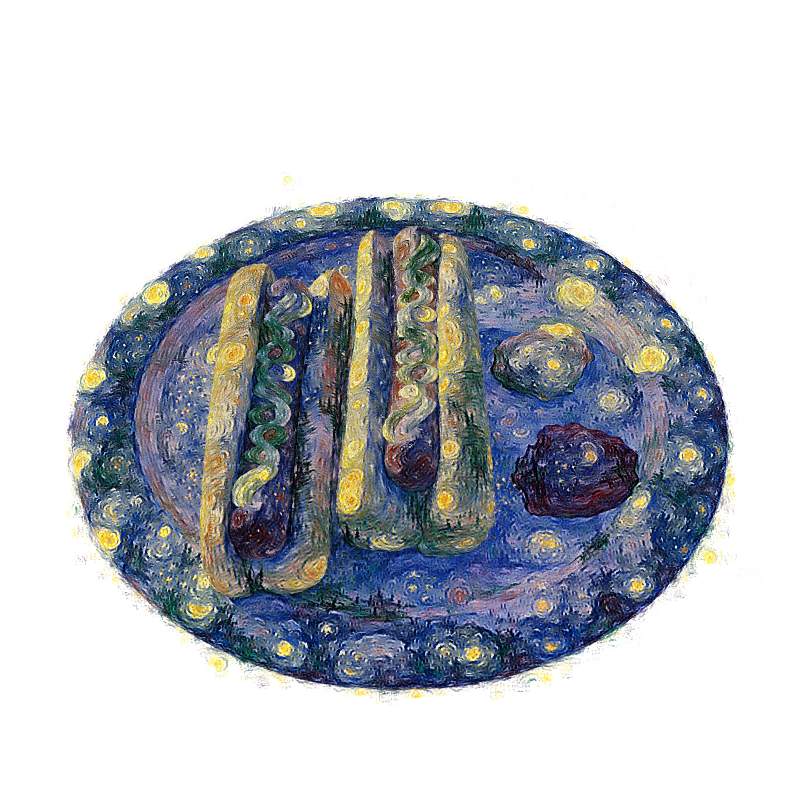}  \\ 
      \includegraphics[trim={100 0 0 0},clip, width=0.12\textwidth,valign=c]{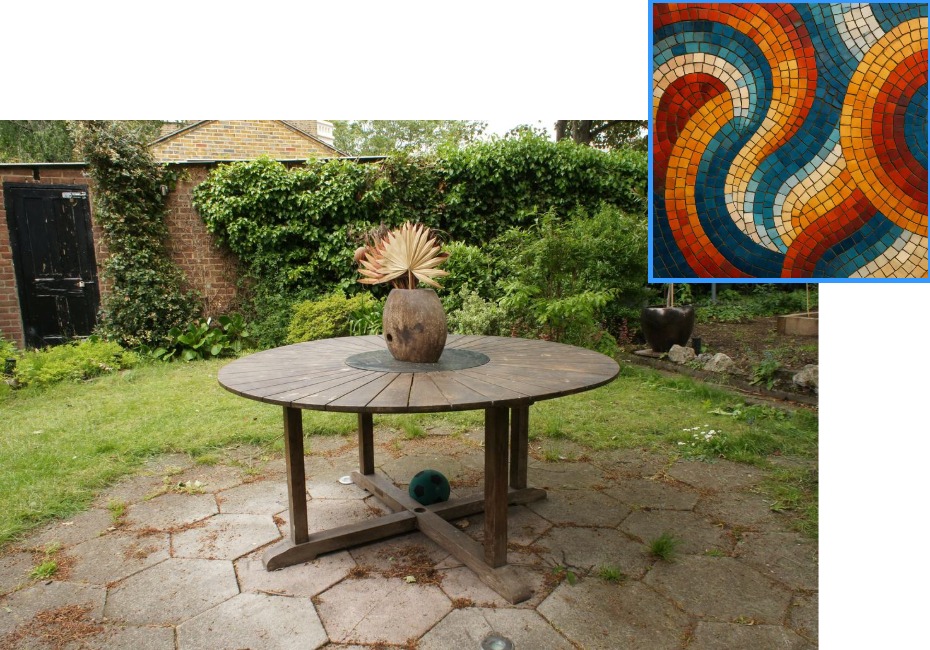}   & 
    \includegraphics[trim={100 0 80 0},clip, width=0.12\textwidth,valign=c]{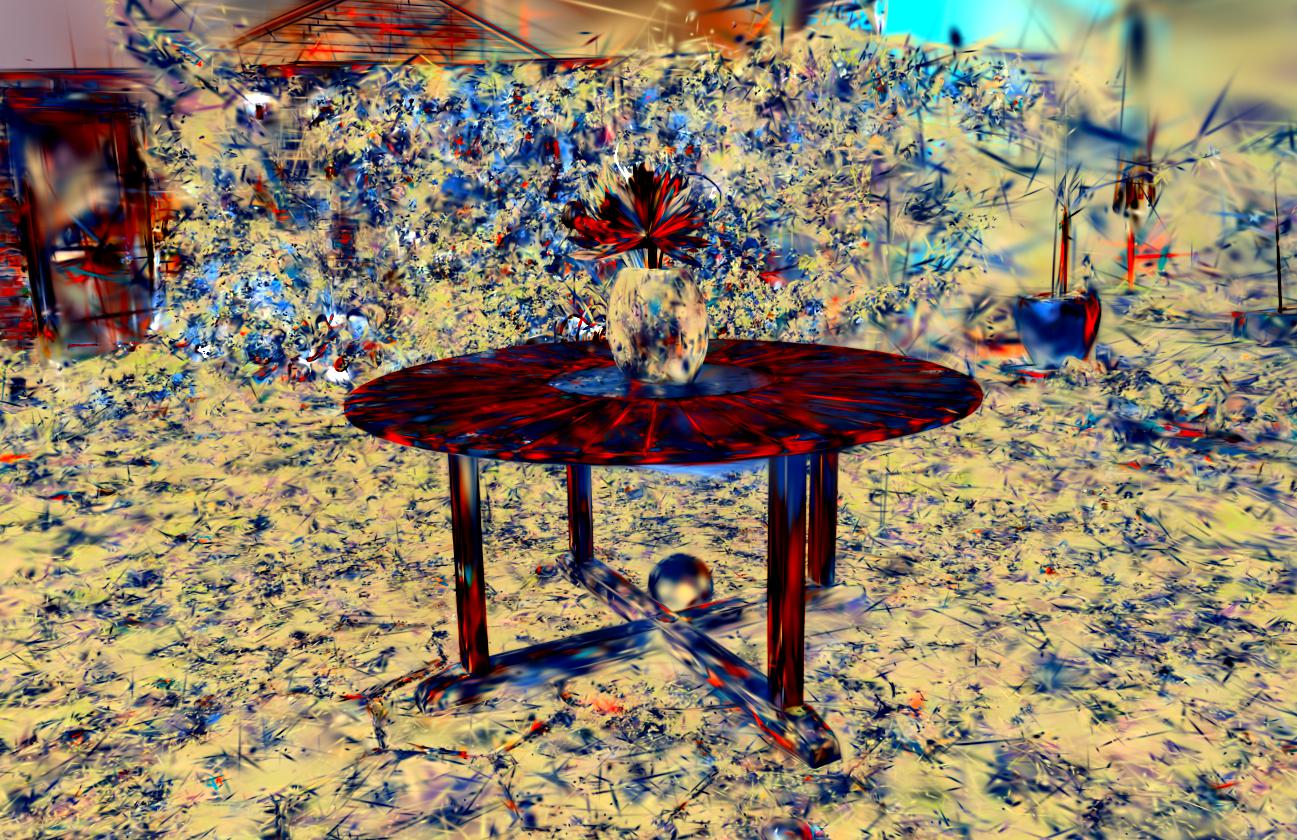}    & 
    \includegraphics[trim={100 0 80 0},clip, width=0.12\textwidth,valign=c]{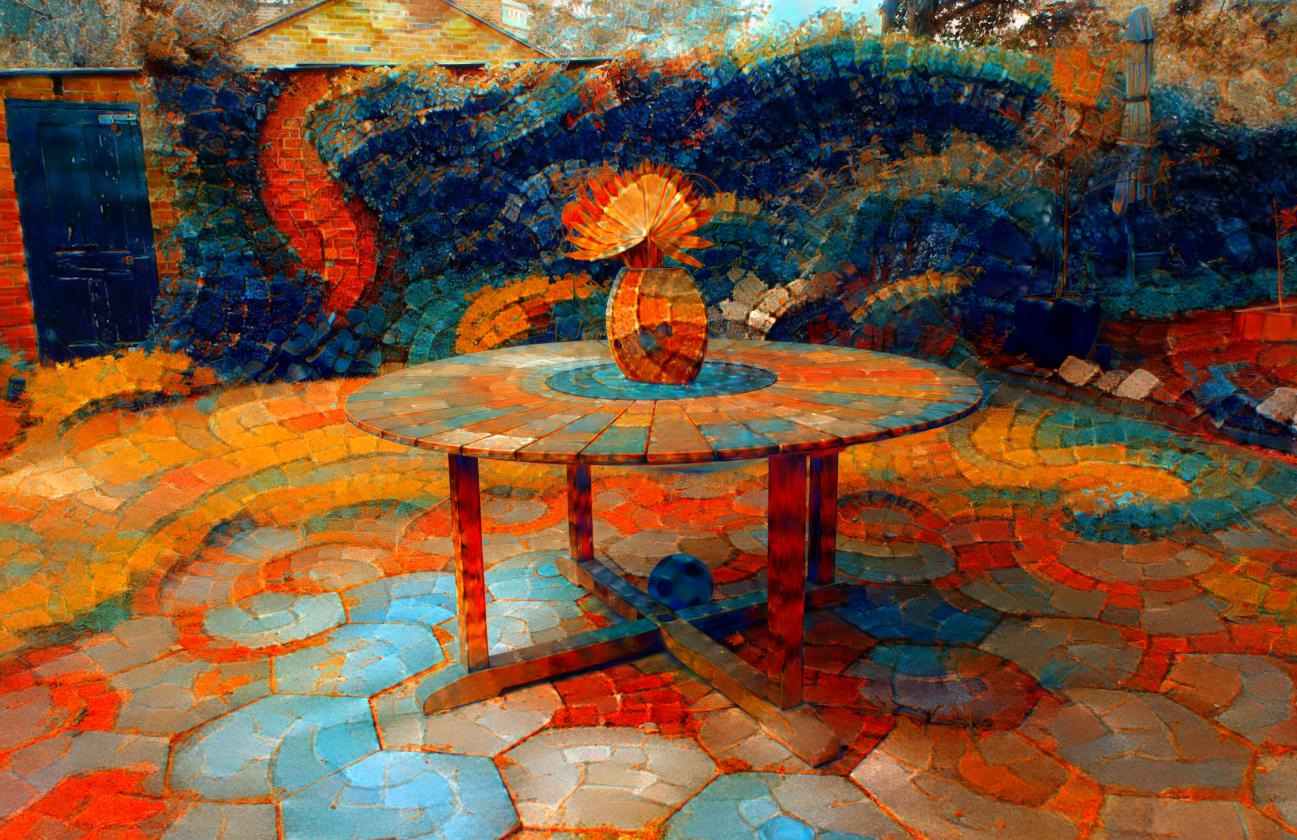}   & 
    \includegraphics[trim={100 0 80 0},clip, width=0.12\textwidth,valign=c]{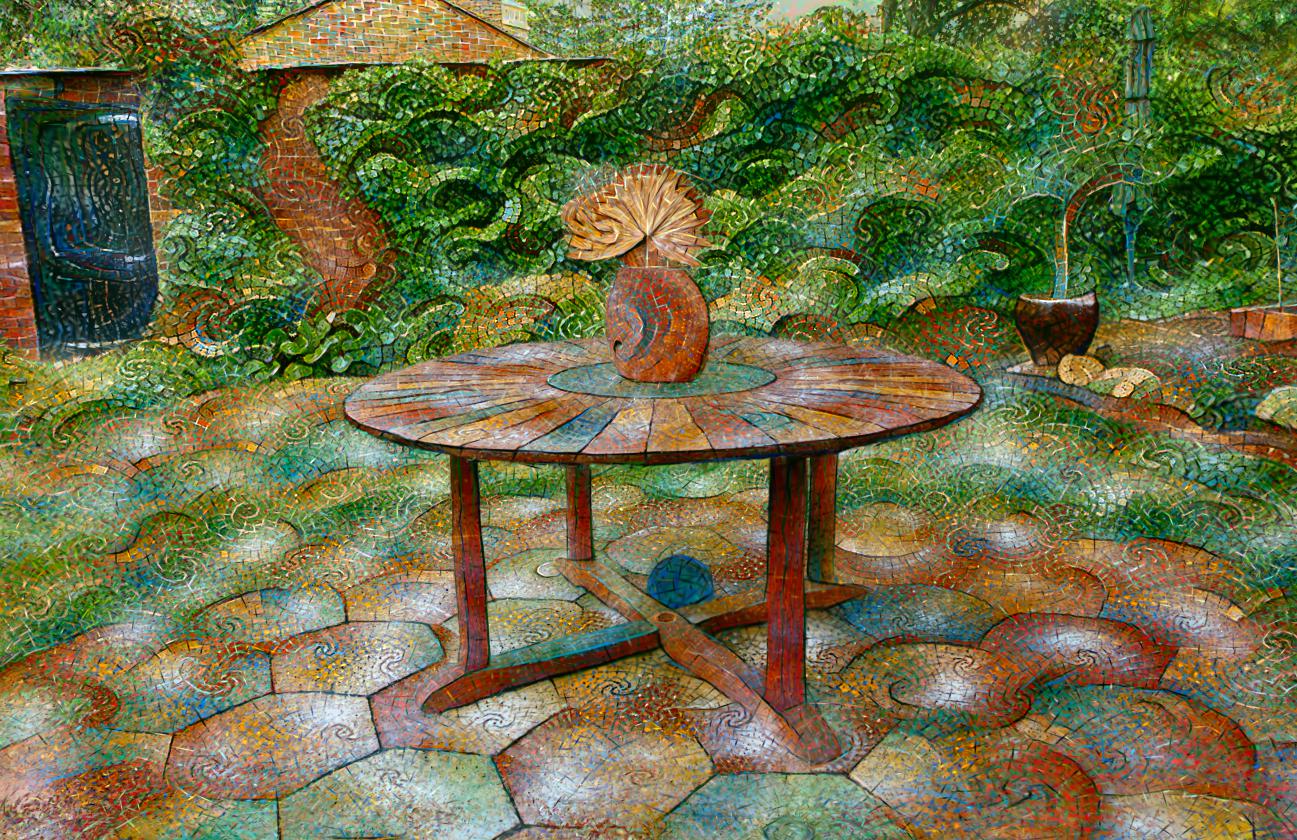}  &
    \includegraphics[trim={100 0 80 0},clip, width=0.12\textwidth,valign=c]{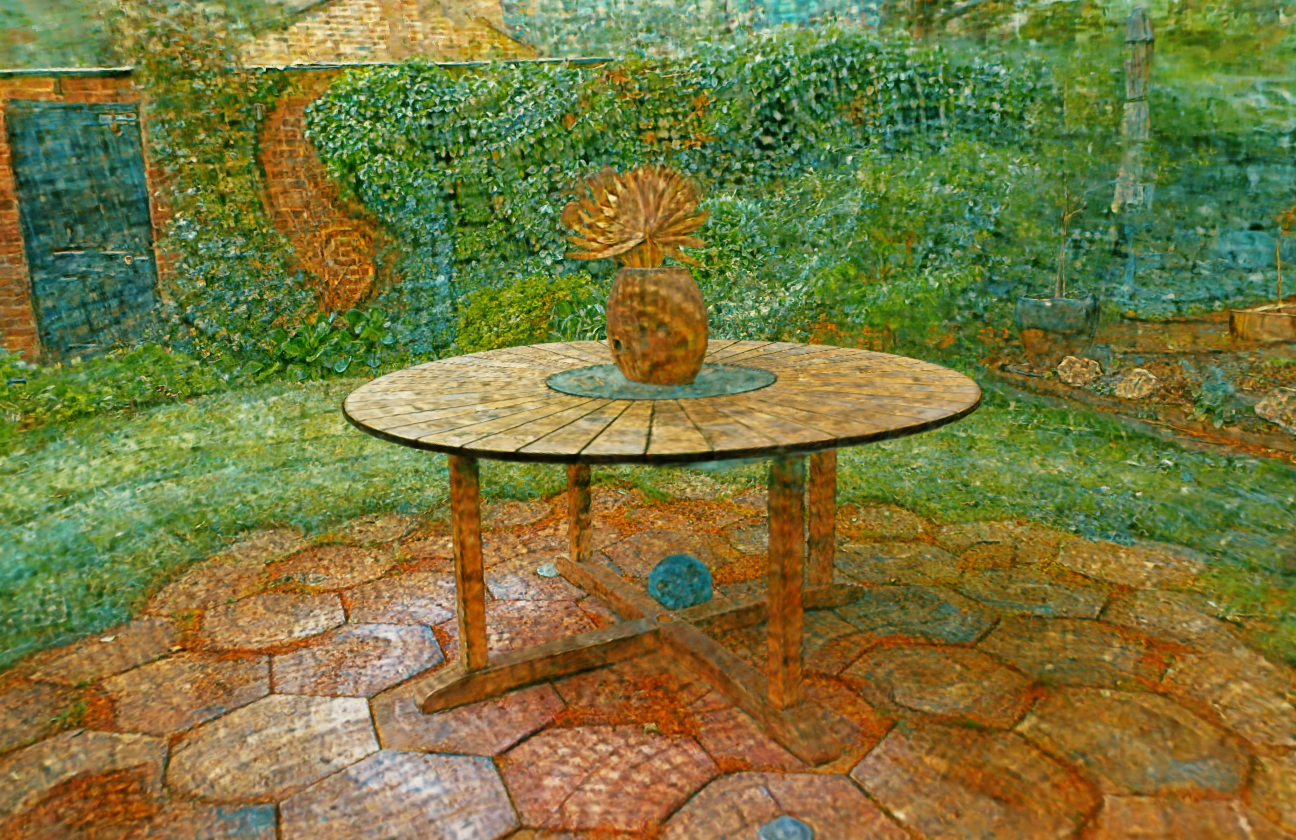}\\
      \includegraphics[trim={100 100 0 0},clip, width=0.12\textwidth,valign=c]{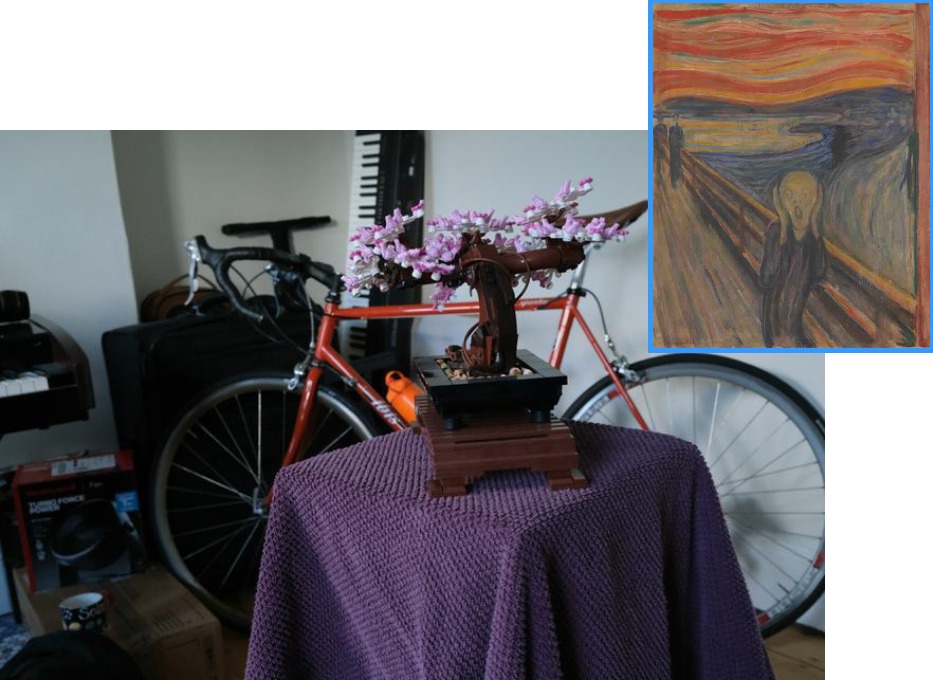}  &  
     \includegraphics[trim={100 0 0 0},clip, width=0.12\textwidth,valign=c]{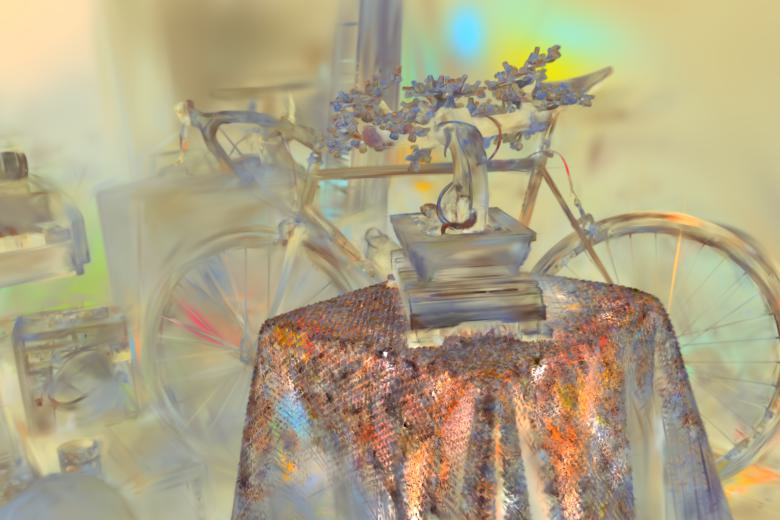}    &  
     \includegraphics[trim={100 0 0 0},clip, width=0.12\textwidth,valign=c]{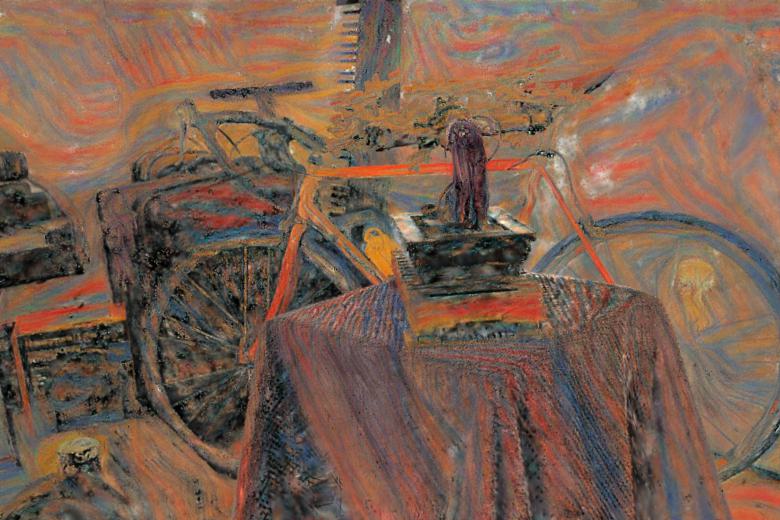}   &  
     \includegraphics[trim={100 0 0 0},clip, width=0.12\textwidth,valign=c]{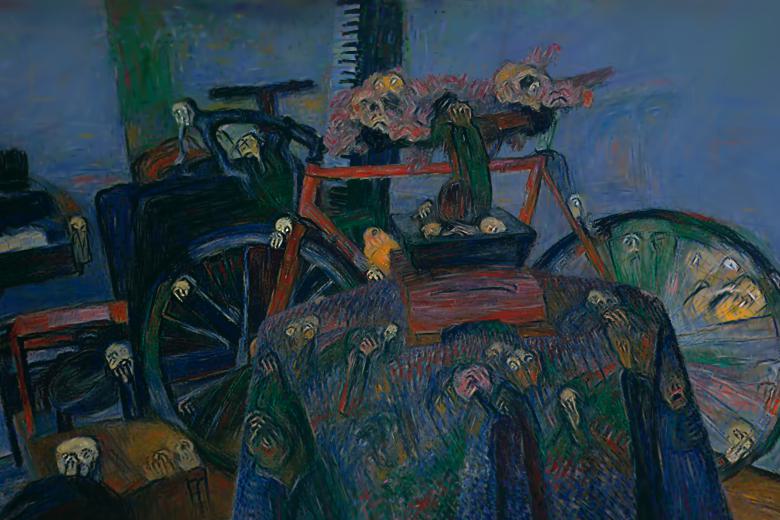}  &
     \includegraphics[trim={100 0 0 0},clip, width=0.12\textwidth,valign=c]{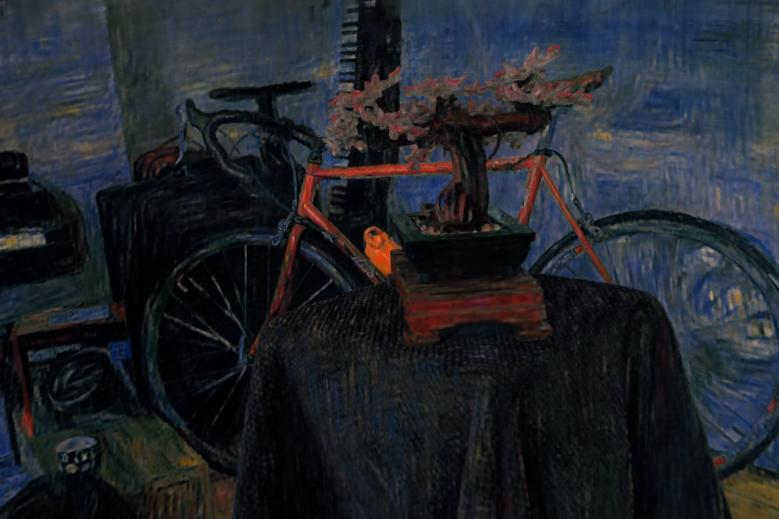}\\
    \end{tabular}
}
\caption{Comparison of various 3D style transfer methods using image conditioning. Our model effectively captures fine style details.}
\label{fig:Image2Style}
\end{figure}

Developing a universal framework that can handle diverse data modalities and be conditioned by both text prompts and reference images is of paramount importance. Such a unified approach bridges the gap between disparate data structures and opens the way toward truly universal foundation models. To this end, we introduce \our{}, a framework that establishes a truly universal domain for multimodal style transfer entirely based on Implicit Neural Representations. Unlike explicit point-based methods, \our{} treats the underlying neural network as a shared canvas for stylized optimization. It naturally supports arbitrary dimensionalities ranging from single images and videos to complex dynamic 4D environments. Our approach leverages robust vision models to guide the optimization process seamlessly using both textual descriptions and visual references .

In summary, our main contributions are as follows:
\begin{itemize}
    \item We propose \our{}, a novel framework for multimodal style transfer that unifies 2D, video, 3D, and 4D domains using continuous neural fields.
    \item We highlight the fundamental advantages of Implicit Neural Representations over explicit Gaussian Splatting for diverse spatial and temporal tasks.
    \item We demonstrate high-quality stylization guided by both text and image references across all visual modalities, paving the way for more generalized multimodal architectures.
\end{itemize}

\begin{figure}[t]
\centering
\small
\setlength{\tabcolsep}{1pt}
\resizebox{\columnwidth}{!}{%
    \begin{tabular}{ccccc}
    GT \& Style &  I-GS2GS~\cite{haque2023instruct} &  DGE~\cite{chen2024dge}  &  CLIPGaussian~\cite{howil2026clipgaussian} & \textbf{\our}\\
      \includegraphics[trim={50 130 0 0},clip, width=0.12\textwidth,valign=c]{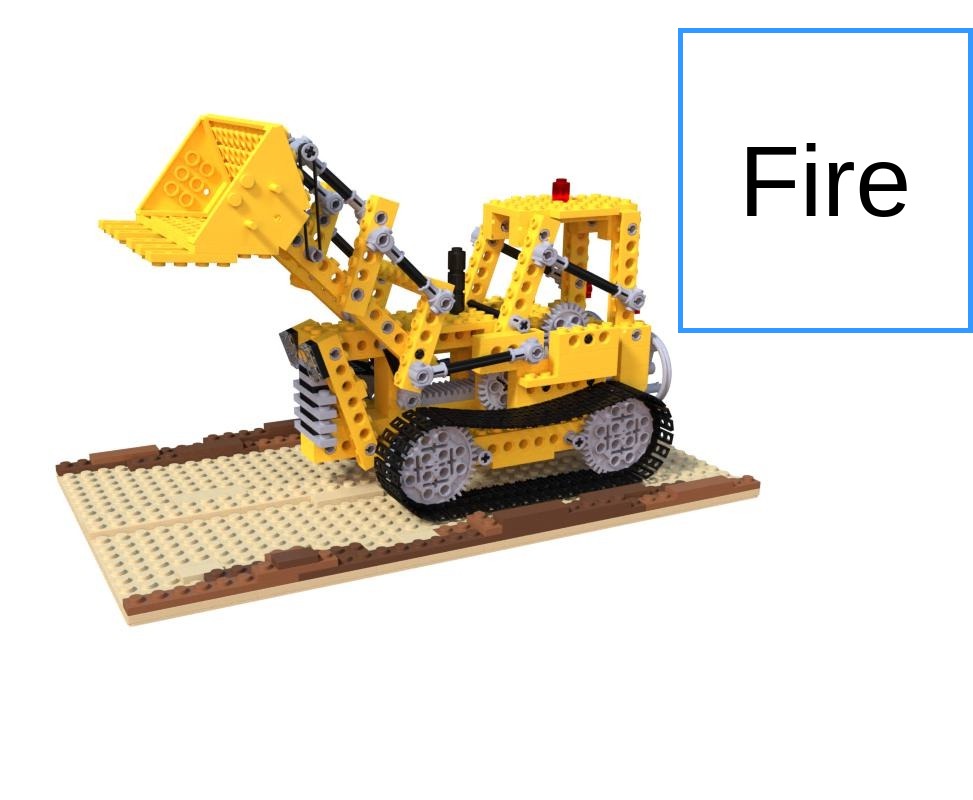}    & 
    \includegraphics[trim={50 130 50 100},clip, width=0.1\textwidth,valign=c]{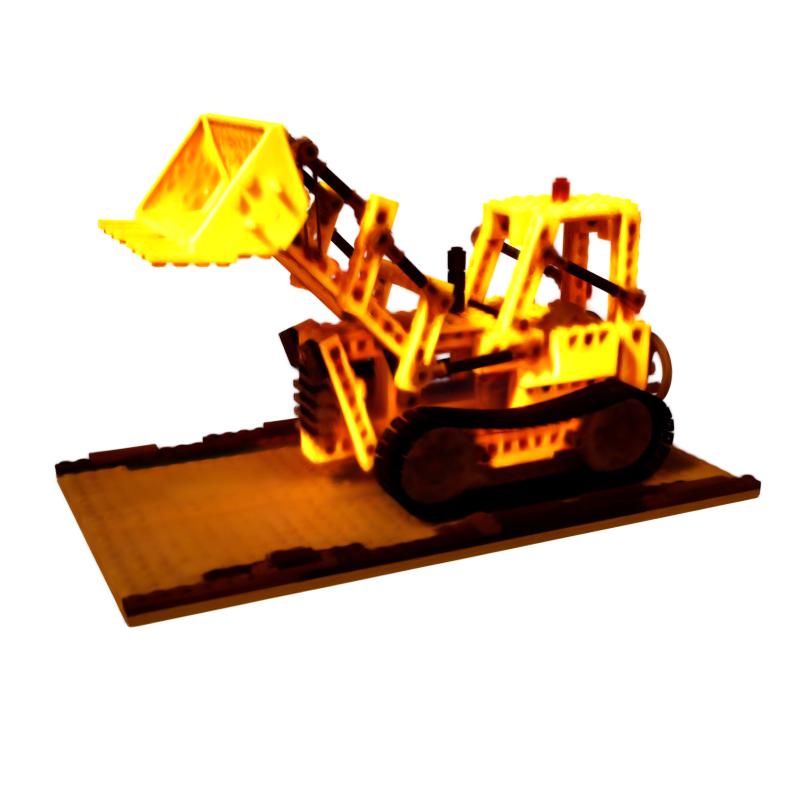}    & 
    \includegraphics[trim={50 130 50 100},clip, width=0.12\textwidth,valign=c]{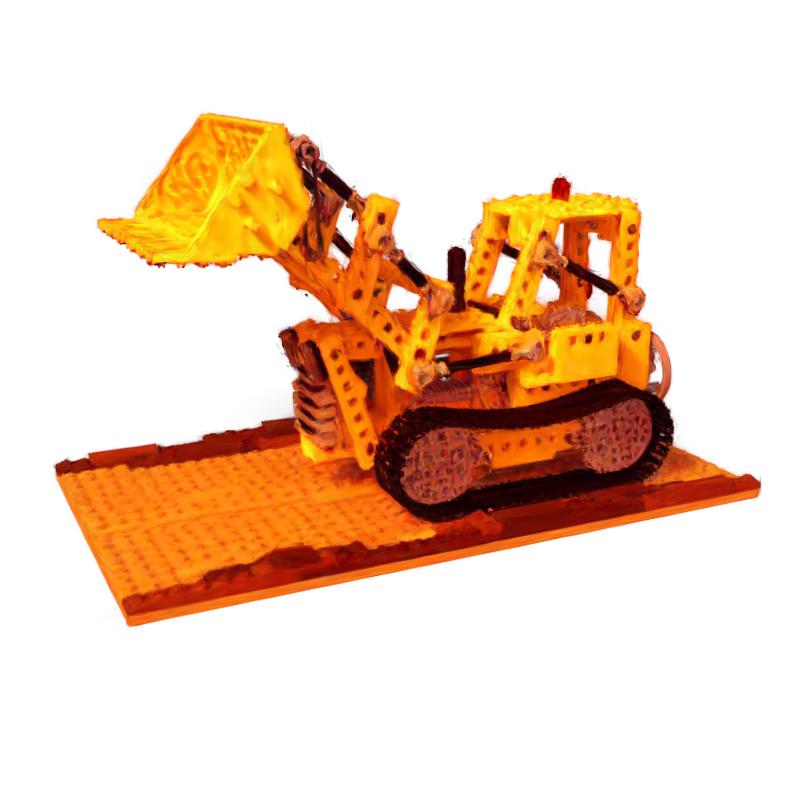}   & 
    \includegraphics[trim={50 130 50 100},clip, width=0.12\textwidth,valign=c]{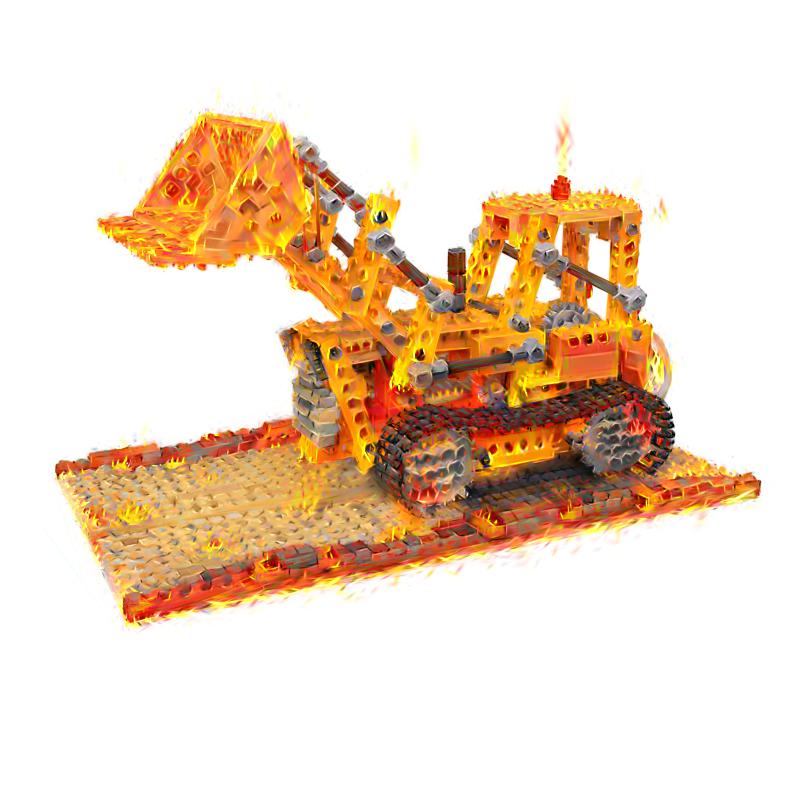}  &
    \includegraphics[trim={50 130 50 100},clip, width=0.12\textwidth,valign=c]{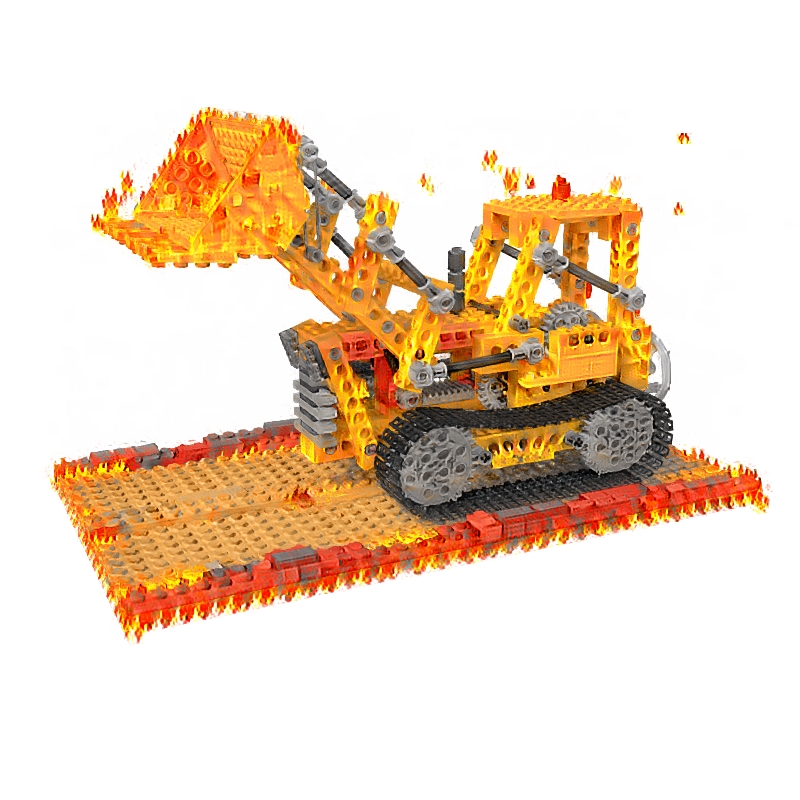} \\
       \includegraphics[trim={50 100 0 0},clip, width=0.12\textwidth,valign=c]{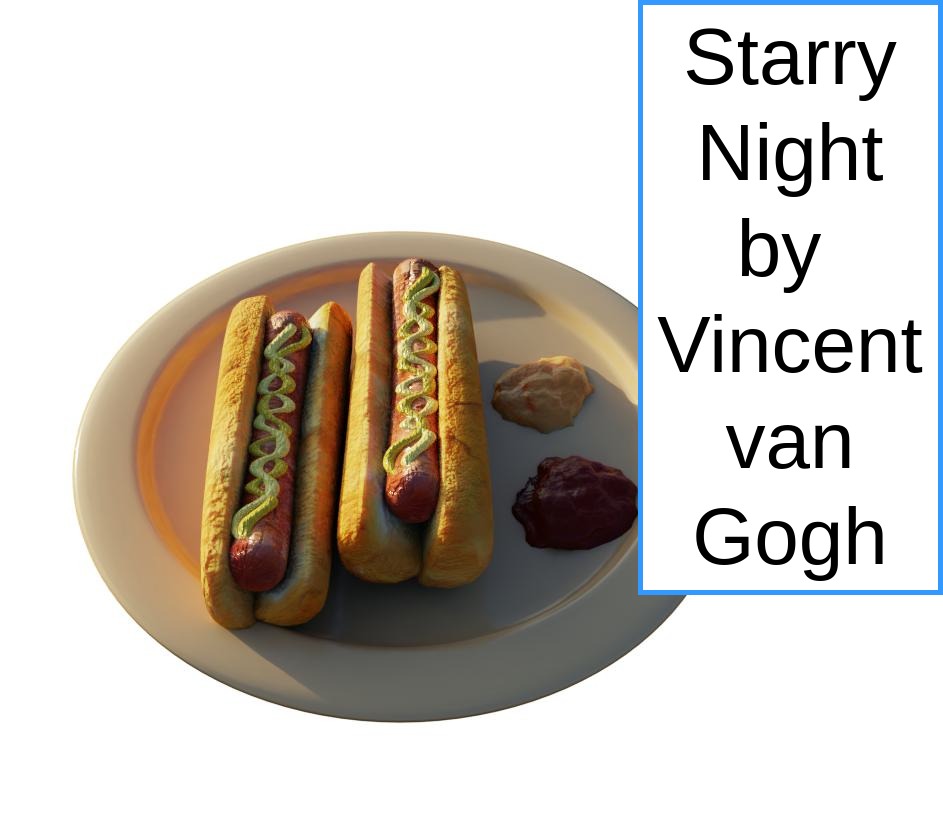}  &  
     \includegraphics[trim={50 100 50 190},clip, width=0.12\textwidth,valign=c]{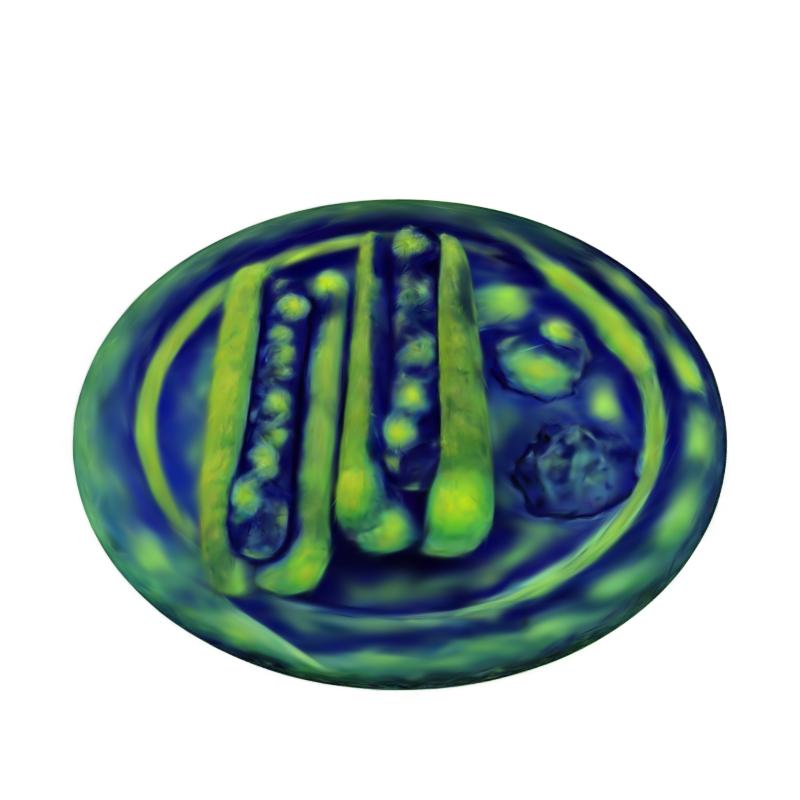}     &  
     \includegraphics[trim={50 100 50 190},clip, width=0.12\textwidth,valign=c]{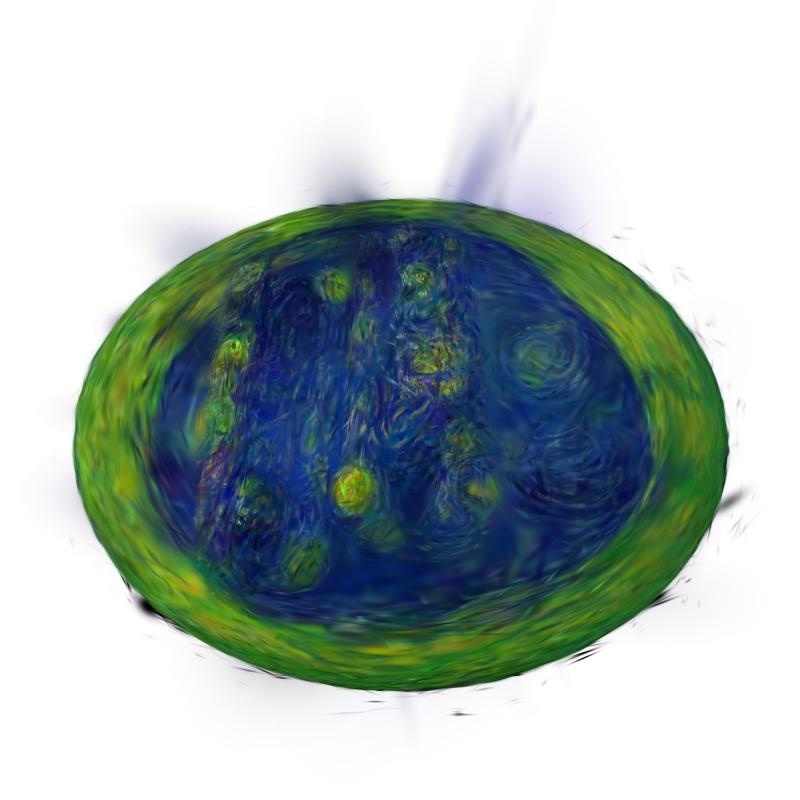}    &  
     \includegraphics[trim={50 100 50 190},clip, width=0.12\textwidth,valign=c]{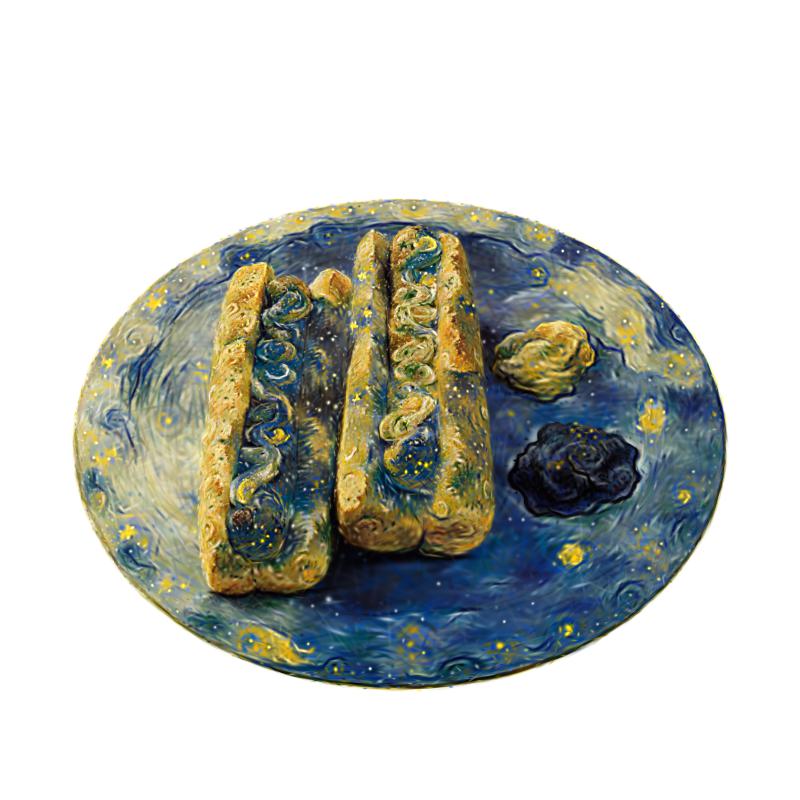}  &
      \includegraphics[trim={50 100 50 190},clip, width=0.12\textwidth,valign=c]{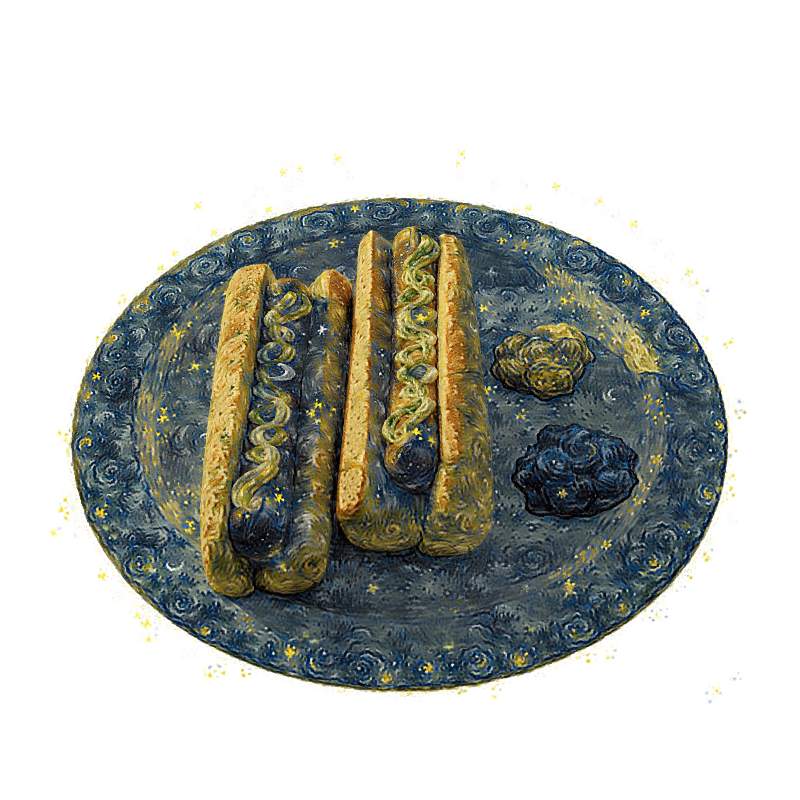}\\
      \includegraphics[trim={100 0 0 0},clip, width=0.12\textwidth,valign=c]{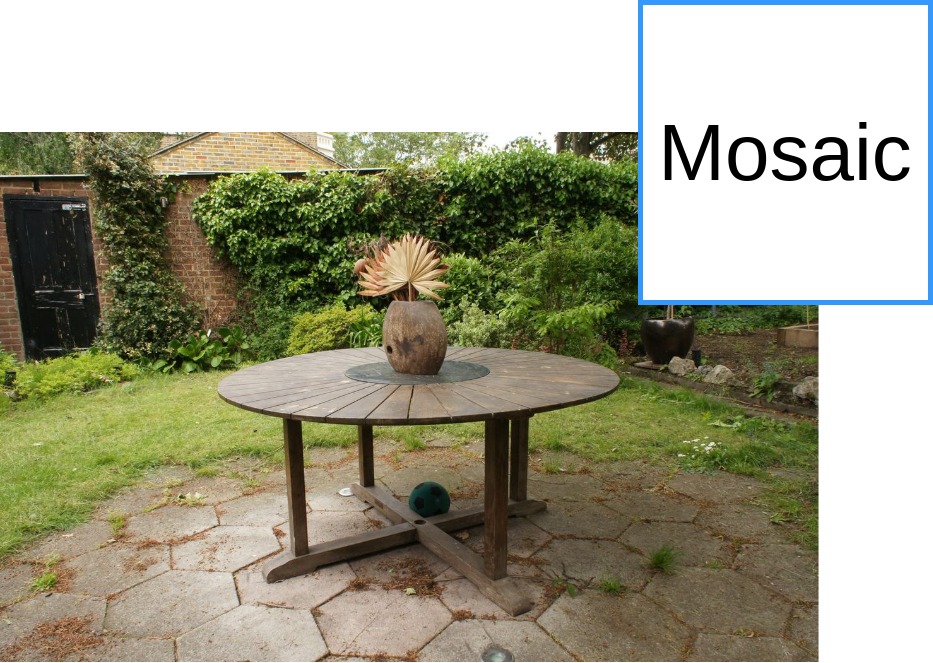}   & 
    \includegraphics[trim={100 0 80 0},clip, width=0.12\textwidth,valign=c]{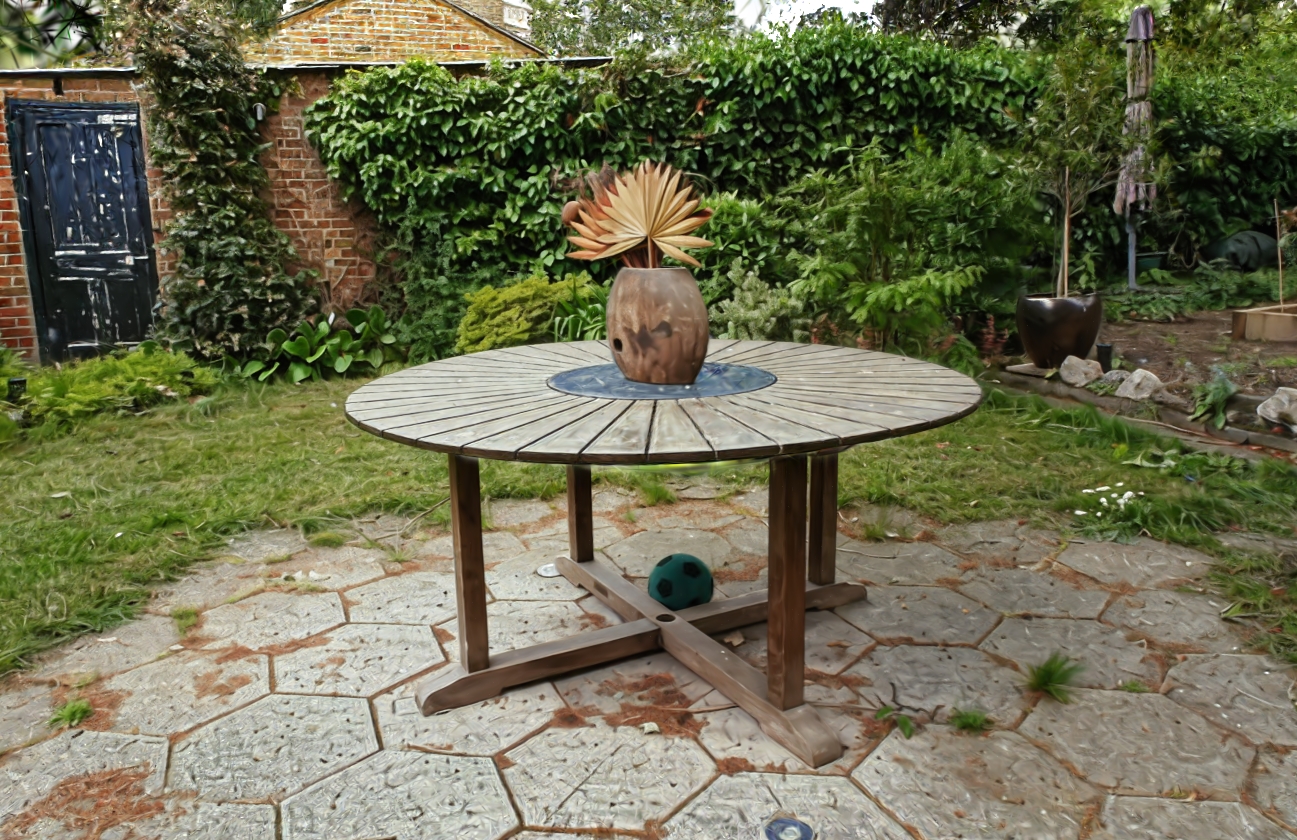}    & 
    \includegraphics[trim={100 0 80 0},clip, width=0.12\textwidth,valign=c]{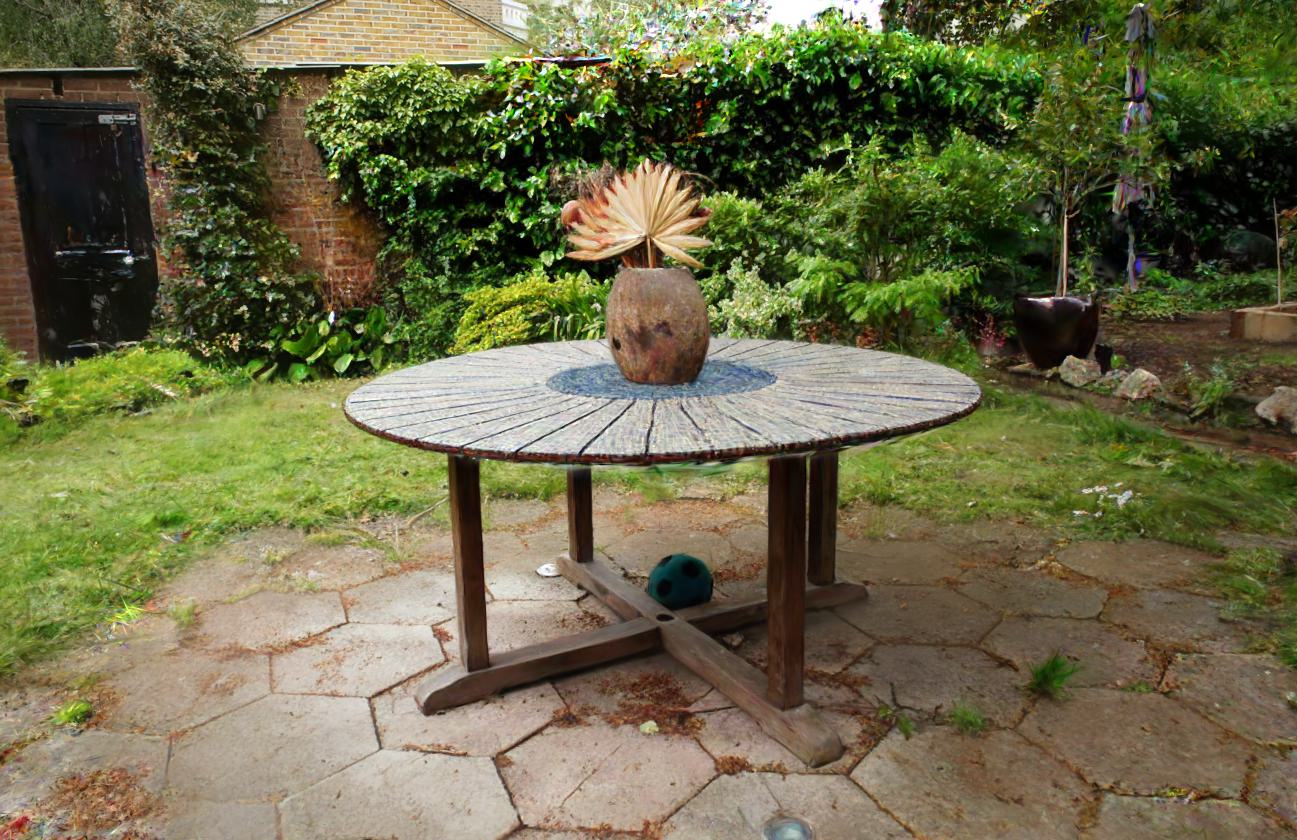}   & 
    \includegraphics[trim={100 0 80 0},clip, width=0.12\textwidth,valign=c]{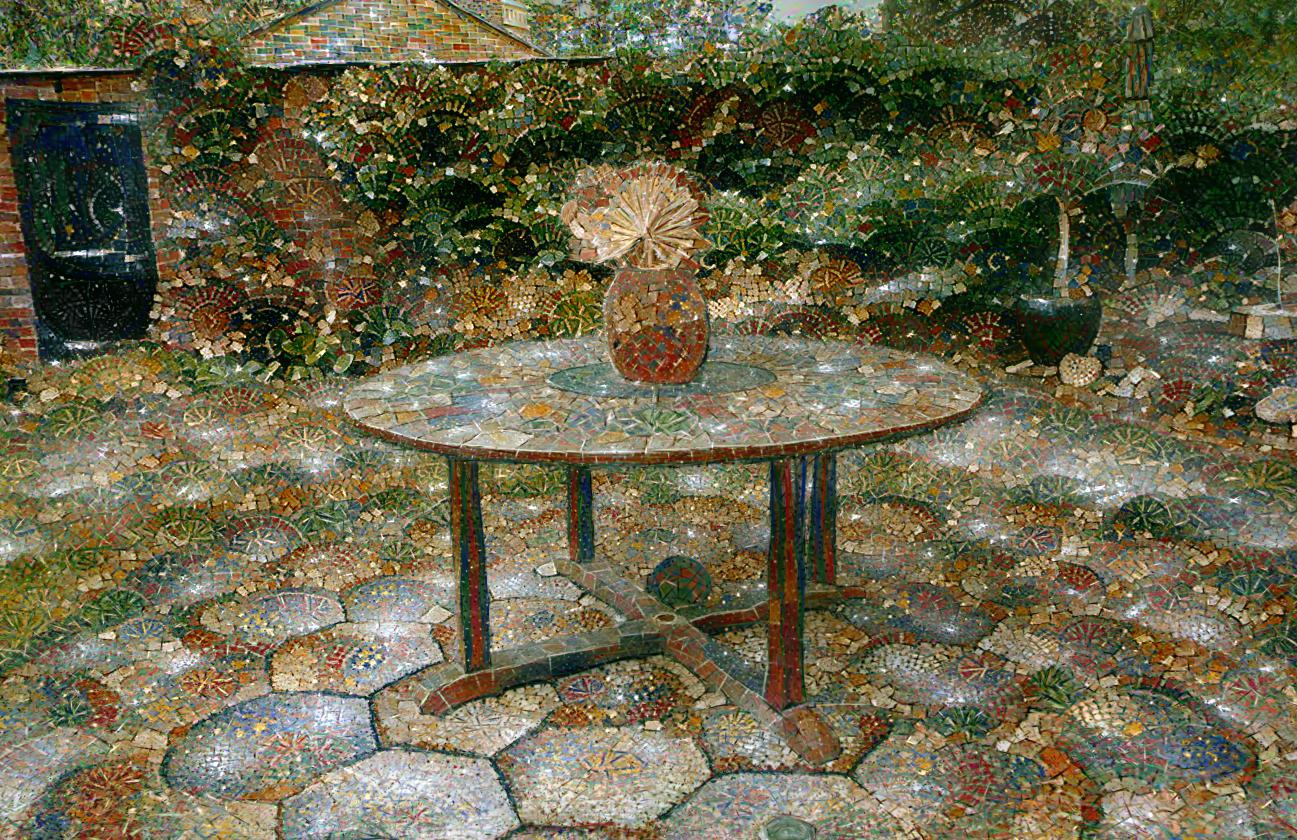}  & 
    \includegraphics[trim={100 0 80 0},clip, width=0.12\textwidth,valign=c]{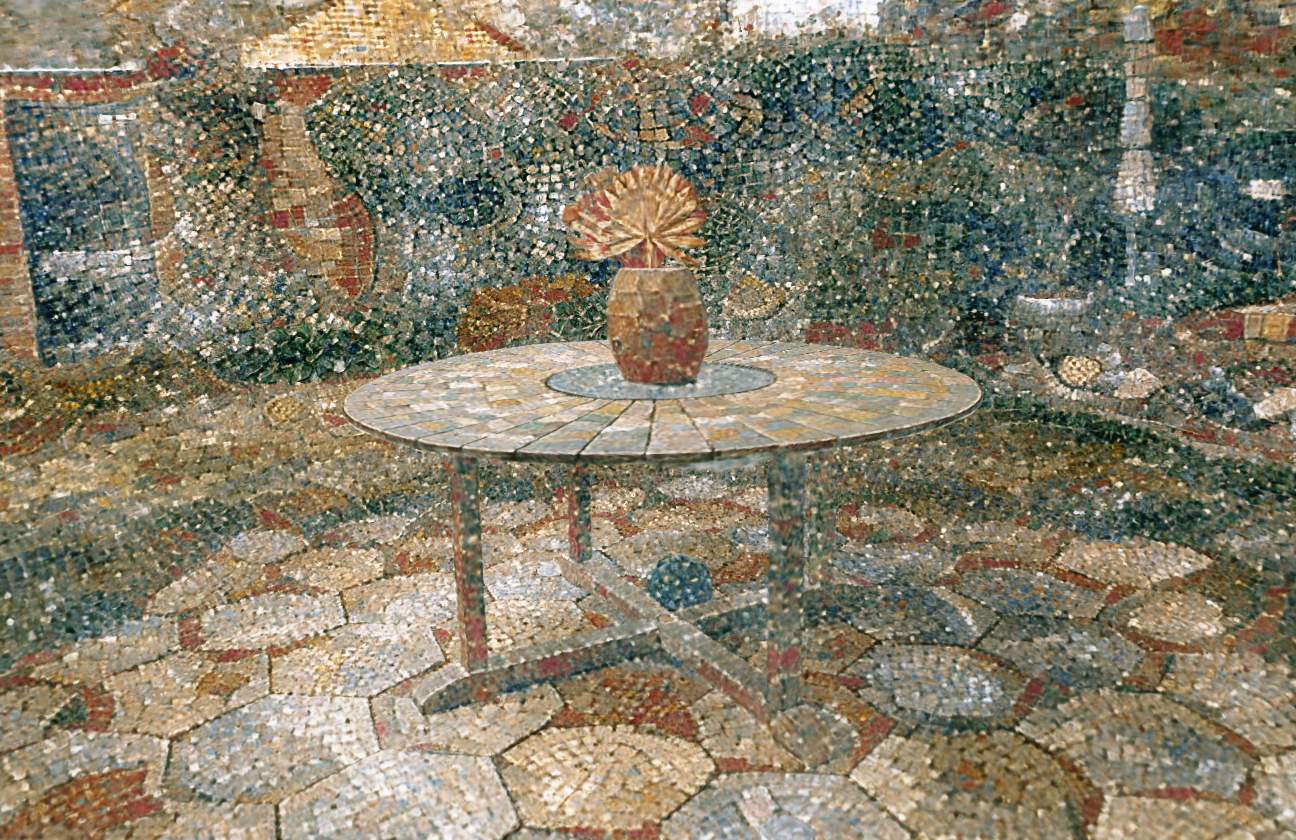} 
    \\
      \includegraphics[trim={100 100 0 0},clip, width=0.12\textwidth,valign=c]{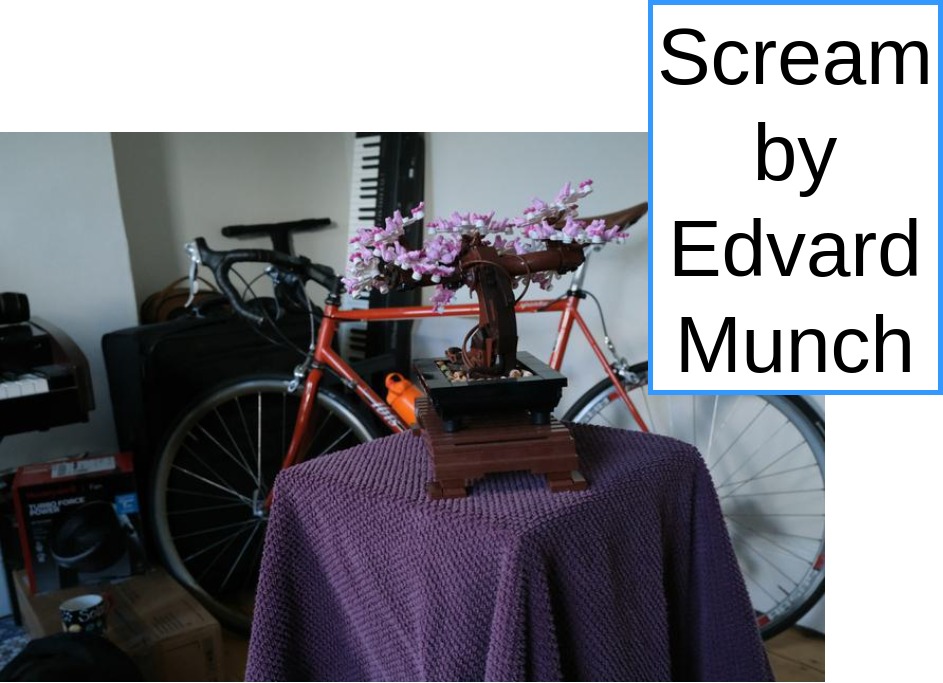}  &  
     \includegraphics[trim={100 0 0 0},clip, width=0.12\textwidth,valign=c]{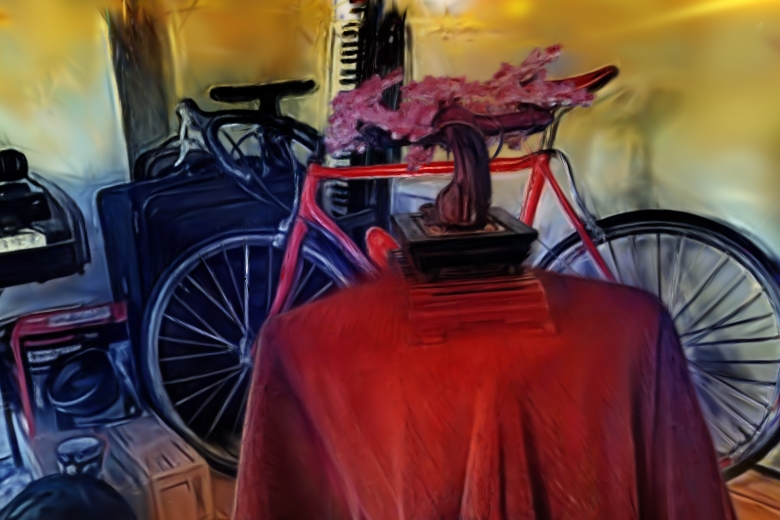}    &  
     \includegraphics[trim={100 0 0 0},clip, width=0.12\textwidth,valign=c]{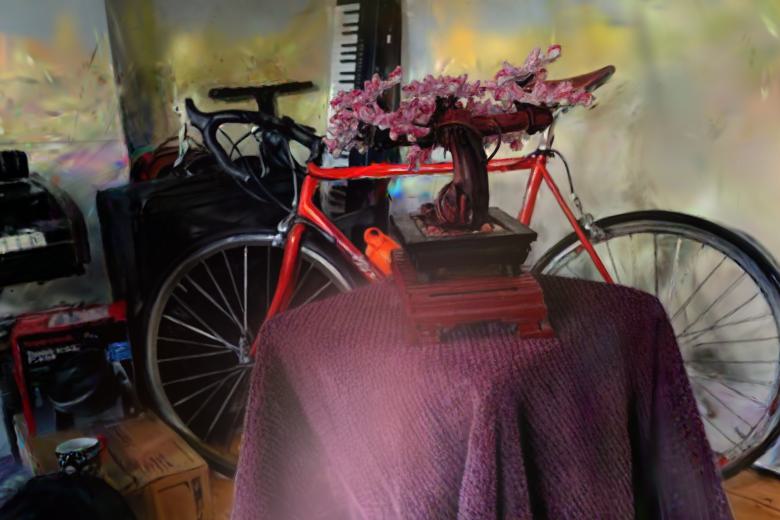}   &  
     \includegraphics[trim={100 0 0 0},clip, width=0.12
    \textwidth,valign=c]{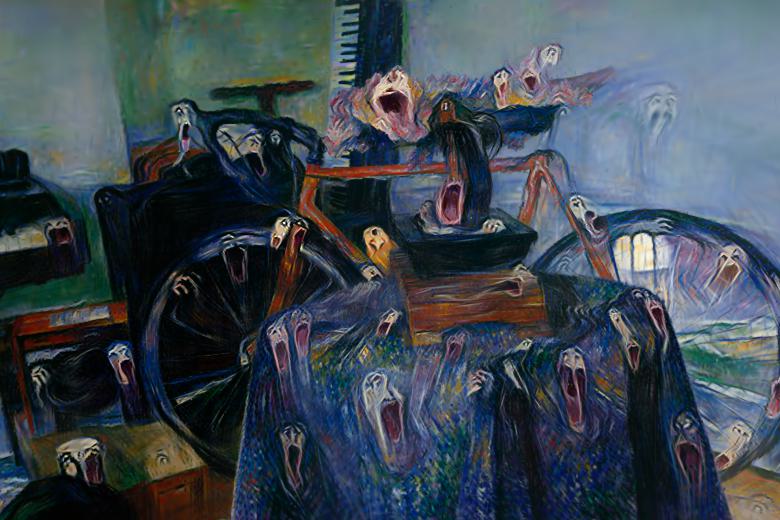}   &
     \includegraphics[trim={100 0 0 0},clip, width=0.12
    \textwidth,valign=c]{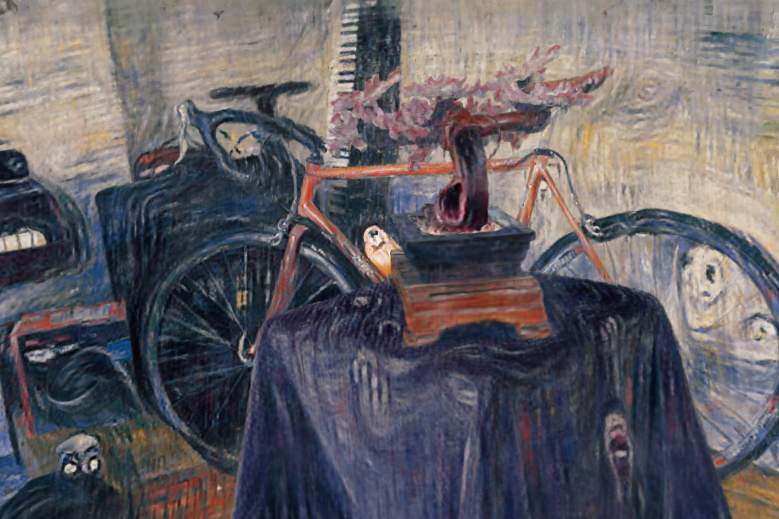}
    \end{tabular}
}
\caption{Comparison of 3D style transfer obtained by
text conditioning. Similarly to CLIPGaussian~\cite{howil2026clipgaussian}, our model captures both local textural details and the overall global aesthetic.}
\label{fig:Text2Style}
\end{figure}

\begin{figure*}[t]
    \centering
 \includegraphics[width=\textwidth]{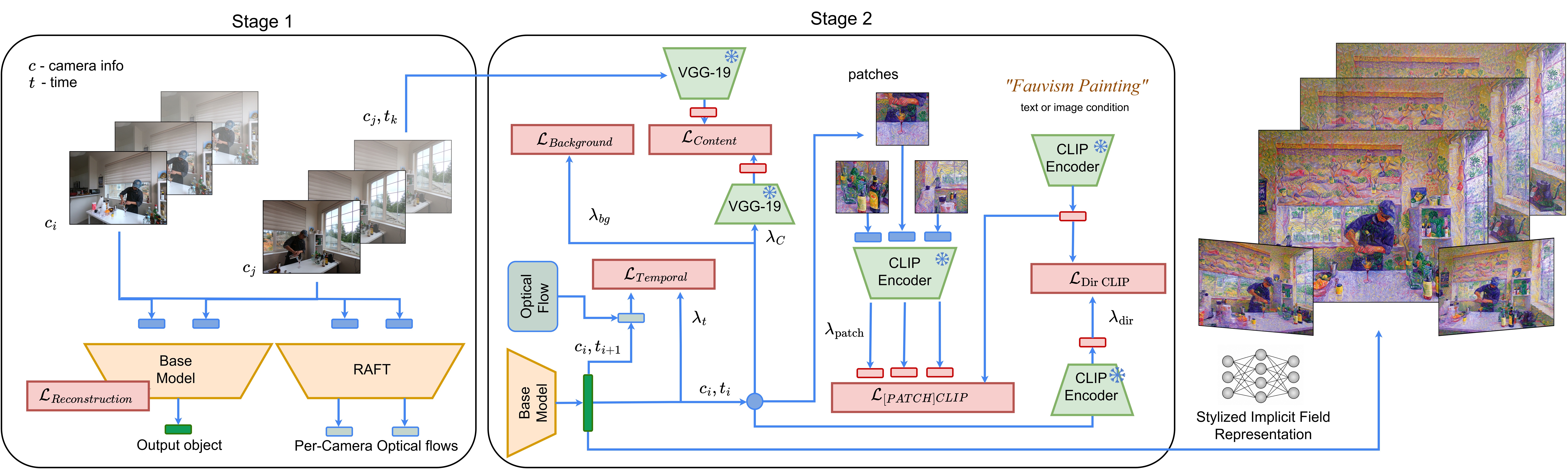}
    \caption{
    \our{} architecture in the case of a 4D dynamic scene. The process is divided into two stages. In the first stage, we train the underlying INR model, (in this case HexPlane) and calculate optical flow for each of the input cameras. The second stage consists of stylizing the INR model using VGG and CLIP embeddings of the entire model output, or of random patches from that output. We utilize optical flow to warp successive frames to calculate a temporal loss that enforces frame-to-frame coherence. The same approach would be used for video models. A very similar structure, without the optical flow and temporal loss calculation is used for static scenes.
    }
\label{fig:schema} 
\end{figure*}

\section{Related Work}

This section reviews the foundational research that contextualizes our work, focusing on the evolution of implicit and explicit visual representations, multimodal style transfer techniques, and recent advancements in unified frameworks.

\subsection{Implicit and Explicit Visual Representations}
Implicit Neural Representations have fundamentally transformed how we model visual data. For 2D images, coordinate based networks such as SIREN~\cite{sitzmann2020implicit}, LIIF~\cite{chen2021learning}, Finer~\cite{liu2024finer}, and FreSh~\cite{kania2025fresh} have demonstrated exceptional capabilities in continuous image parameterization and super resolution. Similarly, video representations have benefited from this paradigm through architectures like NeRV~\cite{chen2021nerv}, PNeRV~\cite{zhao2024pnerv}, DNeRV~\cite{zhao2023dnerv} and HNeRV~\cite{chen2023hnerv}, which encode entire video sequences into neural weights for efficient compression. In spatial domains, NeRF~\cite{mildenhall2020nerf} established the standard for 3D view synthesis, while methods like D-NeRF~\cite{pumarola2021d} and HexPlane~\cite{cao2023hexplane} extended these principles to dynamic 4D scenes. Parallel to these developments, explicit representations based on Gaussian Splatting have been adapted for lower dimensional tasks. MiraGe~\cite{waczynskamirage} and GaINeR~\cite{jakubowska2025gainer} explored point based modeling for 2D images, and VeGaS~\cite{smolak2025vegas} introduced it for video. While visually effective, applying explicit point clouds to these dense continuous domains often results in parameter counts matching the raw pixel resolution. This makes continuous Implicit Neural Representations a more naturally suited architecture for cross dimensional unification.

\subsection{Multimodal Style Transfer on Implicit Fields}
The task of visual style transfer has a rich history~\cite{gatys2016image, jing2019neural}. With the rise of Implicit Neural Representations, this task naturally expanded into the 3D domain. Several prominent methods have been developed to transfer artistic styles onto 3D scenes using reference images. ARF~\cite{zhang2022arf}, SNeRF~\cite{nguyen2022snerf}, StylizedNeRF~\cite{huang2022stylizednerf}, and StyleRF~\cite{liu2023stylerf} successfully extract perceptual features to stylize underlying neural fields while maintaining multi view consistency. Furthermore, the integration of vision language models enabled text guided 3D stylization. Instruct-NeRF2NeRF~\cite{haque2023instruct} and NeRF-Art~\cite{wang2023nerf} demonstrated how textual prompts can effectively manipulate the visual appearance of volumetric scenes. However, these methods remain strictly confined to the 3D spatial domain. A unified framework that extends these multimodal stylization capabilities across 2D, video, and 4D implicit representations has remained largely unexplored.

\subsection{Gaussian Splatting Stylization and Unified Frameworks}
Following the introduction of 3D Gaussian Splatting, a parallel line of research emerged focusing on explicit point based style transfer. Numerous methods, including StyleGaussian~\cite{liu2024stylegaussian}, ReGS~\cite{mei2024regs}, InstantStyleGaussian~\cite{yu2024instantstylegaussian}, Style3D~\cite{song2024style3d}, StyleSplat~\cite{jain2024stylesplat}, and G-Style~\cite{kovacs2024G}, achieved rapid 3D stylization by leveraging efficient rasterization. Expanding on this explicit formulation, CLIPGaussian~\cite{howil2026clipgaussian} proposed a unified backbone capable of handling multimodal style transfer across images, videos, 3D scenes, and 4D dynamics using Gaussian Splatting. While pioneering in its universality, it inherits the structural limitations of point based modeling in lower dimensions. \our{} fills this critical gap by proposing a continuous alternative. By utilizing Implicit Neural Representations, \our{} achieves the same universal and multimodal style transfer across all arbitrary dimensionalities while benefiting from the natural compression and continuity inherent to neural fields.

\section{Method}
\our{} is designed for style transfer in four different data modalities: images, videos, 3D, and 4D scenes. Similarly to CLIPGaussian~\cite{howil2026clipgaussian} provides a universal plug-in method for Gaussian Splatting, \our{} is designed as a universal stylization framework for implicit neural representations (INRs). However, INRs render each view independently, making them particularly vulnerable to temporal inconsistencies when applied to dynamic scenes. We therefore differentiate dynamic (video and 4D scenes) from static (images and 3D scenes) settings, enforcing temporal consistency only in the former.

The core idea of \our{} is to fine-tune a pretrained INR model using an image or text prompt.

\subsection{Base INR Reconstruction}

The first stage builds a neural representation of the original, non-stylized
signal (see Figure~\ref{fig:schema}). We use a different backbone for each modality, since images, videos,
static 3D scenes, and dynamic 4D scenes require different forms of input and
rendering. Each backbone is trained from scratch using its standard
reconstruction procedure and without any style supervision. The purpose of
this stage is to obtain a faithful representation of the source content that
can later be modified without having to learn the underlying signal and its
style at the same time. For video and 4D data, we also prepare the motion
information required for temporal regularization in Stage II.

\paragraph{Images.}
For a single image, we use SIREN~\cite{sitzmann2020implicit}, a coordinate-based
multilayer perceptron with sinusoidal activations. The model represents the
image as a continuous function
$f_{\theta}:[-1,1]^2\rightarrow[0,1]^3$, which maps a normalized pixel
coordinate $(x,y)$ to its RGB value. During reconstruction, the network is
evaluated on the full coordinate grid and optimized using the difference
between the predicted and original pixel colors. Once the reconstruction converges, the learned SIREN parameters
serve as the initialization for image stylization.

\paragraph{Videos.}
For videos, we use the PNeRV~\cite{zhao2024pnerv} and DNeRV~\cite{zhao2023dnerv} variants of neural video
representations. In contrast to the image case, the model must capture not only
the appearance of individual frames but also the changes that occur throughout
the sequence. Conceptually, the model produces a prediction
$\hat{I}_t$ for each position $t$ in the sequence, although its internal input
contains richer frame-level information than a simple mapping
$t\mapsto I_t$.

The reconstruction stage produces an ordered sequence of RGB frames that
closely matches the source video. We additionally run RAFT~\cite{teed2020raft}
between adjacent frames to estimate forward or backward optical flow fields,
together with occlusion masks that indicate regions where the motion estimate
is unreliable. These quantities are not used to reconstruct the base video
model itself, but they are later
required to compare neighboring stylized frames in a common coordinate system.

\paragraph{3D scenes.}
Static 3D scenes are represented with TensoRF~\cite{chen2022tensorf}, using the TensorVMSplit
factorization. Instead of storing the scene in a dense volumetric grid,
TensoRF decomposes its density and appearance fields into compact plane and
line components. For a camera ray with origin $\mathbf{o}$ and direction
$\mathbf{d}$, points are sampled along
$\mathbf{r}(s)=\mathbf{o}+s\mathbf{d}$. The model evaluates density and
appearance features at these points, and volumetric rendering combines them
into the final pixel color.

Training is performed from calibrated camera views and their corresponding
images. For every view, the model predicts an RGB image and a depth map, while
the reconstruction objective encourages the rendered RGB values to match the
observed ones. After training, the scene can be rendered from both the original
and novel viewpoints. This differentiable rendering process is important for
Stage II, because style losses computed on a rendered image can be propagated
back to the underlying 3D representation.

\paragraph{4D dynamic scenes.}
For dynamic 4D content, we use HexPlane~\cite{cao2023hexplane}. HexPlane extends
the factorized representation used for static scenes by introducing
space--time planes, allowing scene properties to vary with both position and
time. The model receives camera rays together with a timestamp $t$, and can be
viewed as a rendering function
$F_{\theta}(\mathbf{o},\mathbf{d},t)$ that produces the RGB image and depth map
for a selected camera and moment in the sequence.

During reconstruction, HexPlane learns the time-dependent density and
appearance of the scene from synchronized multi-view video. Because the same
representation is shared across cameras and timestamps, it can render the
scene from novel viewpoints while preserving its motion. As for ordinary
video, we also precompute RAFT optical flow and occlusion masks between
neighboring frames from each camera. These motion cues are stored together
with the reconstructed model and later used to align consecutive stylized
renders.

At the end of Stage I, each modality is represented by a trained base model
that reproduces the original content. For dynamic inputs, the model is
accompanied by precomputed flow fields and occlusion masks. Stage II starts
from these reconstructed representations and modifies their appearance by
fine-tuning the model parameters under a shared set of perceptual and
style-based objectives.

\subsection{Multimodal Stylization}

Stage II starts from the modality-specific representation obtained in
Stage I. Given the reconstructed parameters $\theta_0$, we initialize the
stylized model as $\theta \leftarrow \theta_0$. The trainable parameters of the underlying representation are then updated to modify its appearance, without requiring full reconstruction of the signal.

Before optimization, the target style $\mathcal{S}$ is provided either as a
text prompt or as a reference image. It is encoded once using the corresponding
frozen CLIP text or image encoder, and since both encoders map their inputs into
the same embedding space, the remaining training procedure is identical for
the two conditioning modes.

At each optimization step, we select a source image $I_l$ and generate the
corresponding render $R_l$ using the current model parameters. Depending on the
modality, $I_l$ represents the original image, a video frame, a training view
of a static 3D scene, or a frame associated with a particular camera and time
in a 4D scene. The render $R_l$ is generated using the same spatial
coordinates, frame index, camera viewpoint, or camera--time pair, respectively.

The full render $R_l$ is passed through the frozen CLIP image encoder to provide
global style supervision. We also sample a set of random crops
$P_l=\{p_i(R_l)\}_{i=1}^{m}$, which provide local supervision for fine-scale
properties such as textures, brush strokes, and repeated patterns. In parallel,
both $R_l$ and its corresponding source image $I_l$ are processed by the frozen
VGG-19 network to obtain features used for preserving the original content and
spatial structure.

For video and 4D scenes, we additionally render a neighboring timestamp. In
the 4D case, the camera viewpoint is kept fixed while the time changes. The
precomputed optical flow and occlusion masks from Stage I are then used to
align the neighboring renders and provide temporal supervision in regions
where the correspondence is reliable.

The global CLIP, patch-level CLIP, content, background, and, when applicable,
temporal terms are combined into the total objective described below. The
resulting gradient is backpropagated through the differentiable renderer to
the trainable parameters $\theta$. Repeating this process gradually transfers
the target appearance while preserving the content, geometry, and motion
captured during Stage I.

\begin{figure}[t]
\centering
\small
\setlength{\tabcolsep}{1pt}
\resizebox{\columnwidth}{!}{%
    \begin{tabular}{cccccc}
    Style & Original &  CLIPstyler~\cite{kwon2022clipstyler}  &  FastCLIPstyler~\cite{Suresh_2024_WACV}   &  CLIPGaussian~\cite{howil2026clipgaussian} &  \textbf{OmniStyle-INR} \\
      \fbox{
      \begin{minipage}[c][0.08\textwidth][c]{0.12\textwidth}
        \centering
         Fire
      \end{minipage}
    }
      &
      \includegraphics[trim={0 0 0 0},clip, width=0.15\textwidth,valign=c]{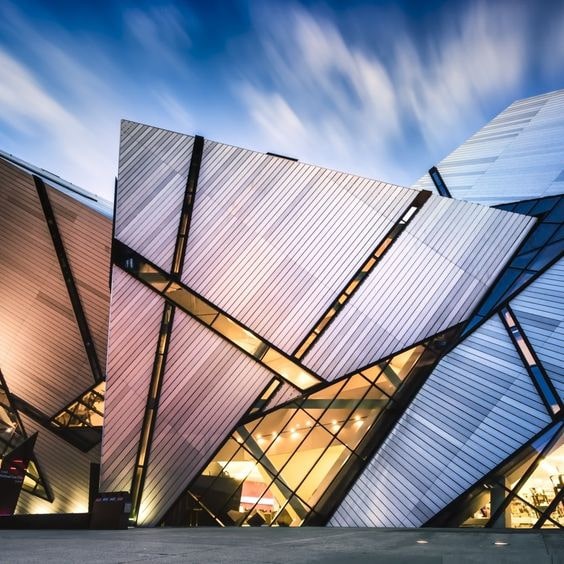}   & 
    \includegraphics[trim={0 0 0 0},clip, width=0.15\textwidth,valign=c]{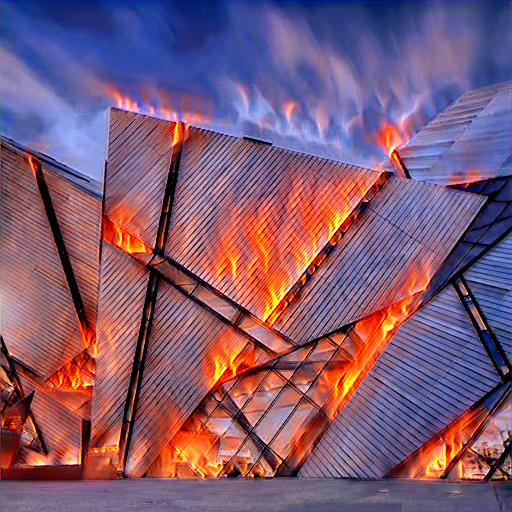}    & 
    \includegraphics[trim={0 0 0 0},clip, width=0.15\textwidth,valign=c]{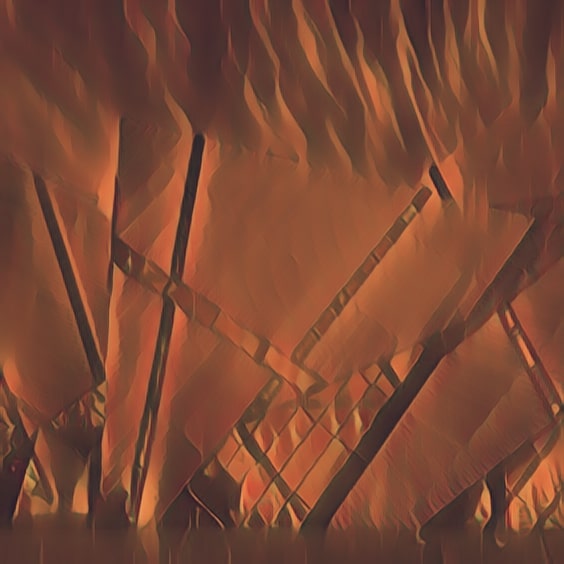}   & 
    \includegraphics[trim={0 0 0 0},clip, width=0.15\textwidth,valign=c]{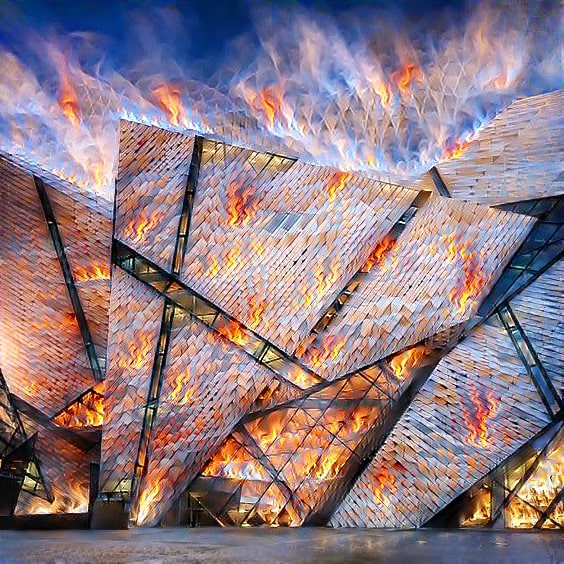}  &
    \includegraphics[trim={0 0 0 0},clip, width=0.15\textwidth,valign=c]{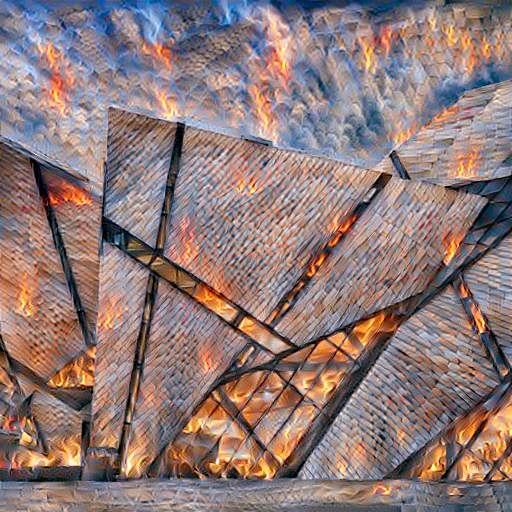}  \\
    \fbox{
      \begin{minipage}[c][0.08\textwidth][c]{0.12\textwidth}
        \centering
          Starry Night by Vincent van Gogh
      \end{minipage}
    }
    &  
      \includegraphics[trim={0 0 0 0},clip, width=0.15\textwidth,valign=c]{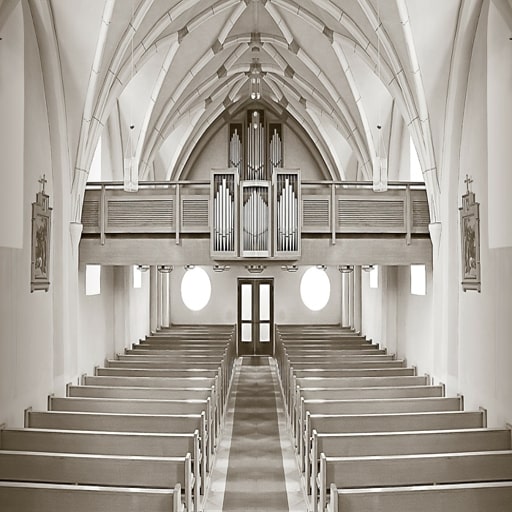}   & 
     \includegraphics[trim={0 0 0 0},clip, width=0.15\textwidth,valign=c]{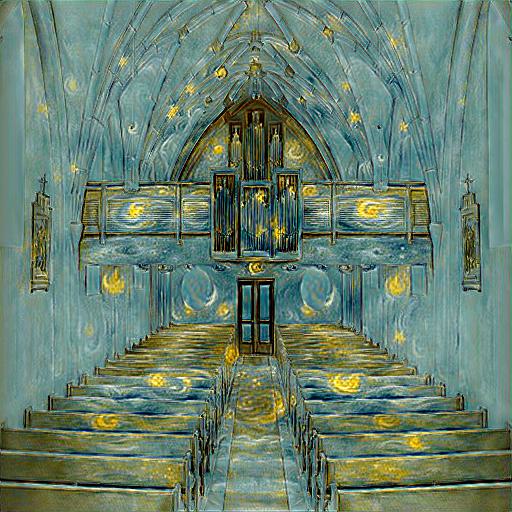}     &  
     \includegraphics[trim={0 0 0 0},clip, width=0.15\textwidth,valign=c]{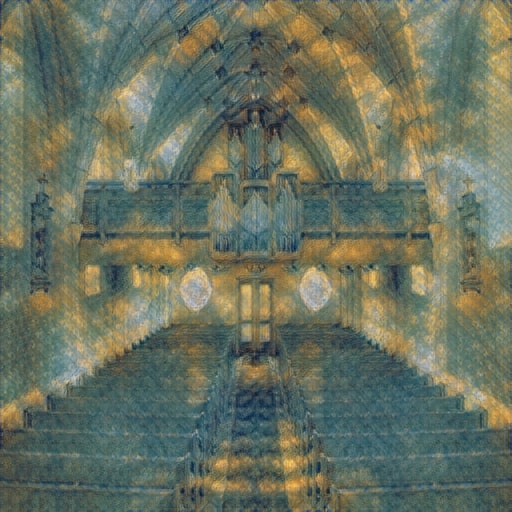}    &  
     \includegraphics[trim={0 0 0 0},clip, width=0.15\textwidth,valign=c]{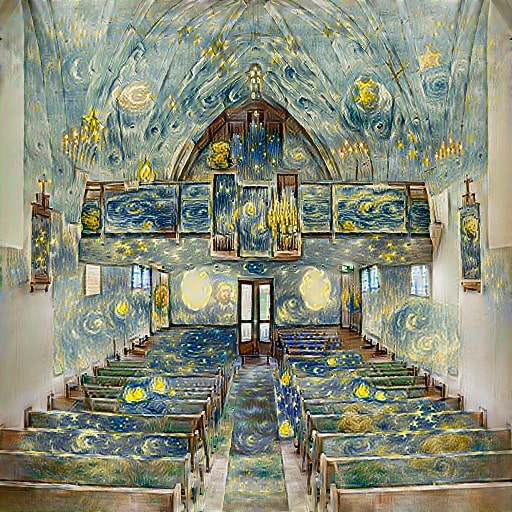}  &
     \includegraphics[trim={0 0 0 0},clip, width=0.15\textwidth,valign=c]{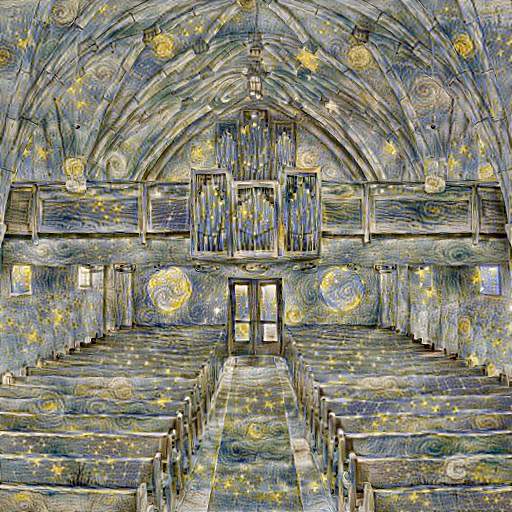}  \\
    \fbox{
      \begin{minipage}[c][0.08\textwidth][c]{0.12\textwidth}
        \centering
          Mosaic
      \end{minipage}
    }
      &
    \includegraphics[trim={0 0 0 0},clip, width=0.15\textwidth,valign=c]{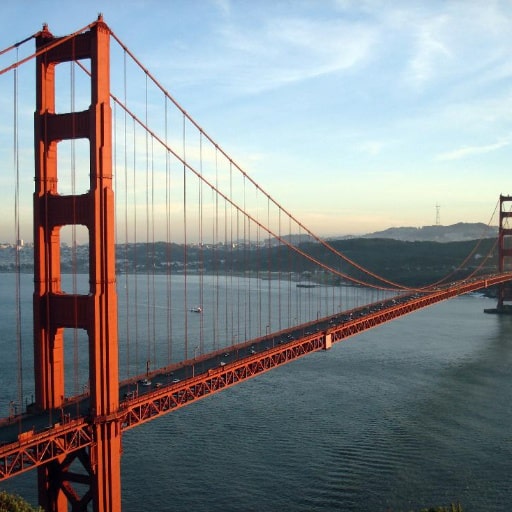}   & 
    \includegraphics[trim={0 0 0 0},clip, width=0.15\textwidth,valign=c]{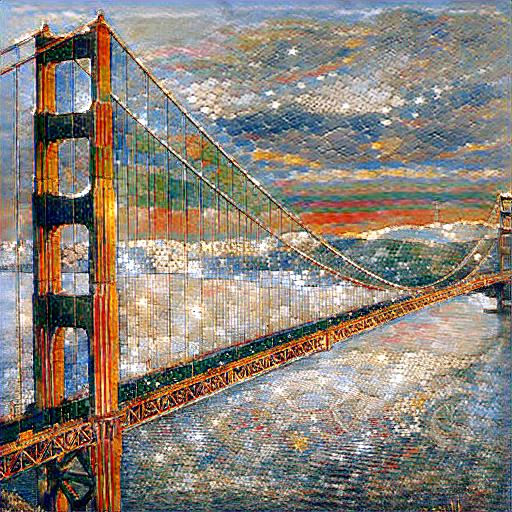}    & 
    \includegraphics[trim={0 0 0 0},clip, width=0.15\textwidth,valign=c]{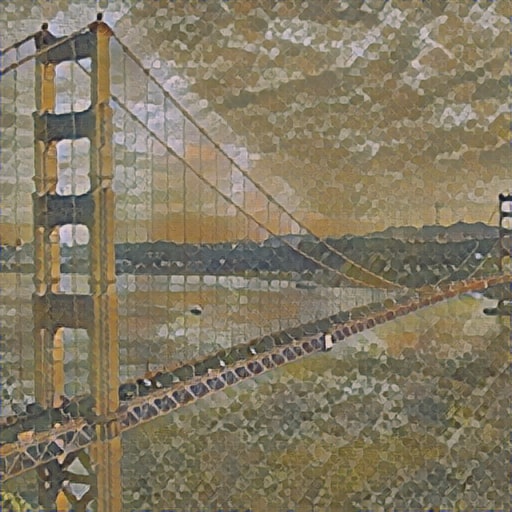}   & 
    \includegraphics[trim={0 0 0 0},clip, width=0.15\textwidth,valign=c]{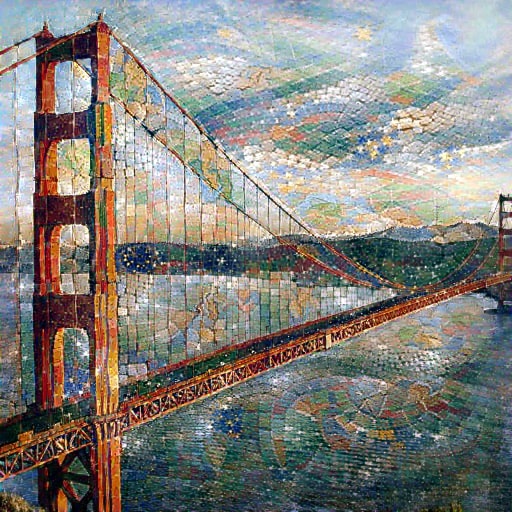}  &
    \includegraphics[trim={0 0 0 0},clip, width=0.15\textwidth,valign=c]{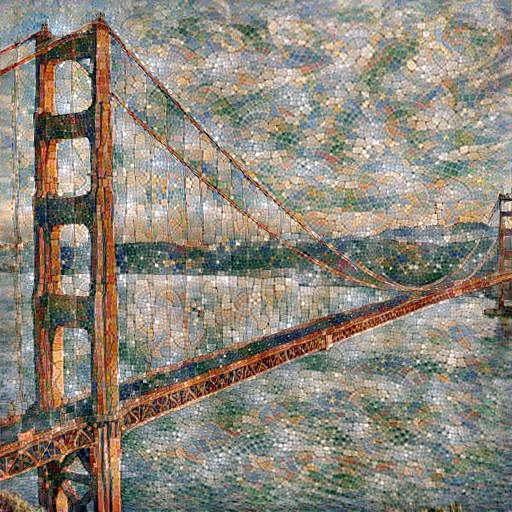}  \\
    \fbox{
      \begin{minipage}[c][0.08\textwidth][c]{0.12\textwidth}
        \centering
          Scream by Edvard Munch
      \end{minipage}
    }&  
      \includegraphics[trim={0 0 0 0},clip, width=0.15\textwidth,valign=c]{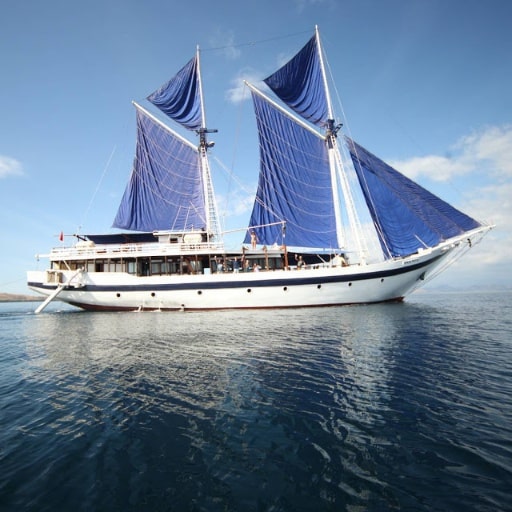}   &  
     \includegraphics[trim={0 0 0 0},clip, width=0.15\textwidth,valign=c]{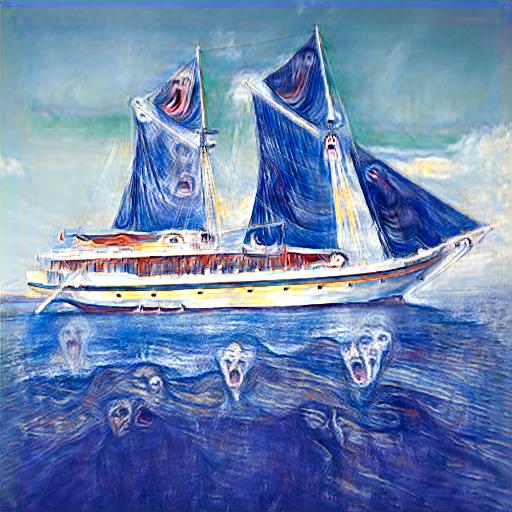}    & 
     \includegraphics[trim={0 0 0 0},clip, width=0.15\textwidth,valign=c]{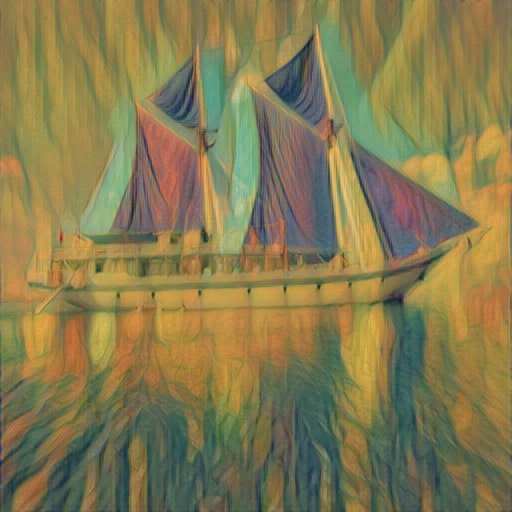}   &  
     \includegraphics[trim={0 0 0 0},clip, width=0.15\textwidth,valign=c]{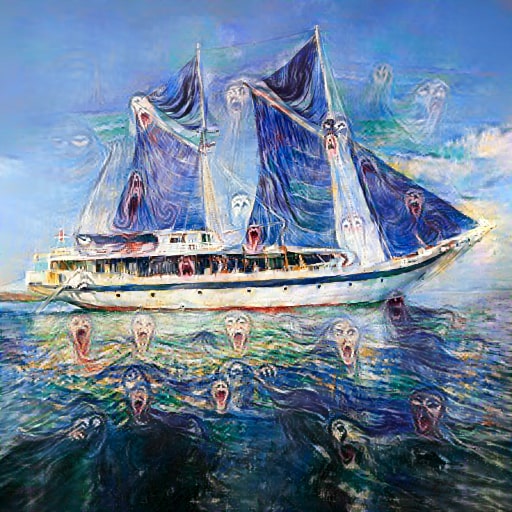}  &
     \includegraphics[trim={0 0 0 0},clip, width=0.15\textwidth,valign=c]{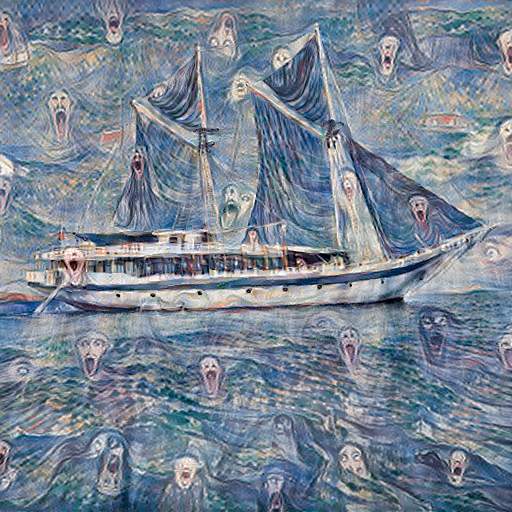}  \\
    \end{tabular}
}
\caption{Comparison of image style transfer using text condition on the MS-COCO \cite{lin2014microsoftcoco} dataset. Our model stylizes local details uniformly across the entire image, rather than concentrating stylization at the center.} 
\label{fig:2D_comp_text_full}
\end{figure}

\subsection{\our{} Loss function}

Our loss function, similarly to CLIPGaussian~\cite{howil2026clipgaussian}, uses two pretrained models, CLIP and VGG-19, which we will refer to as $\Phi_{CLIP}$ or $\Phi_{VGG}$ respectively. $\Phi_{CLIP}$ is used for style transfer, and $\Phi_{VGG}$ embeddings are used to ensure content preservation.

The total loss $\mathcal{L}_{loss}$ is formulated as a weighted loss of four terms  for static scenes (images and 3D)
$$
\mathcal{L}_{loss} = \lambda_c \mathcal{L}_{c} + \lambda_p \mathcal{L}_{patch} + \lambda_d \mathcal{L}_{d} + \lambda_b \mathcal{L}_{b}
$$
And of five terms for dynamic scenes (video and 4D):
$$
\mathcal{L}_{loss} = \lambda_c \mathcal{L}_{c} + \lambda_p \mathcal{L}_{patch} + \lambda_d \mathcal{L}_{d} + \lambda_b \mathcal{L}_{b} + \lambda_t \mathcal{L}_{temp}
$$

where $\lambda_c, \lambda_p, \lambda_d, \lambda_b, \lambda_t$ are hyperparameters controlling the relative contribution of each loss term.

$\mathcal{L}_{c}$ or \textbf{Content Loss} is responsible for ensuring the stylized output retains the structure of the original scene. $\mathcal{L}_{c}$ is calculated as a mean squared error (\textbf{MSE}) between the
\texttt{conv4\_2} and  \texttt{conv5\_2} features of the
original image $I_l$ and the rendered image $R_l$, just like in~\cite{howil2026clipgaussian}:
$$
\mathcal{L}_c(R_l, I_l) =
MSE( \Phi_{VGG}\left(R_l \right) , \Phi_{VGG}\left( I_l \right) ).
$$

\begin{figure}[t]
\centering
\small
\setlength{\tabcolsep}{1pt}
\resizebox{\columnwidth}{!}{%
    \begin{tabular}{cccccc}
    Style & Original &  AdaIN~\cite{Huang_2017_ICCV} &  $ \text{StyTr}^2 $
     ~\cite{Deng_2022_CVPR} &  CLIPGaussian~\cite{howil2026clipgaussian} &  \textbf{OmniStyle-INR} \\
    \includegraphics[trim={0 0 0 0},clip, width=0.10\textwidth,valign=c]{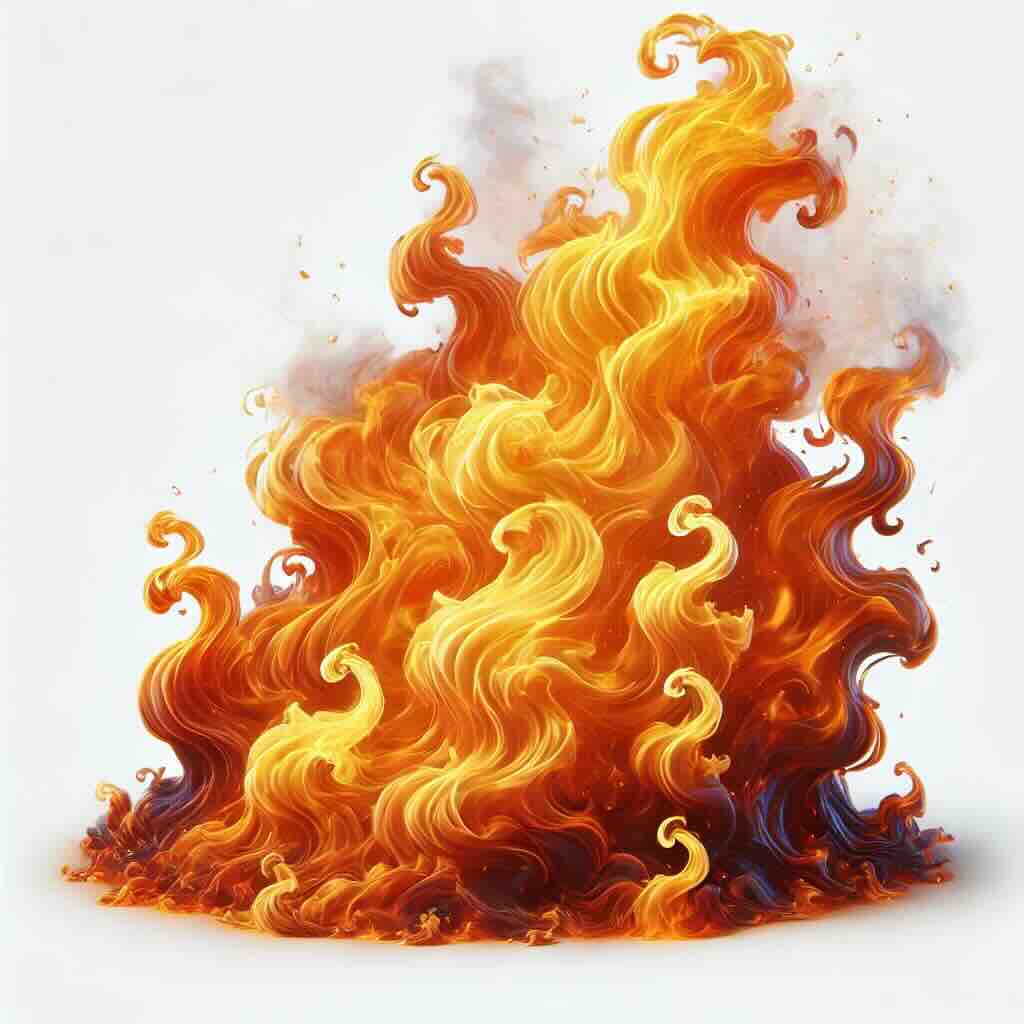}
      &
      \includegraphics[trim={0 0 0 0},clip, width=0.12\textwidth,valign=c]{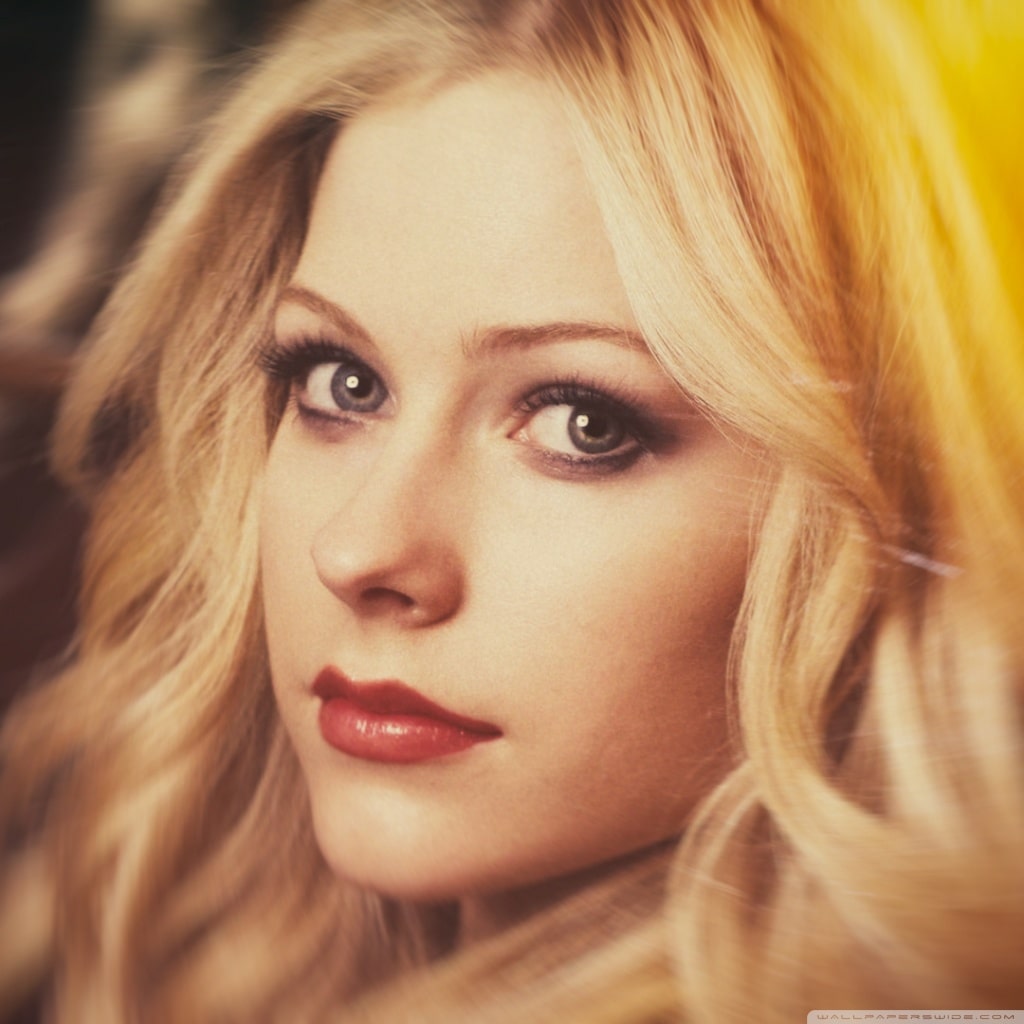}   & 
    \includegraphics[trim={0 0 0 0},clip, width=0.12\textwidth,valign=c]{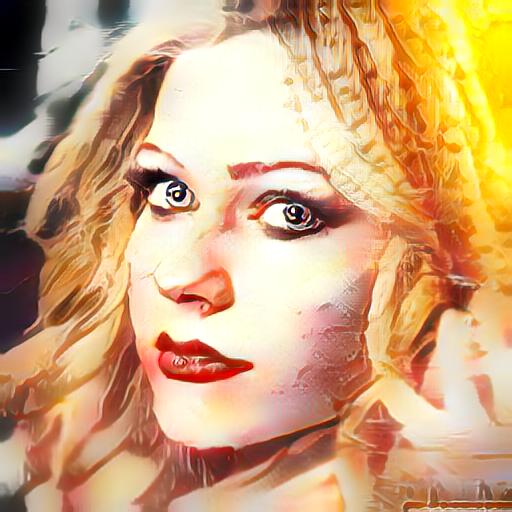}    & 
    \includegraphics[trim={0 0 0 0},clip, width=0.12\textwidth,valign=c]{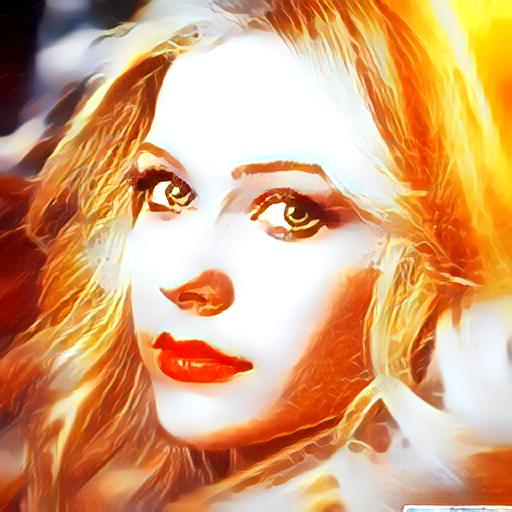}   & 
    \includegraphics[trim={0 0 0 0},clip, width=0.12\textwidth,valign=c]{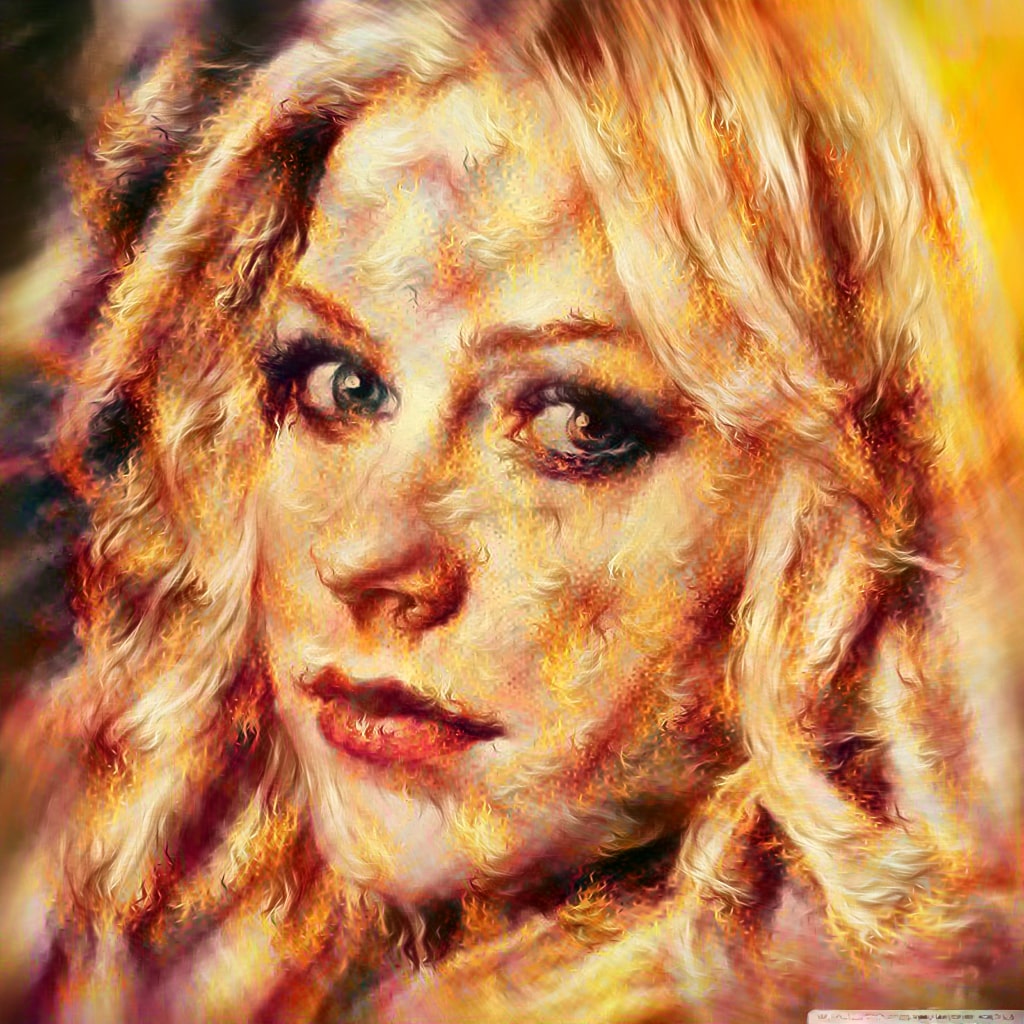}  &
    \includegraphics[trim={0 0 0 0},clip, width=0.12\textwidth,valign=c]{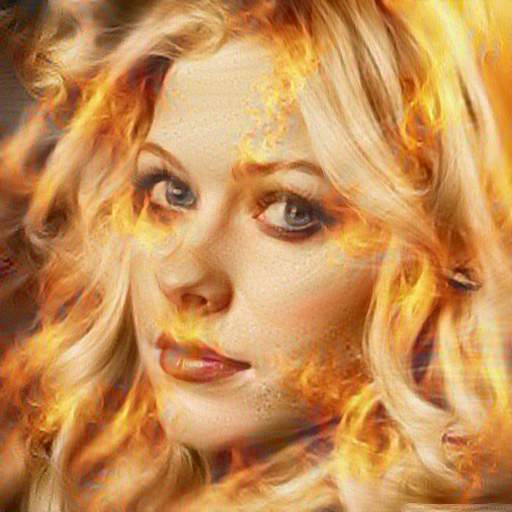}  \\
    \includegraphics[trim={0 0 0 0},clip, width=0.10\textwidth,valign=c]{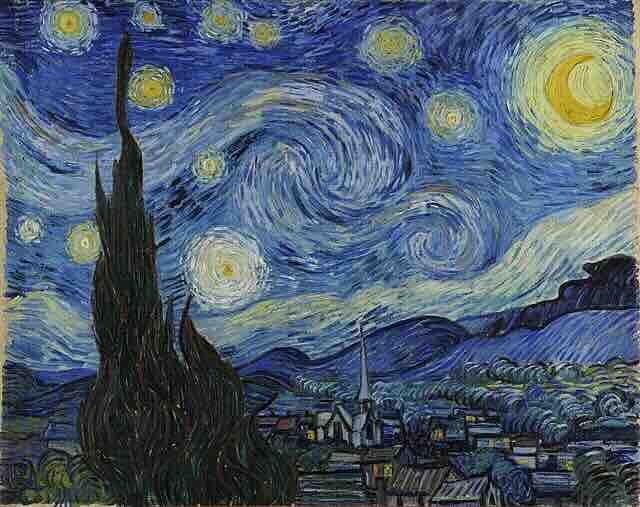}&  
      \includegraphics[trim={0 0 0 0},clip, width=0.12\textwidth,valign=c]{imgs/2D/comp_text/ours/8.jpg}   &  
     \includegraphics[trim={0 0 0 0},clip, width=0.12\textwidth,valign=c]{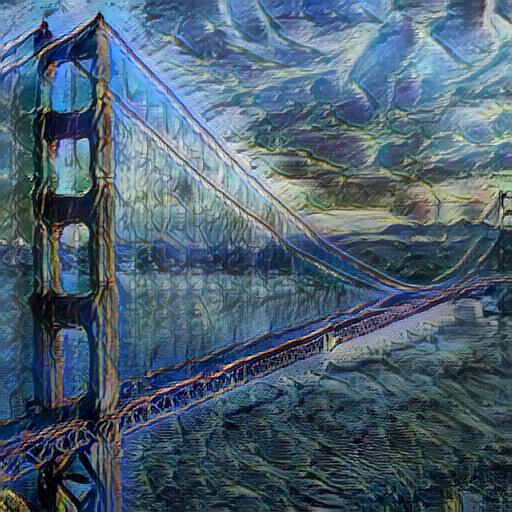}   &  
     \includegraphics[trim={0 0 0 0},clip, width=0.12\textwidth,valign=c]{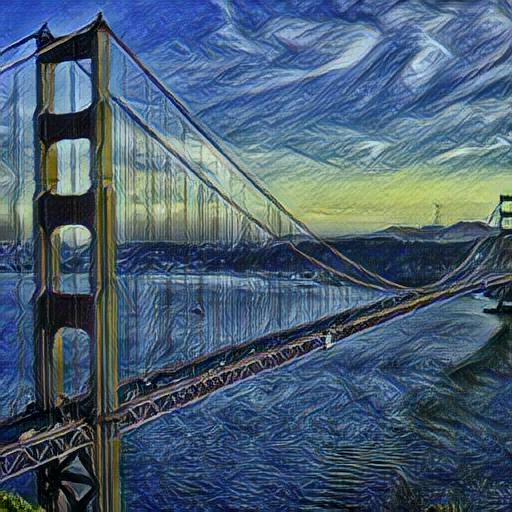}    &  
     \includegraphics[trim={0 0 0 0},clip, width=0.12\textwidth,valign=c]{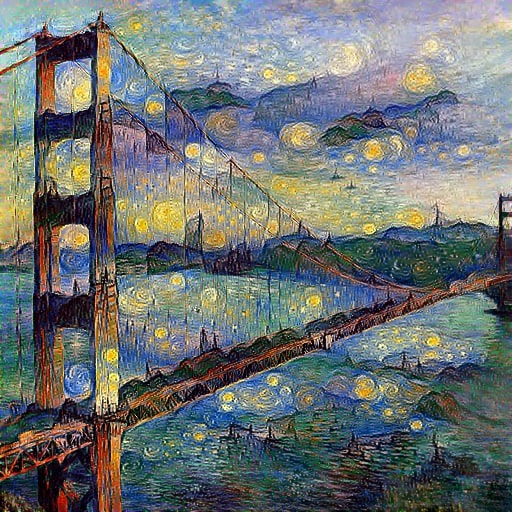}  &
     \includegraphics[trim={0 0 0 0},clip, width=0.12\textwidth,valign=c]{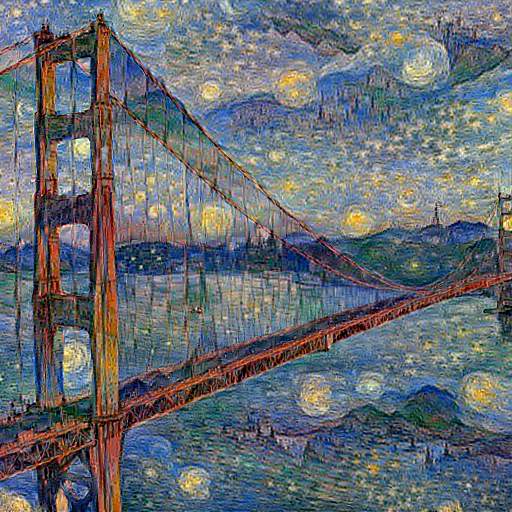}  \\
    
    \includegraphics[trim={0 0 0 0},clip, width=0.10\textwidth,valign=c]{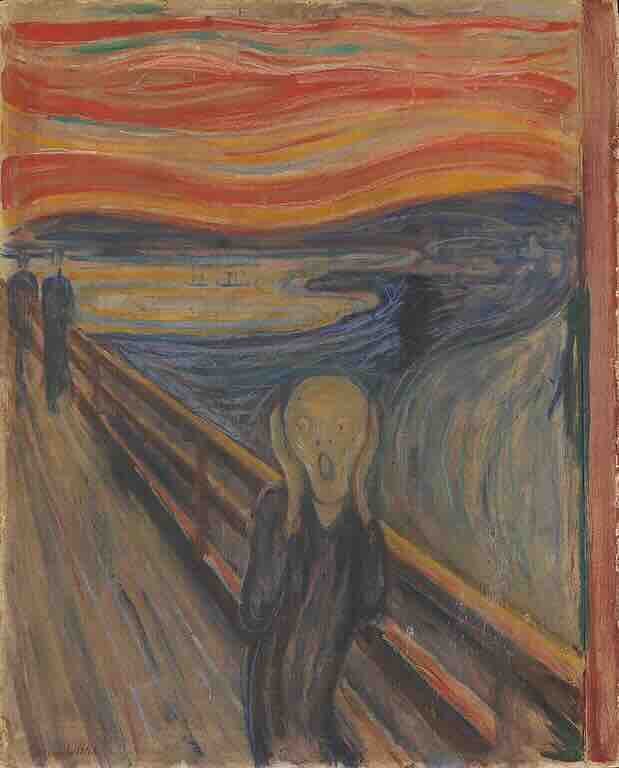}&  
    \includegraphics[trim={0 0 0 0},clip, width=0.12\textwidth,valign=c]{imgs/2D/comp_text/ours/17.jpg}   &  
    \includegraphics[trim={0 0 0 0},clip, width=0.12\textwidth,valign=c]{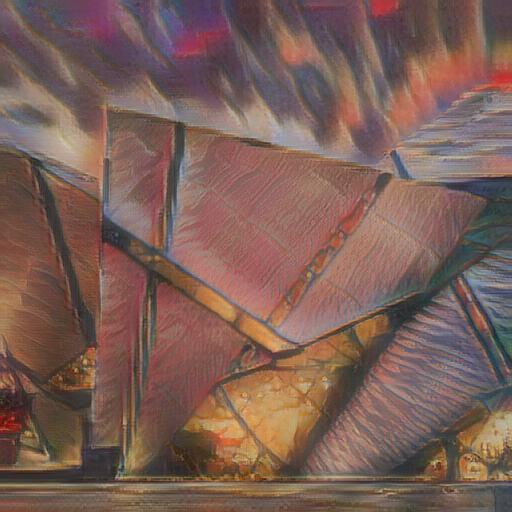}   &  
    \includegraphics[trim={0 0 0 0},clip, width=0.12\textwidth,valign=c]{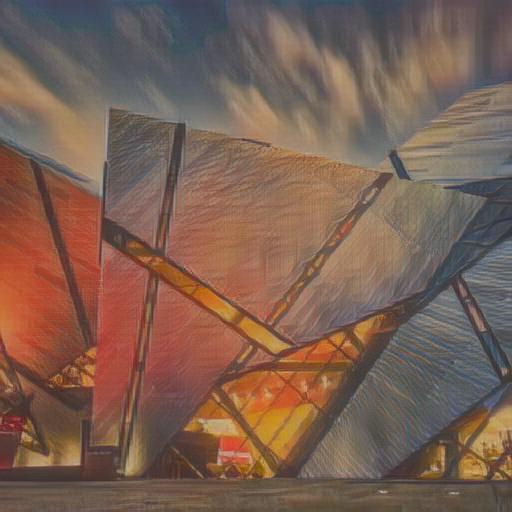}    &  
    \includegraphics[trim={0 0 0 0},clip, width=0.12\textwidth,valign=c]{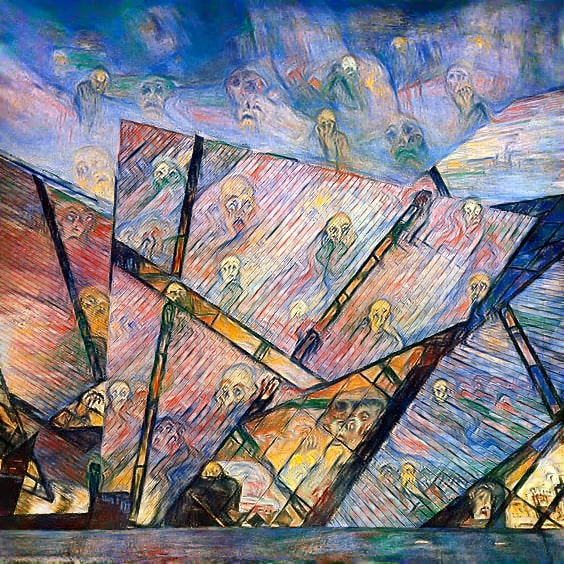}  &
    \includegraphics[trim={0 0 0 0},clip, width=0.12\textwidth,valign=c]{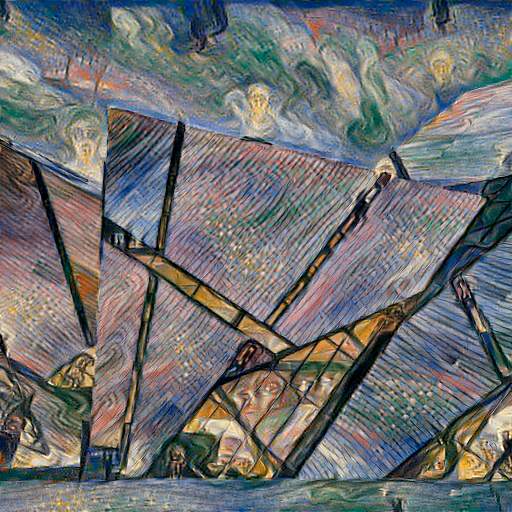}  \\
    
    \includegraphics[trim={0 0 0 0},clip, width=0.10\textwidth,valign=c]{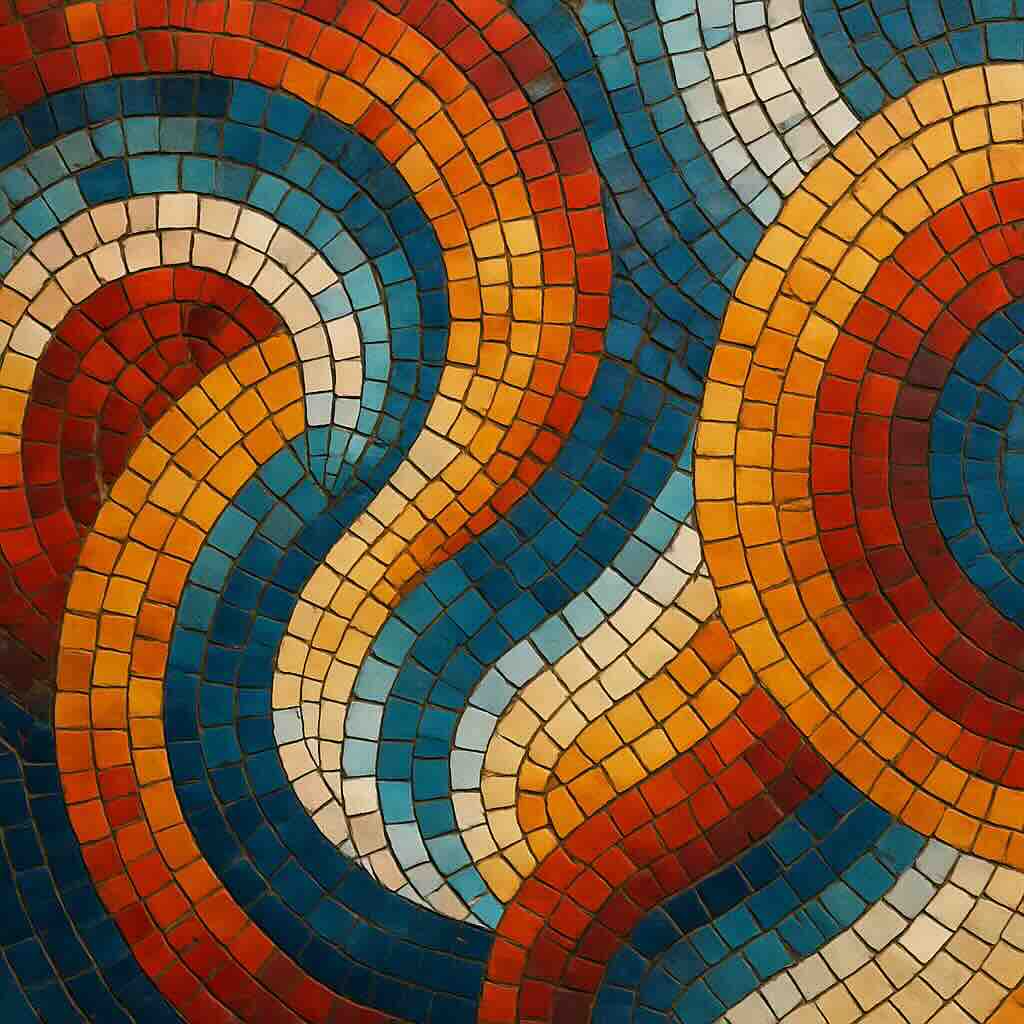}&  
    \includegraphics[trim={0 0 0 0},clip, width=0.12\textwidth,valign=c]{imgs/2D/comp_text/ours/11_resized.jpg}   &  
    \includegraphics[trim={0 0 0 0},clip, width=0.12\textwidth,valign=c]{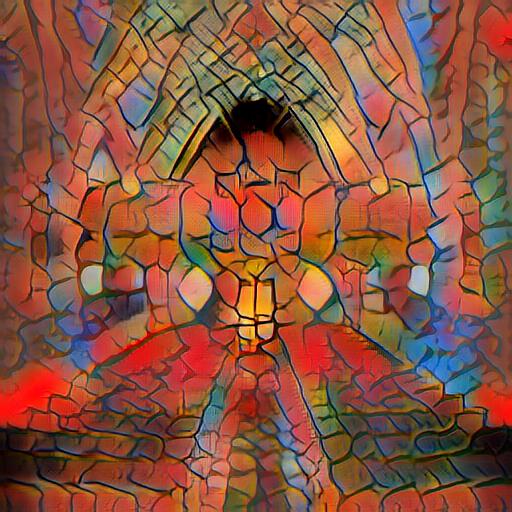}   &  
    \includegraphics[trim={0 0 0 0},clip, width=0.12\textwidth,valign=c]{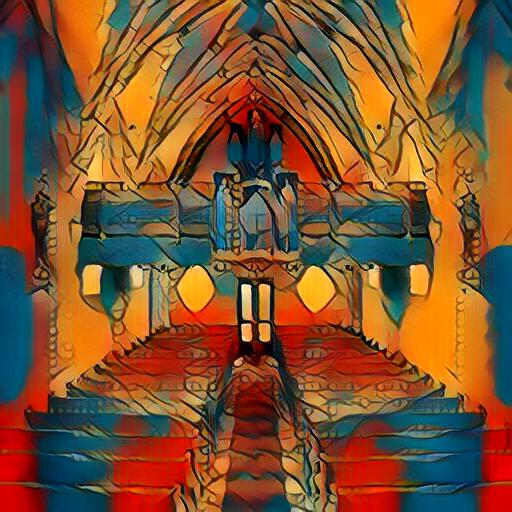}    &  
    \includegraphics[trim={0 0 0 0},clip, width=0.12\textwidth,valign=c]{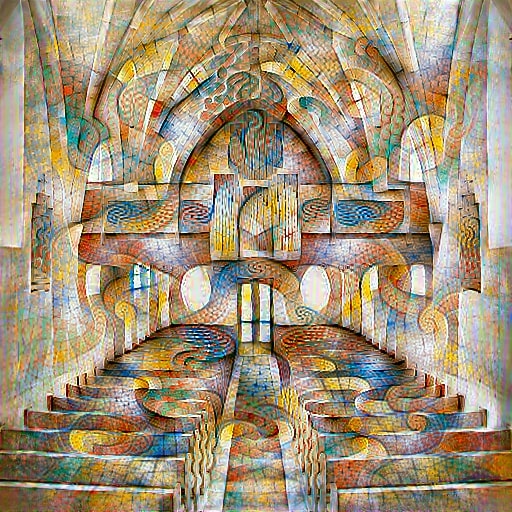}  &
    \includegraphics[trim={0 0 0 0},clip, width=0.12\textwidth,valign=c]{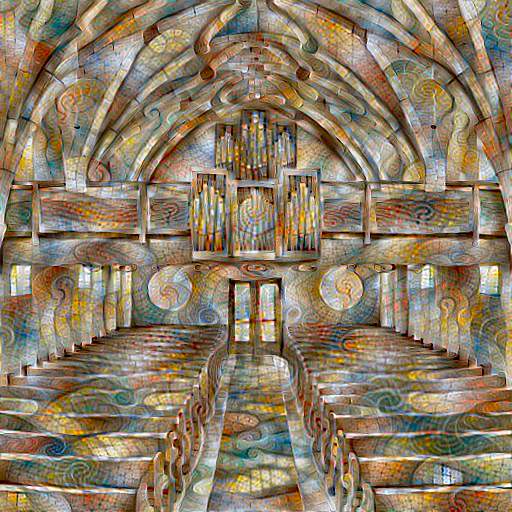}  \\
    \end{tabular}
}
\caption{Comparison of image style transfer using image condition, on the MS-COCO \cite{lin2014microsoftcoco} dataset.}
\label{fig:2D_comp_image_full}
\end{figure}

$\mathcal{L}_{d}$ or \textbf{Directional CLIP loss} aligns the change in the rendered image's CLIP embedding space with the change from \textit{"Photo"} to the target style prompt or image, as in other style transfer models ~\cite{gatys2016image}:

\[
\begin{aligned}
\mathcal{L}_d(R_l, I_l)
&= 1 - \cos\Bigl(
    \Phi_{\mathrm{CLIP}}(R_l) - \Phi_{\mathrm{CLIP}}(I_l), \\
&\qquad\qquad
    \Phi_{\mathrm{CLIP}}(\mathcal{S})
    - \Phi_{\mathrm{CLIP}}(\text{``Photo''})
\Bigr).
\end{aligned}
\]

$\mathcal{L}_{p}$ or \textbf{Patch CLIP loss} is an average Directional CLIP loss calculated on random patches $P$, responsible for style transfer on the local level:

$$
\mathcal{L}_{p}(R_l, I_l) = \frac{1}{n}\sum_{i=1}^n L_{d}( p_i(R_l)), I_l).
$$

$\mathcal{L}_{b}$ or \textbf{Background loss} is a weighted average over a background mask. For synthetic datasets, this loss encourages the background to remain clean and free of stylization artifacts.

$\mathcal{L}_{t}$ or \textbf{Temporal loss} enforces temporal consistency across consecutive frames in dynamic scenes (Video and 4D). We pre-compute optical flow and occlusion masks using RAFT~\cite{teed2020raft} for each adjacent frame pair. During training, we warp the next frame back to the current frame's coordinates using the estimated flow and compute a masked Mean Squared Error (MSE) difference between the current rendered frame and the warped next frame, excluding occluded pixels where flow is unreliable. This encourages the model to produce temporally stable, flicker-free results that move consistently with the original scene.

$$
\mathcal{L}_{t} = MSE(warp(R_{l+1}), R_l)
$$

\begin{table}[t]
    \centering
    \caption{Quantitative comparison of video style transfer using
    CLIP-based metrics. Larger values are better.}
    \label{tab:comp_video_clip_davis}

    \scriptsize
    \setlength{\tabcolsep}{2pt}

    \resizebox{\columnwidth}{!}{%
    \begin{tabular}{l|cccc}
        \toprule
        Model
        & \textit{CLIP-S} $\uparrow$
        & \textit{CLIP-SIM} $\uparrow$
        & \textit{CLIP-F} $\uparrow$
        & \textit{CLIP-CONS} $\uparrow$ \\
        \midrule

        \multicolumn{5}{c}{\textit{Text-conditioned}} \\
        \midrule

        Rerender~\cite{yang2023rerender}
        & 19.40
        & 9.83
        & \tbest{98.23}
        & -0.03 \\

        Text2Video~\cite{khachatryan2023text2video}
        & \sbest{26.05}
        & \sbest{24.99}
        & 93.63
        & \tbest{0.03} \\

        CLIPGaussian~\cite{howil2026clipgaussian}
        & \best{\textbf{26.25}}
        & \tbest{24.53}
        & \sbest{99.00}
        & \best{\textbf{1.92}} \\

        \our{}
        & \tbest{25.83}
        & \best{\textbf{26.63}}
        & \best{\textbf{99.41}}
        & \sbest{0.66} \\

        \midrule
        \multicolumn{5}{c}{\textit{Image-conditioned}} \\
        \midrule

        ViSt3D~\cite{pande2023vist3d}
        & 55.92
        & 2.75
        & \sbest{99.18}
        & \sbest{3.28} \\

        AdaAttN~\cite{liu2021adaattn}
        & 57.71
        & -1.04
        & 97.56
        & 1.08 \\

        ReReVST~\cite{wang2020consistent}
        & 61.50
        & 2.12
        & \tbest{98.85}
        & \tbest{2.51} \\

        UniST~\cite{gu2023two}
        & 65.93
        & 3.85
        & \best{\textbf{99.36}}
        & \best{\textbf{5.16}} \\

        CCPL~\cite{wu2022ccpl}
        & \tbest{68.89}
        & \tbest{8.20}
        & 97.92
        & -0.02 \\

        CLIPGaussian~\cite{howil2026clipgaussian}
        & \sbest{74.31}
        & \sbest{17.60}
        & \sbest{99.18}
        & 1.27 \\

        \our{}
        & \best{\textbf{78.74}}
        & \best{\textbf{17.70}}
        & \sbest{99.18}
        & 0.72 \\

        \bottomrule
    \end{tabular}
    }
\end{table}

\section{Experiments}

Each of the four modalities has been tested on both a text prompt and a style image input. As the modalities are fundamentally different, the baselines are different for each one. We provide hyperparameter details and additional results alongside ablation studies in the appendix.

Video and 2D style transfer was trained and tested on a 11GB NVIDIA RTX2080Ti GPU, and for 3D and 4D scenes a 96GB NVIDIA GH200 GPU was used. The implementation is available on Github \footnote{\url{https://github.com/PiJayson/OmniStyle-INR}}

\subsection{Images}

For evaluating  \our{} on images, we used a subset of images from MS-COCO~\cite{lin2014microsoftcoco}, as it was used to evaluate style transfer in similar models ~~\cite{gatys2016image, Huang_2017_ICCV, howil2026clipgaussian}. Figure~\ref{fig:2D_comp_text_full} shows a comparison of text-based style transfer with CLIPstyler~\cite{kwon2022clipstyler}, FastCLIPstyler~\cite{Suresh_2024_WACV} and CLIPGaussian~\cite{howil2026clipgaussian}. Our results are more semantically faithful to the target prompt across all styles. Unlike CLIPGaussian~\cite{howil2026clipgaussian}, which tends to concentrate stylization near the image center, our method achieves a more spatially uniform and coherent style transfer. This discrepancy may be attributed to the padding strategy employed in CLIPGaussian, which could introduce a center bias during stylization. The same can be seen in the image-based style transfer comparison in Figure~\ref{fig:2D_comp_image_full}, where we compare our results with AdaIN~\cite{Huang_2017_ICCV}, StyTr$^2$~\cite{Deng_2022_CVPR}, and CLIPGaussian~\cite{howil2026clipgaussian}. Our visual results compare favorably against CLIPGaussian~\cite{howil2026clipgaussian}, suggesting that INR-based representations are worth exploring further for 2D style transfer. We have to acknowledge that our method cannot compete with the quality of massive diffusion models. However, we argue that this is a fair trade-off, as our outputs remain structurally faithful to the original content.

An ablation study with calculated metrics is provided in the appendix.

\begin{table}[t]
    \centering
    \caption{Quantitative comparison of video style transfer consistency
    on the DAVIS dataset~\cite{davis}, using RMSE and LPIPS.
    Smaller values are better.}
    \label{tab:comp_video_cons_davis}

    \scriptsize
    \setlength{\tabcolsep}{2pt}

    \resizebox{\columnwidth}{!}{%
    \begin{tabular}{l|cccc}
        \toprule
        Model
        & \multicolumn{2}{c}{Short-range consistency}
        & \multicolumn{2}{c}{Long-range consistency} \\
        \midrule
        & LPIPS $\downarrow$
        & RMSE $\downarrow$
        & LPIPS $\downarrow$
        & RMSE $\downarrow$ \\
        \midrule

        Original Videos
        & 0.042
        & 0.034
        & 0.070
        & 0.055 \\

        \midrule
        \multicolumn{5}{c}{\textit{Text-conditioned}} \\
        \midrule

        Rerender~\cite{yang2023rerender}
        & \best{\textbf{0.062}}
        & \sbest{0.040}
        & \sbest{0.132}
        & \sbest{0.077} \\

        Text2Video~\cite{khachatryan2023text2video}
        & 0.261
        & 0.183
        & 0.235
        & 0.166 \\

        CLIPGaussian~\cite{howil2026clipgaussian}
        & \tbest{0.084}
        & \tbest{0.057}
        & \tbest{0.152}
        & \tbest{0.095} \\

        \our{}
        & \sbest{0.080}
        & \best{\textbf{0.026}}
        & \best{\textbf{0.106}}
        & \best{\textbf{0.036}} \\

        \midrule
        \multicolumn{5}{c}{\textit{Image-conditioned}} \\
        \midrule

        ViSt3D~\cite{pande2023vist3d}
        & 0.081
        & \sbest{0.043}
        & 0.121
        & \sbest{0.063} \\

        AdaAttN~\cite{liu2021adaattn}
        & 0.087
        & 0.059
        & 0.116
        & 0.083 \\

        ReReVST~\cite{wang2020consistent}
        & \sbest{0.072}
        & \tbest{0.047}
        & \sbest{0.100}
        & 0.068 \\

        UniST~\cite{gu2023two}
        & \best{\textbf{0.062}}
        & \tbest{0.047}
        & \best{\textbf{0.088}}
        & \tbest{0.066} \\

        CCPL~\cite{wu2022ccpl}
        & 0.102
        & 0.065
        & 0.132
        & 0.093 \\

        CLIPGaussian~\cite{howil2026clipgaussian}
        & 0.086
        & 0.057
        & 0.157
        & 0.090 \\

        \our{}
        & \tbest{0.078}
        & \best{\textbf{0.028}}
        & \tbest{0.105}
        & \best{\textbf{0.040}} \\

        \bottomrule
    \end{tabular}%
    }
\end{table}

\subsection{Videos} \label{experiment-video}

\our{} for video style transfer was evaluated on the DAVIS dataset~\cite{davis}, which is a high-resolution video dataset usually used for video object segmentation tasks, but was also used for style transfer evaluation~\cite{howil2026clipgaussian}. As with image style transfer, we evaluate \our{} under both image- and text-conditioned settings. The image-conditioned comparison against CCPL~\cite{wu2022ccpl}, UniST~\cite{gu2023two}, and CLIPGaussian~\cite{howil2026clipgaussian} is shown in Figure~\ref{fig:comp_video_image}, while the text-conditioned comparison against Text2Video~\cite{khachatryan2023text2video}, Rerender~\cite{yang2023rerender}, and CLIPGaussian~\cite{howil2026clipgaussian} is presented in Figure~\ref{fig:comp_video_text}. Both figures display the first and last frames of each video. Our method consistently produces finer style details while remaining more faithful to the original video content.

Following the evaluation protocol adopted in CLIPGaussian~\cite{howil2026clipgaussian}, we evaluate both image- and text-based stylization using CLIP-based metrics. Specifically, we adopt CLIP Directional Similarity (CLIP-SIM)~\cite{gal2022stylegannada} and CLIP-S~\cite{hessel2021clipscore} to measure transfer quality, while CLIP Directional Consistency (CLIP-CONS)~\cite{haque2023instruct} and CLIP-F~\cite{wu2023tuneavideo} assess temporal coherence and content preservation, following the same protocol as CLIPGaussian~\cite{howil2026clipgaussian}. Full metric definitions and evaluation details are provided in the appendix. 

The quantitative results are reported in Table~\ref{tab:comp_video_clip_davis}. Both \our{} and CLIPGaussian~\cite{howil2026clipgaussian} achieve strong style transfer performance, while remaining competitive in terms of content retention.

For consistency evaluation, we adopt short-range and long-range metrics following StyleRF~\cite{liu2023stylerf}, computing masked RMSE and LPIPS after warping frames via optical flow. We additionally employ the Farneback optical flow-based metric from ViSt3D~\cite{pande2023vist3d}. Full implementation details and metric definitions are provided in the appendix. The quantitative comparison with other style transfer models is presented in Table~\ref{tab:comp_video_clip_davis}. While our architecture is largely similar to CLIPGaussian~\cite{howil2026clipgaussian}, the key difference lies in our explicit temporal loss, which significantly improves frame-to-frame consistency. Our method achieves the best RMSE scores among all baselines, including CLIPGaussian, while remaining competitive on LPIPS.

\begin{table}[t]
    \centering
    \caption{Quantitative comparison of 3D style transfer using
    CLIP-based metrics. Larger values are better.}
    \label{tab:3Dclip_metrics}

    \scriptsize
    \setlength{\tabcolsep}{2pt}

    \resizebox{\columnwidth}{!}{%
    \begin{tabular}{l|cccc}
        \toprule
        Model
        & \textit{CLIP-S} $\uparrow$
        & \textit{CLIP-SIM} $\uparrow$
        & \textit{CLIP-F} $\uparrow$
        & \textit{CLIP-CONS} $\uparrow$ \\
        \midrule

        \multicolumn{5}{c}{\textit{Text-conditioned}} \\
        \midrule

        I-GS2GS~\cite{igs2gs}
        & 16.80
        & 12.03
        & \sbest{99.19}
        & \best{\textbf{13.53}} \\

        DGE~\cite{chen2024dge}
        & \tbest{17.59}
        & \tbest{12.27}
        & \best{\textbf{99.31}}
        & \sbest{12.46} \\

        CLIPGaussian~\cite{howil2026clipgaussian}
        & \best{\textbf{26.86}}
        & \best{\textbf{26.31}}
        & \tbest{98.80}
        & \tbest{2.34} \\

        \our{}
        & \sbest{25.32}
        & \sbest{21.16}
        & 98.56
        & 2.08 \\

        \midrule
        \multicolumn{5}{c}{\textit{Image-conditioned}} \\
        \midrule

        StyleGaussian~\cite{liu2024stylegaussian}
        & 63.69
        & 13.07
        & 98.87
        & 1.36 \\

        SGSST~\cite{galerne2025sgsst}
        & 66.57
        & 16.24
        & 97.54
        & 0.91 \\

        ABC-GS~\cite{liu2025abc}
        & 68.68
        & 16.29
        & \sbest{99.10}
        & \best{\textbf{2.11}} \\

        G-Style~\cite{kovacs2024G}
        & \best{\textbf{76.94}}
        & \best{\textbf{24.94}}
        & \tbest{98.94}
        & 1.31 \\

        CLIPGaussian~\cite{howil2026clipgaussian}
        & \sbest{72.65}
        & \sbest{20.72}
        & 98.78
        & \sbest{1.77} \\

        \our{}
        & \tbest{69.71}
        & \tbest{19.04}
        & \best{\textbf{99.61}}
        & \tbest{1.65} \\

        \bottomrule
    \end{tabular}%
    }
    \vspace{-0.3cm}
\end{table}

\subsection{3D Scenes} \label{3d_scenes_exp}

Style transfer on 3D scenes using \our{} was evaluated on two datasets: NeRF-Synthetic dataset~\cite{mildenhall2020nerf} consisting of blender objects with no background, and Mip-NeRF~\cite{barron2022mipnerf360}, which provides photorealistic 360-degree scenes. TensoRF \cite{chen2022tensorf}, which serves as the base 3D style transfer model for \our{}, does not inherently support 360-degree scenes. To address this, we adapt it using space contraction. Further details are provided in the appendix.

We evaluated our model using the CLIP-based metrics mentioned in section \ref{experiment-video} on the \textit{lego} and \textit{hotdog} objects and two scenes \textit{garden}, \textit{bonsai}. The average values are presented in Table \ref{tab:3Dclip_metrics}, where we compare with other 
style transfer methods for 3D scenes, like: Instruct-GS2GS \cite{haque2023instruct}, DGE \cite{chen2024dge}, and CLIPGaussian \cite{howil2026clipgaussian} for text based style transfer. \our{} produces similar values as CLIPGaussian \cite{howil2026clipgaussian}, outperforming the other two methods in CLIP-S and CLIP-SIM metrics. 

We report the same metrics on the same 3D scenes for image-based style transfer, comparing against StyleGaussian \cite{liu2024stylegaussian}, SGSST \cite{galerne2025sgsst}, ABC-GS \cite{liu2025abc}, and G-Style \cite{kovacs2024G}.
 Our method achieves the highest CLIP-F score among all image-conditioned methods, indicating superior content fidelity, while remaining competitive on the other metrics.

A visual comparison of 3D style transfer methods is available in Figure~\ref{fig:Image2Style} for image-conditioned stylization, and in Figure~\ref{fig:Text2Style} for text-conditioned stylization.
 
An ablation study with metrics and renders is provided in the appendix.

\begin{figure}[t]
\centering
    \includegraphics[width=\linewidth]{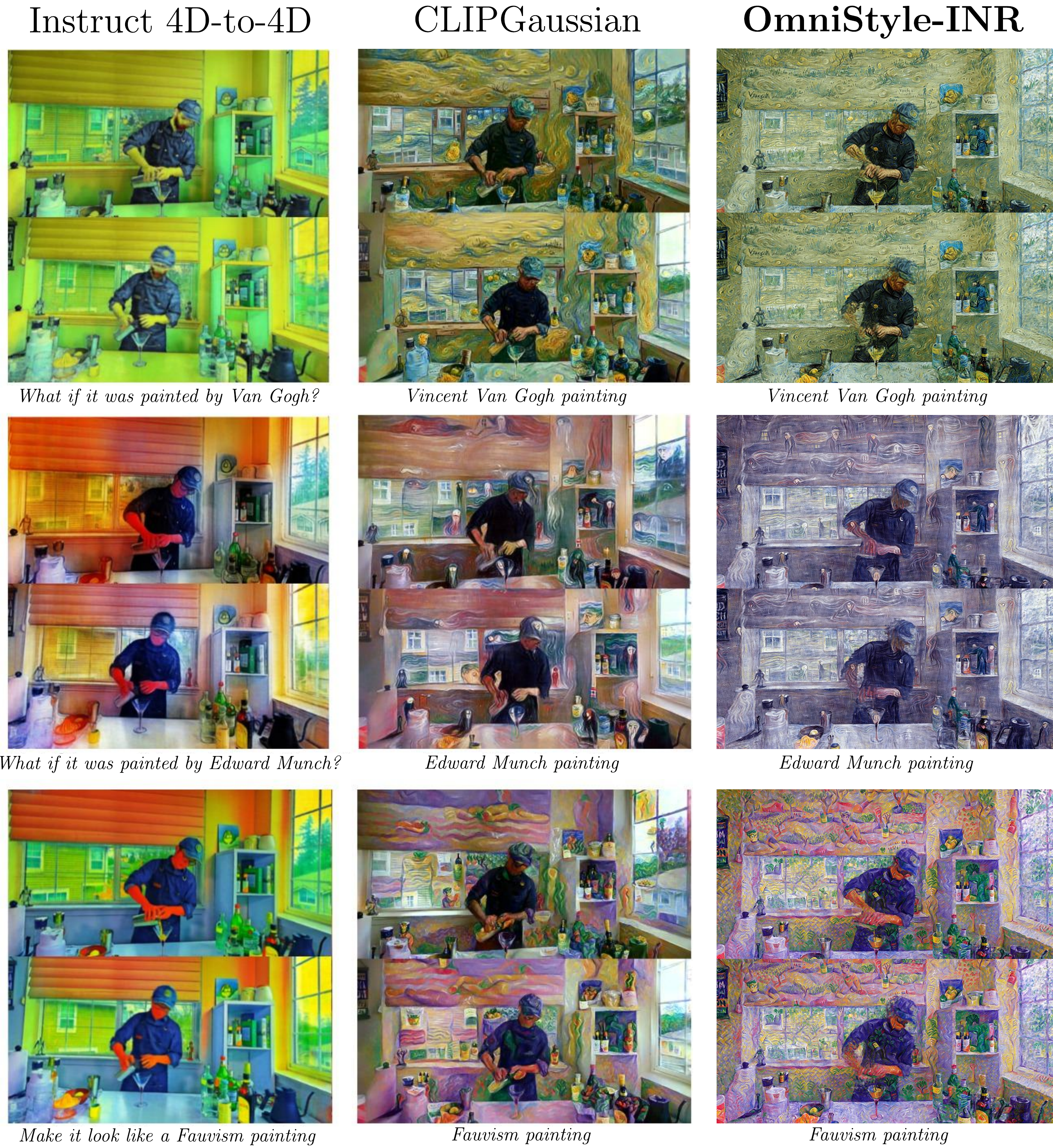}
    \caption{
    Text-conditioned 4D style transfer comparison on the \texttt{coffee\_martini}
sequence from the N3DV dataset~\cite{Li_2022_CVPR}.
    }
    \label{fig:4d-text-st}
    \vspace{-0.2cm}
\end{figure}

\begin{figure*}[t]
\centering
    \includegraphics[width=\linewidth]{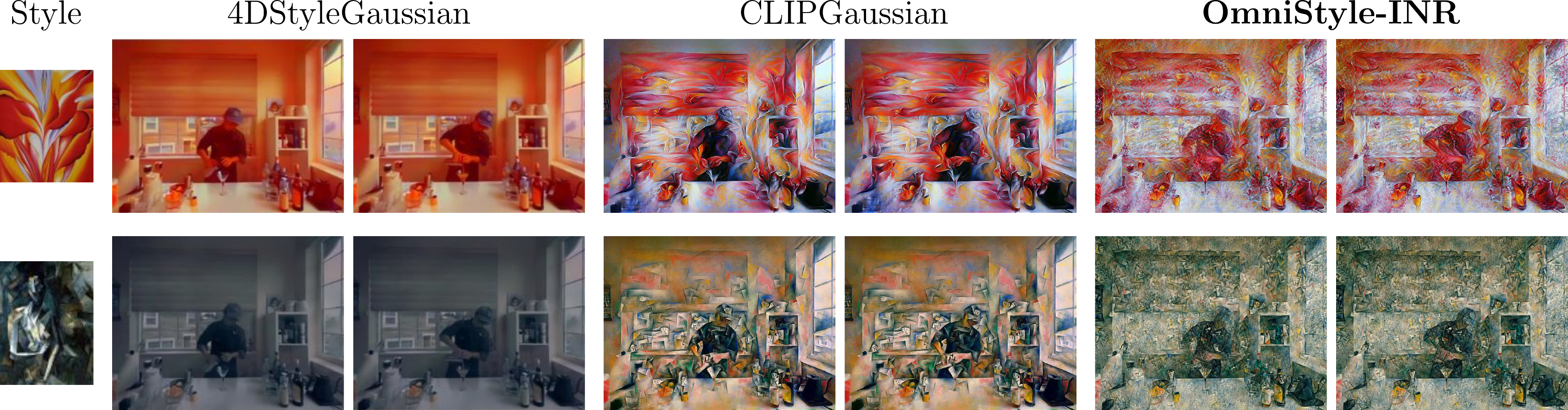}
    \caption{
    Image-conditioned 4D style transfer comparison on the \texttt{coffee\_martini}
sequence from the Neural 3D Video dataset (DyNeRF)~\cite{Li_2022_CVPR}.
    }
    \label{fig:4d-image-st}
    \vspace{-0.3cm}
\end{figure*}

\subsection{4D Dynamic Scenes}

We evaluate \our{} for dynamic 4D content on the \texttt{coffee\_martini}
sequence from the Neural 3D Video (DyNeRF) dataset~\cite{Li_2022_CVPR}, which
provides time-synchronized, calibrated multi-view recordings of a complex
dynamic scene. Starting from the HexPlane base reconstruction, we apply our
style objective to obtain a stylized dynamic scene that remains consistent
across frames. As in the other modalities, both text-prompt and style-image
conditioning are supported, here additionally driven by the temporal term
$\mathcal{L}_t$ that aligns consecutive frames.

For image-conditioned stylization we compare against
4DStyleGaussian~\cite{liang20254dstylegaussian} and
CLIPGaussian~\cite{howil2026clipgaussian} in Figure~\ref{fig:4d-image-st}. For
text-conditioned stylization, Figure~\ref{fig:4d-text-st} compares \our{}
against the instruction-guided, diffusion-based Instruct
4D-to-4D~\cite{Mou_2024_CVPR} and CLIPGaussian. Across both settings \our{}
transfers fine-grained style (detail patterns, texture, and color) while
preserving the motion and structure of the original scene. As shown in these
comparisons, \our{} achieves stylization quality on par with that of
CLIPGaussian.

Quantitatively, we report the four CLIP-based metrics on the rendered dynamic
sequences and study the sensitivity to the directional and patch weights
$\lambda_d$ and $\lambda_p$ finding the
patch loss to be the primary driver of stylization while a small content weight
is retained to stabilize temporal consistency. A full loss-component ablation,
which isolates the contribution of each term of our objective, is provided in the appendix.

\begin{figure}[t]
\centering
\small
\setlength{\tabcolsep}{2pt}

\resizebox{\columnwidth}{!}{%
    \begin{tabular}{
        c
        @{\hspace{3mm}}
        c
        @{\hspace{8mm}}
        c
        @{\hspace{4mm}}
        c
        @{\hspace{1mm}}
        c
    }
        Style \& GT
        & CCPL~\cite{wu2022ccpl}
        & UniST~\cite{gu2023two}
        & CLIPGaussian~\cite{howil2026clipgaussian}
        & \textbf{\our{}} \\
    \end{tabular}

}
\includegraphics[
    width=\columnwidth
]{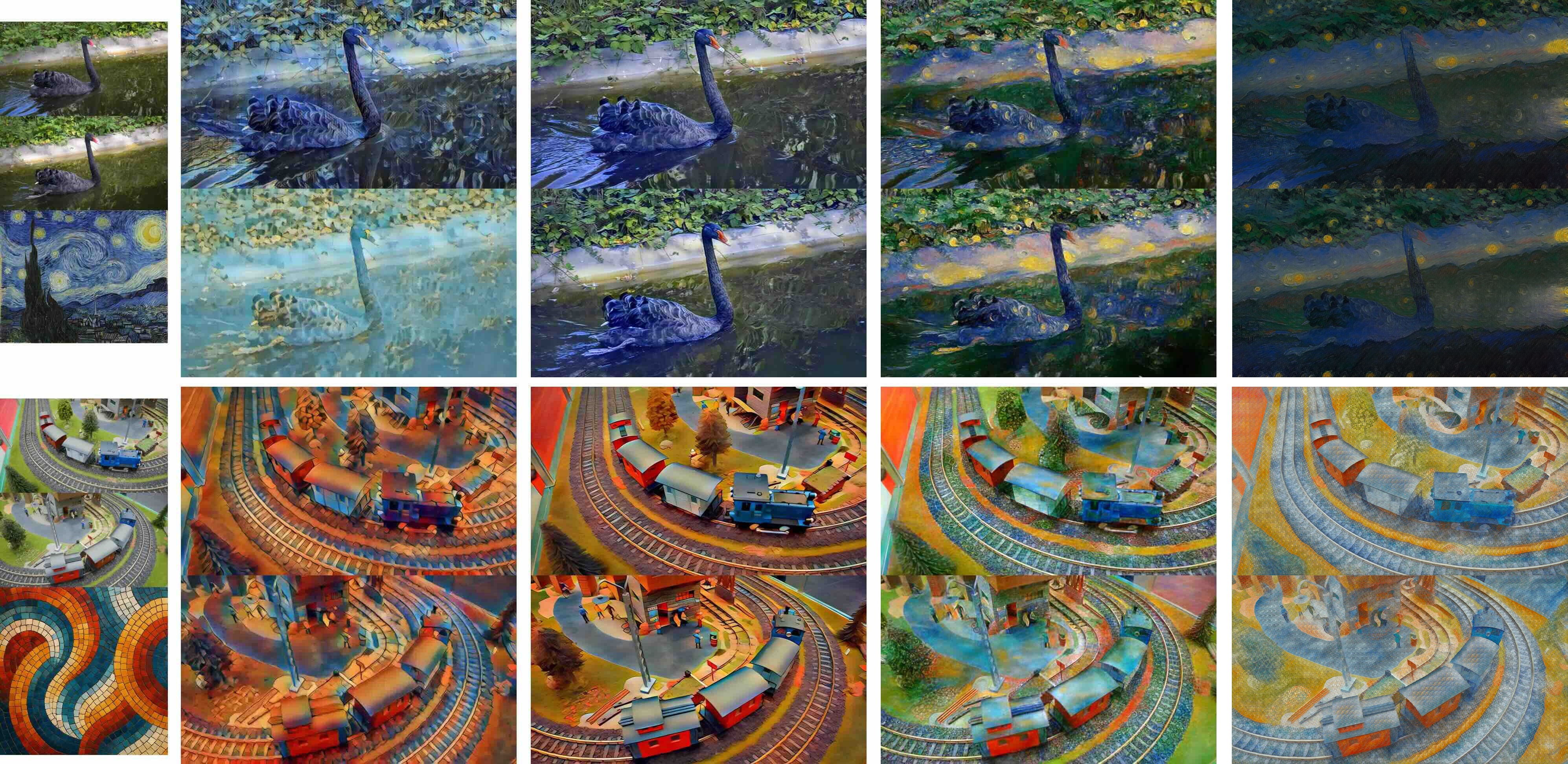}
\caption{Comparison of video style transfer using image conditioning
on the DAVIS dataset~\cite{davis}.}
\label{fig:comp_video_image}
\vspace{-0.3cm}
\end{figure}

\begin{figure}[t]
    \centering
    \scriptsize
    \setlength{\tabcolsep}{2pt}

    \resizebox{\columnwidth}{!}{%
    \begin{tabular}{ccccc}
        Style \& GT
        & Text2Video~\cite{khachatryan2023text2video}
        & Rerender~\cite{yang2023rerender}
        & CLIPGaussian~\cite{howil2026clipgaussian}
        & \textbf{\our{}} \\
    \end{tabular}%
    }

    \includegraphics[
        width=\columnwidth
    ]{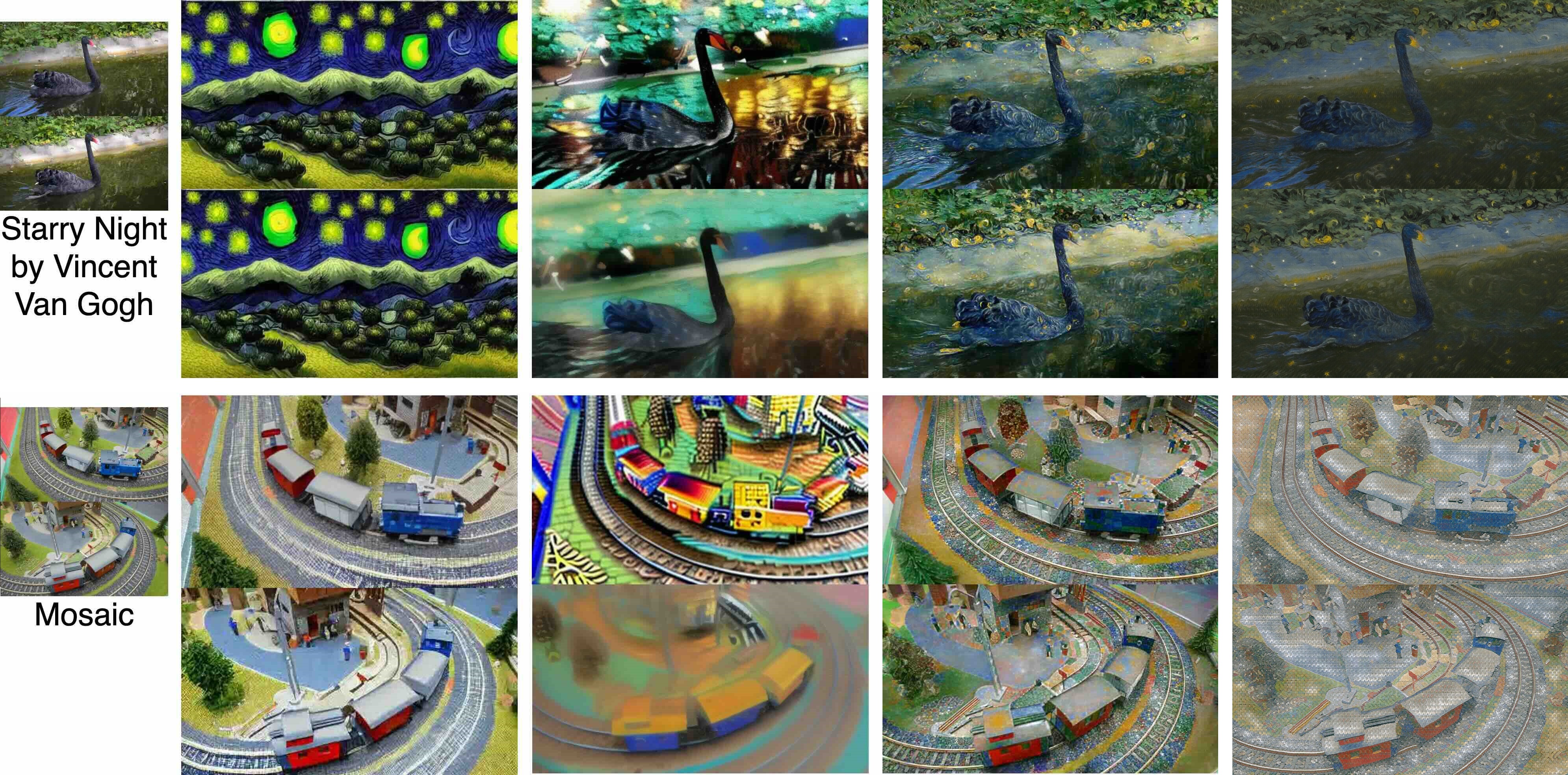}

    \caption{Comparison of video style transfer using text conditioning
    on the DAVIS dataset~\cite{davis}.}
    \label{fig:comp_video_text}
    \vspace{-0.3cm}
\end{figure}

\section{Conclusion}
\our{} is designed as a plug-in style transfer module for implicit neural representations, distinguishing itself from CLIPGaussian \cite{howil2026clipgaussian} through the incorporation of a temporal loss for dynamic scenes. Our framework supports both image- and text-conditioned stylization across multiple modalities, including images, videos, 3D scenes, and 4D dynamic scenes.

In our experiments, we observe that \our{} achieves comparable performance to CLIPGaussian \cite{howil2026clipgaussian} in most settings, while surpassing it on key metrics—particularly in terms of temporal consistency—due to our explicit temporal regularization.

\paragraph{Limitations}
While unconstrained generation using massive diffusion models may produce higher-quality 2D outputs, our INR-based approach provides a crucial trade-off by strictly preserving structural fidelity without introducing hallucinations. Nevertheless, we argue that our method offers a compelling alternative, particularly for 3D and 4D scenes, where INR-based representations excel at capturing continuous geometry and temporal dynamics. Our framework preserves the exact structure while still delivering visually pleasing style transfer.

\paragraph{Future work}
As a plug-in approach, \our{} can be seamlessly applied to other INR models.The core ideas from our method could generalize to other 3D representations, such as meshes or point clouds. Future work includes adapting our framework for interactive editing applications, exploring user-guided stylization with fine-grained control, and developing lightweight edge-deployable models.

\section*{Acknowledgments}
The work of P. Spurek, was supported by the National Centre of Science (Poland) Grant No. 2023/50/E/ST6/00068.
We gratefully acknowledge Polish high-performance computing infrastructure PLGrid (HPC Centers: ACK Cyfronet AGH, CI TASK, WCSS) for providing computer facilities and support within computational grant no. PLG/2026/019272.

\FloatBarrier

{
    \small
    \bibliographystyle{ieeenat_fullname}
    \bibliography{main}
}

\clearpage
\FloatBarrier

\appendix

\section{Implementation Details and Evaluation Protocol}

\subsection{Implementation Details}

In this section, we elaborate on several implementation details that were only briefly mentioned in the main sections, including loss calculation, optical flow estimation, backward warping, occlusion handling, full scene rendering and the Mip-NeRF 360 \cite{barron2022mipnerf360} dataset.

\subsubsection{Full scene rendering}

In standard INR-based models—such as NeRF~\cite{mildenhall2020nerf}, Mip-NeRF~\cite{barron2022mipnerf360}, or the baseline TensoRF~\cite{chen2022tensorf} training is entirely ray-based. Instead of rendering a complete image, the model randomly samples a batch of independent rays (typically $4096$ or $8192$) scattered across different views and coordinates. Because the standard reconstruction loss (MSE) is pixel-wise, it requires no spatial context and can be computed on these disconnected points.

For style transfer, however, this ray-based approach is not feasible. Both the VGG content loss and the CLIP-based style losses (directional and patch-based) require full, coherent 2D images. VGG relies on convolutional layers to extract spatial hierarchies and textures, while CLIP processes images as globally correlated sequences of patches. The patch loss additionally requires cropping contiguous regions and applying spatial augmentations such as perspective warps. None of these operations can be computed on a set of isolated, non-contiguous pixels. To calculate these losses, the model must render the entire full-resolution image at a given camera pose.

At base model training iterations, the model renders only a small batch of rays, as is standard in INR models. At style-loss iterations, however, the model must render the full camera frame (e.g., $800 \times 800 = 640{,}000$ rays), which is significantly more computationally expensive. To fit within GPU memory limits, the full frame is rendered in sequential chunks (e.g., $102{,}400$ rays per chunk), which are then stitched back together into a 2D tensor before being passed to VGG and CLIP.

Because of this, we opted for more memory-efficient architectures than NeRF~\cite{mildenhall2020nerf}. Specifically, for memory-intensive modalities, 3D and 4D, we used grid-based representations: TensoRF~\cite{chen2022tensorf} for 3D and HexPlane~\cite{cao2023hexplane} for 4D. Both models employ tensor decomposition to dramatically reduce memory usage compared to pure MLP-based NeRFs. The freed GPU memory then allows us to render full-resolution frames in manageable chunks, making style transfer losses—which require full 2D spatial context—practical within our computational budget.

\subsubsection{Loss calculation}

The CLIP model for loss calculation that we use is the ViT-B/32 architecture~\cite{radford2021learning}, and VGG-19 with torchvision's default ImageNet weights~\cite{simonyan2015very}.

\subsubsection{Image Normalization}
Prior to feature extraction, all rendered images (with pixel values in $[0, 1]$) are normalized using the standard channel-wise statistics for each network:

\begin{itemize}
    \item \textbf{VGG-19:}  \\
    $\mu = [0.485, 0.456, 0.406]$ \\ 
    $\sigma = [0.229, 0.224, 0.225]$
    \item \textbf{CLIP (ViT-B/32):} \\
    $\mu = [0.48145466, 0.4578275, 0.40821073]$ \\
    $\sigma = [0.26862954, 0.26130258, 0.27577711]$
\end{itemize}

\subsubsection{Patch Extraction for Local Style Loss}
To capture local multi-scale texture details, we apply a patching mechanism to the rendered images. We randomly extract $\texttt{patch\_num}$ crops of size $\texttt{patch\_size} \times \texttt{patch\_size}$ pixels using reflection padding. Each crop is then augmented with a random perspective distortion (distortion scale $0.5$, probability $p = 1.0$), and bilinearly resized to $224 \times 224$ pixels to match CLIP's input resolution.

\subsubsection{Style Direction Template Embedding}
For text-conditioned stylization, we compute a style direction vector using prompt engineering. We average text embeddings over the standard 80 ImageNet templates~\cite{radford2021learning} (e.g., \textit{"a photo of the [style\_prompt]"}, \textit{"a rendering of the [style\_prompt]"}) to avoid bias from single-sentence embeddings.

\subsubsection{Optical Flow Estimation}

We estimate optical flow between consecutive frames to guide our temporal consistency loss. For dynamic 4D scenes, we precompute forward flow and occlusion masks offline using RAFT \cite{teed2020raft}. For video data, we compute flow dynamically during training, supporting both RAFT and the Farneback algorithm \cite{farneback2003two}, with flow estimated in both forward and backward directions to facilitate occlusion reasoning.

If the precomputed flow resolution differs from the rendering resolution, we bilinearly interpolate and scale the flow accordingly.

\subsubsection{Occlusion Masks}

Temporal losses should only be applied to regions that remain visible across frames. For dynamic 4D scenes, we directly load precomputed binary occlusion masks. For video data, we compute a forward-backward consistency mask at runtime. Specifically, for a forward flow $\mathbf{f}$ and backward flow $\mathbf{b}$, a pixel at position $\mathbf{x}$ is considered non-occluded if:
$$
\|\mathbf{f}(\mathbf{x}) + \mathbf{b}(\mathbf{x} + \mathbf{f}(\mathbf{x}))\|^2 \leq \alpha \cdot \left( \|\mathbf{f}(\mathbf{x})\|^2 + \|\mathbf{b}(\mathbf{x} + \mathbf{f}(\mathbf{x}))\|^2 \right) + \beta
$$

\subsubsection{Warping Procedure}

Both Video (PNeRV~\cite{zhao2024pnerv} / DNeRV~\cite{zhao2023dnerv}) and 4D (HexPlane~\cite{cao2023hexplane}) backbones use backward warping with bilinear sampling to align a frame at timestep $t+1$ to timestep $t$. We generate a normalized coordinate grid, add the optical flow offsets to each pixel location, and sample the next frame at the warped coordinates. Out-of-bound coordinates are handled by assigning a weight of zero.

Pixels that fail this consistency check, or that warp outside the image boundaries, are masked out and excluded from the temporal loss. 

The temporal consistency loss is then computed as the masked Mean Squared Error between the rendered frame $I_t$ and the warped frame $W(I_{t+1})$:

\[
\begin{aligned}
\mathcal{L}_{\text{temp}} 
&= \frac{1}{\sum M} \sum_{x,y} M(x,y) \cdot \left\| I_t(x,y) - W\left(I_{t+1}\right)(x,y) \right\|^2
\end{aligned}
\]

where $M$ is the combined occlusion and boundary validity mask.
\subsubsection{Space contraction for Mip-NeRF 360 scenes}

The original TensoRF \cite{chen2022tensorf} is designed for bounded scenes, placing a finite 3D grid around the scene and discarding anything outside this region. While effective for object-centric or indoor captures, this formulation breaks for 360° outdoor scenes, where the background extends to infinity and cannot be trivially bounded.

To address this limitation, we integrate the scene contraction proposed in Mip-NeRF 360 \cite{barron2022mipnerf360}. The contraction function maps points from unbounded space into a finite volume, allowing the tensor grid of TensoRF to represent both foreground and background structures. Formally, the contraction is defined as follows:

$$
f(\mathbf{x}) = 
\begin{cases}
\mathbf{x} & \text{if } \|\mathbf{x}\| \leq 1 \\[6pt]
\left(2 - \frac{1}{\|\mathbf{x}\|}\right) \frac{\mathbf{x}}{\|\mathbf{x}\|} & \text{if } \|\mathbf{x}\| > 1
\end{cases}
\label{eq:contraction}
$$

Intuitively, points inside the unit sphere ($\|\mathbf{x}\| \leq 1$) remain unchanged, preserving the local geometry near the camera. Points outside the unit sphere ($\|\mathbf{x}\| > 1$) are compressed along their original viewing direction, with their distance from the origin remapped by $(2 - 1/\|\mathbf{x}\|)$. This non-linear mapping squashes the entire infinite exterior into the finite spherical shell between radius $1$ and $2$, ensuring that all scene content fits within TensoRF's bounded representation.

In addition to the contraction, we replace the standard linear ray sampling with inverse-depth (disparity) sampling, which naturally allocates more samples near the camera and fewer in the far field. This sampling strategy aligns well with the contraction's redistribution of space, improving rendering quality in both near and distant regions.

On the data side, we implement a custom dataset loader that handles 360° COLMAP captures, normalizes camera poses so that cameras sit at approximately unit distance from the scene center, and sets appropriate near and far bounds for each scene.

\subsection{Evaluation Metrics}

\subsubsection{CLIP-based Metrics}

We evaluate stylization quality, content preservation, and temporal consistency using CLIP-based and optical-flow-based metrics. We denote the source image as $I$, rendered frame by $R$, the CLIP image encoder by $\Phi_{\mathrm{CLIP}}^{I}$, and the CLIP embedding of a text or image style condition $\mathcal{S}$ by $\Phi_{\mathrm{CLIP}}(\mathcal{S})$. Unless stated otherwise, all CLIP features are computed with frozen encoders and normalized before cosine similarity is measured. Larger values are better for all CLIP-based metrics, while lower values are better for RMSE and LPIPS.

\paragraph{CLIP-S.}
CLIP-S measures the semantic agreement between the stylized output and the target style condition:
\[
\mathrm{CLIP\text{-}S}(R,\mathcal{S})
=
100 \cdot
\cos\left(
\Phi_{\mathrm{CLIP}}^{I}(R),
\Phi_{\mathrm{CLIP}}(\mathcal{S})
\right).
\]
The score is averaged over all evaluated images, frames, or rendered views.

\paragraph{CLIP-SIM.}
CLIP-SIM measures whether the CLIP-space change from the original content to the stylized output follows the direction implied by the target style:
\[
\begin{aligned}
\mathrm{CLIP\text{-}SIM}(R,I,\mathcal{S})
&=
100 \cdot
\cos\Bigl(
\Phi_{\mathrm{CLIP}}^{I}(R) - \Phi_{\mathrm{CLIP}}^{I}(I), \\
&\qquad
\Phi_{\mathrm{CLIP}}(\mathcal{S}) - \Phi_{\mathrm{CLIP}}(\mathcal{S}_{0})
\Bigr),
\end{aligned}
\]
where $\mathcal{S}_{0}$ is a neutral source-domain condition, such as \textit{``Photo''}.

\paragraph{CLIP-F.}
CLIP-F measures content fidelity between the original input and the stylized output in CLIP image-feature space:
\[
\mathrm{CLIP\text{-}F}(R,I)
=
100 \cdot
\cos\left(
\Phi_{\mathrm{CLIP}}^{I}(R),
\Phi_{\mathrm{CLIP}}^{I}(I)
\right).
\]

\paragraph{CLIP-CONS.}
CLIP-CONS measures temporal or multi-view consistency by comparing the CLIP-space changes between neighboring source renders and neighboring stylized renders:
\[
\begin{aligned}
\mathrm{CLIP\text{-}CONS}
&=
100 \cdot
\cos\Bigl(
\Phi_{\mathrm{CLIP}}^{I}(R_{t+1}) - \Phi_{\mathrm{CLIP}}^{I}(R_t), \\
&\qquad
\Phi_{\mathrm{CLIP}}^{I}(I_{t+1}) - \Phi_{\mathrm{CLIP}}^{I}(I_t)
\Bigr).
\end{aligned}
\]

\subsubsection{Additional Consistency Metrics}

\paragraph{Optical-flow-based consistency.}
For video and 4D dynamic scenes, we additionally evaluate temporal consistency with optical-flow-based RMSE and LPIPS. Given two stylized frames $R_t$ and $R_{t+\Delta}$, we warp $R_{t+\Delta}$ into the coordinate system of $R_t$ using optical flow estimated on the original sequence and remove unreliable pixels with an occlusion mask $M_t$. The masked RMSE is computed as
\[
\mathrm{RMSE}
=
\sqrt{
\frac{
\sum_{\mathbf{x}} M_t(\mathbf{x})
\left\|
R_t(\mathbf{x}) -
W(R_{t+\Delta})(\mathbf{x})
\right\|_2^2
}{
\sum_{\mathbf{x}} M_t(\mathbf{x})
}
},
\]
where $W(\cdot)$ denotes flow-based warping. LPIPS is computed in the same aligned setting. Short-range consistency uses nearby frames, while long-range consistency uses a larger temporal offset.

\paragraph{Farneback consistency.}
We additionally measure motion consistency using Farneback optical flow \cite{farneback2003two}. Let
$F_t^{I}$ be the flow estimated between original frames $(I_t, I_{t+1})$ and
$F_t^{R}$ the flow estimated between stylized frames $(R_t, R_{t+1})$. We compute
the average endpoint error between the two flow fields:
\[
\mathrm{EPE}_{\mathrm{FB}}
=
\frac{1}{|\Omega|}
\sum_{\mathbf{x}\in\Omega}
\left\|
F_t^{R}(\mathbf{x}) - F_t^{I}(\mathbf{x})
\right\|_2 .
\]
Lower values indicate that the stylized sequence preserves the apparent motion
of the original video more faithfully.

\section{Additional Results and Ablation Studies}

This section presents additional experimental results and analyses, including extended results across diverse data modalities images, videos, 3D objects, and 4D dynamic scenes. Further .mp4 video files are provided in the supplementary materials.

\subsection{2D Ablation Studies}

We conduct ablation studies to analyze the impact of key hyperparameters in our 2D style transfer framework. These studies are performed on the same prompts and images used in Figures~\ref{fig:2D_comp_text_full} and~\ref{fig:2D_comp_image_full}, and the reported values are averages across all experiments for each respective configuration.

Table~\ref{tab:2d_ablation_lambda} evaluates the effect of the directional loss weight $\lambda_d$ and the patch loss weight $\lambda_p$ under both text and image conditioned settings. For this ablation, $\texttt{patch\_size}$ was set to $125$ and $\texttt{patch\_num}$ to $64$. For text-conditioned stylization, the best performance is achieved at $\lambda_d = 20$ and $\lambda_p = 90$, while image-conditioned stylization performs best at $\lambda_d = 10$ and $\lambda_p = 90$. Qualitative results are shown in Tables~\ref{tab:2d_ablation_text_lambda} and~\ref{tab:2d_ablation_photo_lambda}.

Table~\ref{tab:2d_ablation_patch} ablates the patch size $\texttt{patch\_size}$ and the number of patches $\texttt{patch\_num}$. $\lambda_d$ was set to $10$, and $\lambda_p$ to $90$. For both text and image conditioned style transfer, $\texttt{patch\_size} = 200$ and $\texttt{patch\_num} = 32$ deliver the best overall performance. That is why we selected these values of $\texttt{patch\_size}$ and $\texttt{patch\_num}$ as our default configuration. Visual comparisons are provided in Tables~\ref{tab:2d_ablation_text_patch} and~\ref{tab:2d_ablation_photo_patch}.

\begin{table}[t]
\centering
\caption{Ablation study of $\lambda_d$ and $\lambda_p$ parameters under text-conditioned and image-conditioned settings for 2D style transfer. Larger values are better.}
\label{tab:ablation_2d_lambda_conditioned}
\setlength{\tabcolsep}{2pt}
\begin{tabular}{cc|cc}
    \toprule
        $\lambda_d$ & $\lambda_p$ & \textit{CLIP-S} $\uparrow$ & \textit{CLIP-CONS} $\uparrow$ \\
        \midrule
        \multicolumn{4}{c}{\textit{Text-conditioned}} \\
        \midrule
        & 30  & 26.72 & 21.98 \\
        5  & 90  & 28.47 & 22.19 \\
        & 180 & 28.76 & 22.12 \\
        \midrule
        & 30  & 26.15 & 21.21 \\
        10 & 90  & 29.25 & \tbest{23.43} \\
        & 180 & \tbest{29.62} & \best{\textbf{24.01}} \\
        \midrule
        & 30  & 27.40 & 21.00 \\
        20 & 90  & \best{\textbf{29.79}} & \sbest{23.74} \\
        & 180 & \sbest{29.67} & 22.23 \\
        \midrule
        \multicolumn{4}{c}{\textit{Image-conditioned}} \\
        \midrule
        & 30  & 66.42 & 3.74 \\
        5  & 90  & 64.47 & 2.24 \\
        & 180 & 66.28 & 3.67 \\
        \midrule
        & 30  & 65.71 & 2.82 \\
        10 & 90  & \best{\textbf{68.85}} & \best{\textbf{5.87}} \\
        & 180 & 66.56 & 3.58 \\
        \midrule
        & 30  & \sbest{67.98} & \sbest{5.07} \\
        20 & 90  & 66.66 & 4.03 \\
        & 180 & \tbest{67.14} & \tbest{4.30} \\
        \bottomrule
\end{tabular}
\label{tab:2d_ablation_lambda}
\end{table}

\begin{table}[t]
\centering
\small
\caption{Ablation study of  $\texttt{patch\_size}$ ($p_{size}$) and  $\texttt{patch\_num}$ ($p_{num}$) under text-conditioned and image-conditioned settings for 2D style transfer. Larger values are better.}
\label{tab:ablation_patch_conditioned}
\setlength{\tabcolsep}{2pt}
    \begin{tabular}{cc|cc}
        \toprule
            $p_{size}$ & $p_{num}$ & \textit{CLIP-S} $\uparrow$ & \textit{CLIP-CONS} $\uparrow$ \\
            \midrule
            \multicolumn{4}{c}{\textit{Text-conditioned}} \\
            \midrule
            & 32  & 22.49 & 14.93 \\
            64  & 64  & 23.57 & 16.87 \\
            & 128 & 23.18 & 15.75 \\
            \midrule
            & 32  & 28.86 & 22.40 \\
            125 & 64  & \tbest{28.92} & \tbest{22.93} \\
            & 128 & 28.89 & 22.90 \\
            \midrule
            & 32  & \best{\textbf{29.26}} & \best{\textbf{25.30}} \\
            200 & 64  & 27.58 & 22.45 \\
            & 128 & \sbest{28.93} & \sbest{24.92} \\
            \midrule
            \multicolumn{4}{c}{\textit{Image-conditioned}} \\
            \midrule
            & 32  & 62.78 & -0.86 \\
            64  & 64  & 61.74 & -1.74 \\
            & 128 & 62.41 & -1.34 \\
            \midrule
            & 32  & 67.26 & 4.46 \\
            125 & 64  & 66.89 & 4.24 \\
            & 128 & 66.39 & 3.66 \\
            \midrule
            & 32  & \best{\textbf{71.37}} & \sbest{8.29} \\
            200 & 64  & \sbest{71.28} & \best{\textbf{8.32}} \\
            & 128 & \tbest{68.31} & \tbest{5.48} \\
            \bottomrule
    \end{tabular}
\label{tab:2d_ablation_patch}
\end{table}

\newenvironment{gridoneimages}[2]{%
  \begin{table}[t]
  \centering
  \caption{Ablation study over $\lambda_{\text{dir}}$ and $\lambda_{\text{path}}$.\label{#2}}
  \setlength{\tabcolsep}{2pt}
  \begin{tabular}{c c c c}
  \toprule
  $\lambda_{\text{dir}}$ / $\lambda_{\text{path}}$ & 30 & 90 & 180 \\
  \midrule
}{
  \bottomrule
  \end{tabular}
  \end{table}
}

\newcommand{\gridonerow}[3]{%
  #3 &
  \includegraphics[width=0.25\linewidth]{#2/grid1_dir#1.0_patch30.0.png} &
  \includegraphics[width=0.25\linewidth]{#2/grid1_dir#1.0_patch90.0.png} &
  \includegraphics[width=0.25\linewidth]{#2/grid1_dir#1.0_patch180.0.png} \\
}

\newenvironment{gridtwoimages}[2]{%
  \begin{table}[t]
  \centering
  \caption{Ablation study over patch size (here denoted as $p_{size}$ and patch number (denoted as $p_{num}$.\label{#2}}
  \setlength{\tabcolsep}{2pt}
  \begin{tabular}{c c c c}
  \toprule
  $p_{size}$ / $p_{num}$ & 32 & 64 & 128 \\
  \midrule
}{
  \bottomrule
  \end{tabular}
  \end{table}
}

\newcommand{\gridtworow}[3]{%
  #3 &
  \includegraphics[width=0.25\linewidth]{#2/grid2_sz#1_num32.png} &
  \includegraphics[width=0.25\linewidth]{#2/grid2_sz#1_num64.png} &
  \includegraphics[width=0.25\linewidth]{#2/grid2_sz#1_num128.png} \\
}

\subsection{3D Experiments}

\begin{figure*}[p]
\centering
\setlength{\tabcolsep}{5pt}
\begin{tabular}{ccccc}
Style &  I-GS2GS \cite{igs2gs} &  DGE \cite{chen2024dge}  &  CLIPGaussian \cite{howil2026clipgaussian} &  \our{} \\
  \fbox{
  \begin{minipage}[c][0.1\textwidth][c]{0.14\textwidth}
    \centering
     Fire
  \end{minipage}
}
  &
\includegraphics[trim={50 130 50 100},clip, width=0.18\textwidth,valign=c]{imgs/3D/comp_text/igs2gs/lego/068_lego_Fire.jpg}    & 
\includegraphics[trim={50 130 50 100},clip, width=0.18\textwidth,valign=c]{imgs/3D/comp_text/dge/lego/068_lego_Fire.jpg}   & 
\includegraphics[trim={50 130 50 100},clip, width=0.18\textwidth,valign=c]{imgs/3D/comp_text/gaussian/lego/068_lego_Fire.jpg}  &
\includegraphics[trim={50 130 50 100},clip, width=0.18\textwidth,valign=c]{imgs/3D/comp_text/ours/lego/068_fire.jpg}  \\
\fbox{
  \begin{minipage}[c][0.1\textwidth][c]{0.14\textwidth}
    \centering
      Starry Night by Vincent van Gogh
  \end{minipage}
}&  
 \includegraphics[trim={50 130 50 100},clip, width=0.18\textwidth,valign=c]{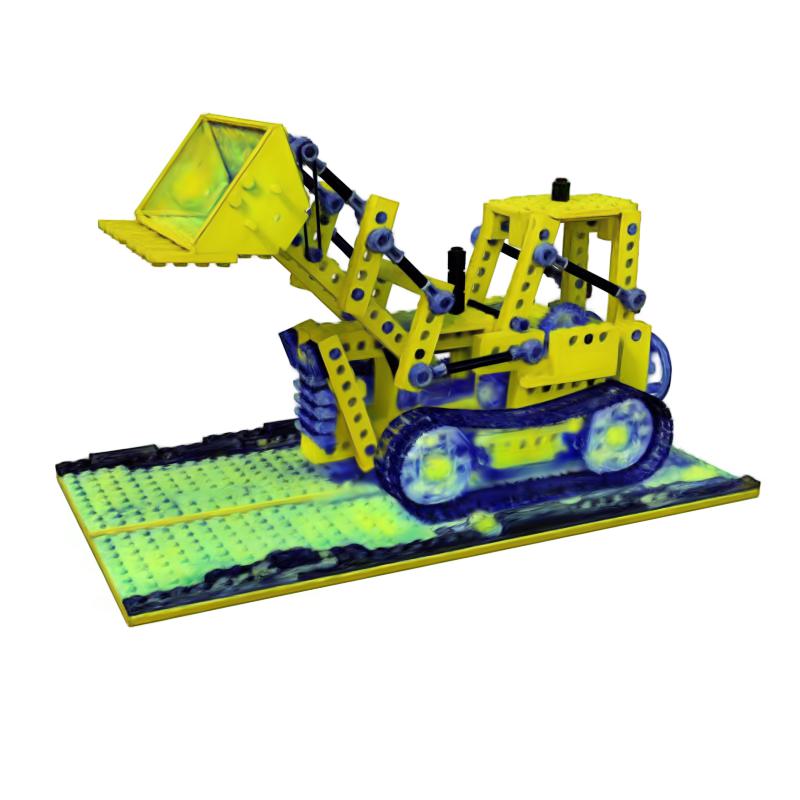}     &  
 \includegraphics[trim={50 130 50 100},clip, width=0.18\textwidth,valign=c]{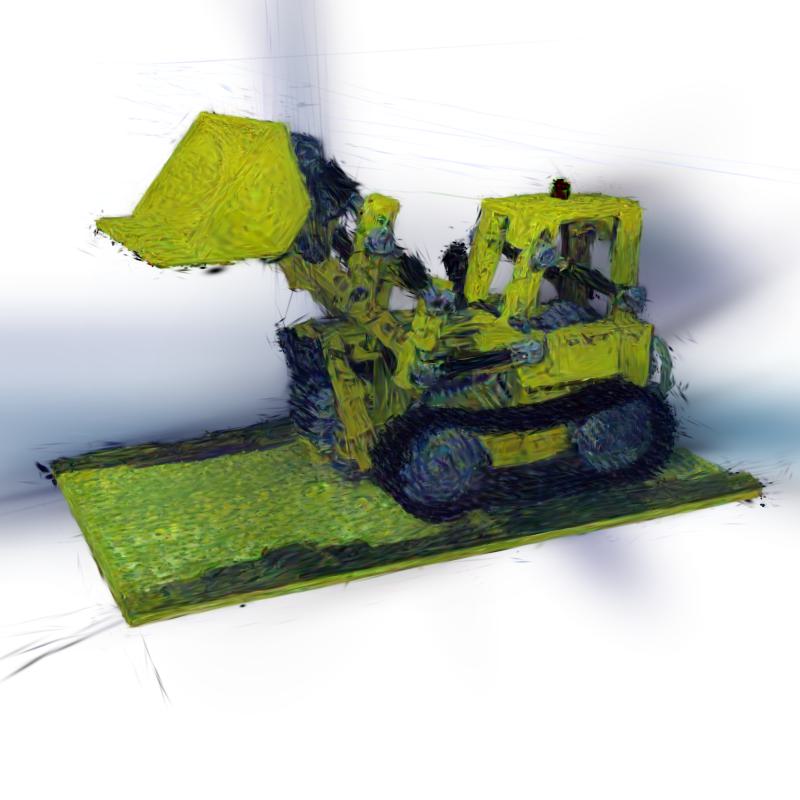}    &  
 \includegraphics[trim={50 130 50 100},clip, width=0.18\textwidth,valign=c]{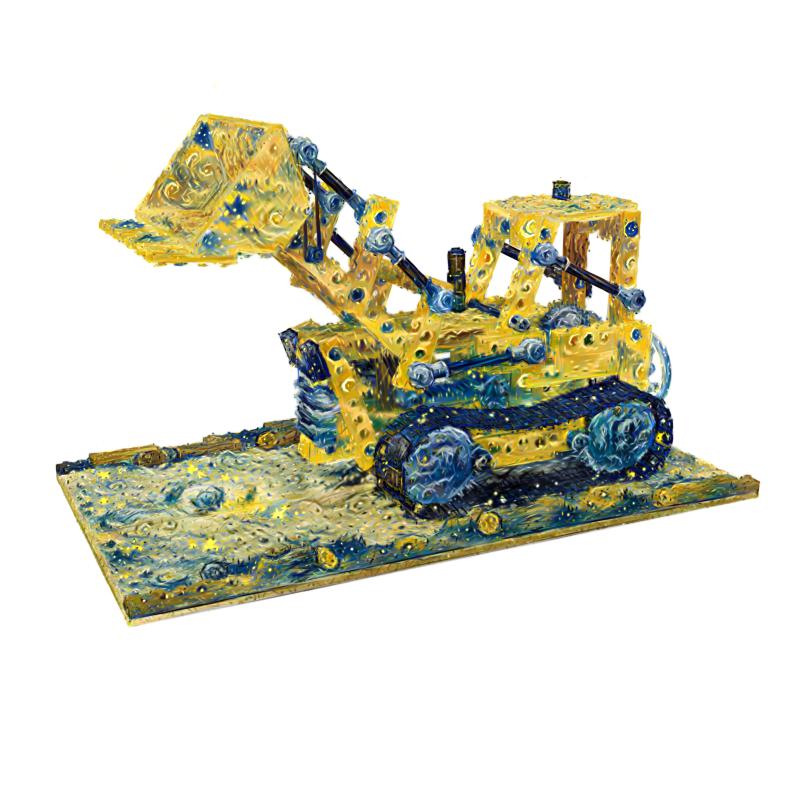}  &
 \includegraphics[trim={50 130 50 100},clip, width=0.18\textwidth,valign=c]{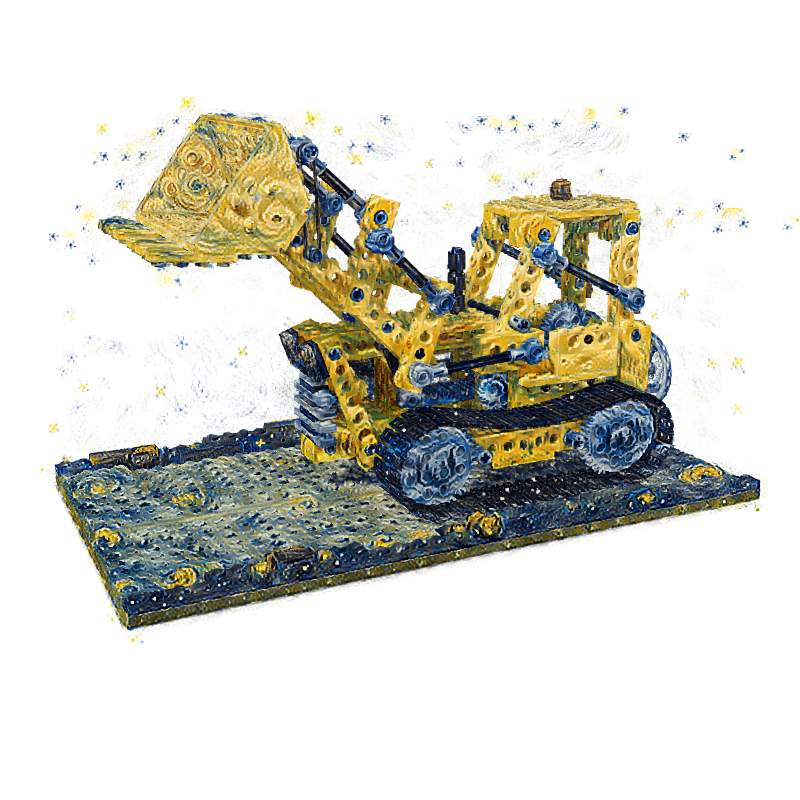}  \\
\fbox{
  \begin{minipage}[c][0.1\textwidth][c]{0.14\textwidth}
    \centering
      Mosaic
  \end{minipage}
}
  &
\includegraphics[trim={50 130 50 100},clip, width=0.18\textwidth,valign=c]{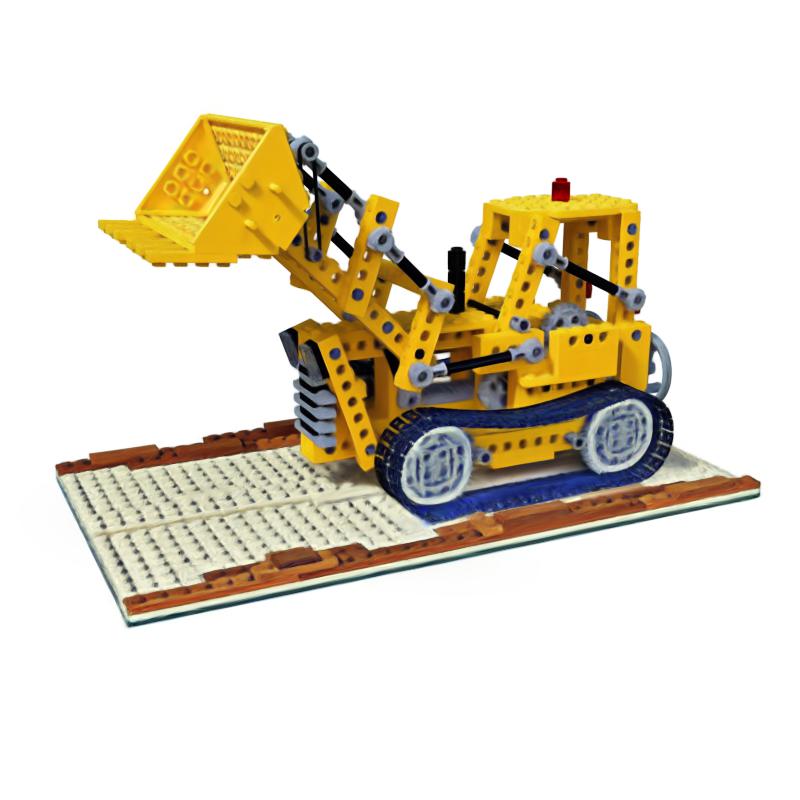}    & 
\includegraphics[trim={50 130 50 100},clip, width=0.18\textwidth,valign=c]{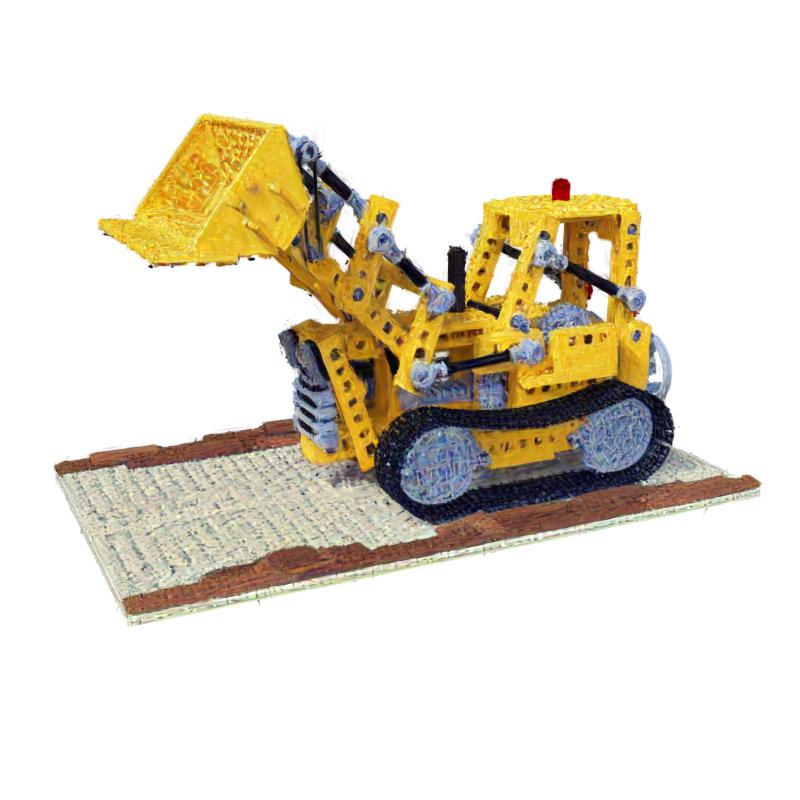}   & 
\includegraphics[trim={50 130 50 100},clip, width=0.18\textwidth,valign=c]{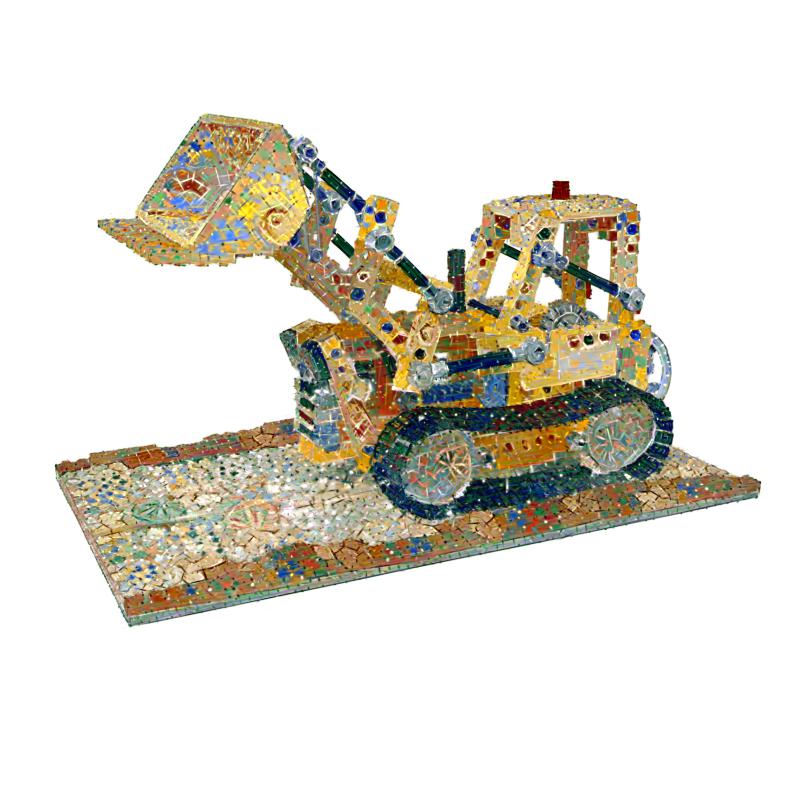}  &
\includegraphics[trim={50 130 50 100},clip, width=0.18\textwidth,valign=c]{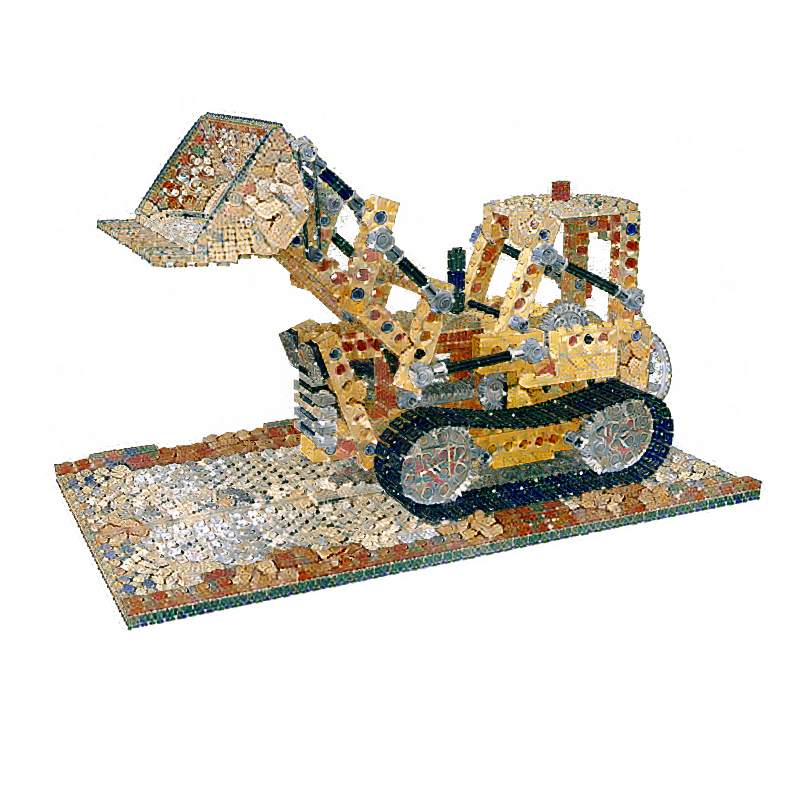}  \\
\fbox{
  \begin{minipage}[c][0.1\textwidth][c]{0.14\textwidth}
    \centering
      Scream by Edvard Munch
  \end{minipage}
}&  
 \includegraphics[trim={50 130 50 100},clip, width=0.18\textwidth,valign=c]{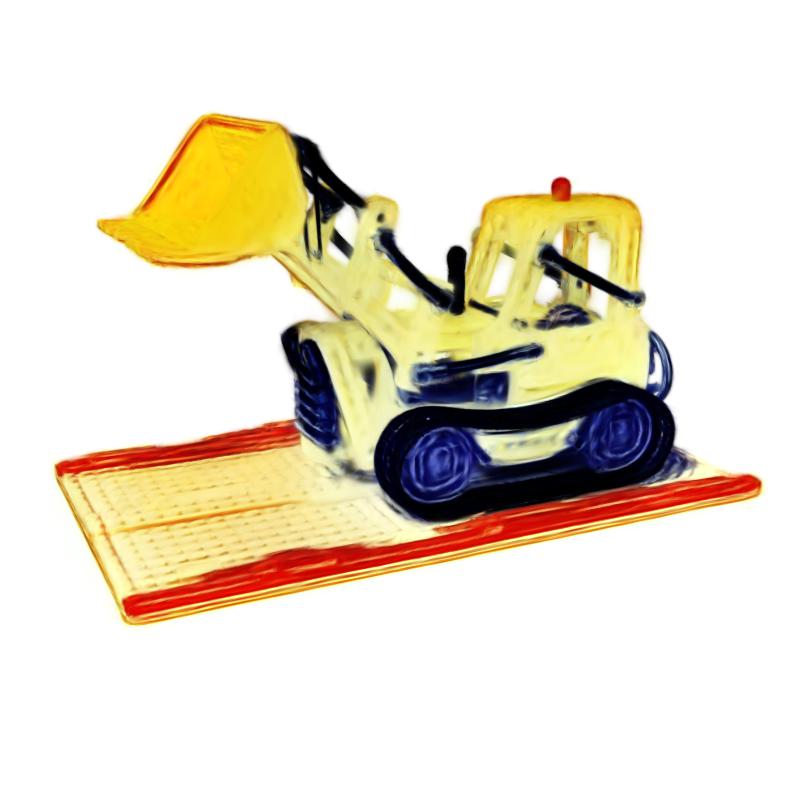}    &  
 \includegraphics[trim={50 130 50 100},clip, width=0.18\textwidth,valign=c]{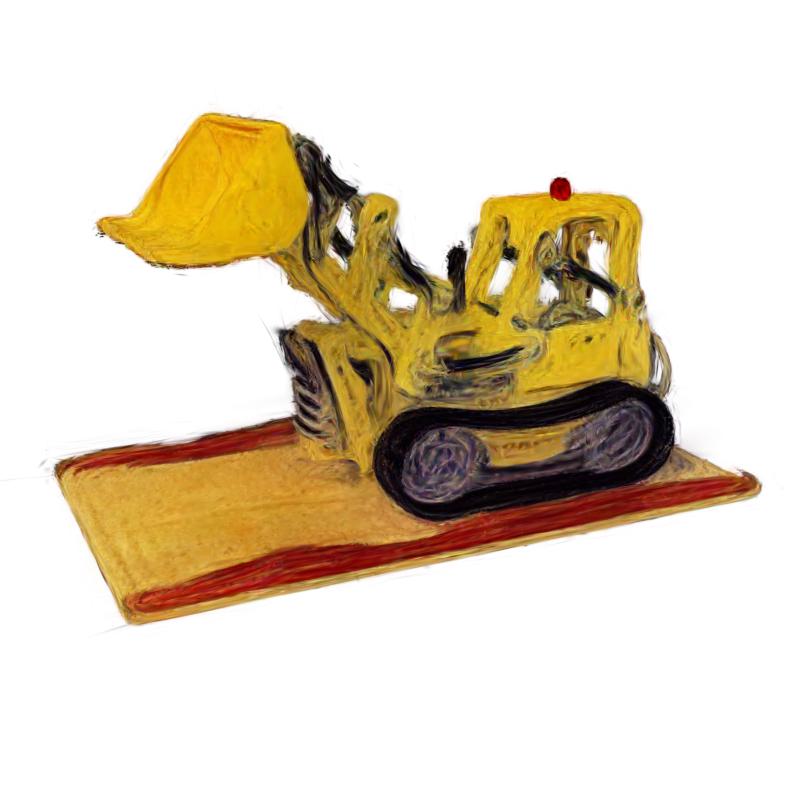}   &  
 \includegraphics[trim={50 130 50 100},clip, width=0.18\textwidth,valign=c]{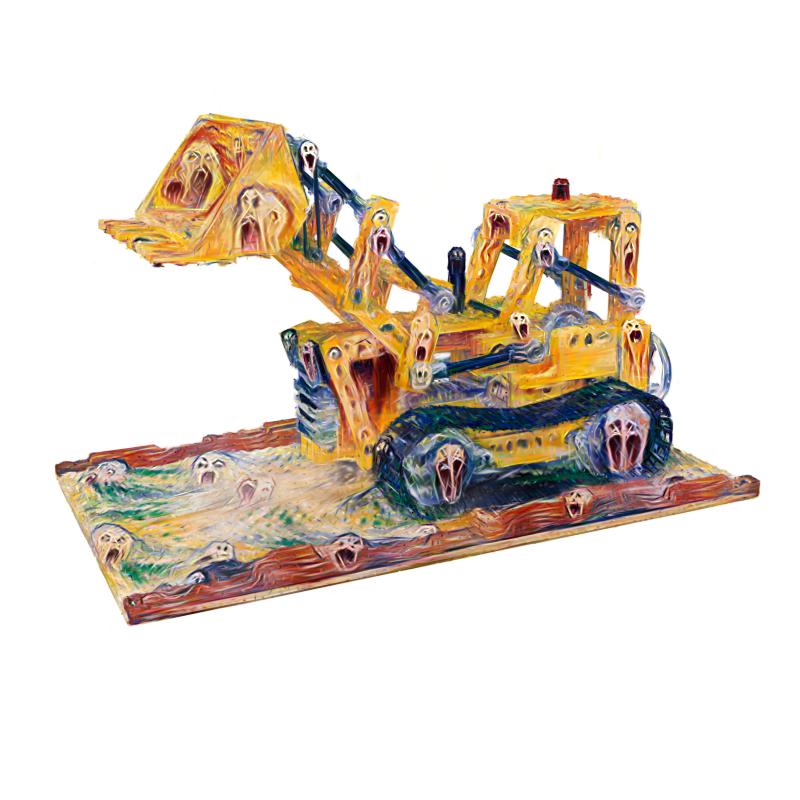}  &
 \includegraphics[trim={50 130 50 100},clip, width=0.18\textwidth,valign=c]{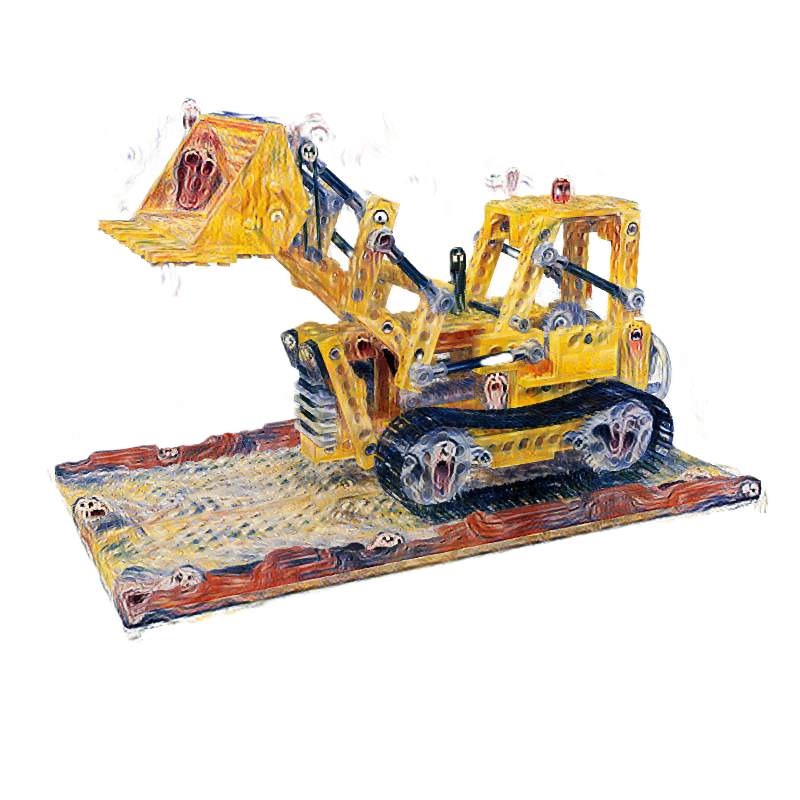}  \\
  \fbox{
  \begin{minipage}[c][0.1\textwidth][c]{0.14\textwidth}
    \centering
     Fire
  \end{minipage}
}
  &
\includegraphics[trim={50 100 50 190},clip, width=0.18\textwidth,valign=c]{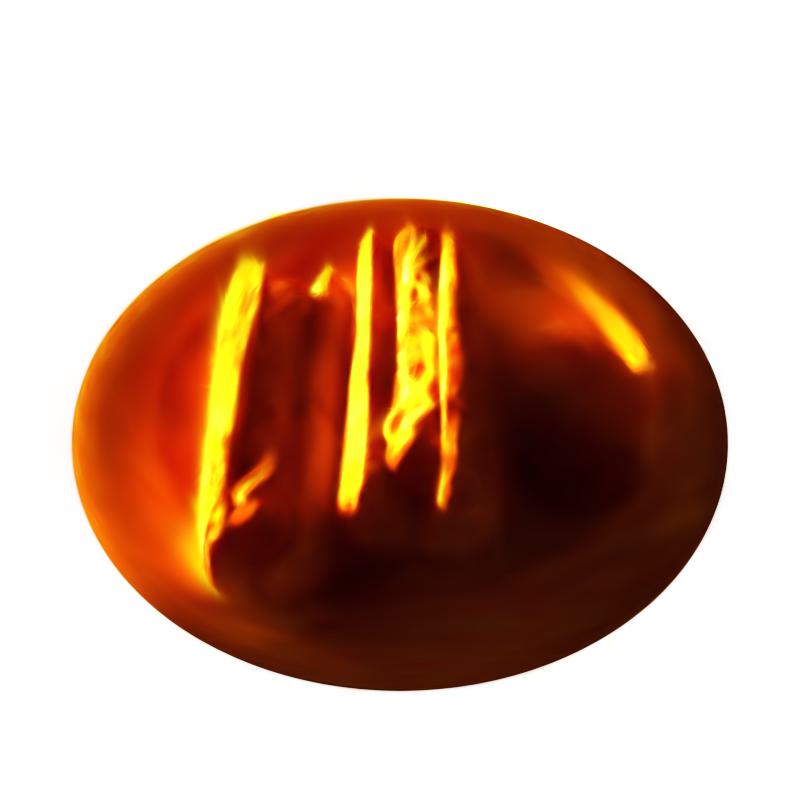}    & 
\includegraphics[trim={50 100 50 190},clip, width=0.18\textwidth,valign=c]{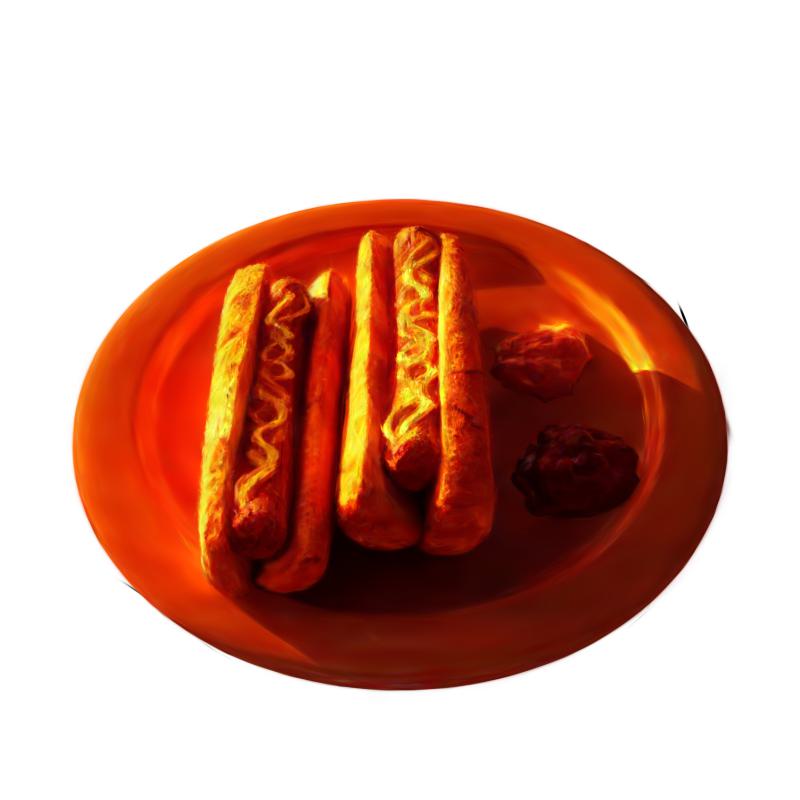}   & 
\includegraphics[trim={50 100 50 190},clip, width=0.18\textwidth,valign=c]{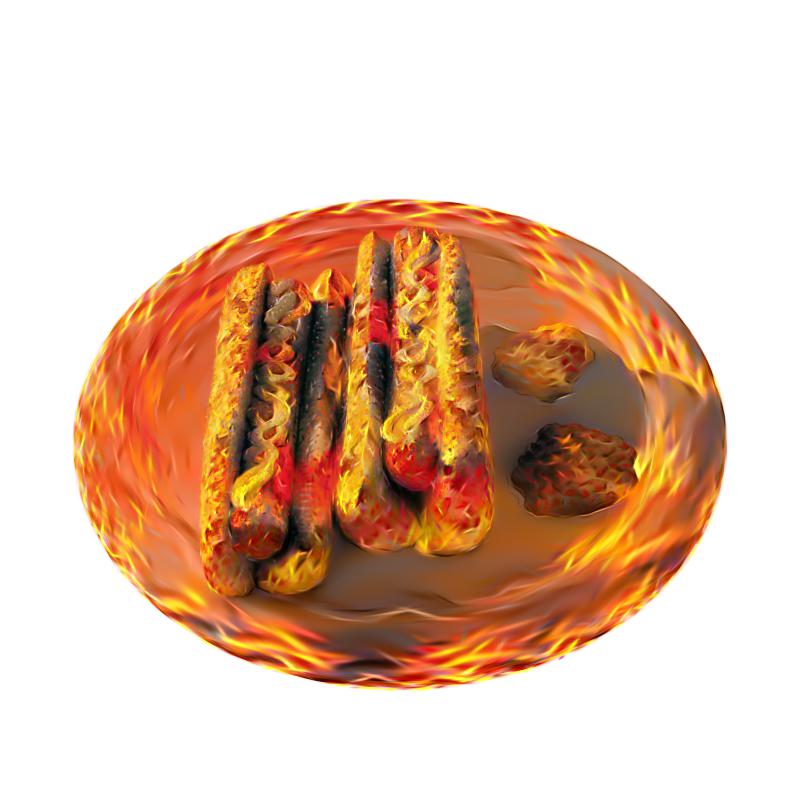}  &
\includegraphics[trim={50 100 50 190},clip, width=0.18\textwidth,valign=c]{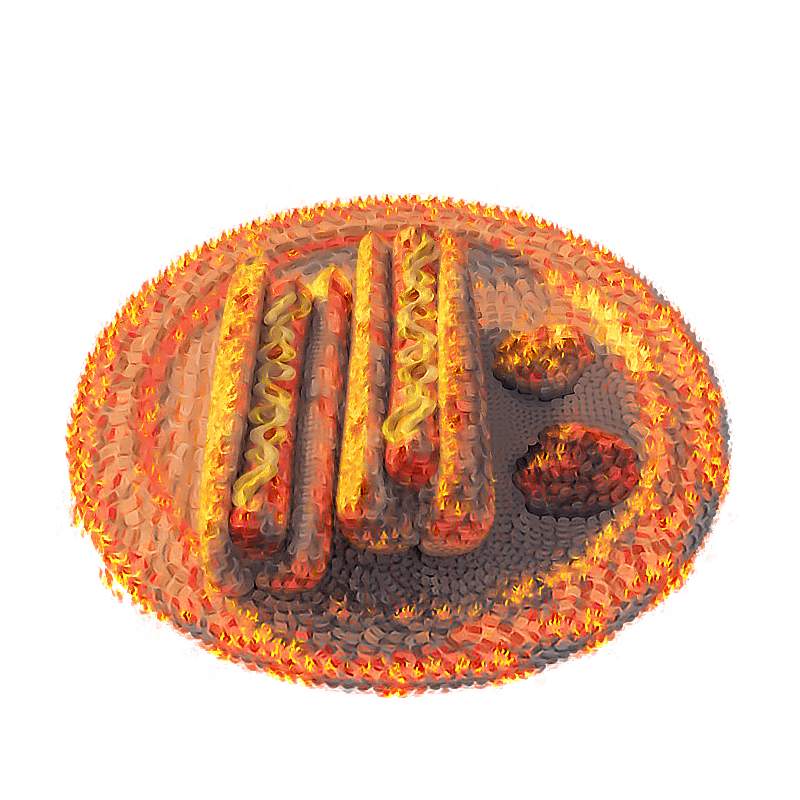}  \\
\fbox{
  \begin{minipage}[c][0.1\textwidth][c]{0.14\textwidth}
    \centering
      Starry Night by Vincent van Gogh
  \end{minipage}
}&  
 \includegraphics[trim={50 100 50 190},clip, width=0.18\textwidth,valign=c]{imgs/3D/comp_text/igs2gs/hotdog/000_hotdog_Starry_Night_by_Vincent_van_Gogh.jpg}     &  
 \includegraphics[trim={50 100 50 190},clip, width=0.18\textwidth,valign=c]{imgs/3D/comp_text/dge/hotdog/000_hotdog_Starry_Night_by_Vincent_van_Gogh.jpg}    &  
 \includegraphics[trim={50 100 50 190},clip, width=0.18\textwidth,valign=c]{imgs/3D/comp_text/gaussian/hotdog/000_hotdog_Starry_Night_by_Vincent_van_Gogh.jpg}  &
 \includegraphics[trim={50 100 50 190},clip, width=0.18\textwidth,valign=c]{imgs/3D/comp_text/ours/hotdog/000_night.jpg}  \\
\fbox{
  \begin{minipage}[c][0.1\textwidth][c]{0.14\textwidth}
    \centering
      Mosaic
  \end{minipage}
}
  &
\includegraphics[trim={50 100 50 190},clip, width=0.18\textwidth,valign=c]{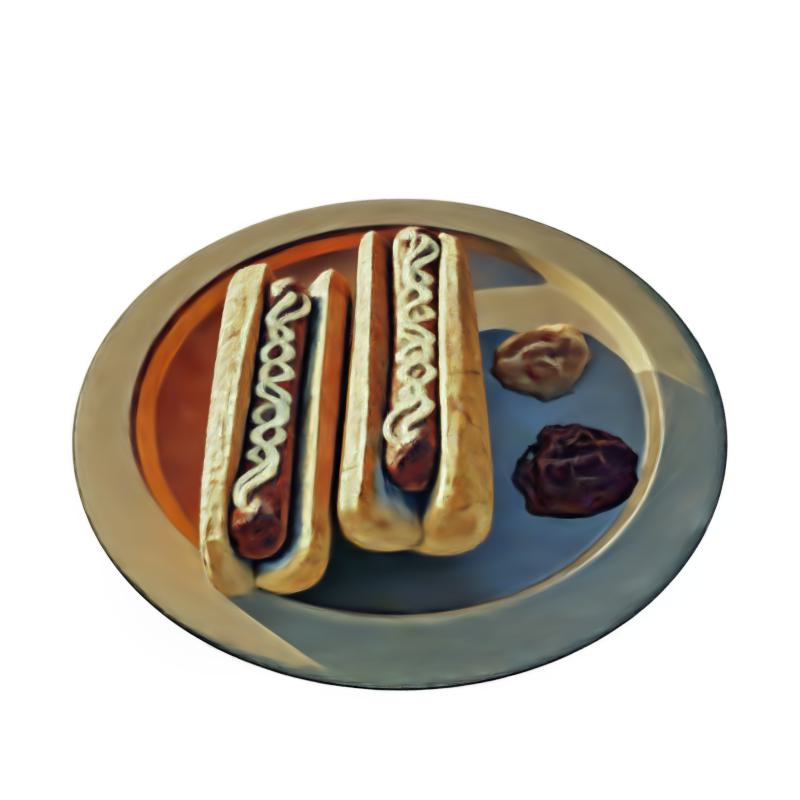}    & 
\includegraphics[trim={50 100 50 190},clip, width=0.18\textwidth,valign=c]{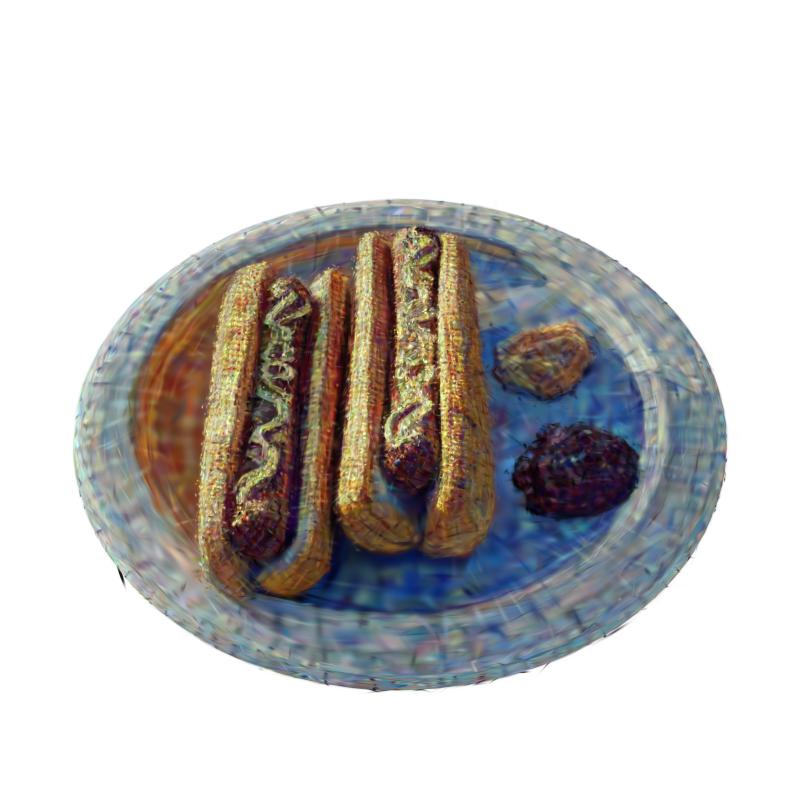}   & 
\includegraphics[trim={50 100 50 190},clip, width=0.18\textwidth,valign=c]{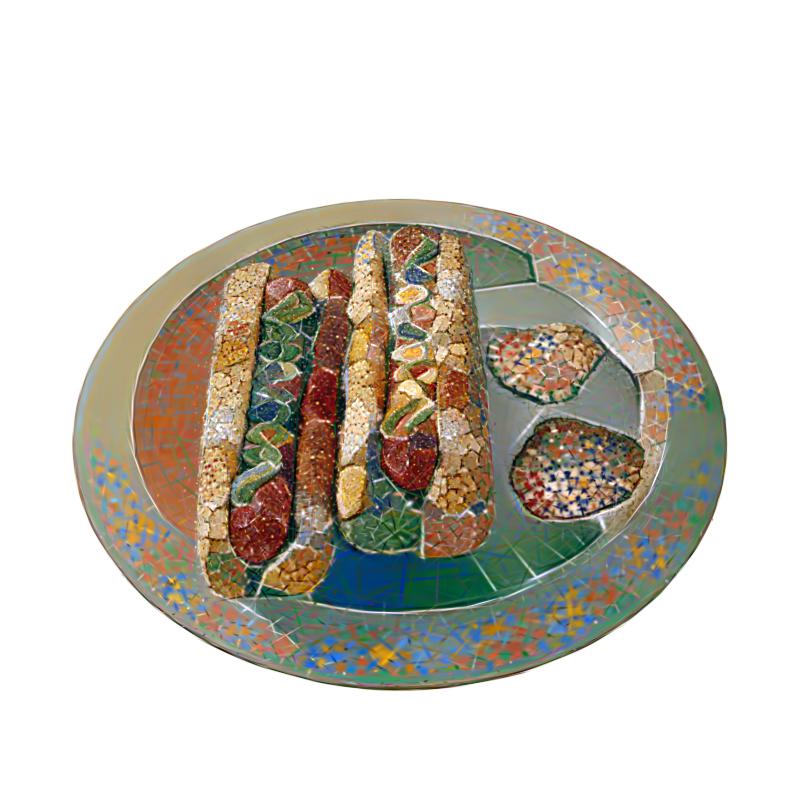}  &
\includegraphics[trim={50 100 50 190},clip, width=0.18\textwidth,valign=c]{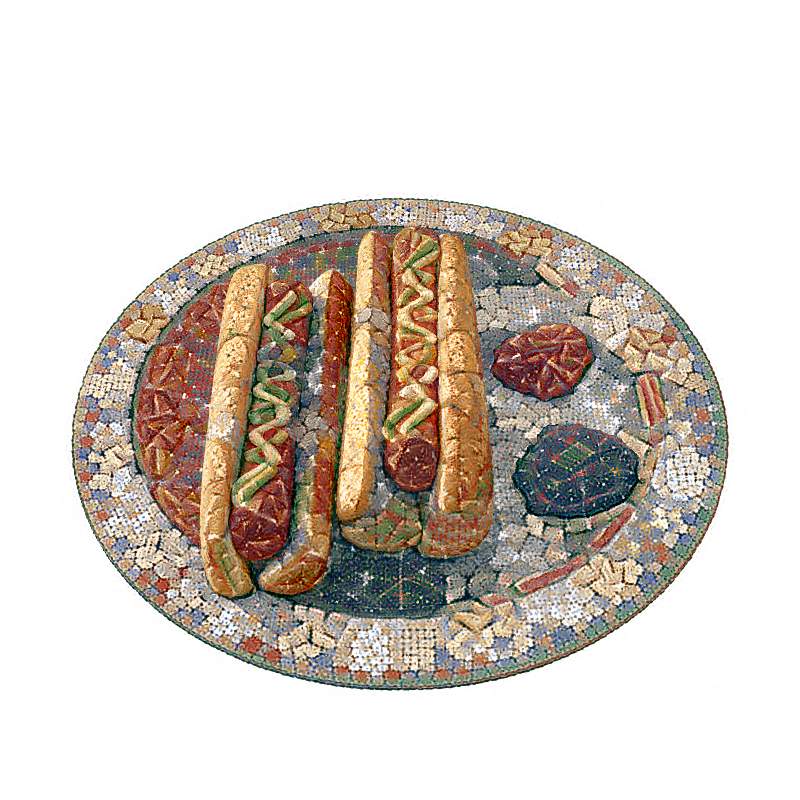}  \\
\fbox{
  \begin{minipage}[c][0.1\textwidth][c]{0.14\textwidth}
    \centering
      Scream by Edvard Munch
  \end{minipage}
}&  
 \includegraphics[trim={50 100 50 190},clip, width=0.18\textwidth,valign=c]{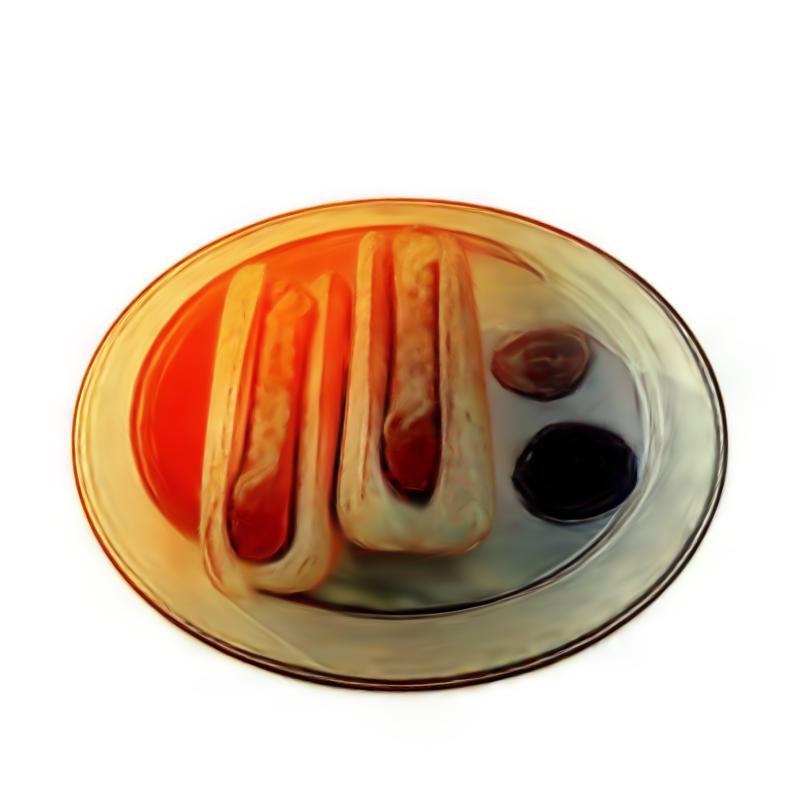}    &  
 \includegraphics[trim={50 100 50 190},clip, width=0.18\textwidth,valign=c]{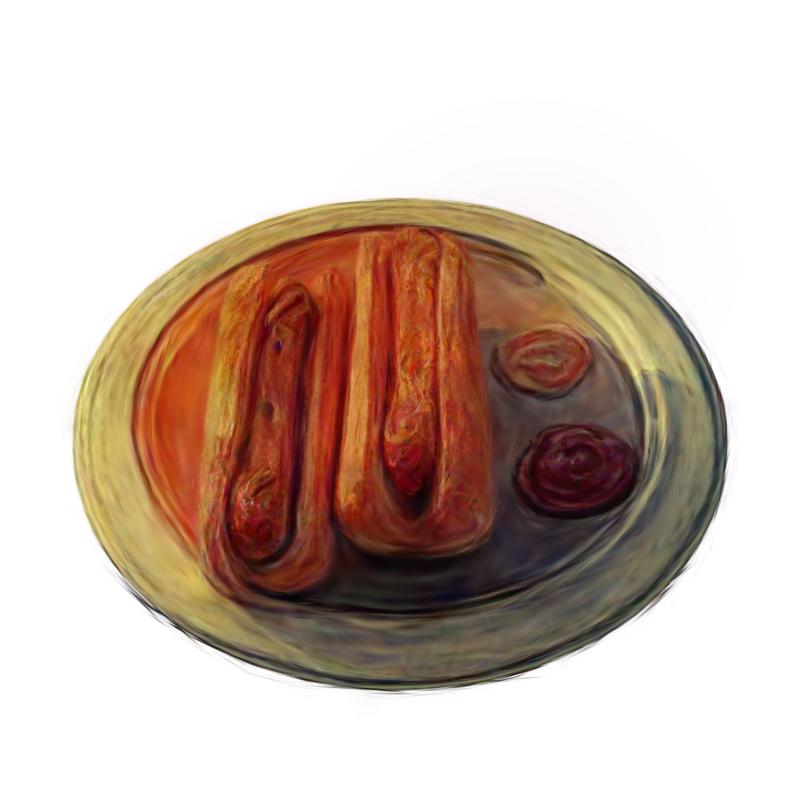}   &  
 \includegraphics[trim={50 100 50 190},clip, width=0.18\textwidth,valign=c]{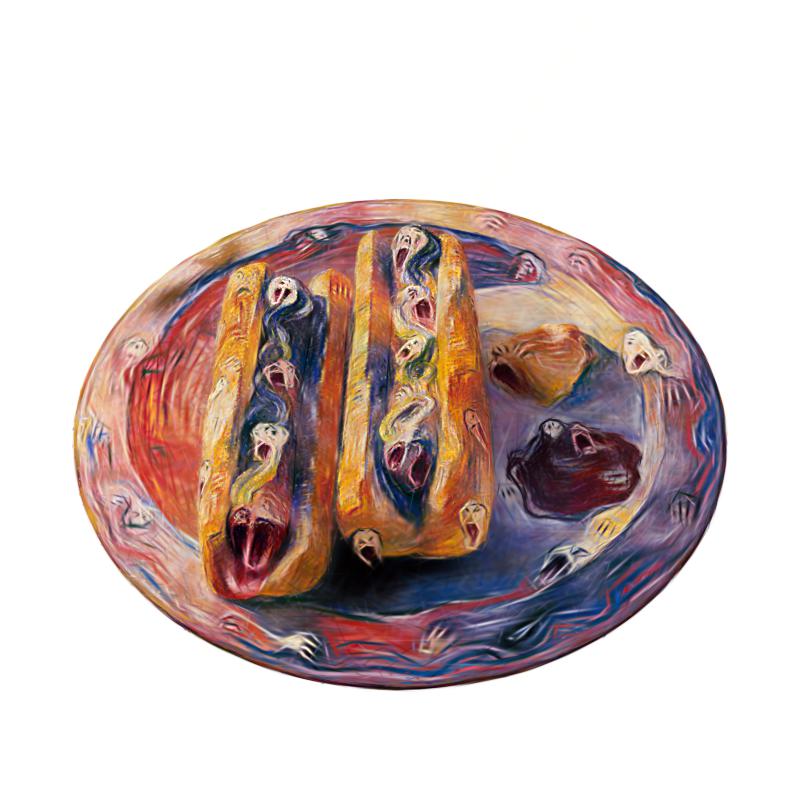}  &
 \includegraphics[trim={50 100 50 190},clip, width=0.18\textwidth,valign=c]{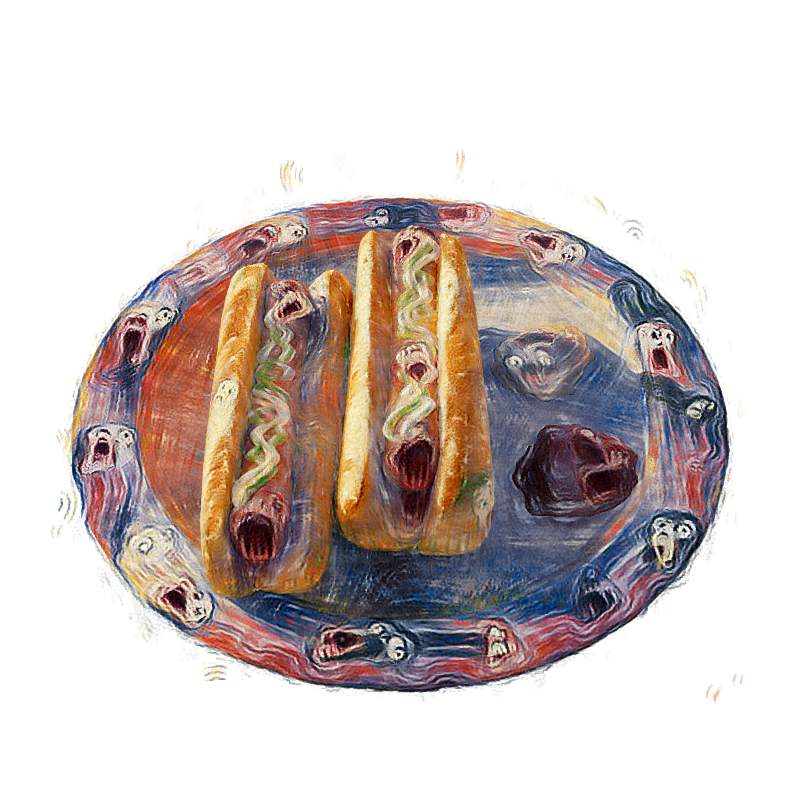}  \\
\end{tabular}
\caption{Full comparison of \our{} (our) and baseline models in 3D style transfer, conditioned by text, on \textit{hotdog} and \textit{lego} objects from NeRF-Synthetic dataset \cite{mildenhall2020nerf}.}
\label{fig:comp_text_obj}
\end{figure*}

\begin{figure*}[p]
\centering
\begin{tabular}{ccccc}
Style &  StyleGaussian \cite{liu2024stylegaussian} &  G-Style \cite{kovacs2024G}  &  CLIPGaussian \cite{howil2026clipgaussian} &  \our{} \\
  \includegraphics[trim={0 0 0 0},clip, width=0.10\textwidth,valign=c]{imgs/styles/fire.jpg}
  &
\includegraphics[trim={50 130 50 100},clip, width=0.18\textwidth,valign=c]{imgs/3D/comp_image/style_gaussian/lego/068_lego_fire.jpg}    & 
\includegraphics[trim={50 130 50 100},clip, width=0.18\textwidth,valign=c]{imgs/3D/comp_image/gstyle/lego/068_lego_fire.jpg}   & 
\includegraphics[trim={50 130 50 100},clip, width=0.18\textwidth,valign=c]{imgs/3D/comp_image/ours/lego/068_lego_fire.jpg}  &
\includegraphics[trim={50 130 50 100},clip, width=0.18\textwidth,valign=c]{imgs/3D/comp_image/omnistyle/lego/068_lego_fire.jpg}  \\
\includegraphics[trim={50 0 50 0},clip, width=0.10\textwidth,valign=c]{imgs/styles/starry_night.jpg}&  
 \includegraphics[trim={50 130 50 100},clip, width=0.18\textwidth,valign=c]{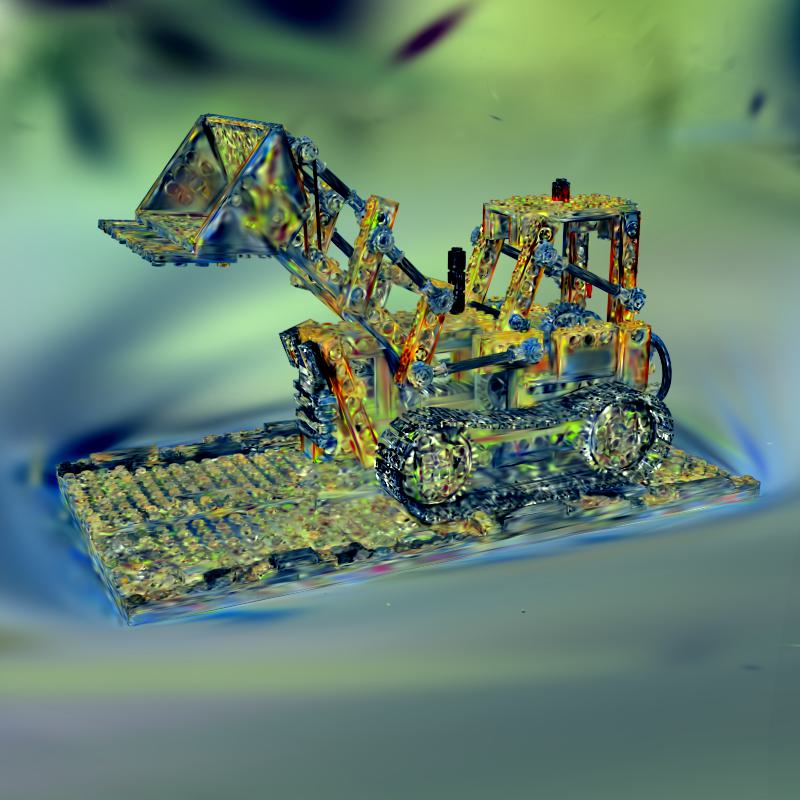}    &  
 \includegraphics[trim={50 130 50 100},clip, width=0.18\textwidth,valign=c]{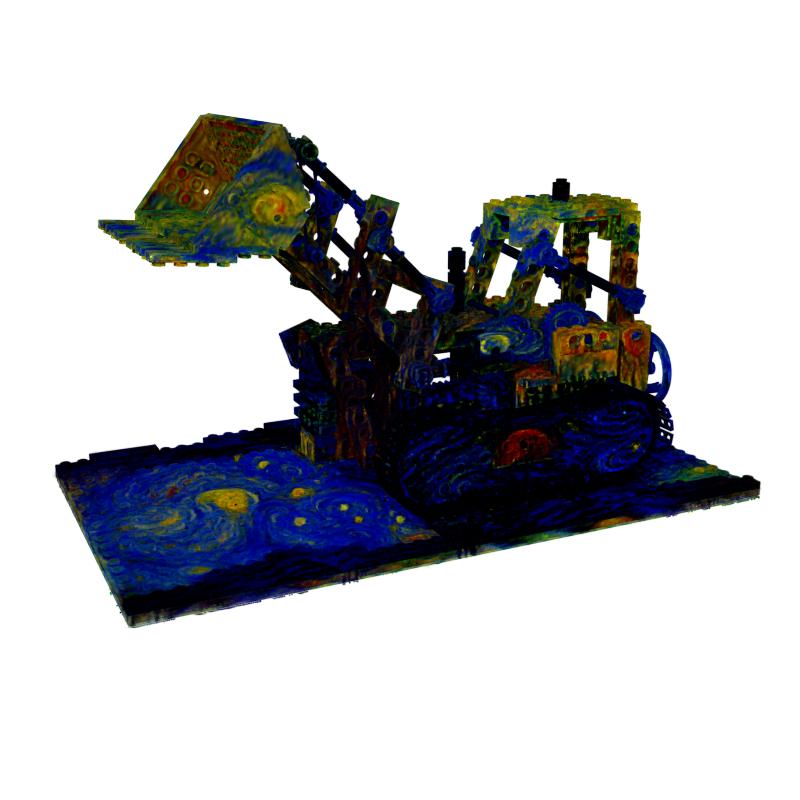}   &  
 \includegraphics[trim={50 130 50 100},clip, width=0.18\textwidth,valign=c]{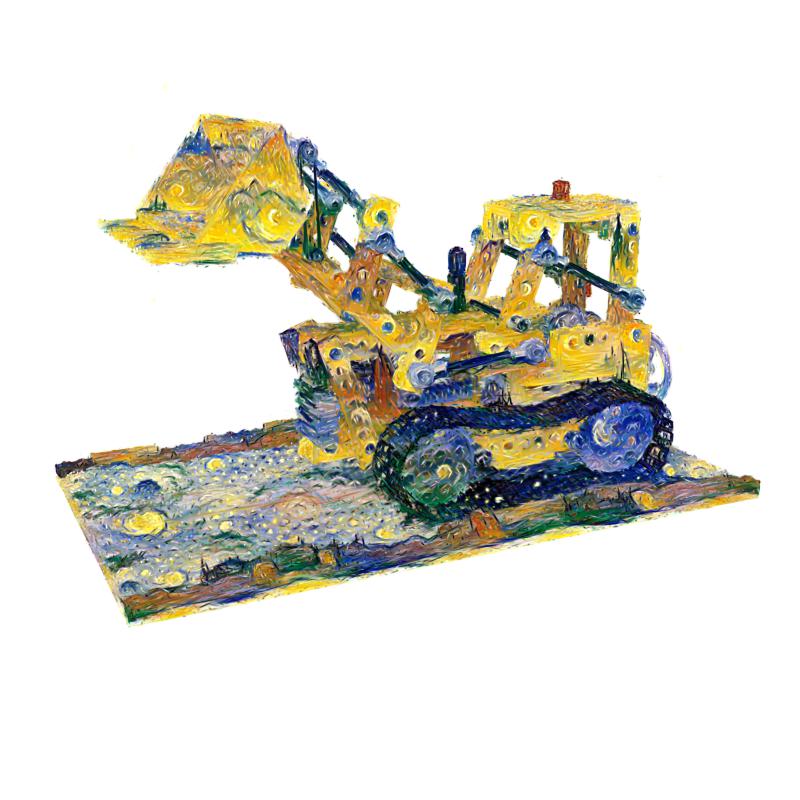}  &
 \includegraphics[trim={50 130 50 100},clip, width=0.18\textwidth,valign=c]{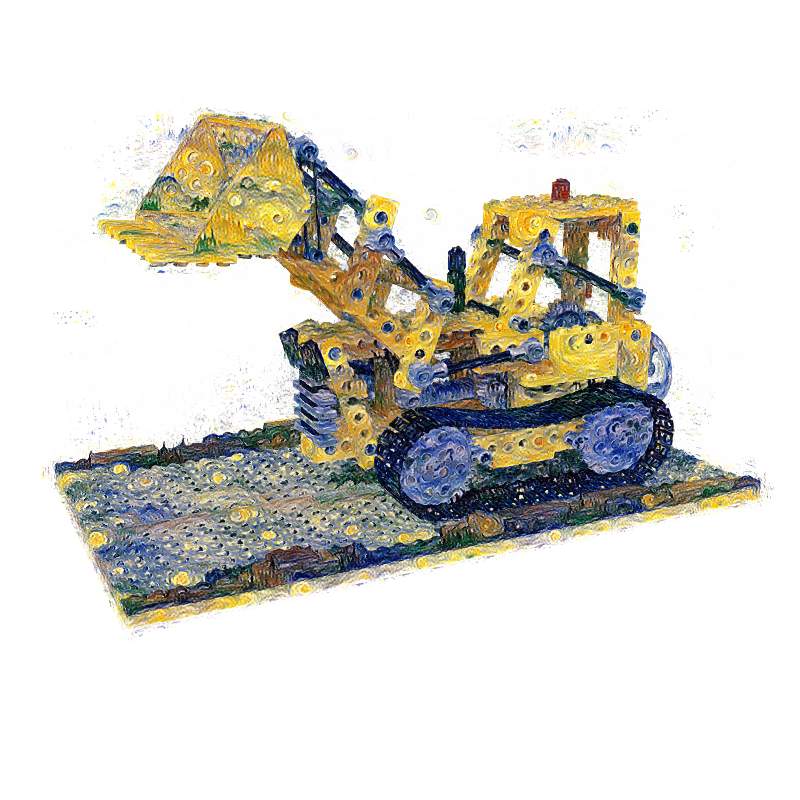}  \\
\includegraphics[trim={0 0 0 0},clip, width=0.10\textwidth,valign=c]{imgs/styles/mosaic.jpg}
  & 
\includegraphics[trim={50 130 50 100},clip, width=0.18\textwidth,valign=c]{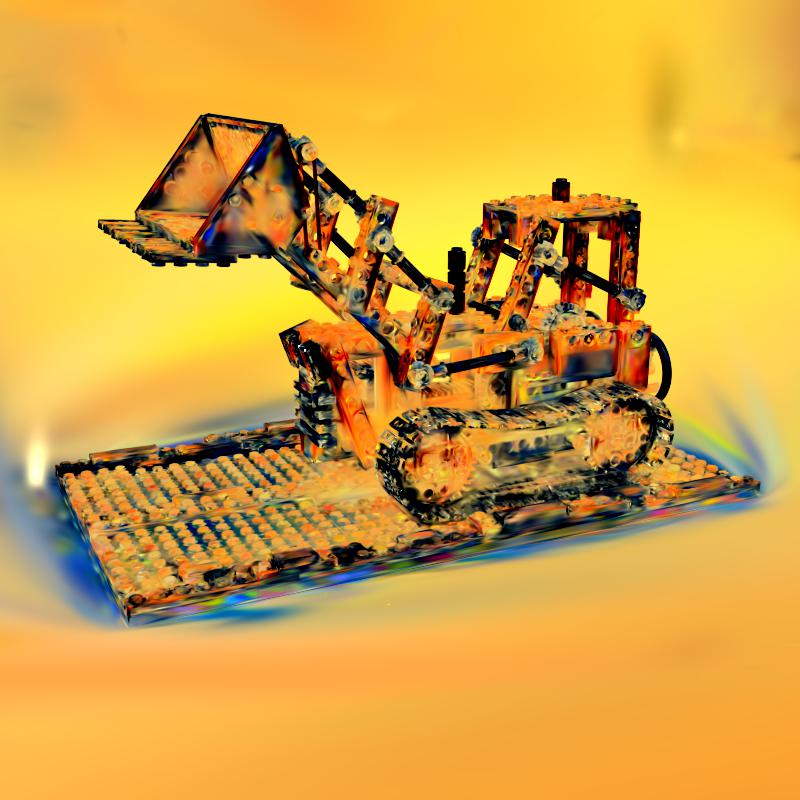}    & 
\includegraphics[trim={50 130 50 100},clip, width=0.18\textwidth,valign=c]{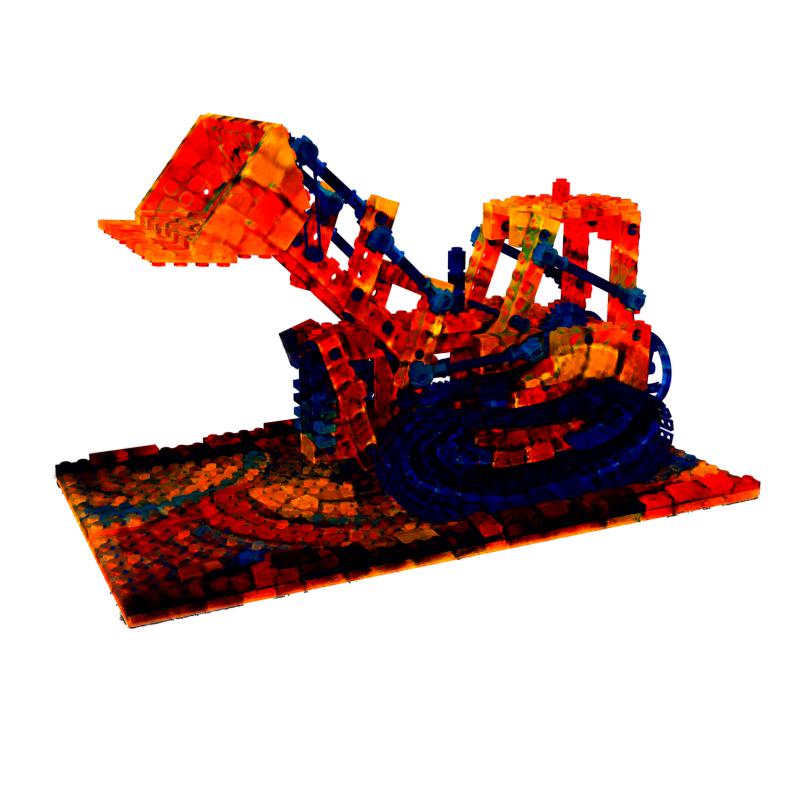}   & 
\includegraphics[trim={50 130 50 100},clip, width=0.18\textwidth,valign=c]{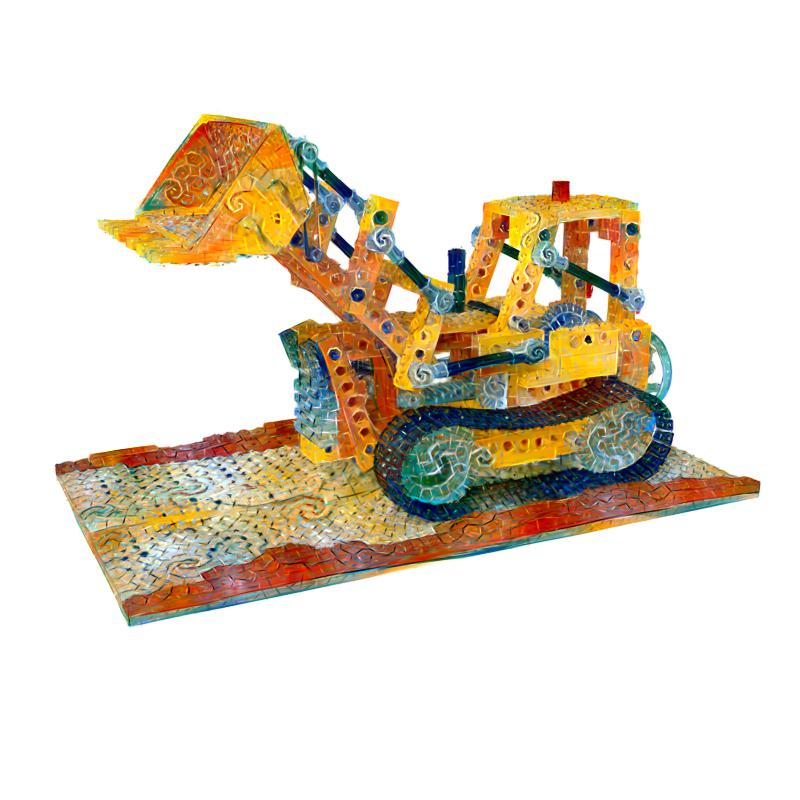}  &
\includegraphics[trim={50 130 50 100},clip, width=0.18\textwidth,valign=c]{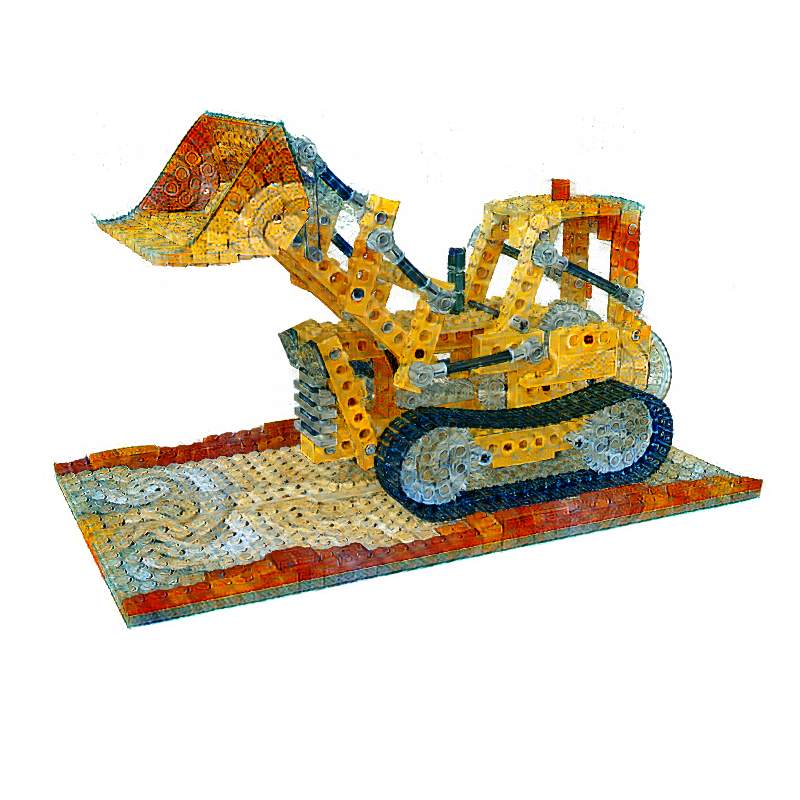}  \\
\includegraphics[trim={0 0 0 180},clip, width=0.10\textwidth,valign=c]{imgs/styles/scream.jpg}&  
 \includegraphics[trim={50 130 50 100},clip, width=0.18\textwidth,valign=c]{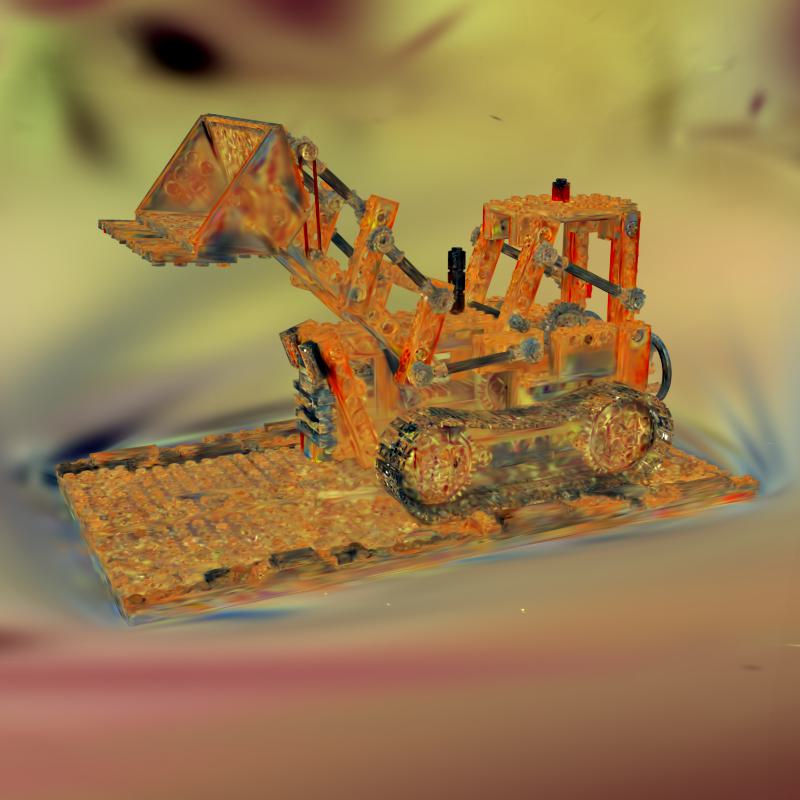}    &  
 \includegraphics[trim={50 130 50 100},clip, width=0.18\textwidth,valign=c]{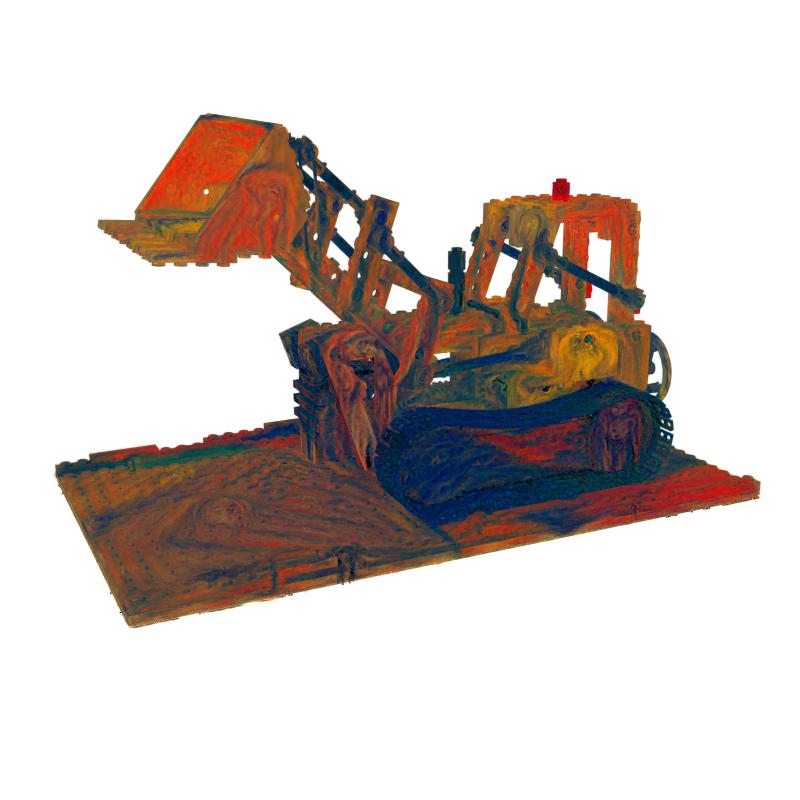}   &  
 \includegraphics[trim={50 130 50 100},clip, width=0.18\textwidth,valign=c]{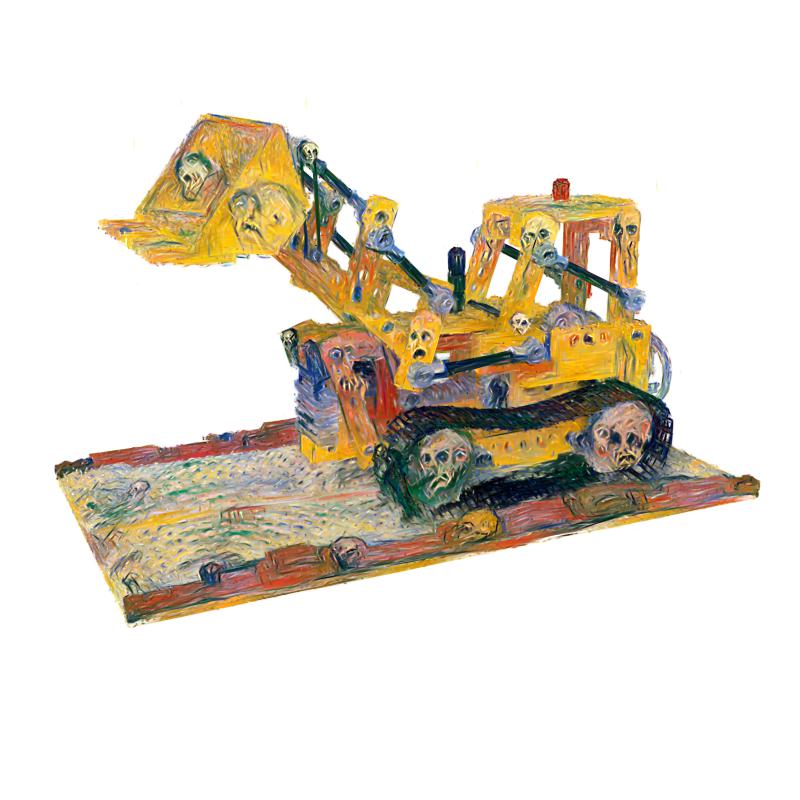}  &
 \includegraphics[trim={50 130 50 100},clip, width=0.18\textwidth,valign=c]{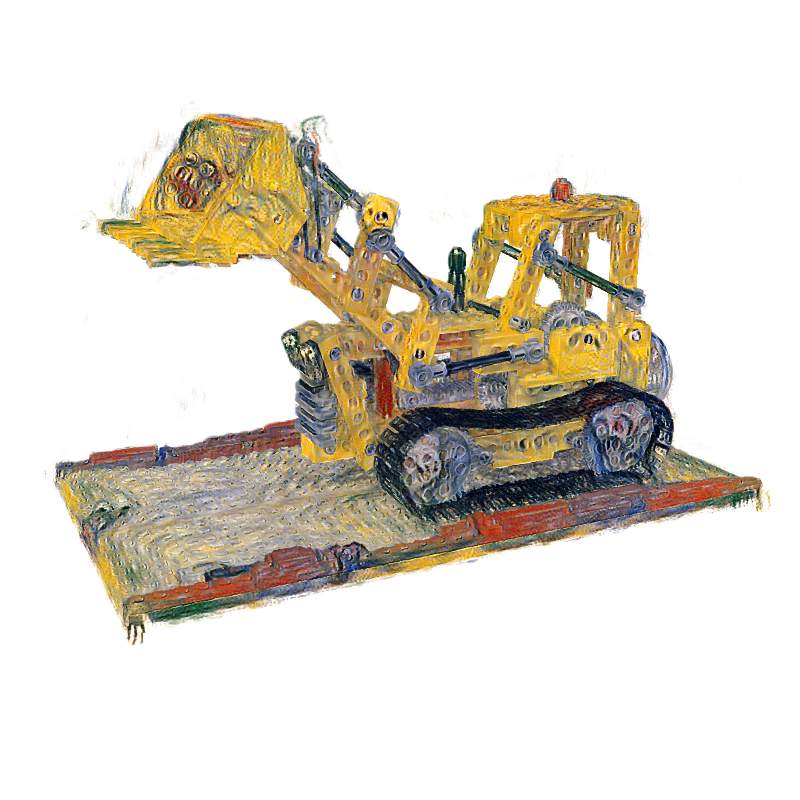}  \\
   \includegraphics[trim={0 0 0 0},clip, width=0.10\textwidth,valign=c]{imgs/styles/fire.jpg}
  &
\includegraphics[trim={50 100 50 190},clip, width=0.18\textwidth,valign=c]{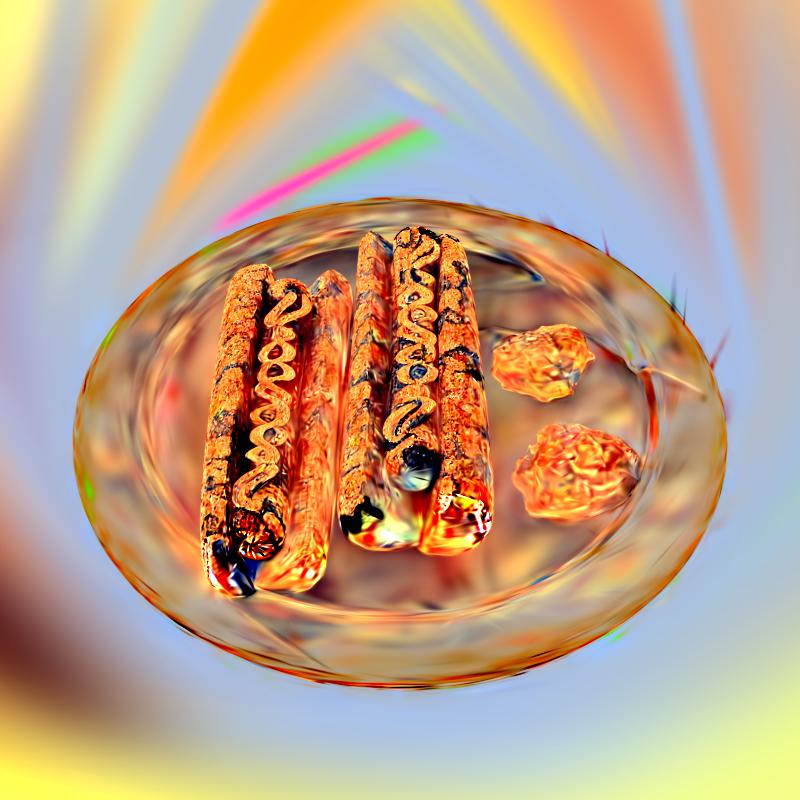}    & 
\includegraphics[trim={50 100 50 190},clip, width=0.18\textwidth,valign=c]{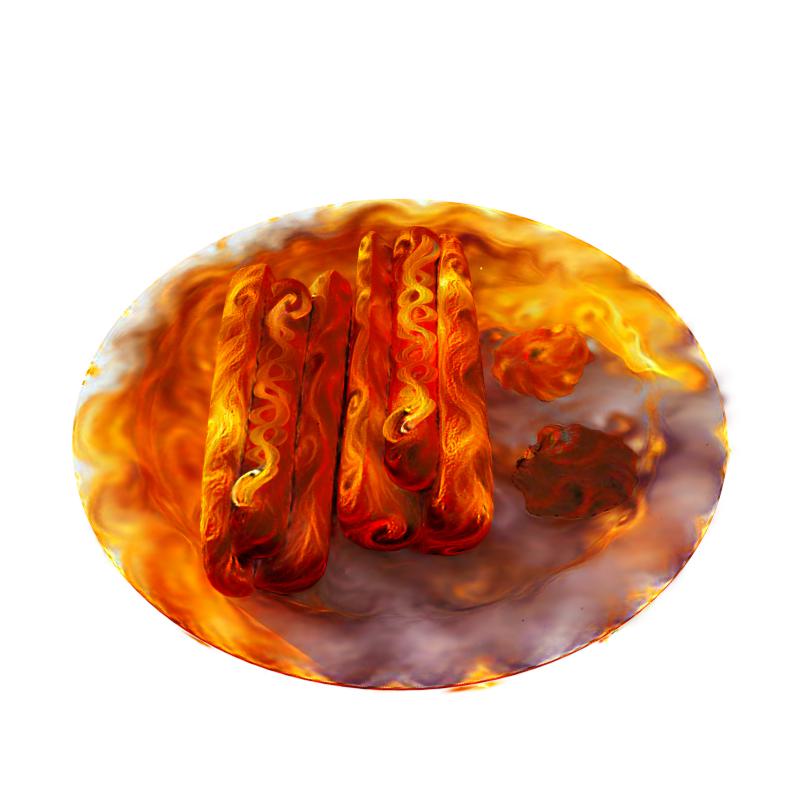}   & 
\includegraphics[trim={50 100 50 190},clip, width=0.18\textwidth,valign=c]{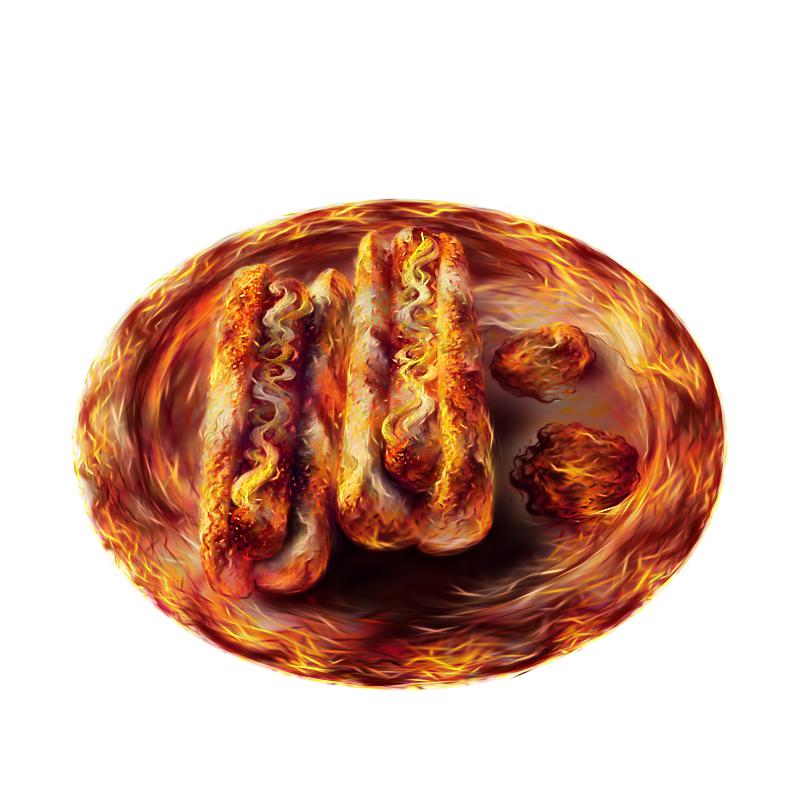}  &
\includegraphics[trim={50 100 50 190},clip, width=0.18\textwidth,valign=c]{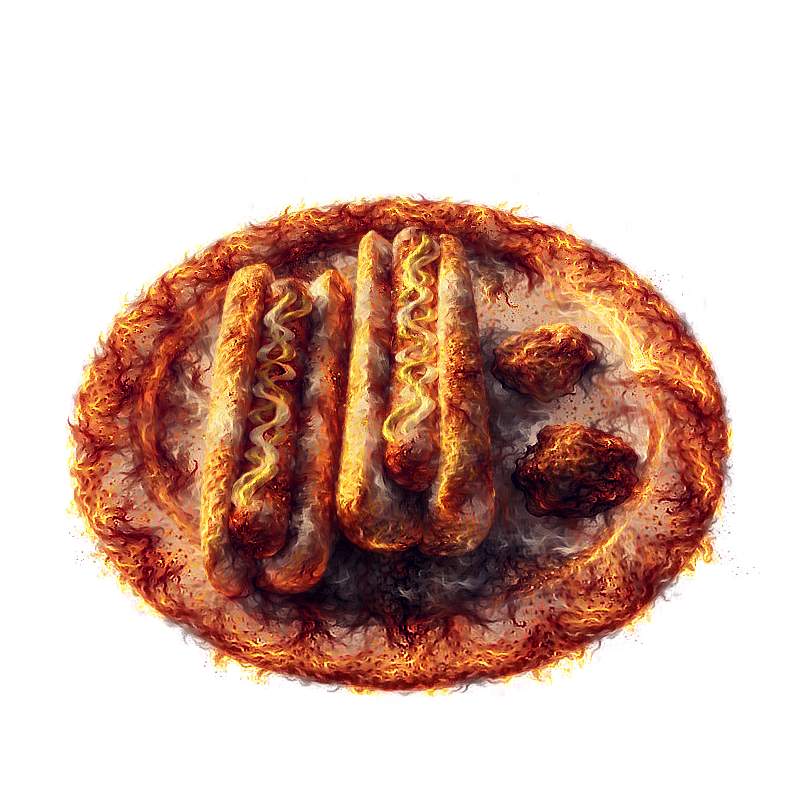}  \\
\includegraphics[trim={50 0 50 0},clip, width=0.10\textwidth,valign=c]{imgs/styles/starry_night.jpg}&   
 \includegraphics[trim={50 100 50 190},clip, width=0.18\textwidth,valign=c]{imgs/3D/comp_image/style_gaussian/hotdog/000_hotdog_starry_night.jpg}    &  
 \includegraphics[trim={50 100 50 190},clip, width=0.18\textwidth,valign=c]{imgs/3D/comp_image/gstyle/hotdog/000_hotdog_starry_night.jpg}   &  
 \includegraphics[trim={50 100 50 190},clip, width=0.18\textwidth,valign=c]{imgs/3D/comp_image/ours/hotdog/000_hotdog_starry_night.jpg}  &
 \includegraphics[trim={50 100 50 190},clip, width=0.18\textwidth,valign=c]{imgs/3D/comp_image/omnistyle/hotdog/000_hotdog_starry_night.jpg}  \\
\includegraphics[trim={0 0 0 0},clip, width=0.10\textwidth,valign=c]{imgs/styles/mosaic.jpg}
  &
\includegraphics[trim={50 100 50 190},clip, width=0.18\textwidth,valign=c]{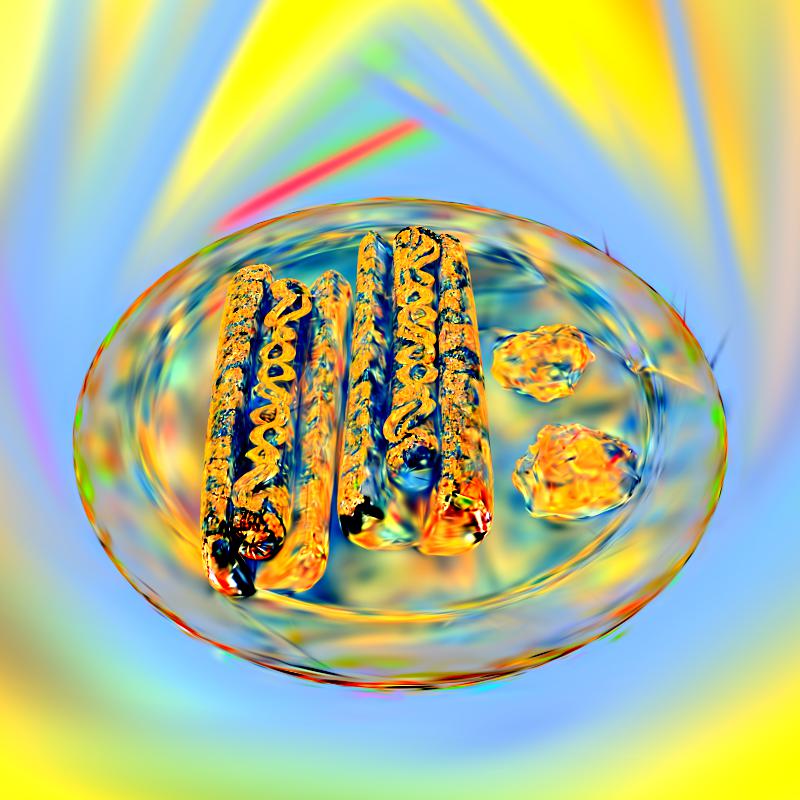}    & 
\includegraphics[trim={50 100 50 190},clip, width=0.18\textwidth,valign=c]{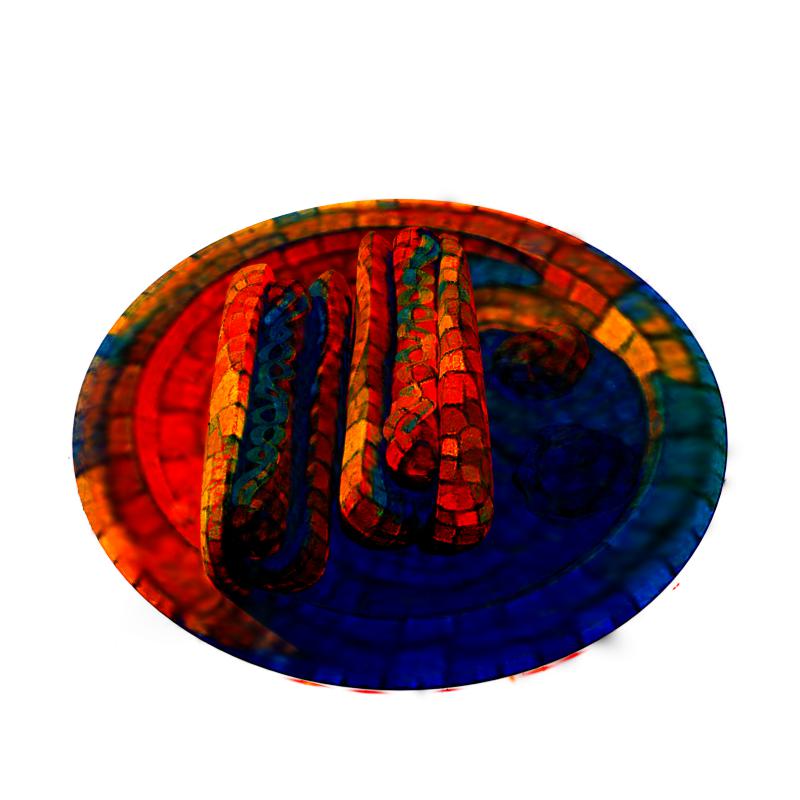}   & 
\includegraphics[trim={50 100 50 190},clip, width=0.18\textwidth,valign=c]{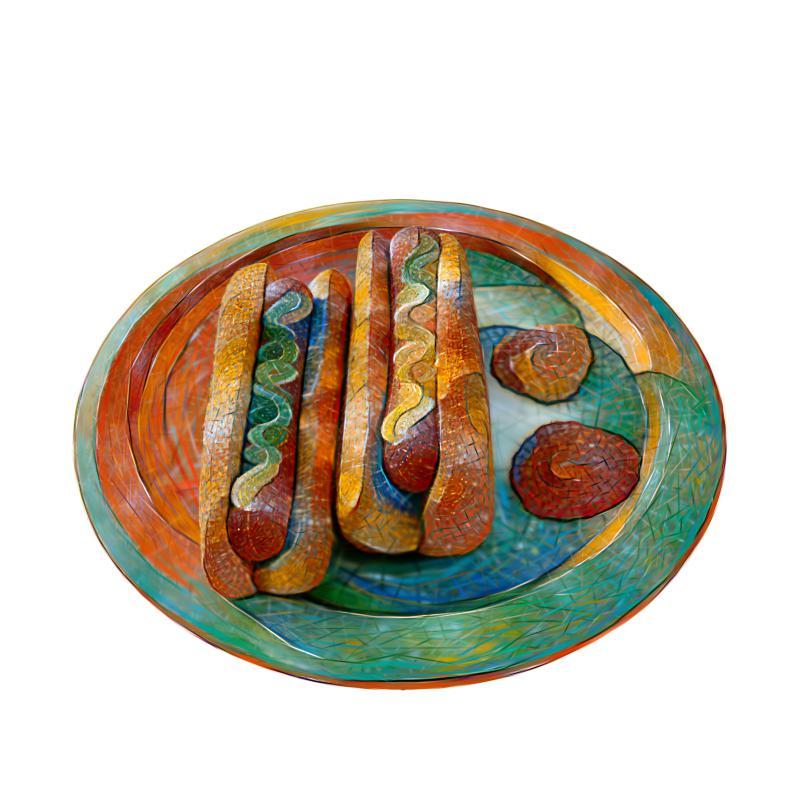}  &
\includegraphics[trim={50 100 50 190},clip, width=0.18\textwidth,valign=c]{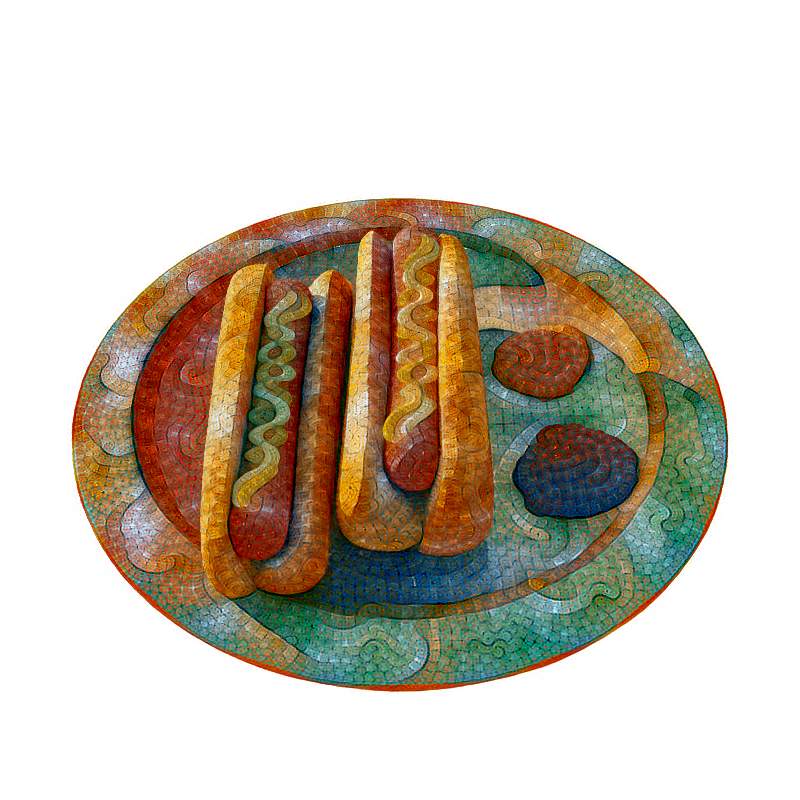}  \\
\includegraphics[trim={0 0 0 180},clip, width=0.10\textwidth,valign=c]{imgs/styles/scream.jpg}&  
 \includegraphics[trim={50 100 50 190},clip, width=0.18\textwidth,valign=c]{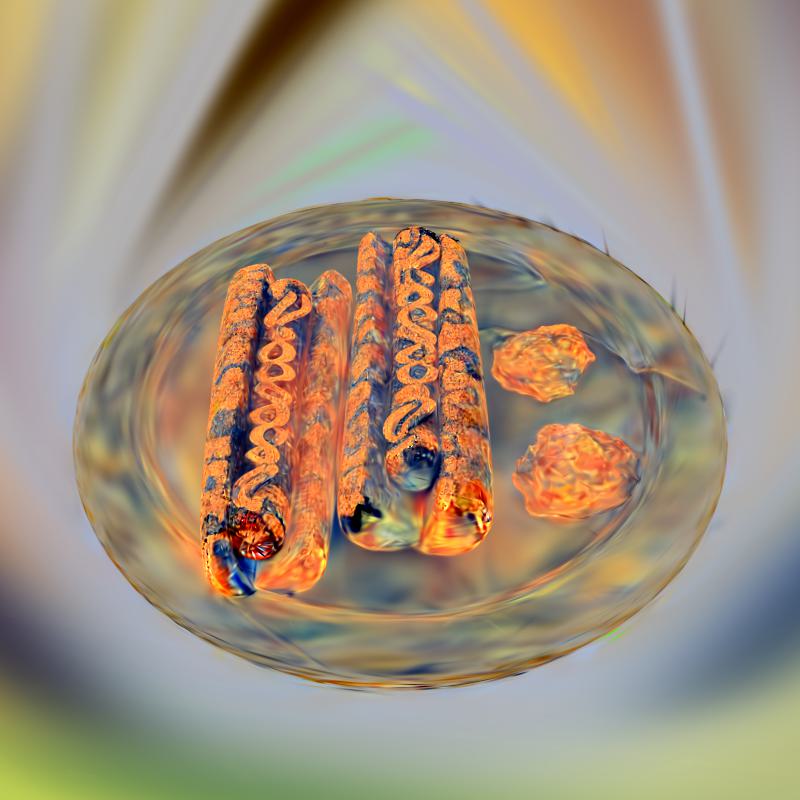}    &  
 \includegraphics[trim={50 100 50 190},clip, width=0.18\textwidth,valign=c]{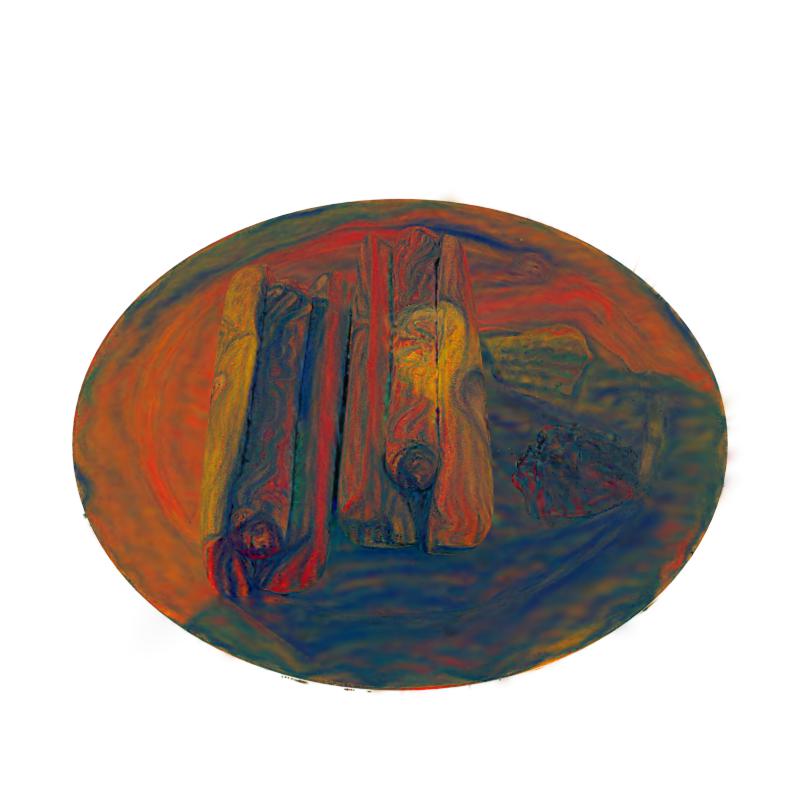}   &  
 \includegraphics[trim={50 100 50 190},clip, width=0.18\textwidth,valign=c]{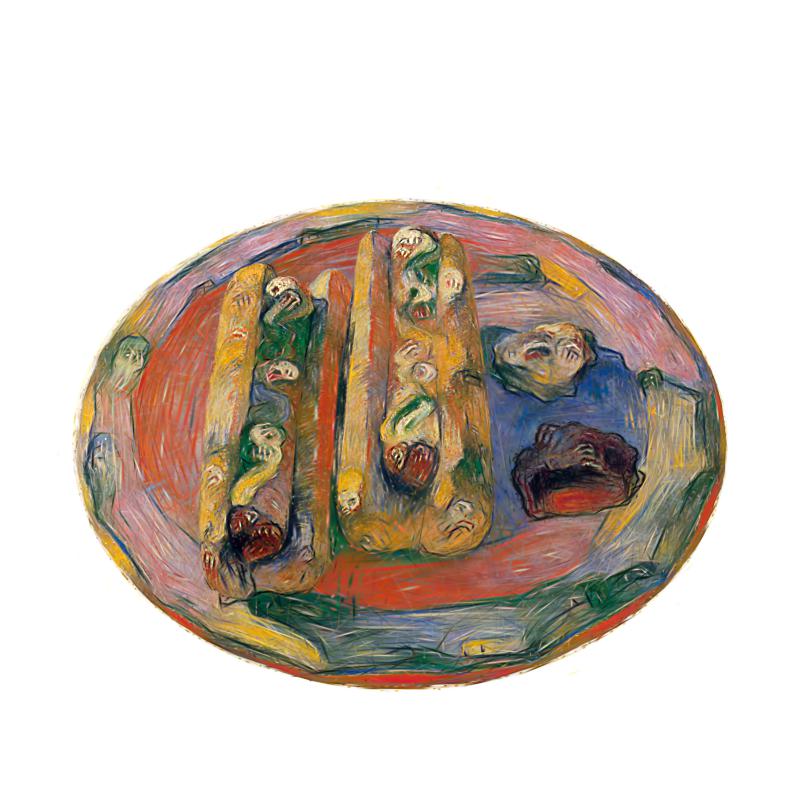}  &
 \includegraphics[trim={50 100 50 190},clip, width=0.18\textwidth,valign=c]{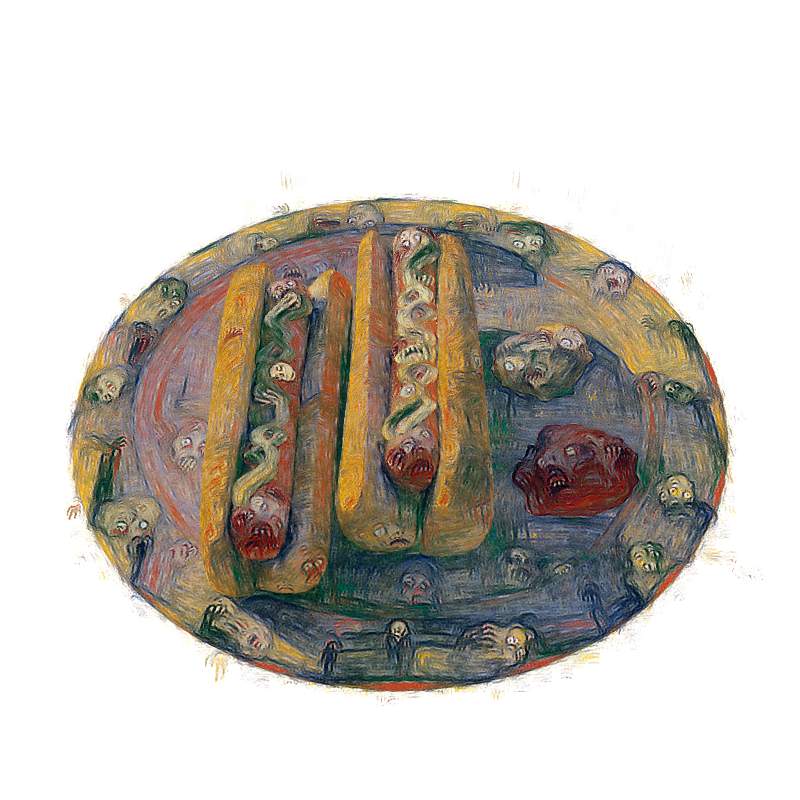}  \\
\end{tabular}
\caption{Full comparison of \our{} (our) and baseline models in 3D style transfer, conditioned by image, on \textit{hotdog} and \textit{lego} objects from NeRF-Synthetic dataset \cite{mildenhall2020nerf}.}
\label{fig:comp_image_obj}
\end{figure*}

As described in Section~\ref{3d_scenes_exp}, we evaluate \our{} on the NeRF-Synthetic dataset~\cite{mildenhall2020nerf} and the Mip-NeRF 360 dataset~\cite{barron2022mipnerf360}. Figure~\ref{fig:comp_text_obj} presents comparisons for text conditioned 3D style transfer on the \textit{lego} and \textit{hotdog} scenes from the NeRF-Synthetic dataset~\cite{mildenhall2020nerf}, comparing against Instruct-GS2GS~\cite{haque2023instruct}, DGE~\cite{chen2024dge}, and CLIPGaussian~\cite{howil2026clipgaussian}. Figure~\ref{fig:comp_image_obj} shows the corresponding image conditioned comparisons, where we compare against StyleGaussian~\cite{liu2024stylegaussian}, SGSST~\cite{galerne2025sgsst}, ABC-GS~\cite{liu2025abc}, and G-Style~\cite{kovacs2024G}.

\subsubsection{3D Ablation Studies}

We conduct ablation studies on our 3D style transfer framework using the same scenes and prompts from the main experiments. The reported values are averages across all scene and prompt combination for each hyperparameter configurations. For each ablation, the complementary hyperparameters are fixed to their middle values from the main experimental grid.

Table~\ref{tab:ablation_lambda_conditioned} evaluates the effect of the directional loss weight $\lambda_d$ and the patch loss weight $\lambda_p$ for both text and image conditioned style transfer, with $\texttt{patch\_size} = 128$ and $\texttt{patch\_num} = 50$. For text conditioned stylization, the best CLIP-S is achieved at $\lambda_d = 48$ and $\lambda_p = 2160$, while image conditioned stylization performs best at $\lambda_d = 96$ and $\lambda_p = 2160$.

Table~\ref{tab:ablation_3d_lambda_conditioned} ablates the patch size $\texttt{patch\_size}$ and the number of patches $\texttt{patch\_num}$, with $\lambda_d = 48$ and $\lambda_p = 1080$. For both text and image conditioned style transfer, $\texttt{patch\_size} = 256$ and $\texttt{patch\_num} = 100$ yield the best overall performance.

\begin{figure*}[ht]
\centering
\begin{tabular}{l l c@{}c c@{}c c@{}c}
\multicolumn{8}{c}{\texttt{patch\_size}} \\
& & \multicolumn{2}{c}{64} & \multicolumn{2}{c}{128} & \multicolumn{2}{c}{256} \\
\multirow{3}{*}{lego} & 25 & 
  \includegraphics[trim={50 100 50 100},clip, width=0.13\textwidth]{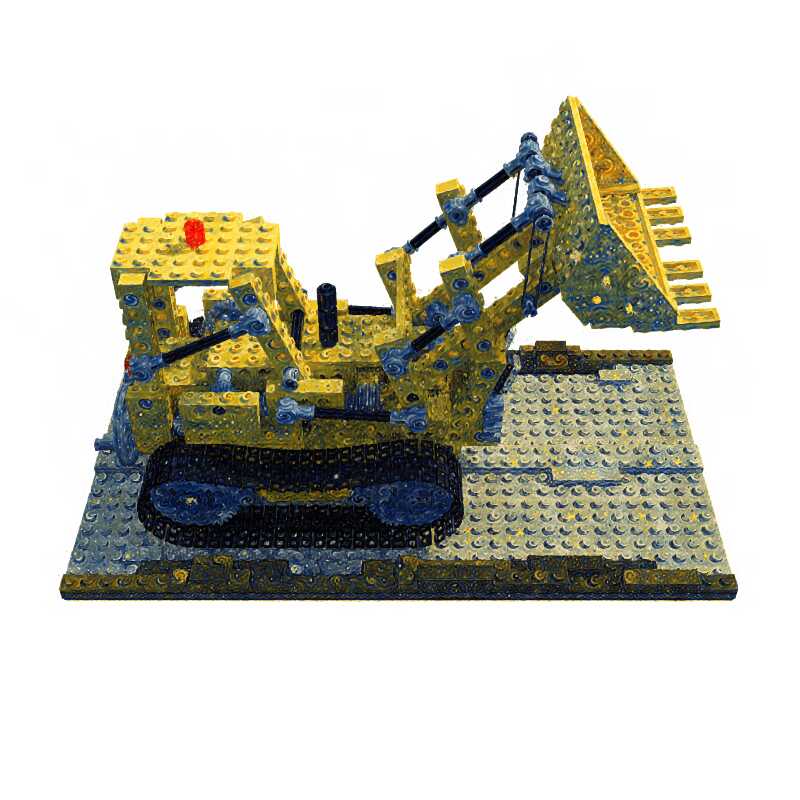} &
  \includegraphics[trim={50 100 50 100},clip, width=0.13\textwidth]{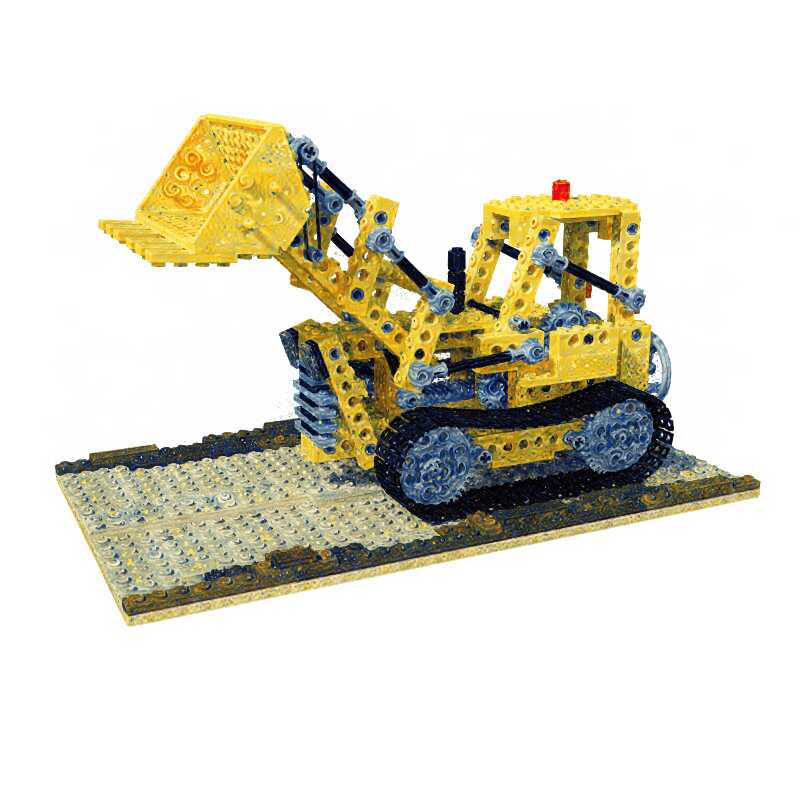} &
  \includegraphics[trim={50 100 50 100},clip, width=0.13\textwidth]{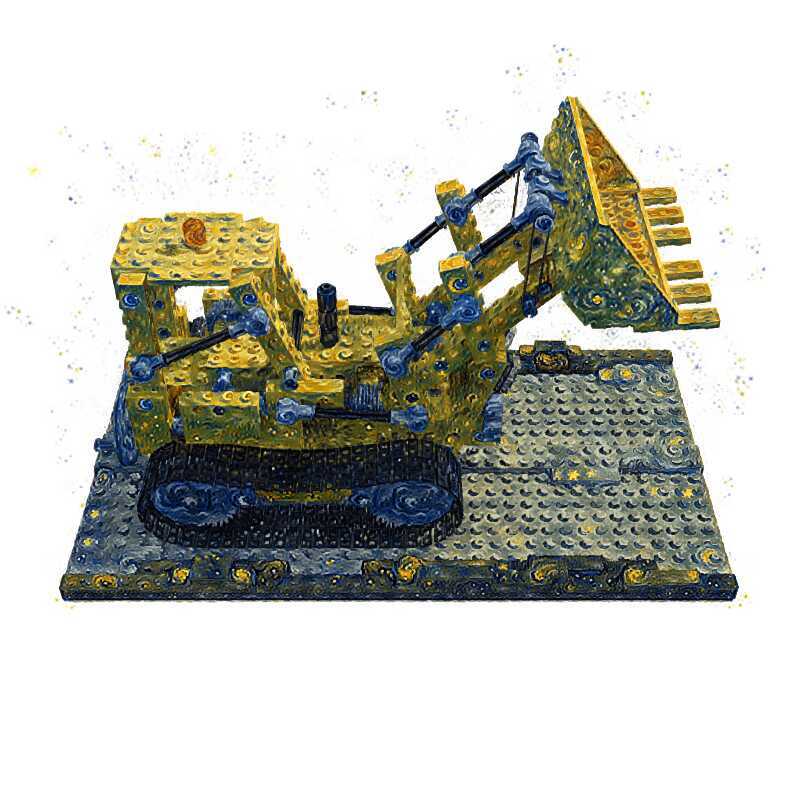} &
  \includegraphics[trim={50 100 50 100},clip, width=0.13\textwidth]{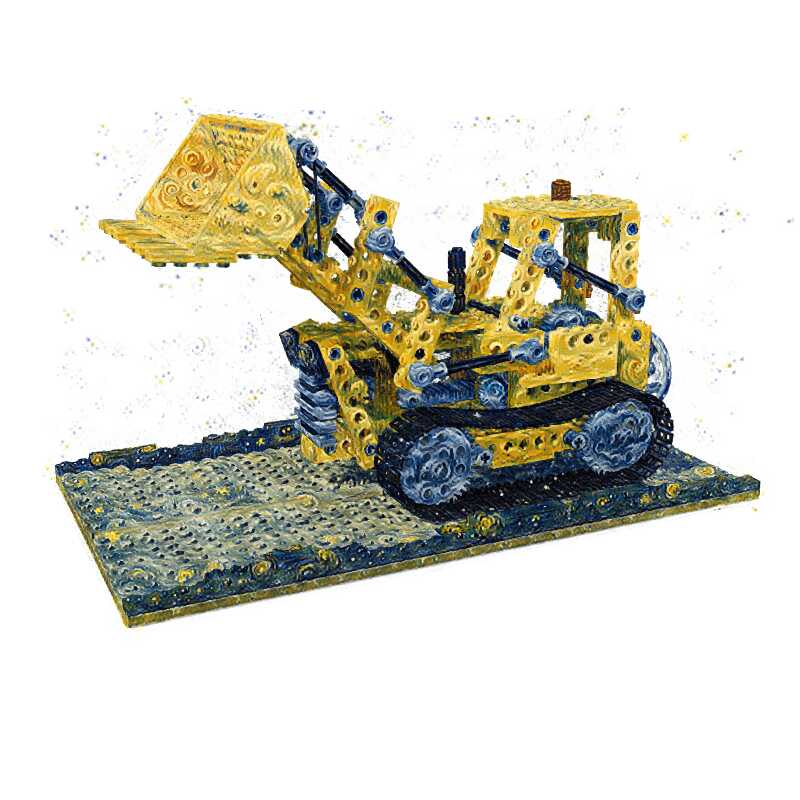} &
  \includegraphics[trim={50 100 50 100},clip, width=0.13\textwidth]{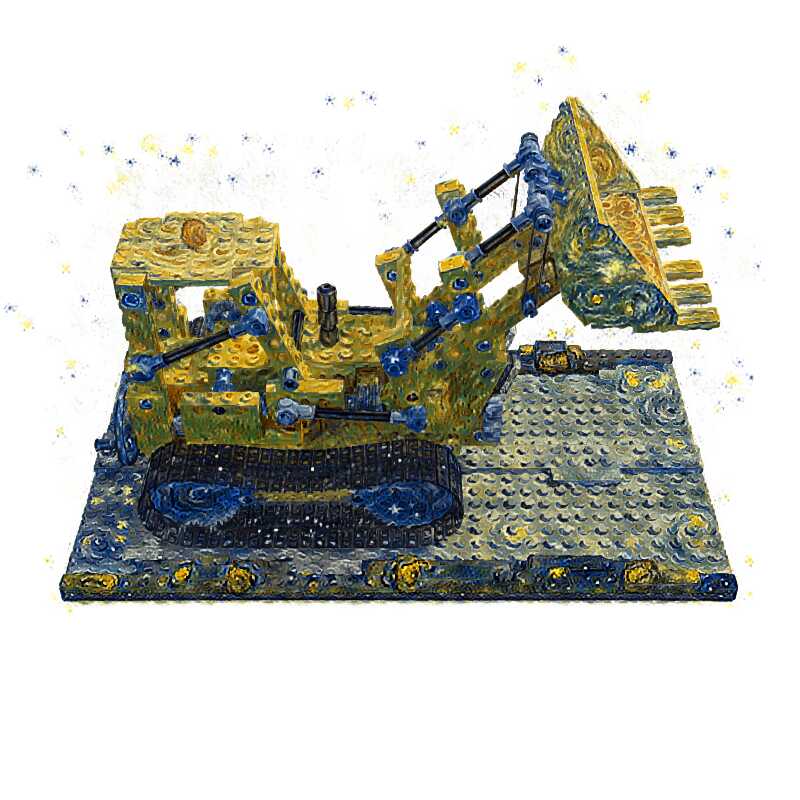} &
  \includegraphics[trim={50 100 50 100},clip, width=0.13\textwidth]{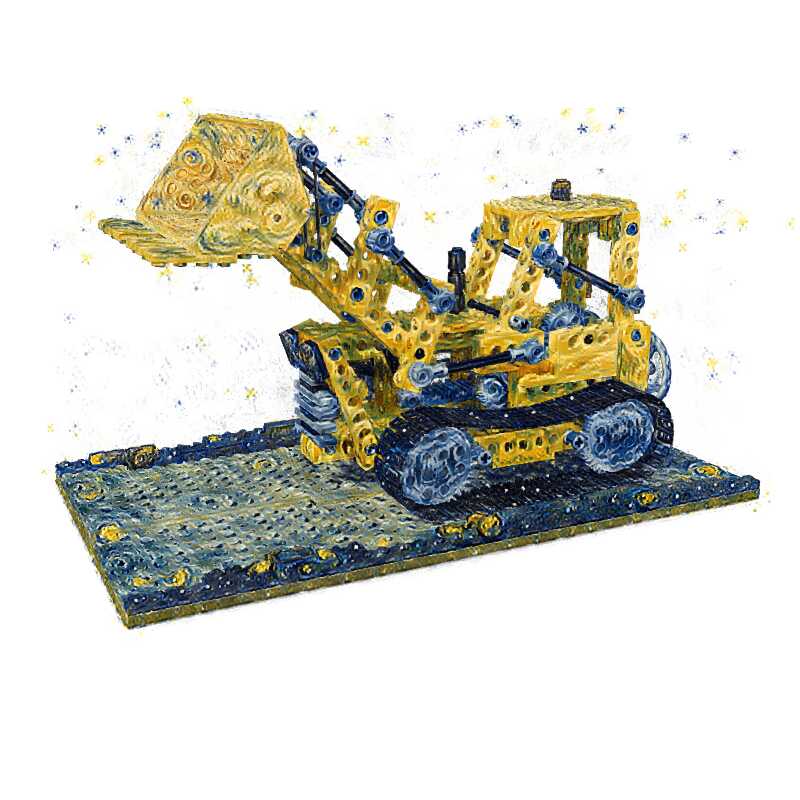} \\
 & 50 & 
  \includegraphics[trim={50 100 50 100},clip, width=0.13\textwidth]{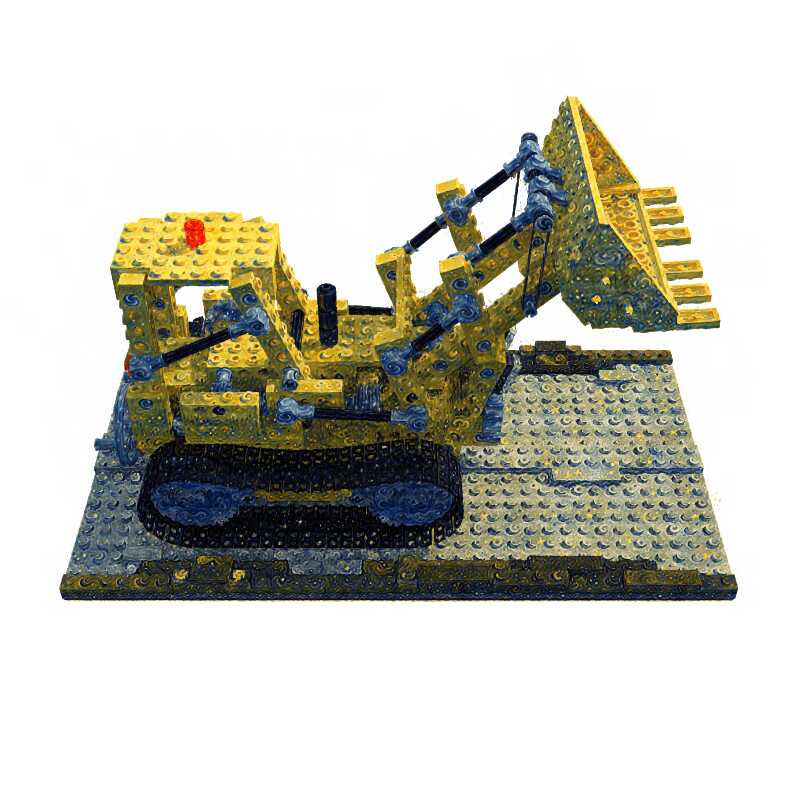} &
  \includegraphics[trim={50 100 50 100},clip, width=0.13\textwidth]{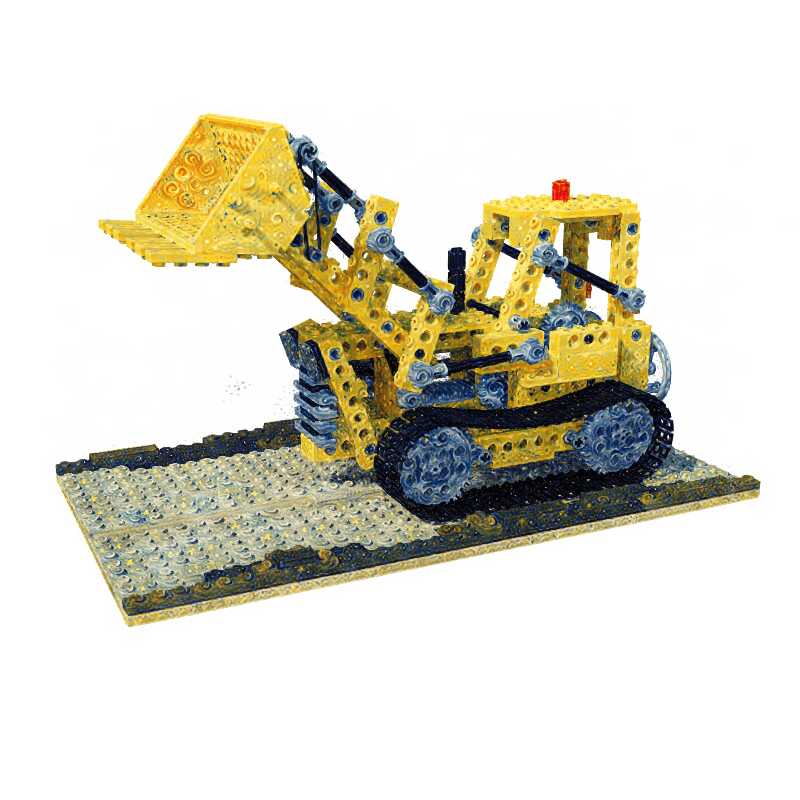} &
  \includegraphics[trim={50 100 50 100},clip, width=0.13\textwidth]{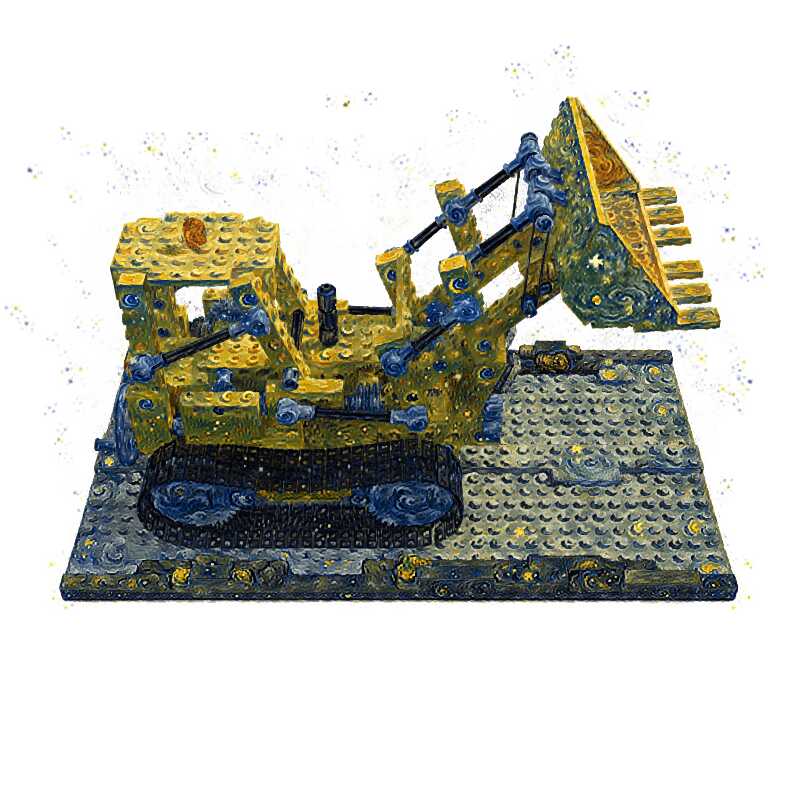} &
  \includegraphics[trim={50 100 50 100},clip, width=0.13\textwidth]{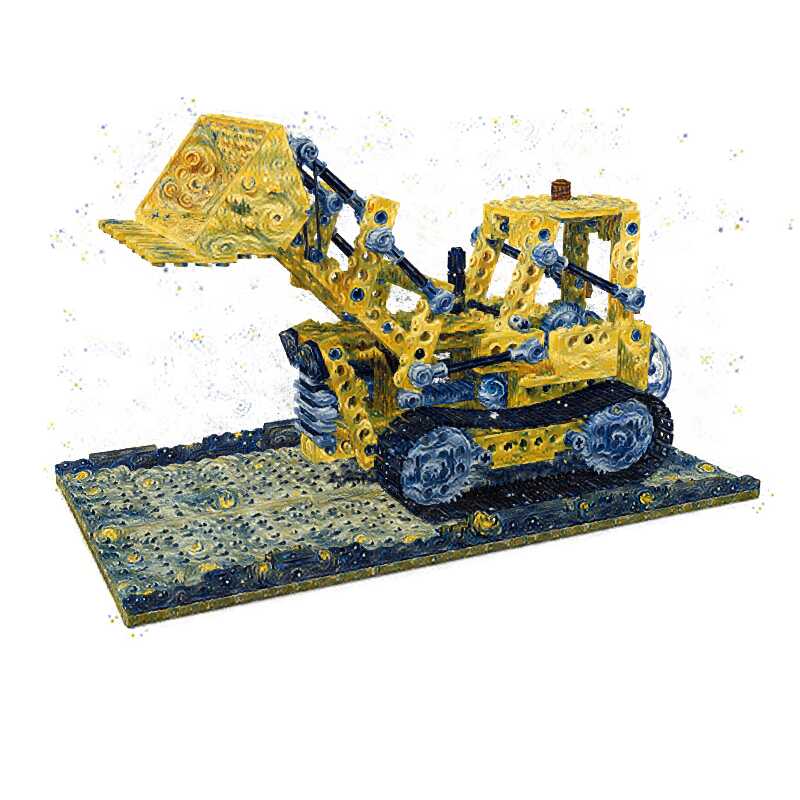} &
  \includegraphics[trim={50 100 50 100},clip, width=0.13\textwidth]{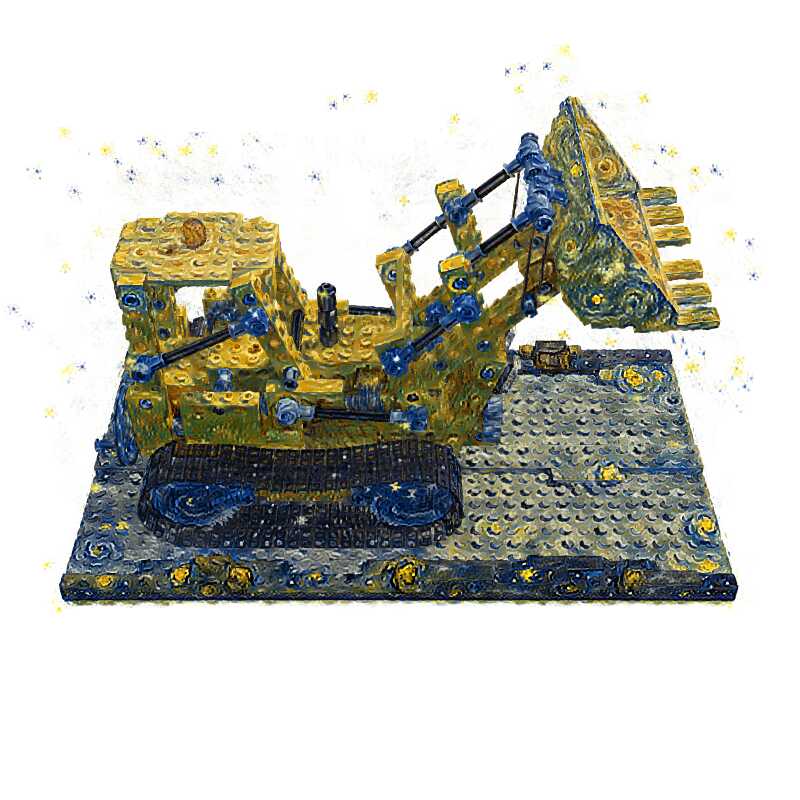} &
  \includegraphics[trim={50 100 50 100},clip, width=0.13\textwidth]{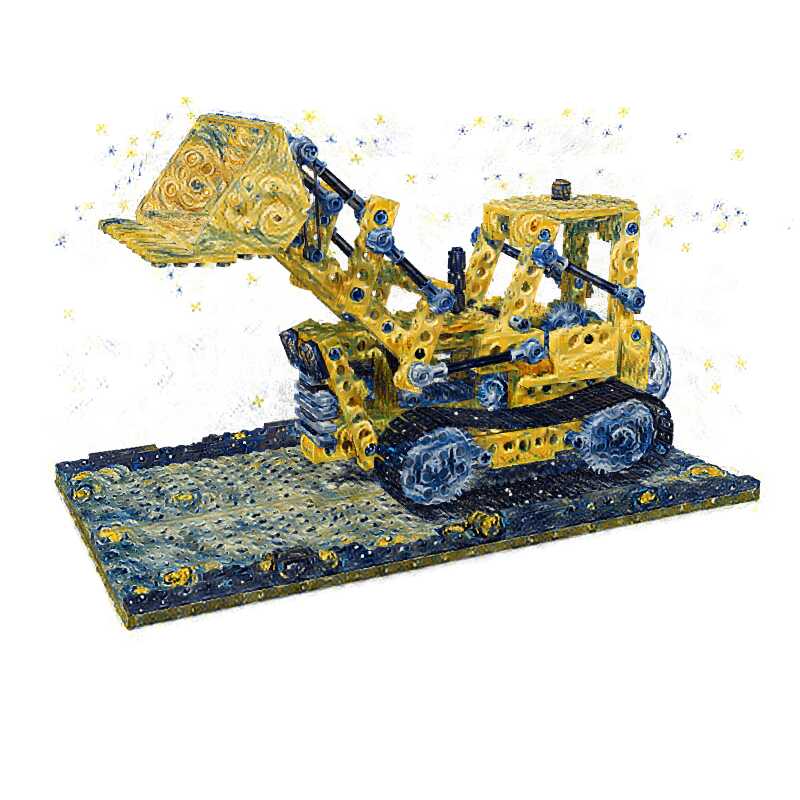} \\
 & 100 & 
  \includegraphics[trim={50 100 50 100},clip, width=0.13\textwidth]{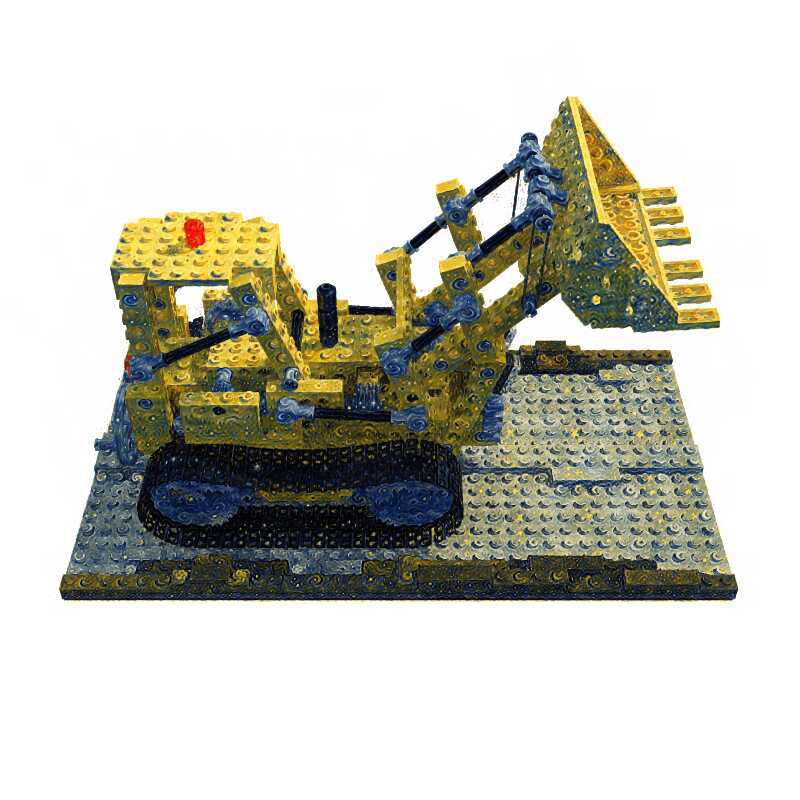} &
  \includegraphics[trim={50 100 50 100},clip, width=0.13\textwidth]{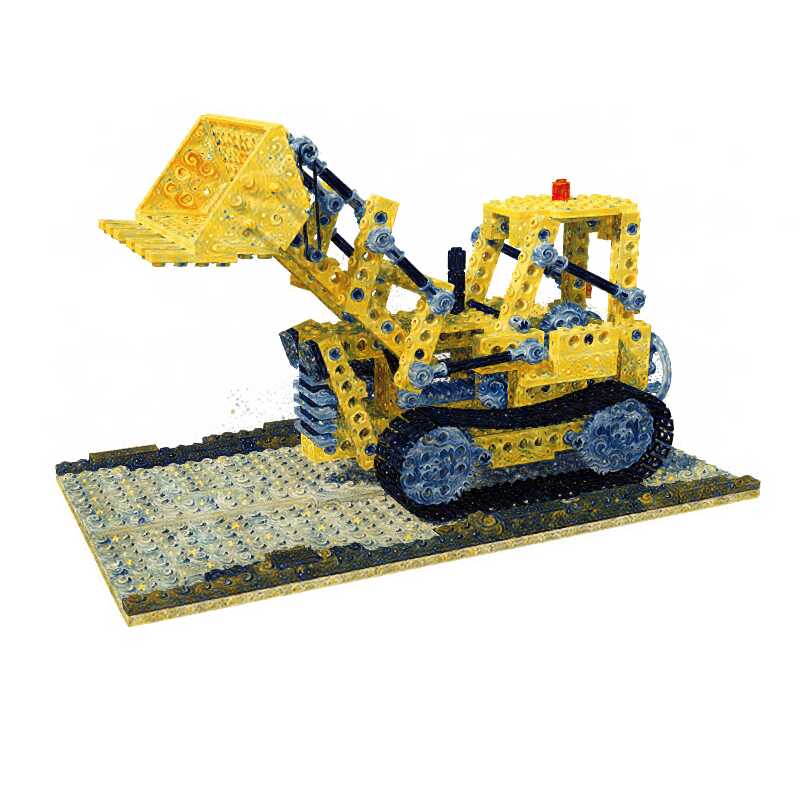} &
  \includegraphics[trim={50 100 50 100},clip, width=0.13\textwidth]{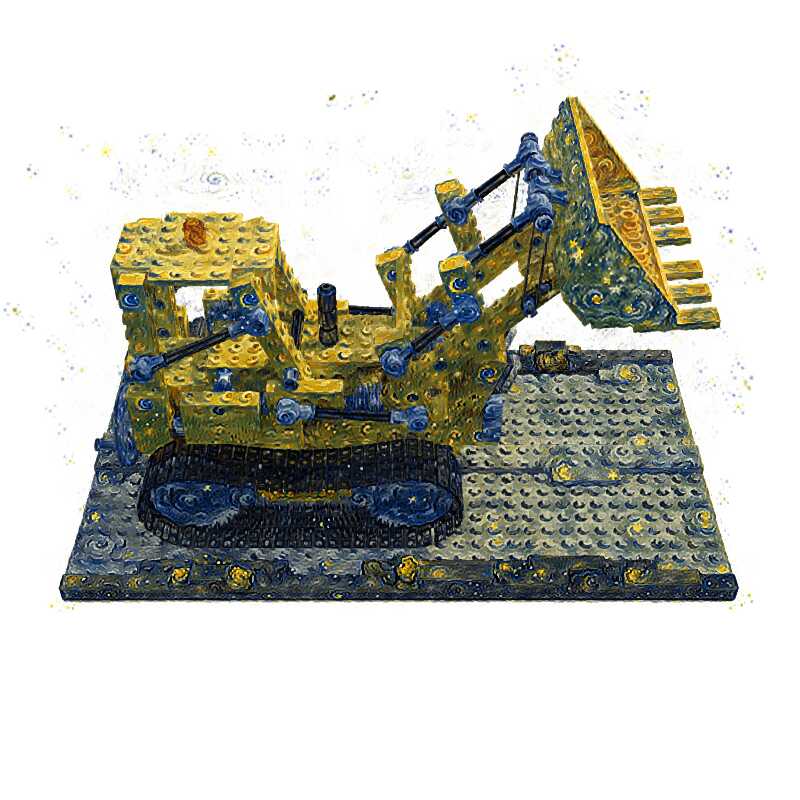} &
  \includegraphics[trim={50 100 50 100},clip, width=0.13\textwidth]{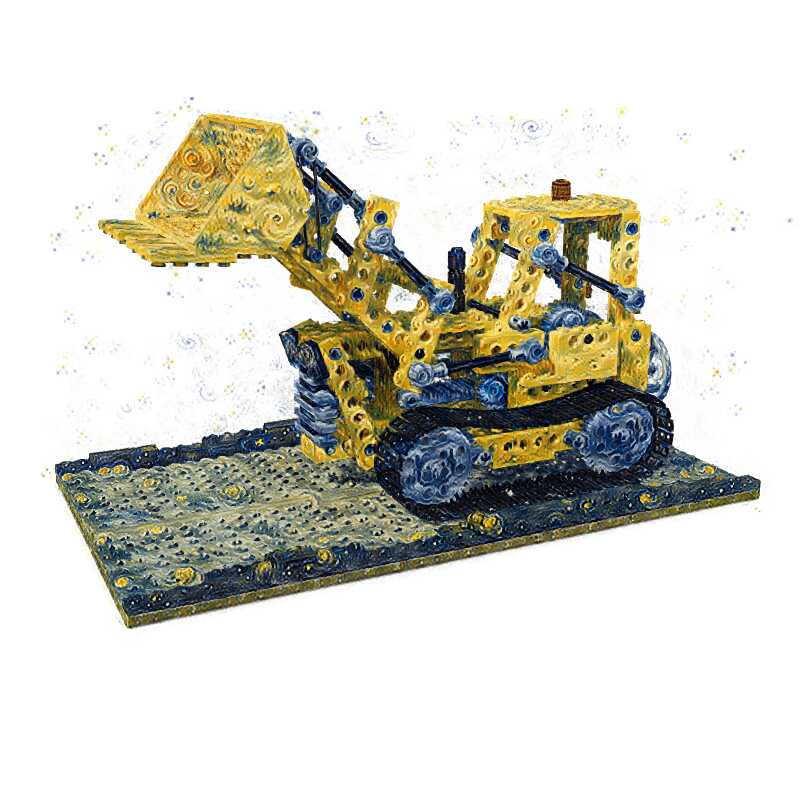} &
  \includegraphics[trim={50 100 50 100},clip, width=0.13\textwidth]{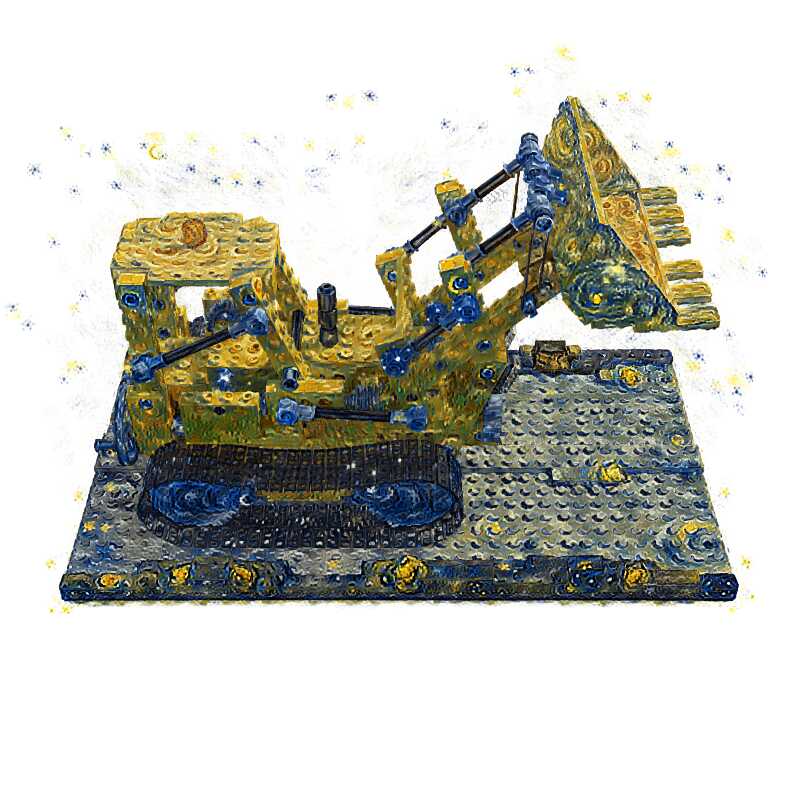} &
  \includegraphics[trim={50 100 50 100},clip, width=0.13\textwidth]{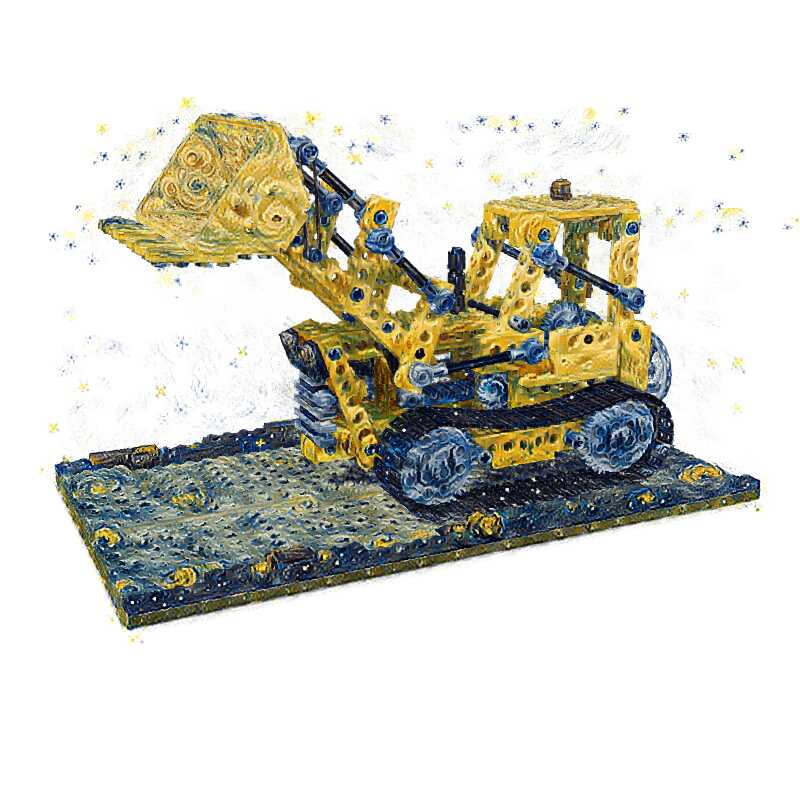} \\
\multirow{3}{*}{hotdog} & 25 & 
  \includegraphics[trim={50 100 50 100},clip, width=0.13\textwidth]{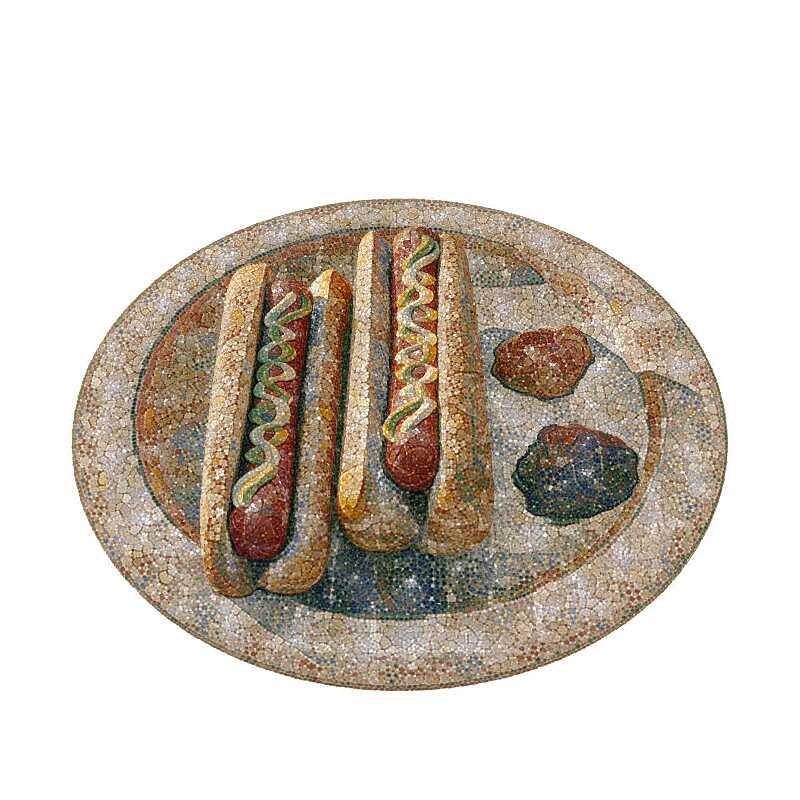} &
  \includegraphics[trim={50 100 50 100},clip, width=0.13\textwidth]{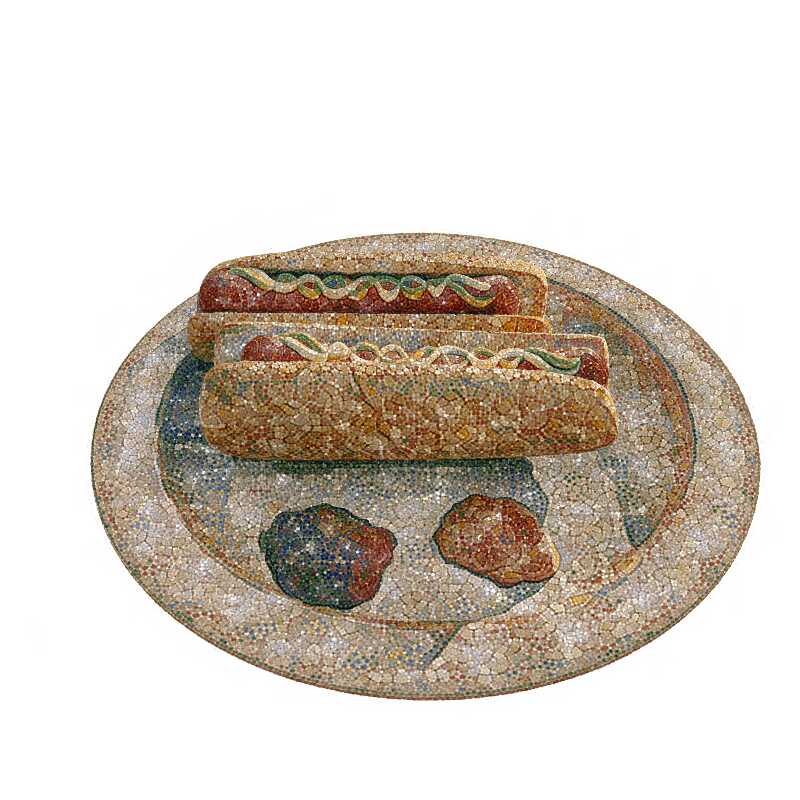} &
  \includegraphics[trim={50 100 50 100},clip, width=0.13\textwidth]{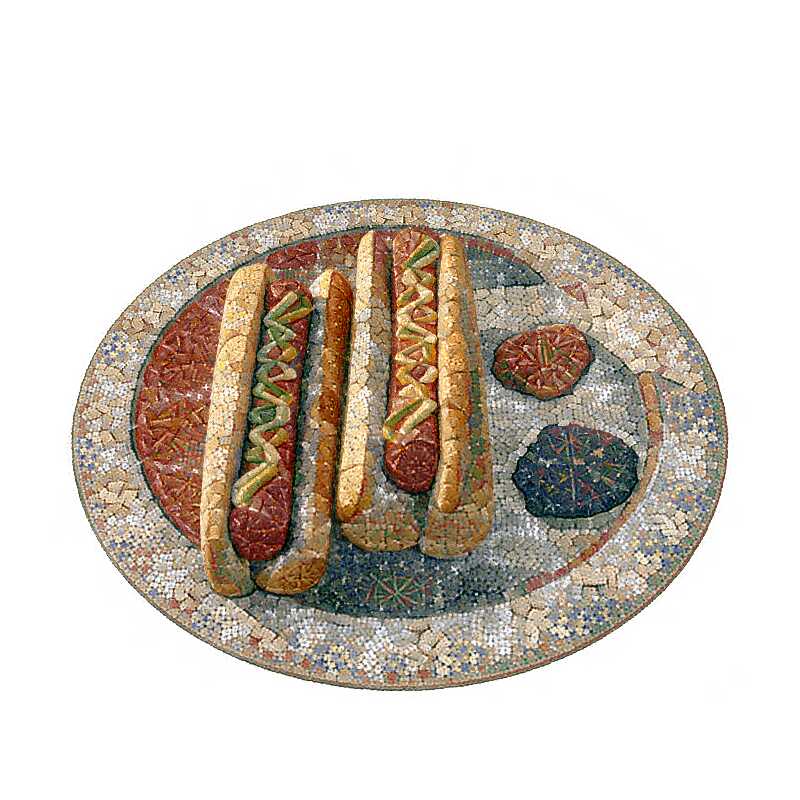} &
  \includegraphics[trim={50 100 50 100},clip, width=0.13\textwidth]{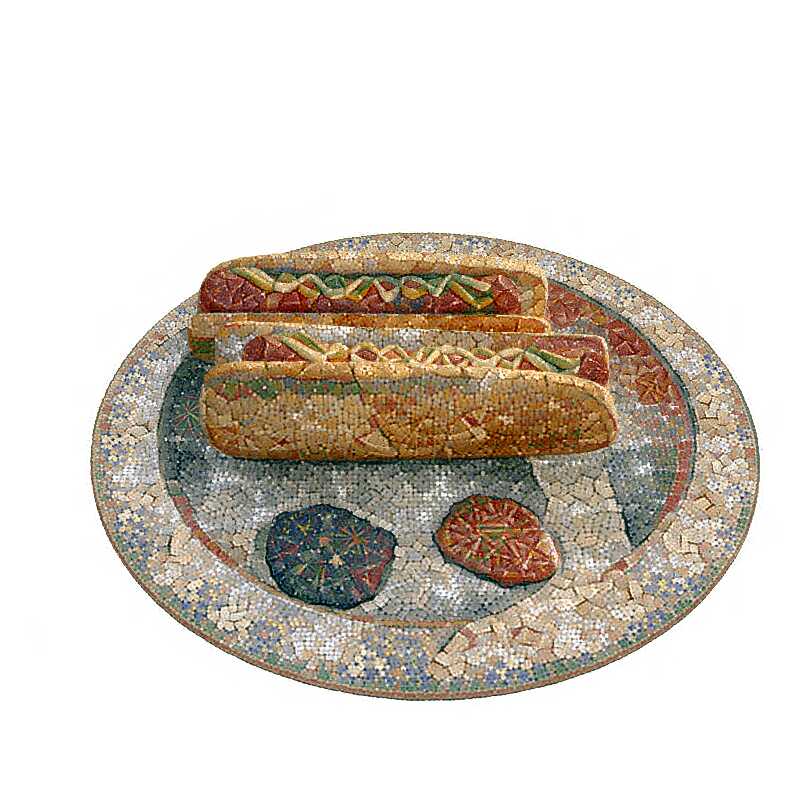} &
  \includegraphics[trim={50 100 50 100},clip, width=0.13\textwidth]{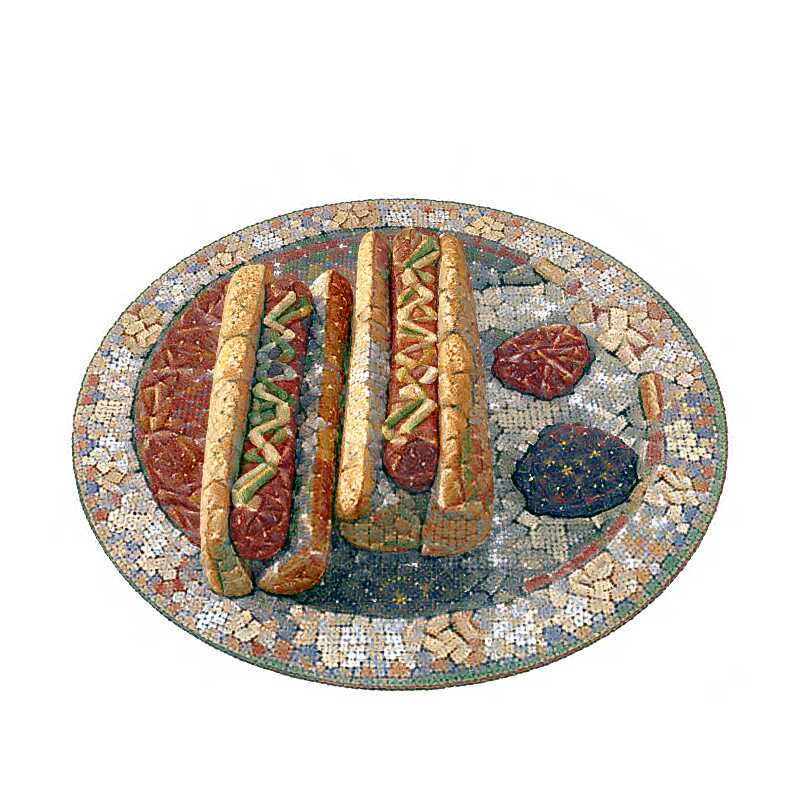} &
  \includegraphics[trim={50 100 50 100},clip, width=0.13\textwidth]{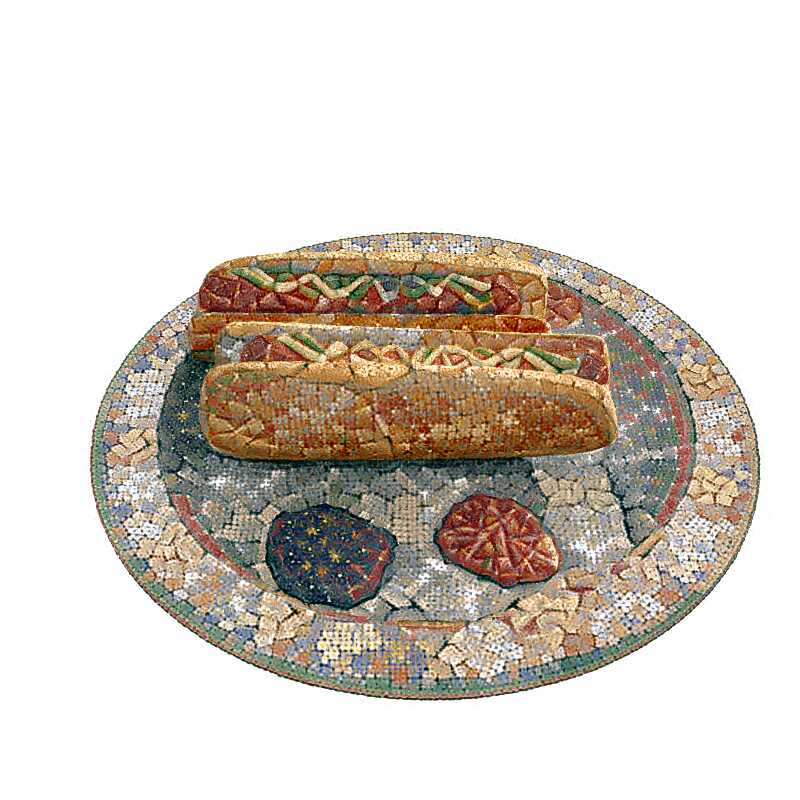} \\
 & 50 & 
  \includegraphics[trim={50 100 50 100},clip, width=0.13\textwidth]{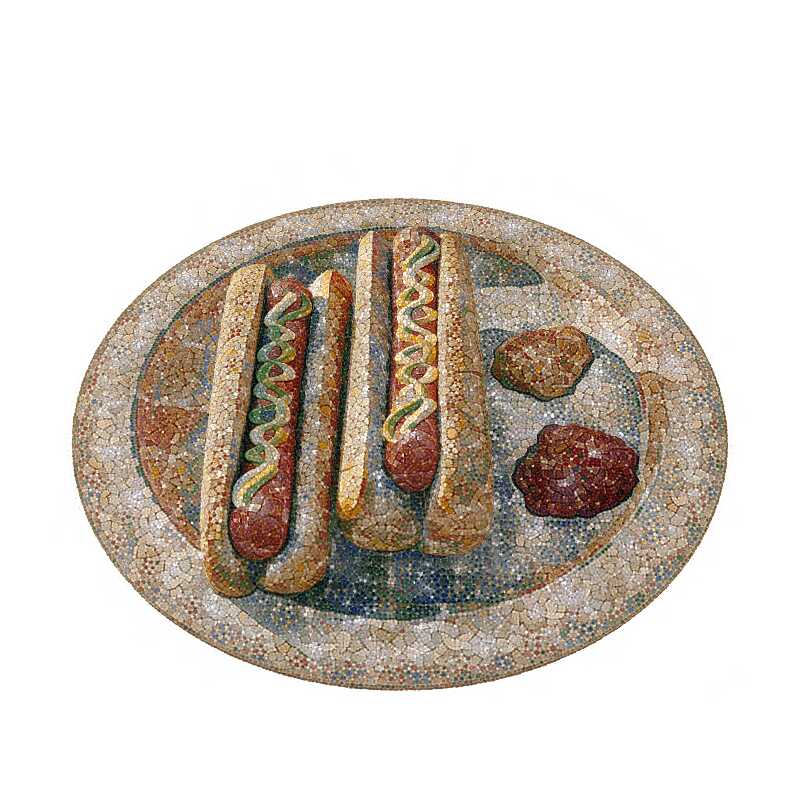} &
  \includegraphics[trim={50 100 50 100},clip, width=0.13\textwidth]{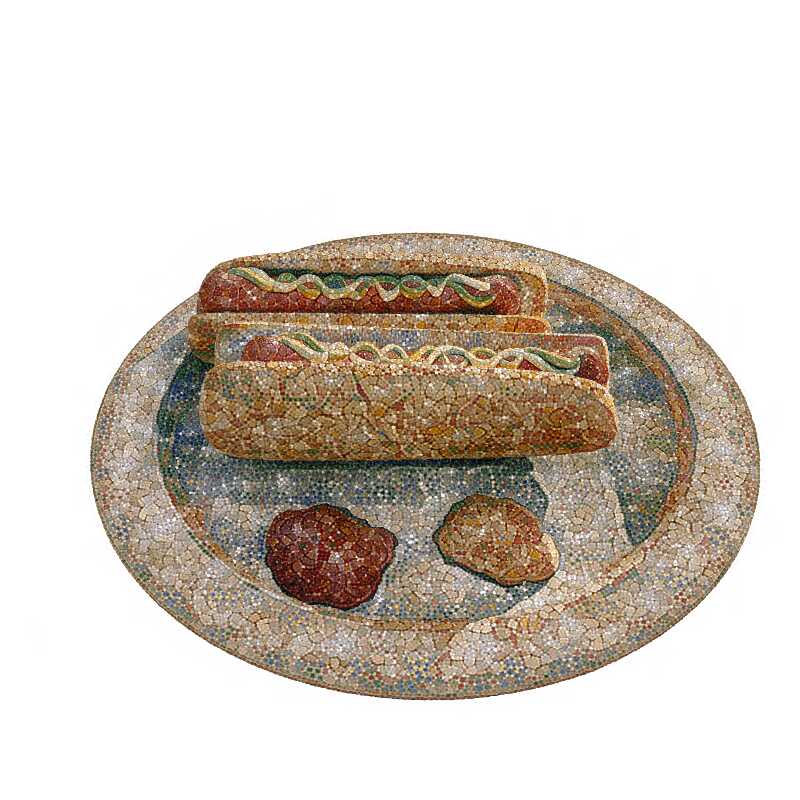} &
  \includegraphics[trim={50 100 50 100},clip, width=0.13\textwidth]{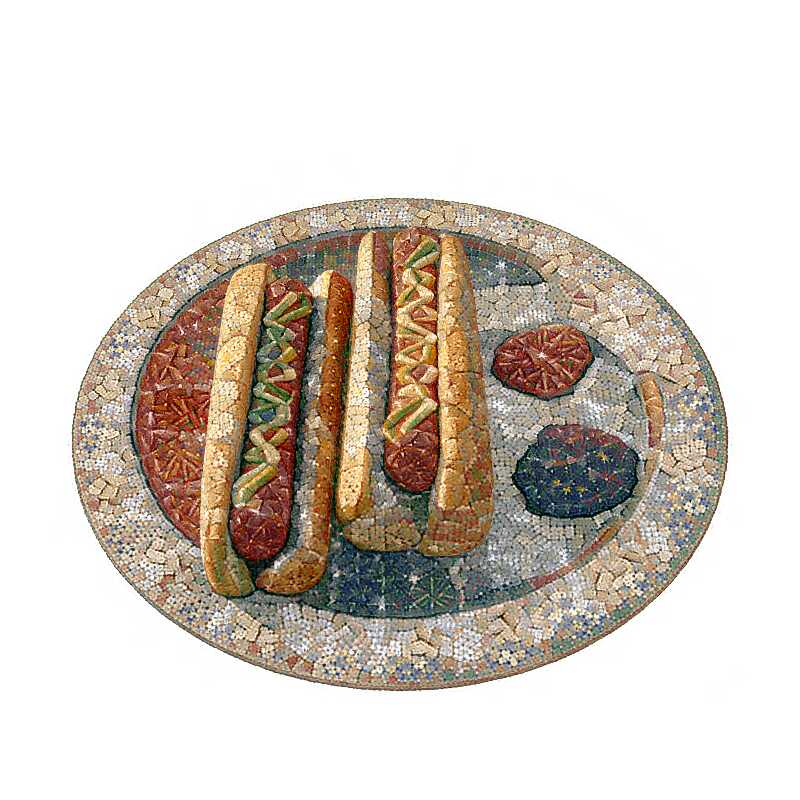} &
  \includegraphics[trim={50 100 50 100},clip, width=0.13\textwidth]{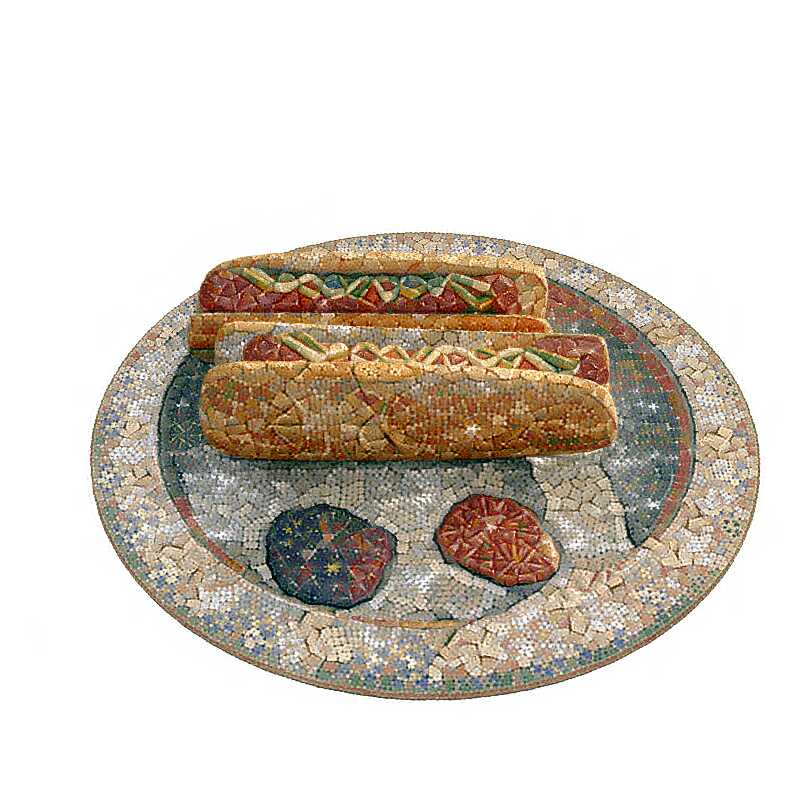} &
  \includegraphics[trim={50 100 50 100},clip, width=0.13\textwidth]{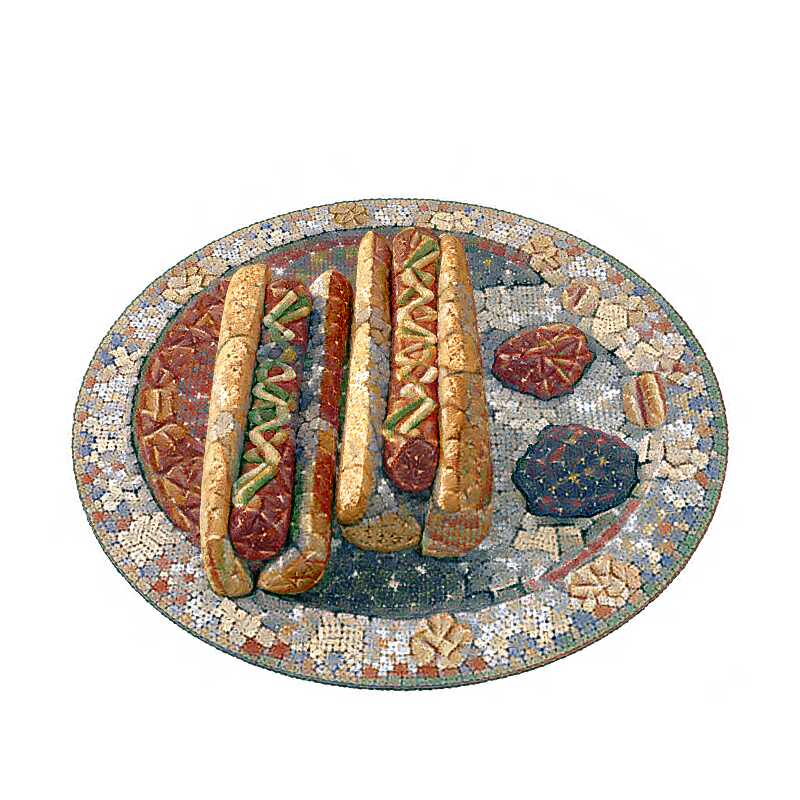} &
  \includegraphics[trim={50 100 50 100},clip, width=0.13\textwidth]{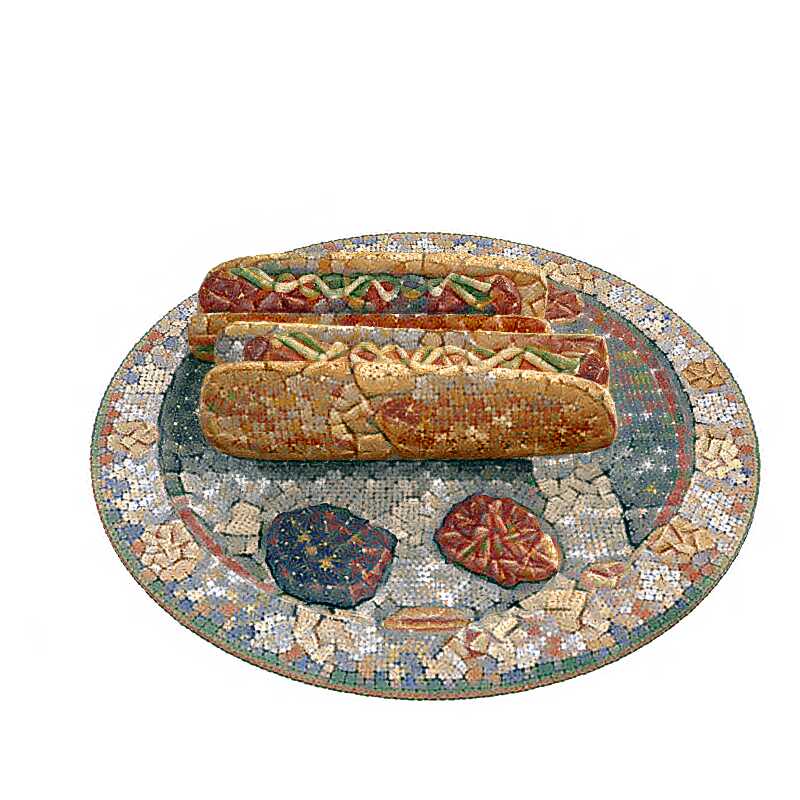} \\
 & 100 & 
  \includegraphics[trim={50 100 50 100},clip, width=0.13\textwidth]{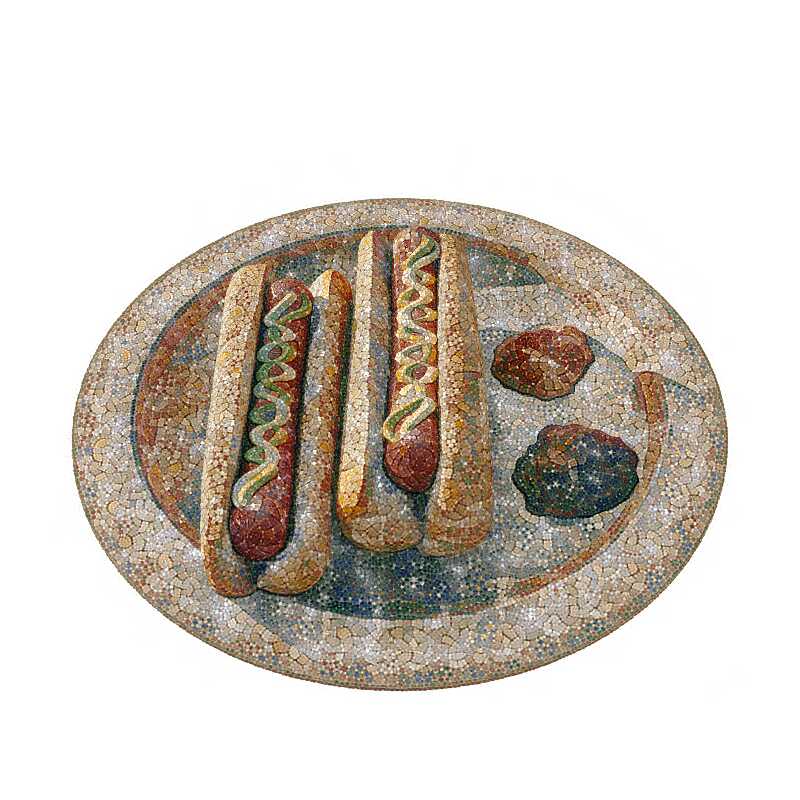} &
  \includegraphics[trim={50 100 50 100},clip, width=0.13\textwidth]{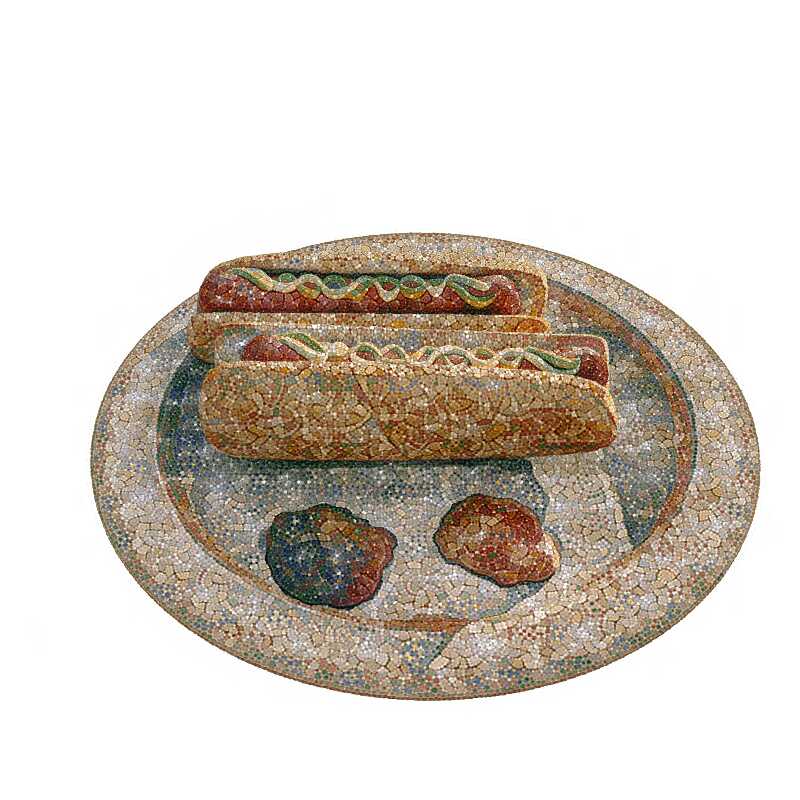} &
  \includegraphics[trim={50 100 50 100},clip, width=0.13\textwidth]{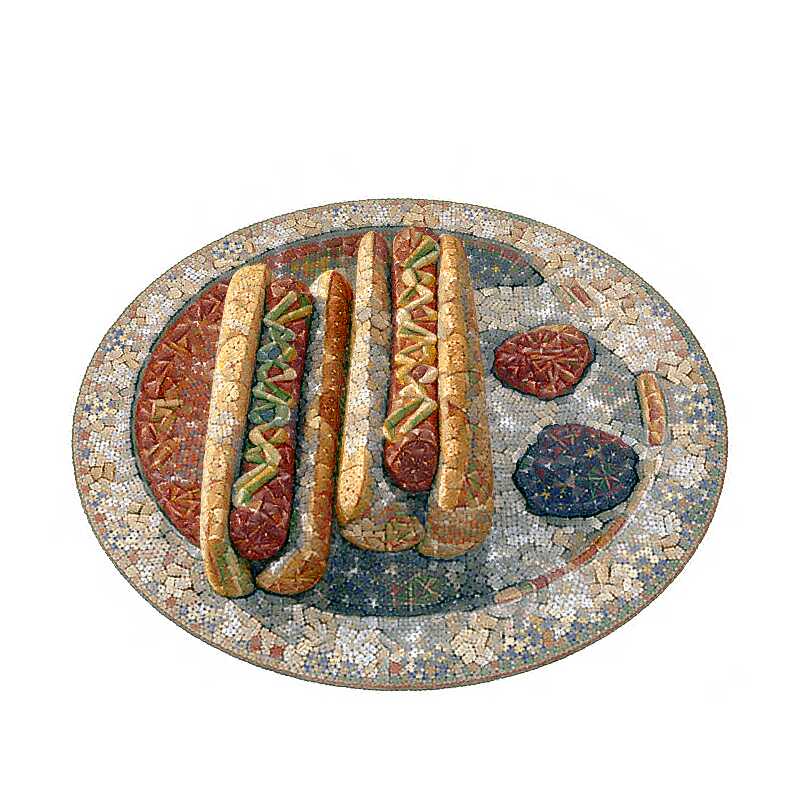} &
  \includegraphics[trim={50 100 50 100},clip, width=0.13\textwidth]{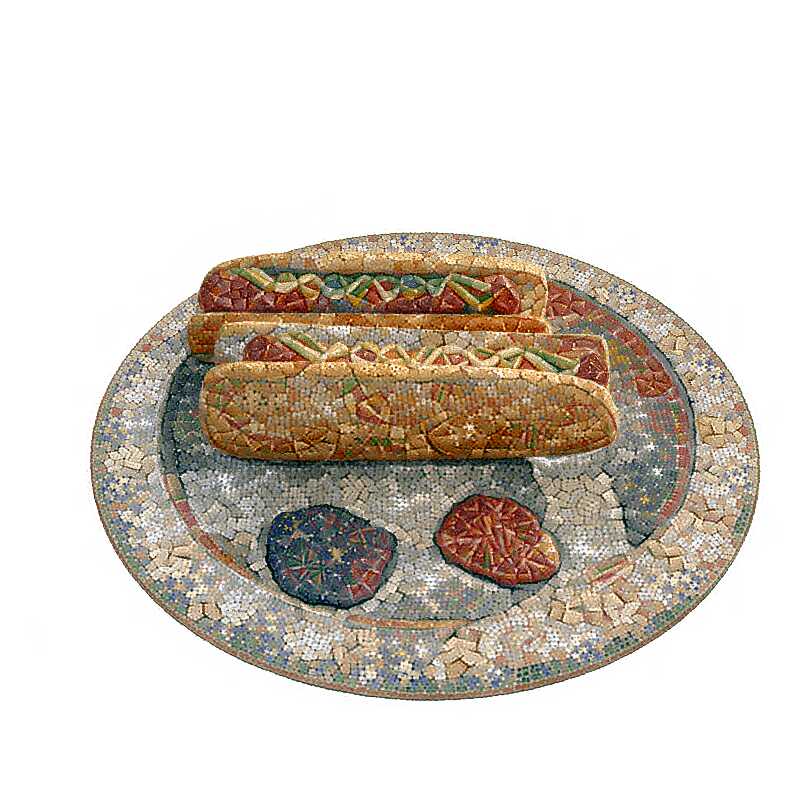} &
  \includegraphics[trim={50 100 50 100},clip, width=0.13\textwidth]{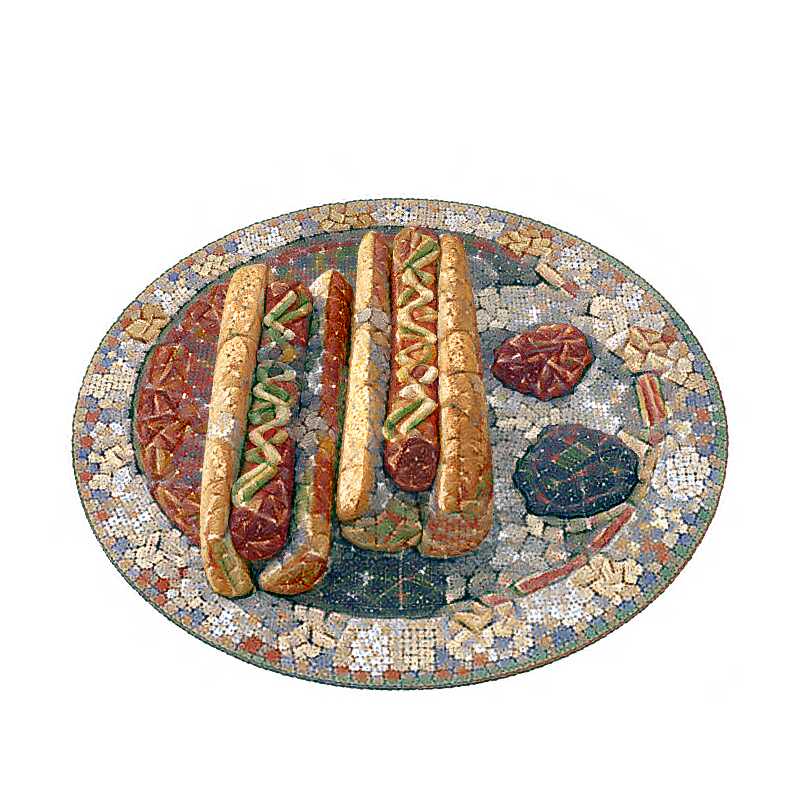} &
  \includegraphics[trim={50 100 50 100},clip, width=0.13\textwidth]{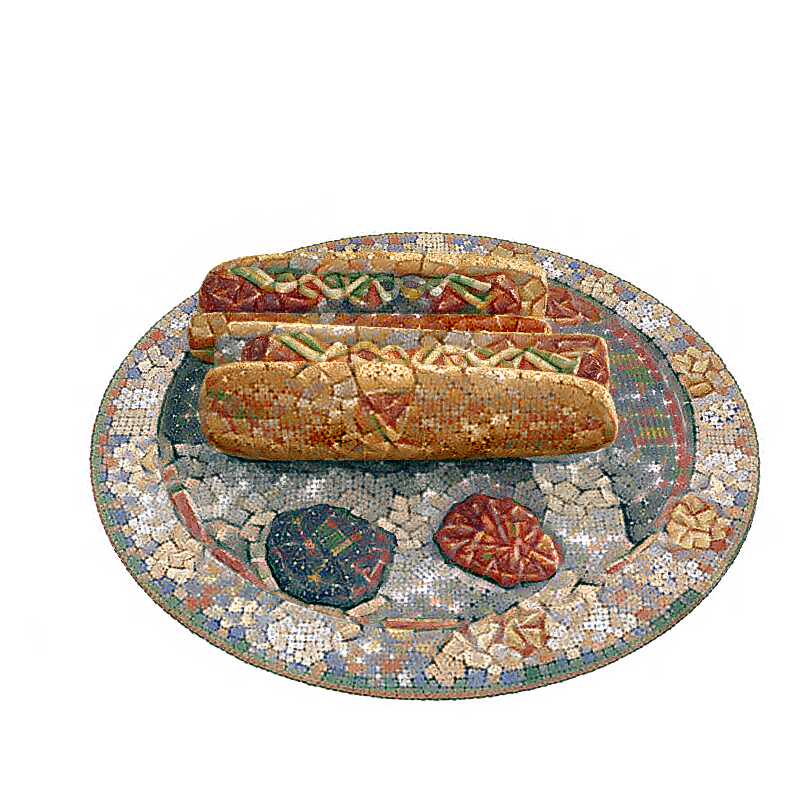} \\
\end{tabular}
\caption{Effect of patch number and patch size on the performance of \our{} on \textit{lego} and \textit{hotdog} objects from NeRF-Synthetic dataset \cite{mildenhall2020nerf}. Objects are stylized with "Starry Night by Vincent van Gogh" and "Mosaic" prompts.}
\label{fig:3D_alb_patch}
\end{figure*}

\begin{figure*}[ht]
\centering
\begin{tabular}{l l c@{}c c@{}c c@{}c}
\multicolumn{8}{c}{$\lambda_{p}$} \\
& & \multicolumn{2}{c}{540} & \multicolumn{2}{c}{1080} & \multicolumn{2}{c}{2160} \\
\multirow{3}{*}{lego} & 24 & 
  \includegraphics[trim={50 100 50 100},clip, width=0.13\textwidth]{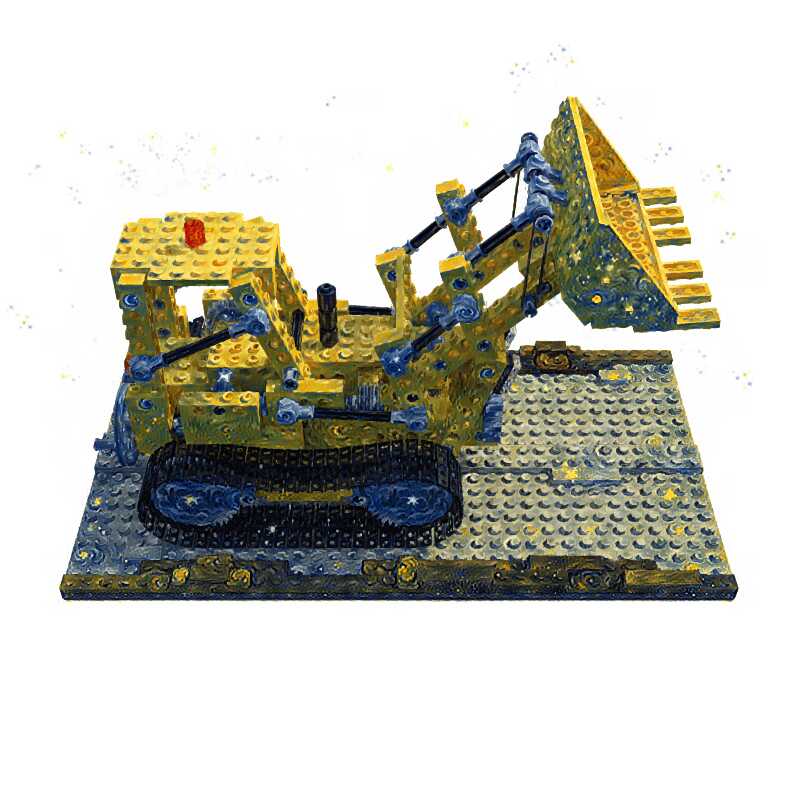} &
  \includegraphics[trim={50 100 50 100},clip, width=0.13\textwidth]{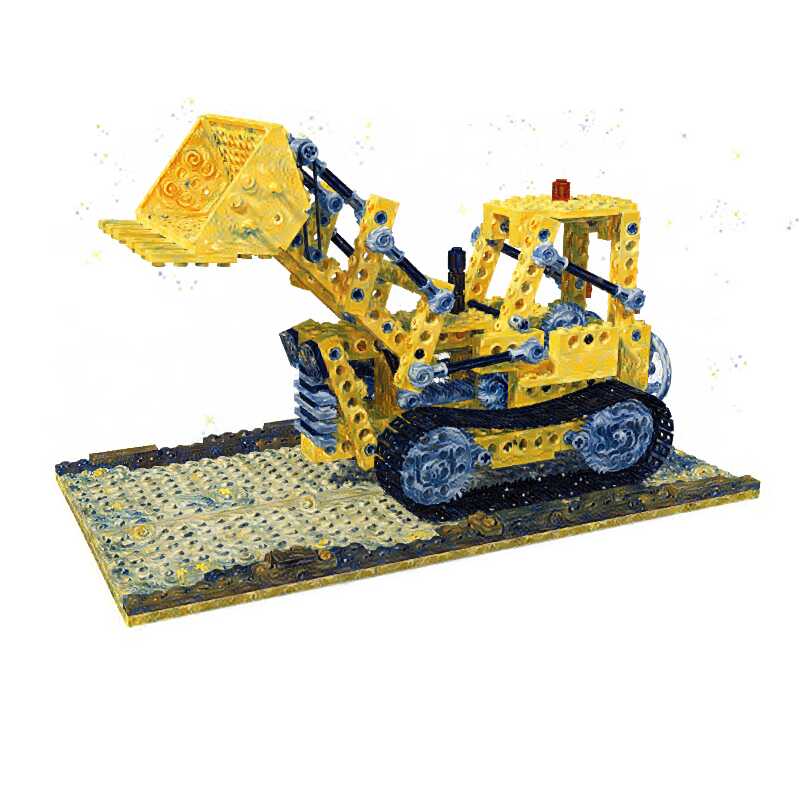} &
  \includegraphics[trim={50 100 50 100},clip, width=0.13\textwidth]{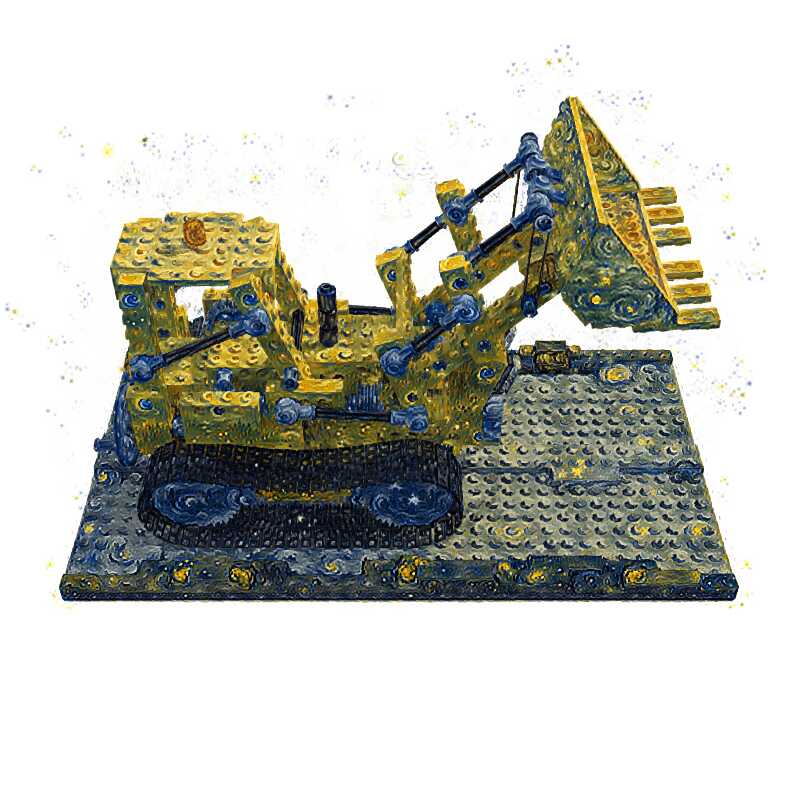} &
  \includegraphics[trim={50 100 50 100},clip, width=0.13\textwidth]{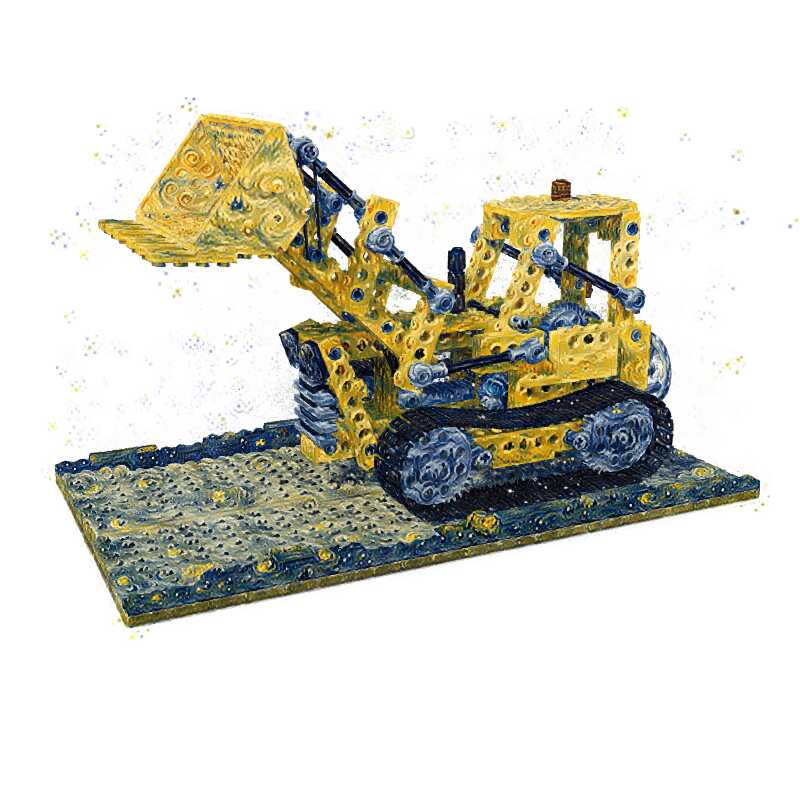} &
  \includegraphics[trim={50 100 50 100},clip, width=0.13\textwidth]{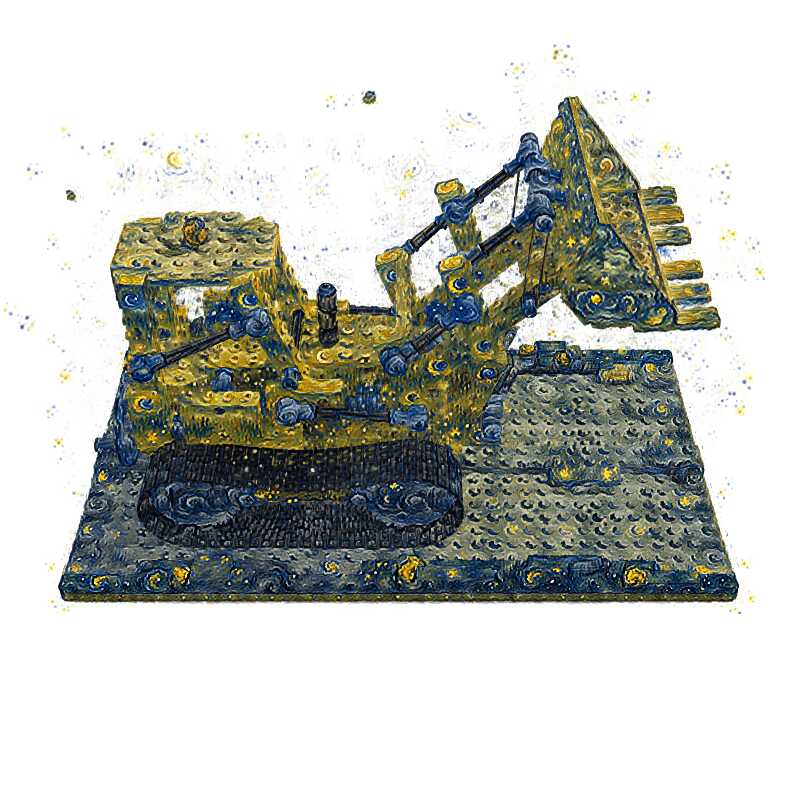} &
  \includegraphics[trim={50 100 50 100},clip, width=0.13\textwidth]{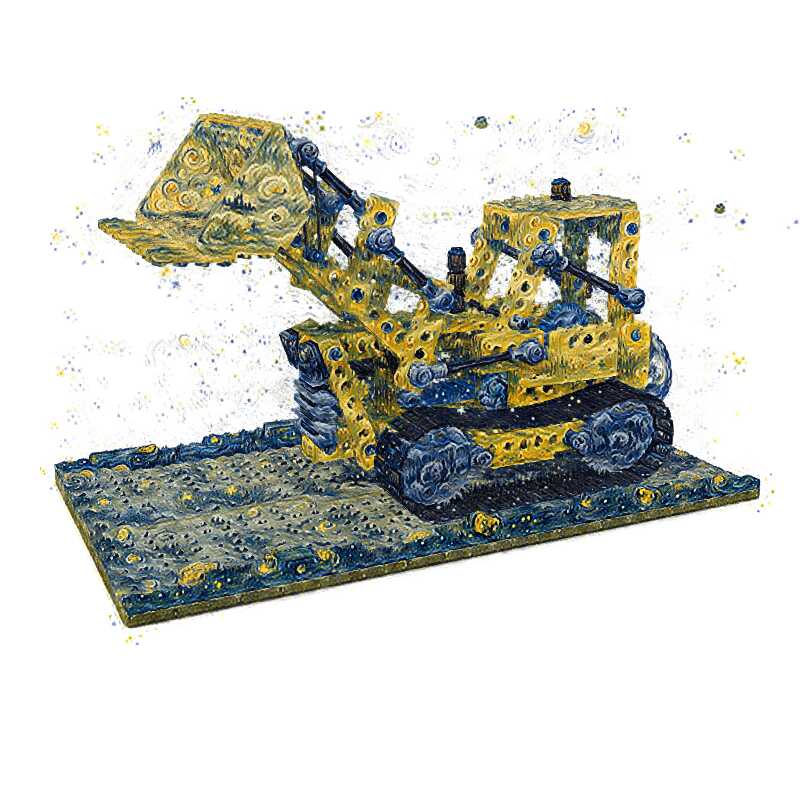} \\
 & 48 & 
  \includegraphics[trim={50 100 50 100},clip, width=0.13\textwidth]{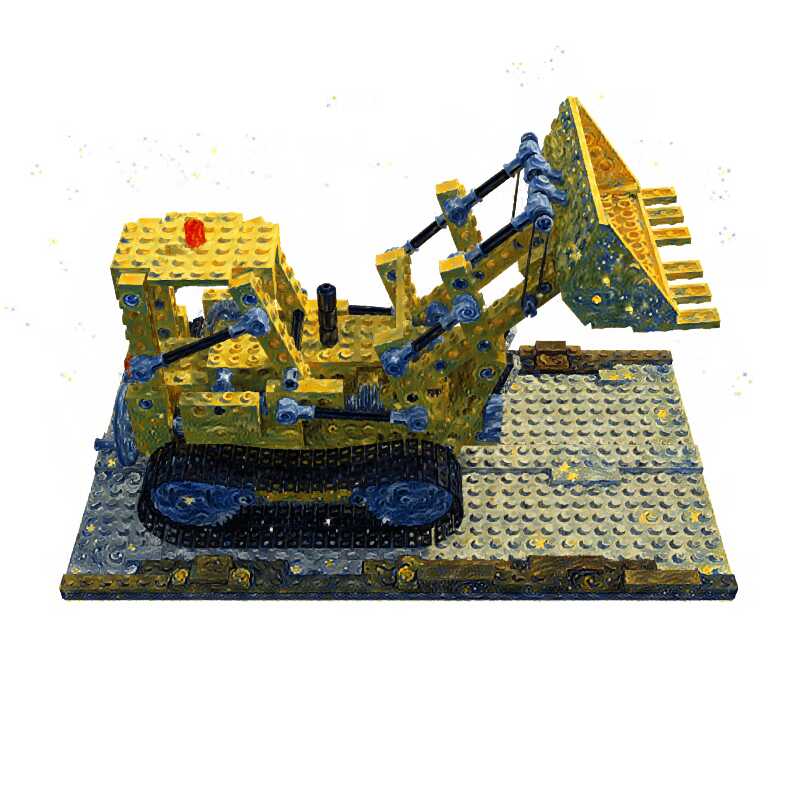} &
  \includegraphics[trim={50 100 50 100},clip, width=0.13\textwidth]{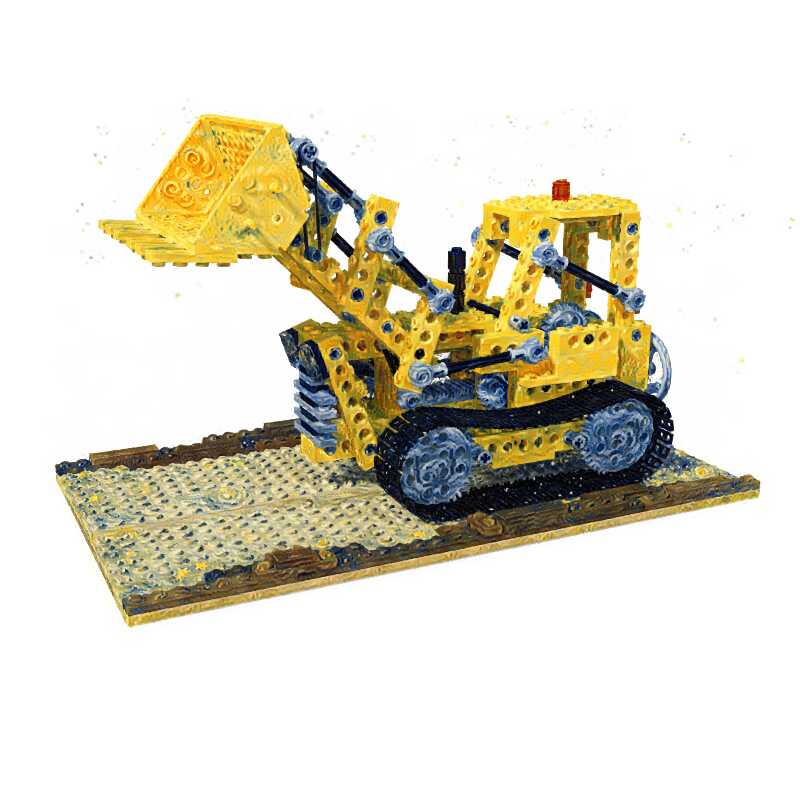} &
  \includegraphics[trim={50 100 50 100},clip, width=0.13\textwidth]{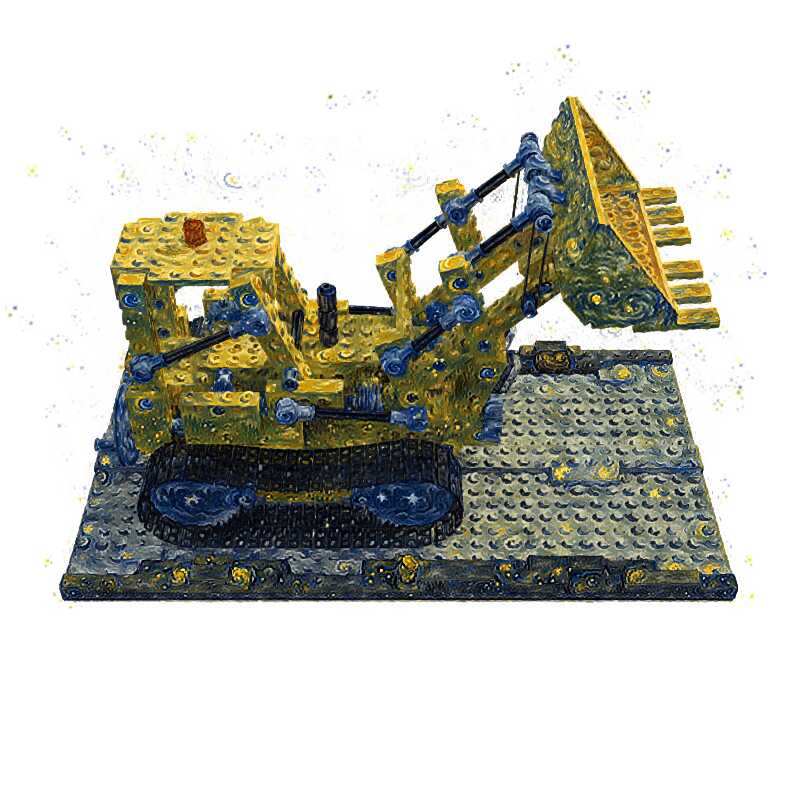} &
  \includegraphics[trim={50 100 50 100},clip, width=0.13\textwidth]{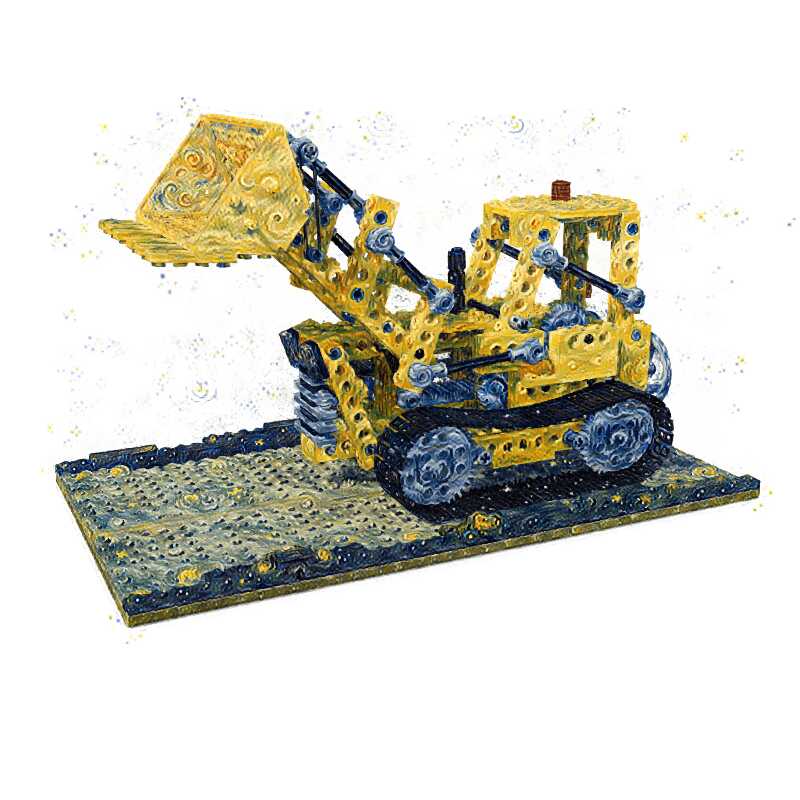} &
  \includegraphics[trim={50 100 50 100},clip, width=0.13\textwidth]{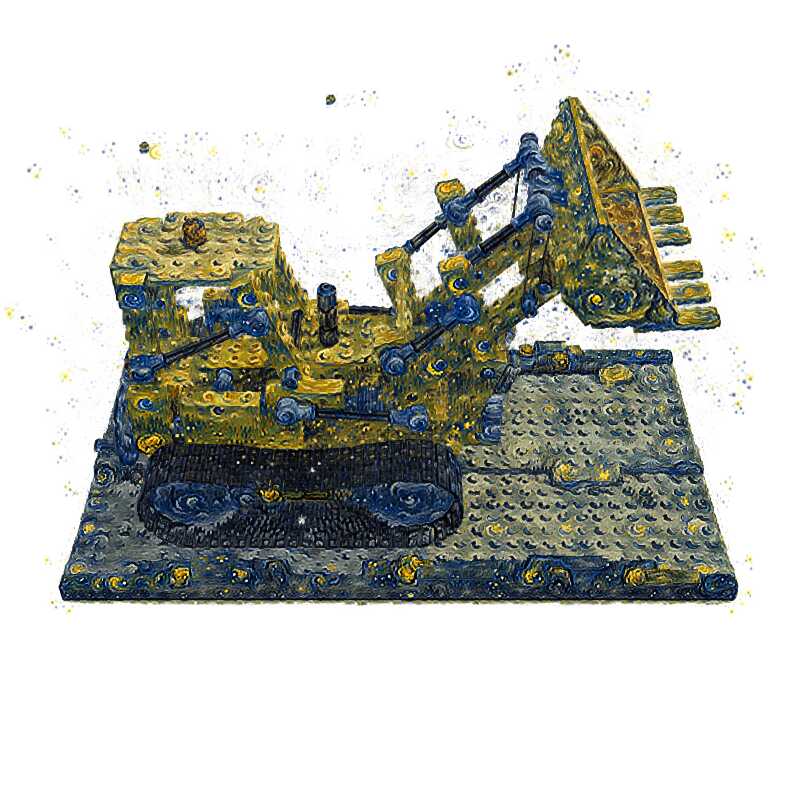} &
  \includegraphics[trim={50 100 50 100},clip, width=0.13\textwidth]{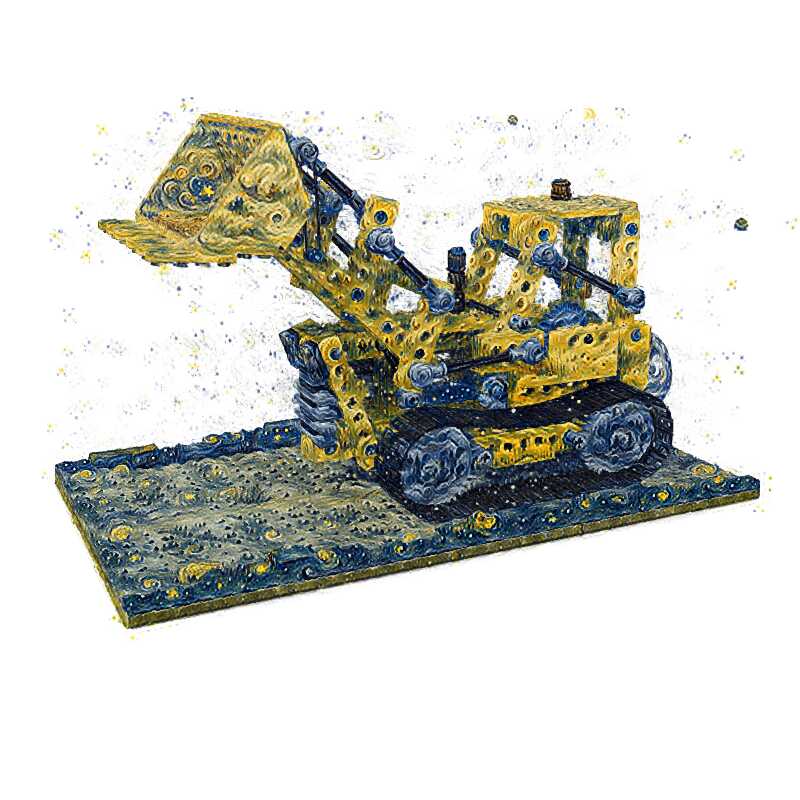} \\
 & 96 & 
  \includegraphics[trim={50 100 50 100},clip, width=0.13\textwidth]{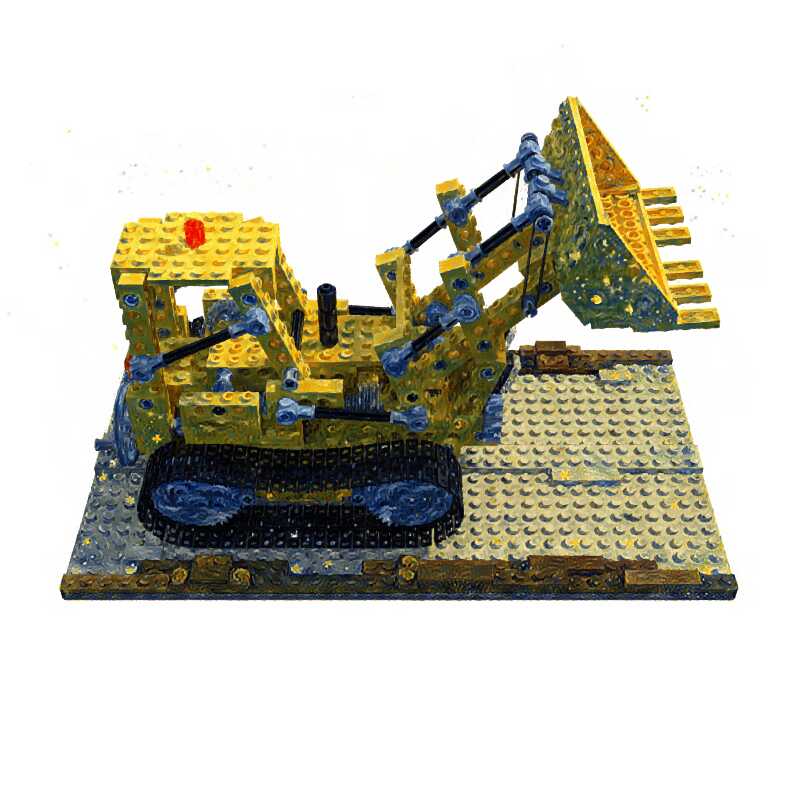} &
  \includegraphics[trim={50 100 50 100},clip, width=0.13\textwidth]{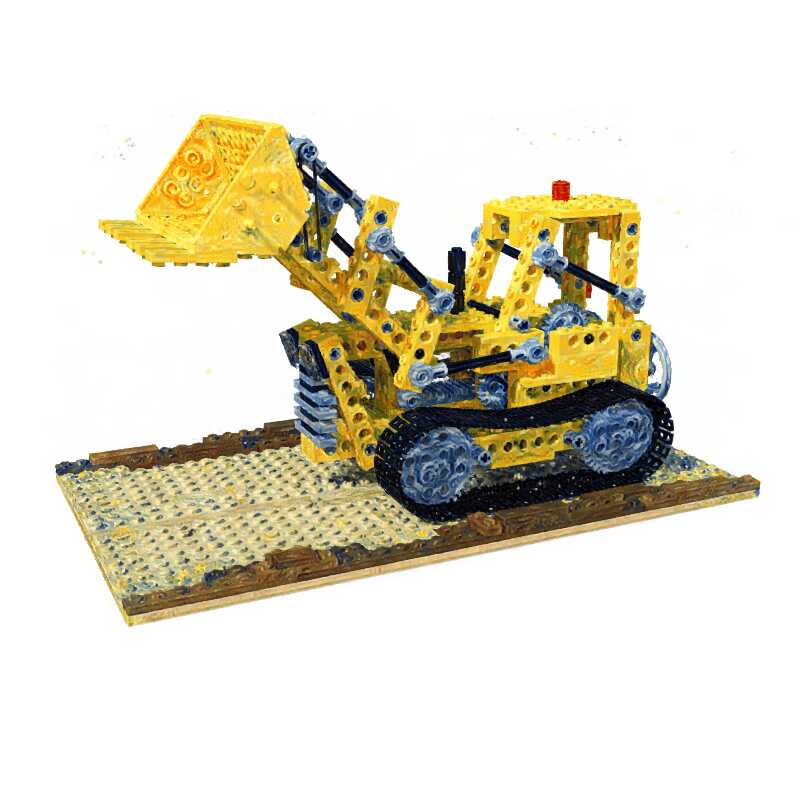} &
  \includegraphics[trim={50 100 50 100},clip, width=0.13\textwidth]{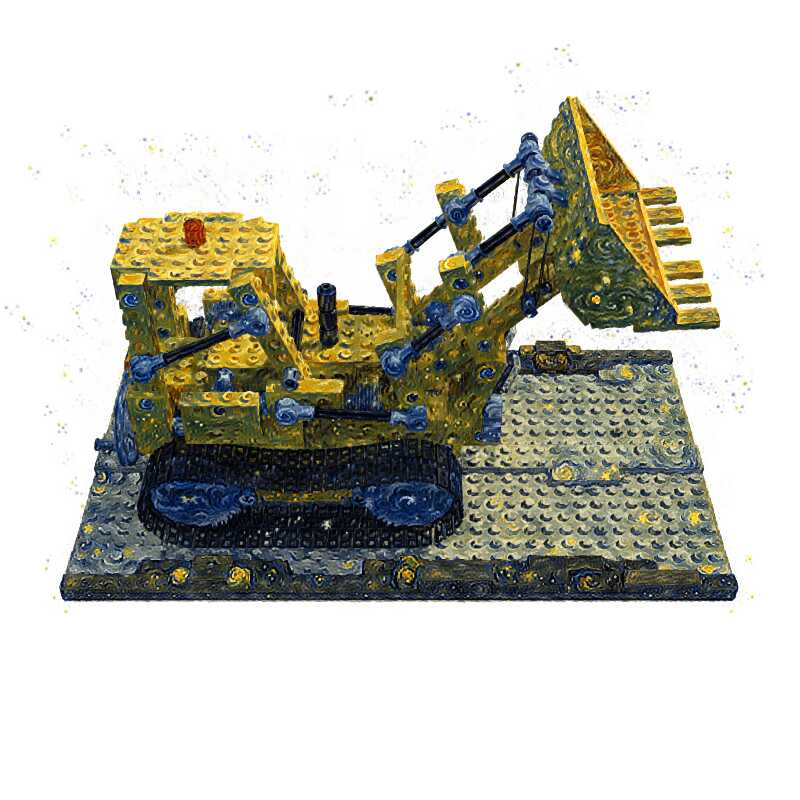} &
  \includegraphics[trim={50 100 50 100},clip, width=0.13\textwidth]{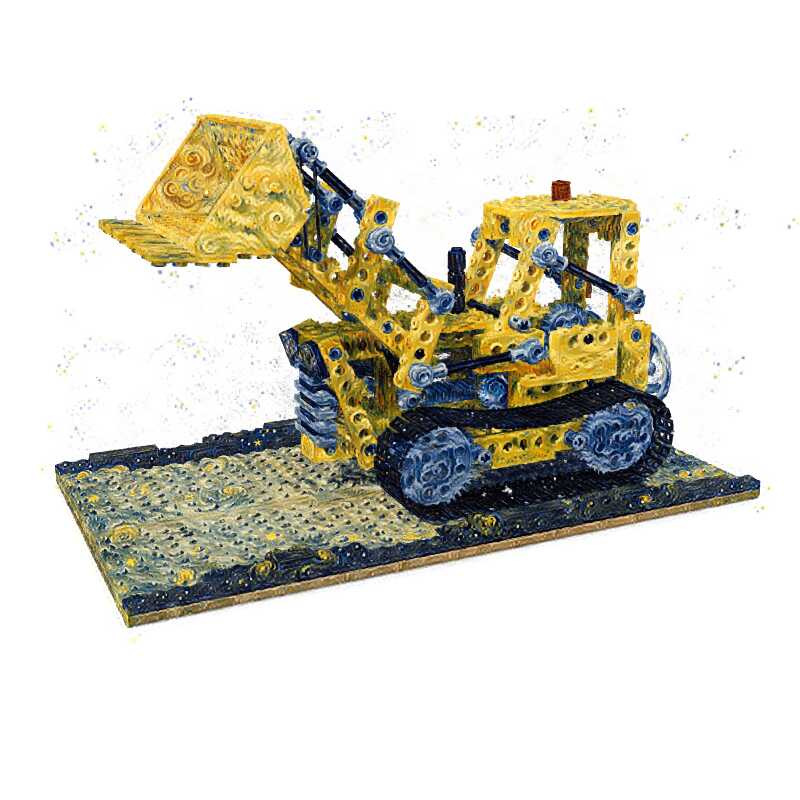} &
  \includegraphics[trim={50 100 50 100},clip, width=0.13\textwidth]{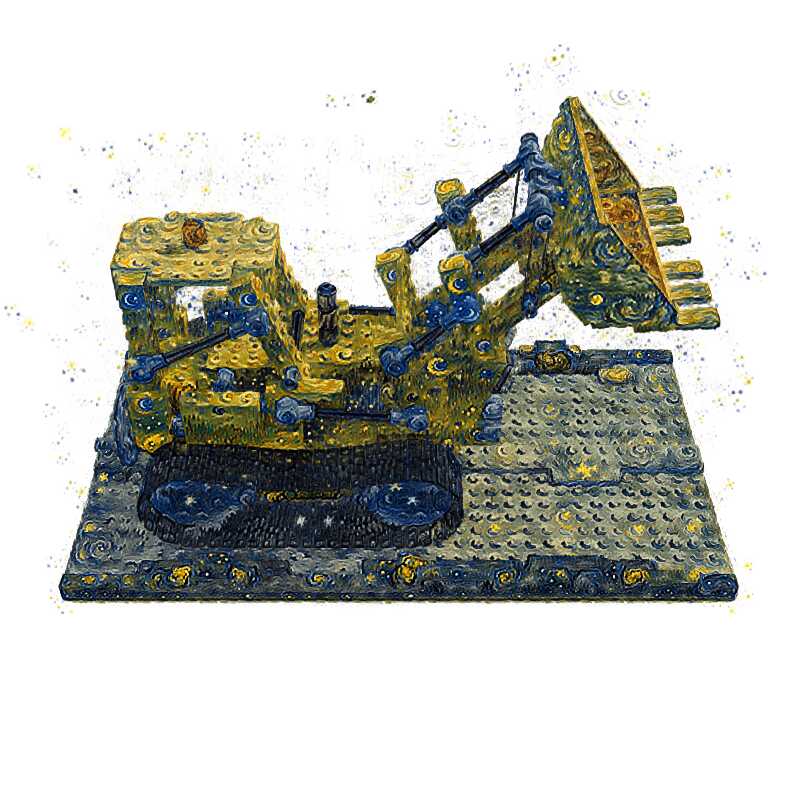} &
  \includegraphics[trim={50 100 50 100},clip, width=0.13\textwidth]{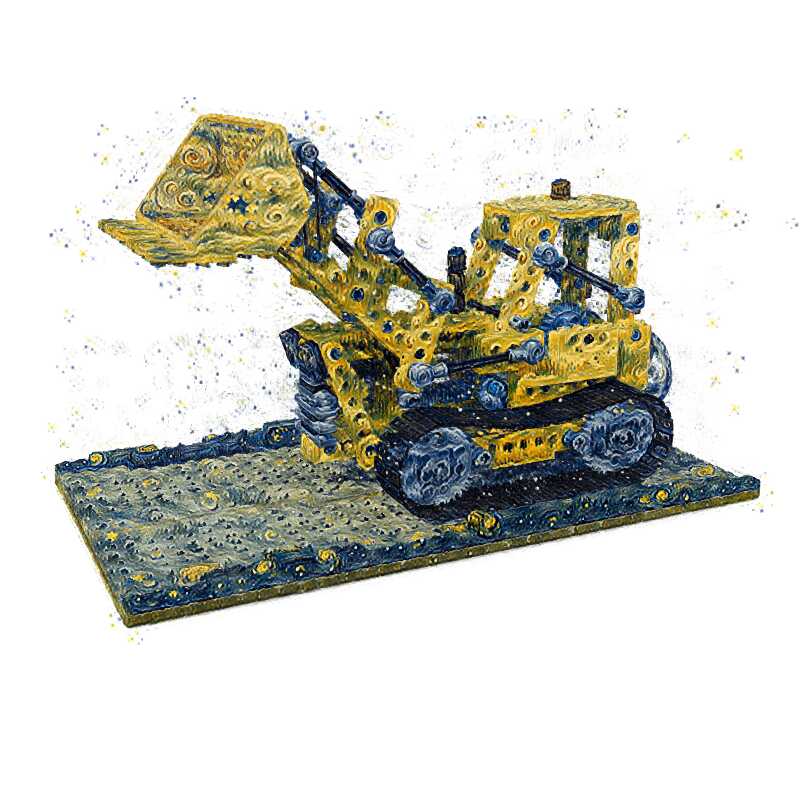} \\
\multirow{3}{*}{hotdog} & 24 & 
  \includegraphics[trim={50 100 50 100},clip, width=0.13\textwidth]{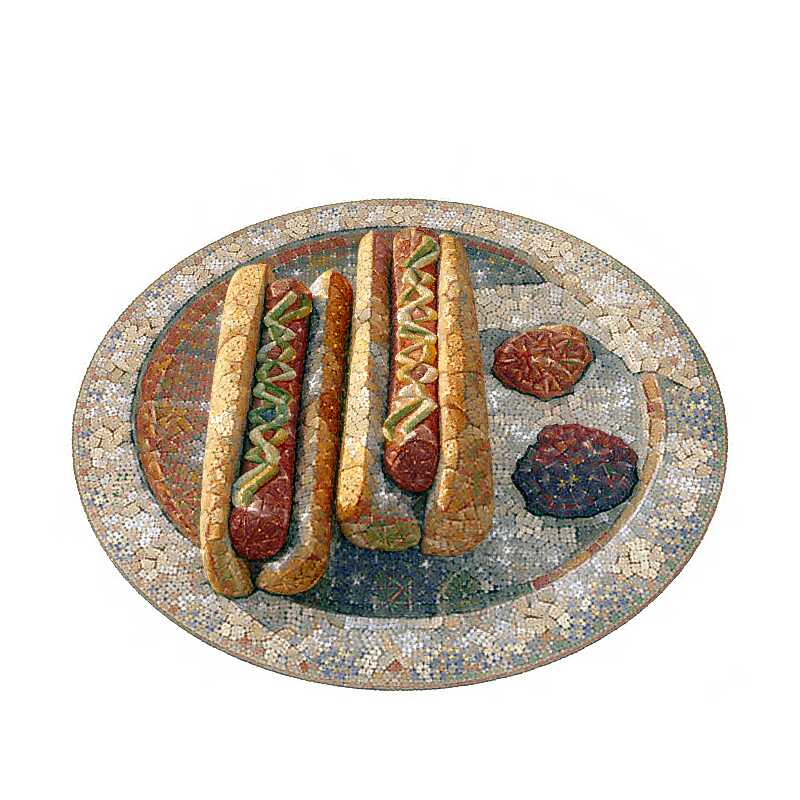} &
  \includegraphics[trim={50 100 50 100},clip, width=0.13\textwidth]{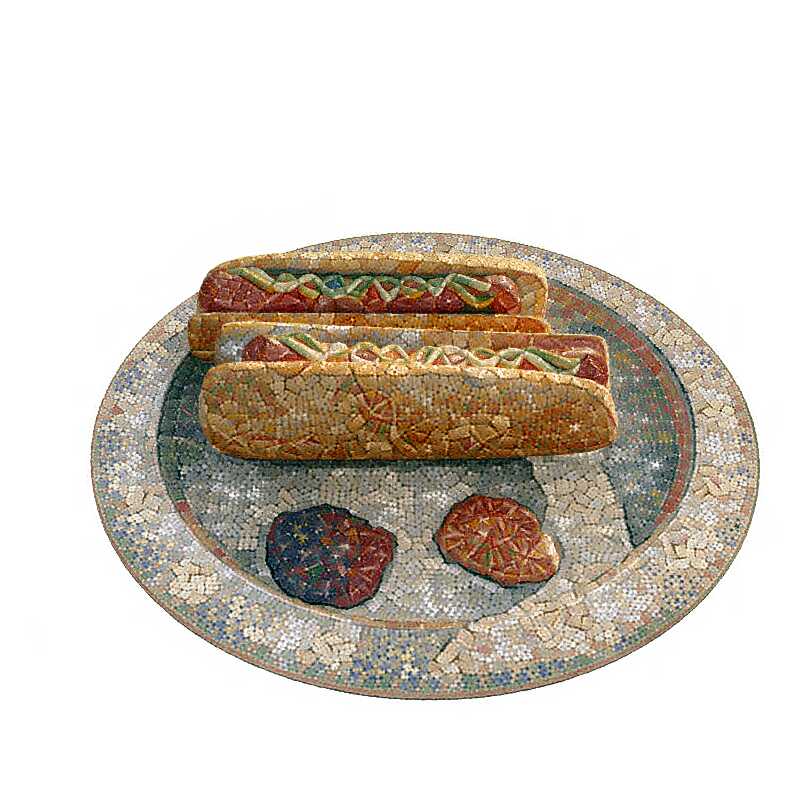} &
  \includegraphics[trim={50 100 50 100},clip, width=0.13\textwidth]{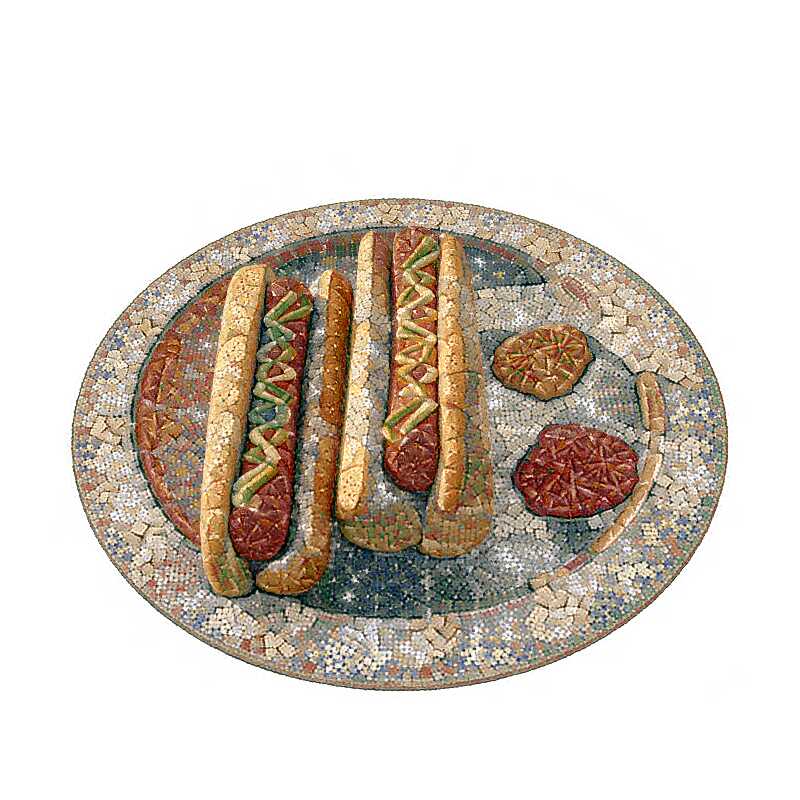} &
  \includegraphics[trim={50 100 50 100},clip, width=0.13\textwidth]{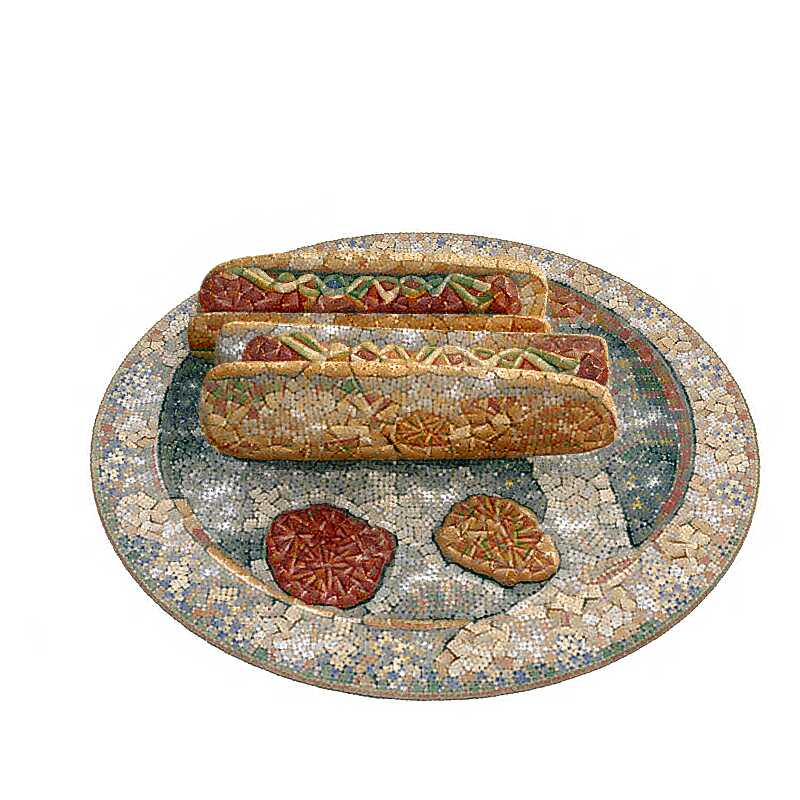} &
  \includegraphics[trim={50 100 50 100},clip, width=0.13\textwidth]{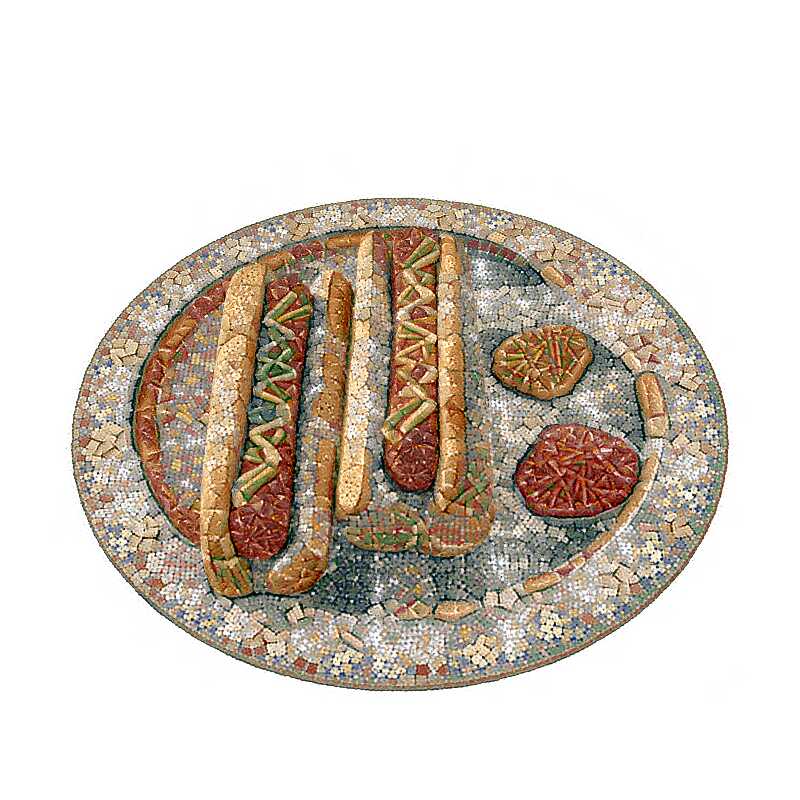} &
  \includegraphics[trim={50 100 50 100},clip, width=0.13\textwidth]{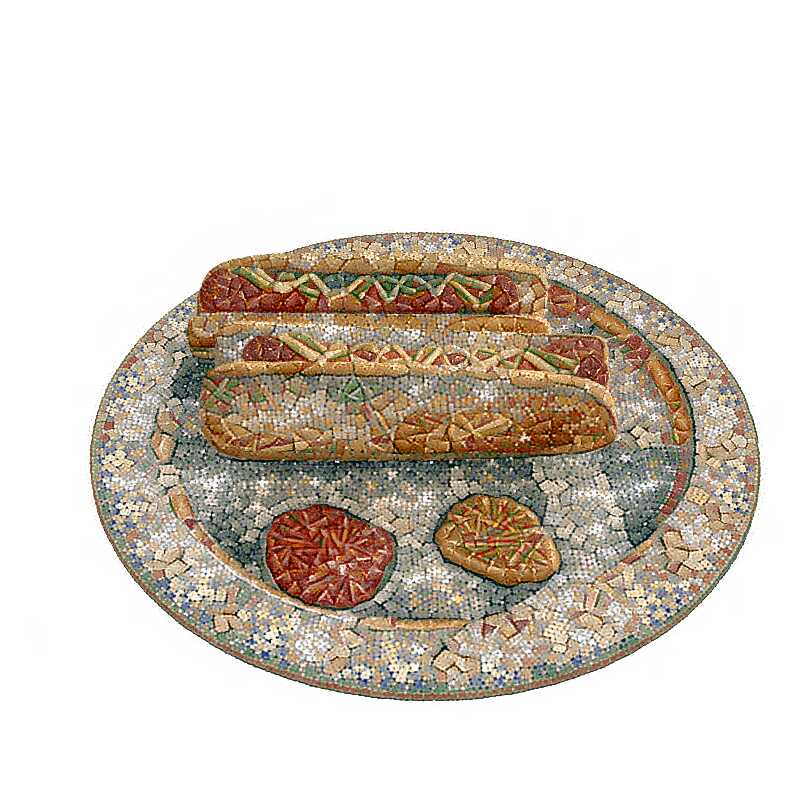} \\
 & 48 & 
  \includegraphics[trim={50 100 50 100},clip, width=0.13\textwidth]{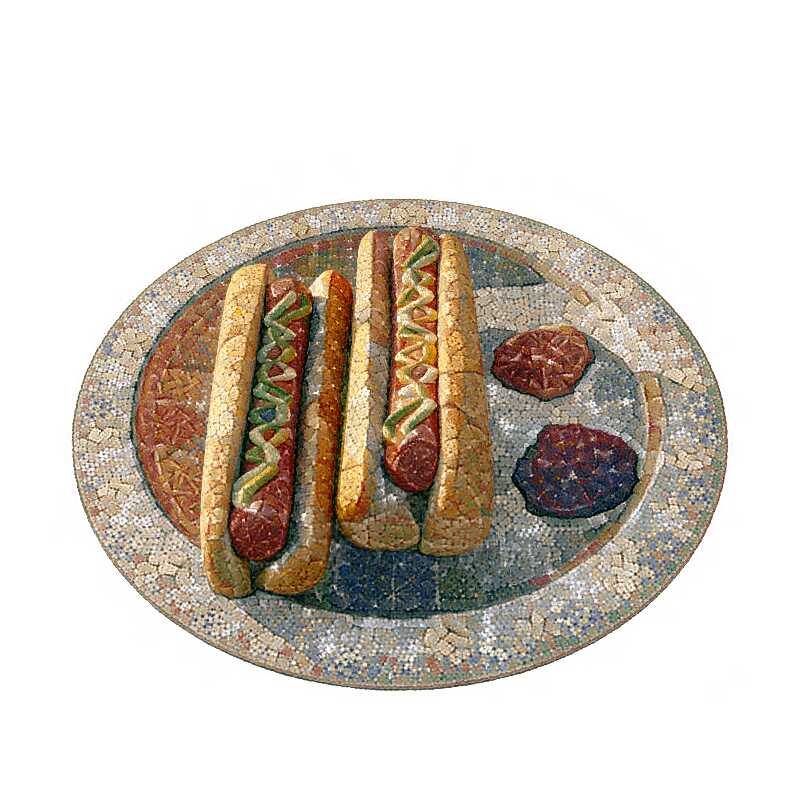} &
  \includegraphics[trim={50 100 50 100},clip, width=0.13\textwidth]{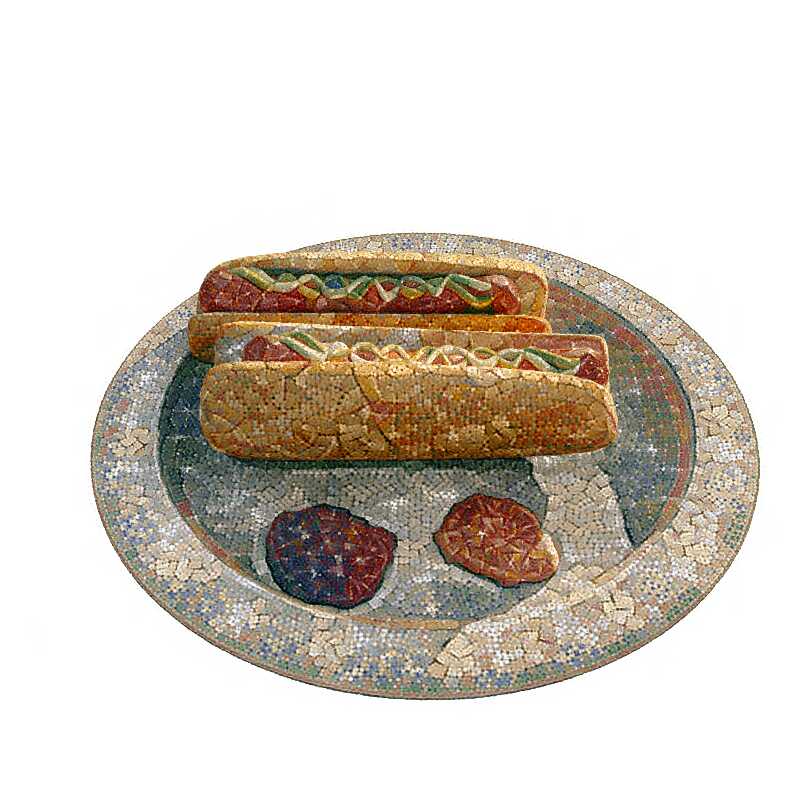} &
  \includegraphics[trim={50 100 50 100},clip, width=0.13\textwidth]{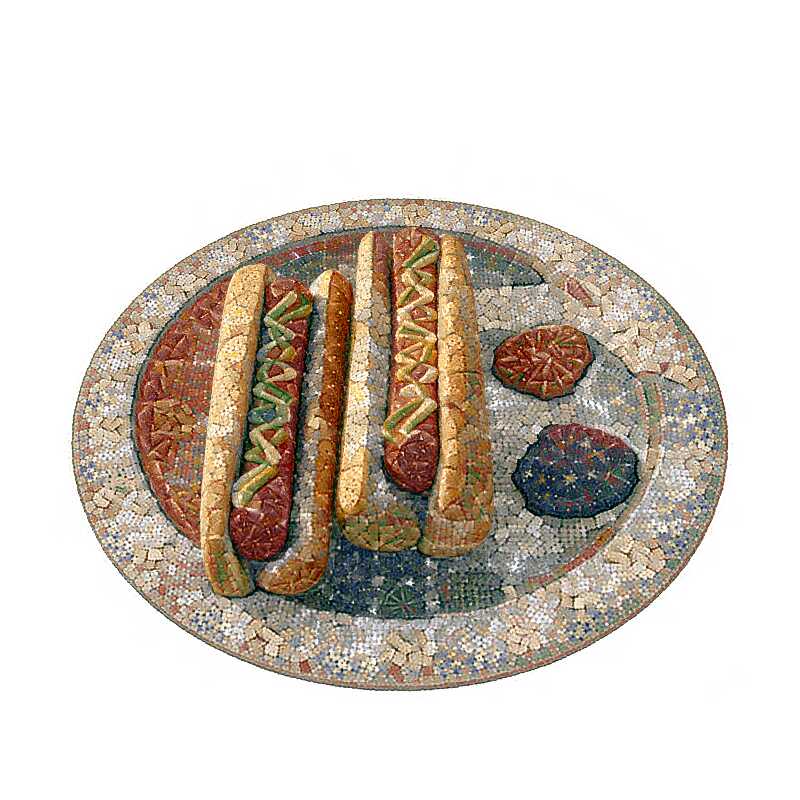} &
  \includegraphics[trim={50 100 50 100},clip, width=0.13\textwidth]{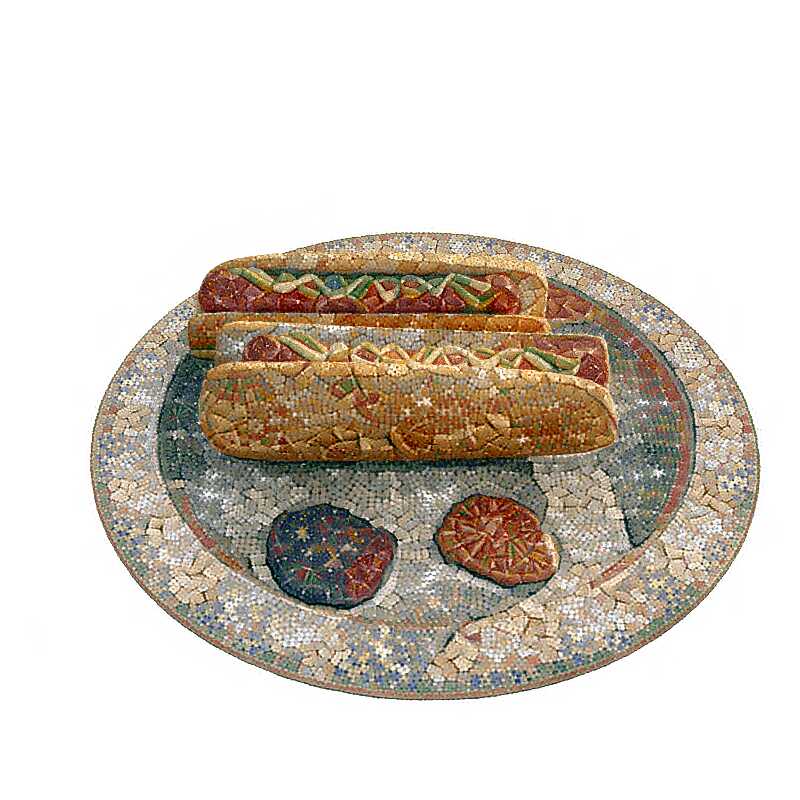} &
  \includegraphics[trim={50 100 50 100},clip, width=0.13\textwidth]{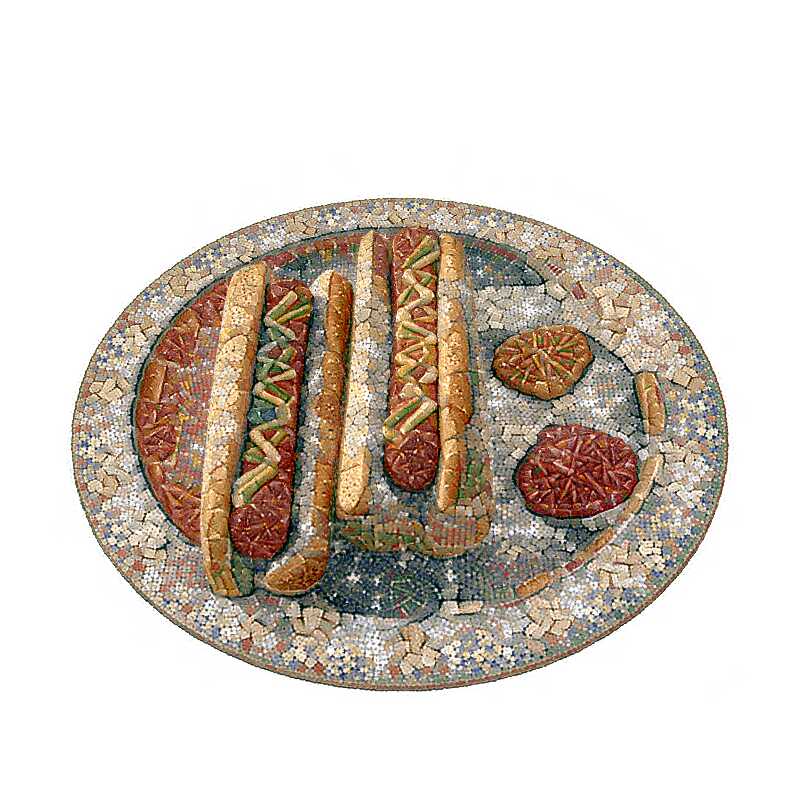} &
  \includegraphics[trim={50 100 50 100},clip, width=0.13\textwidth]{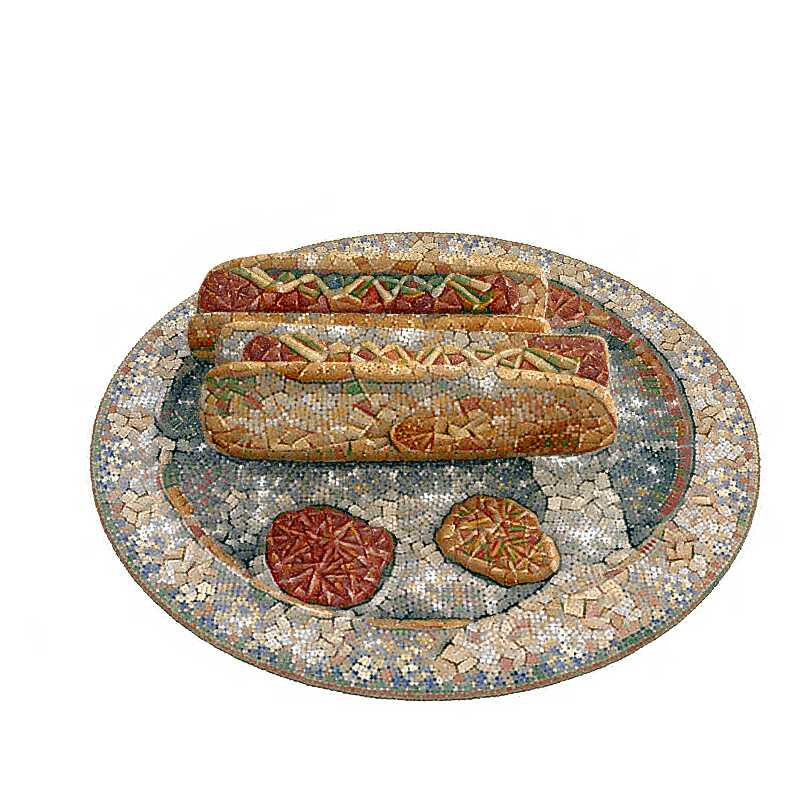} \\
 & 96 & 
  \includegraphics[trim={50 100 50 100},clip, width=0.13\textwidth]{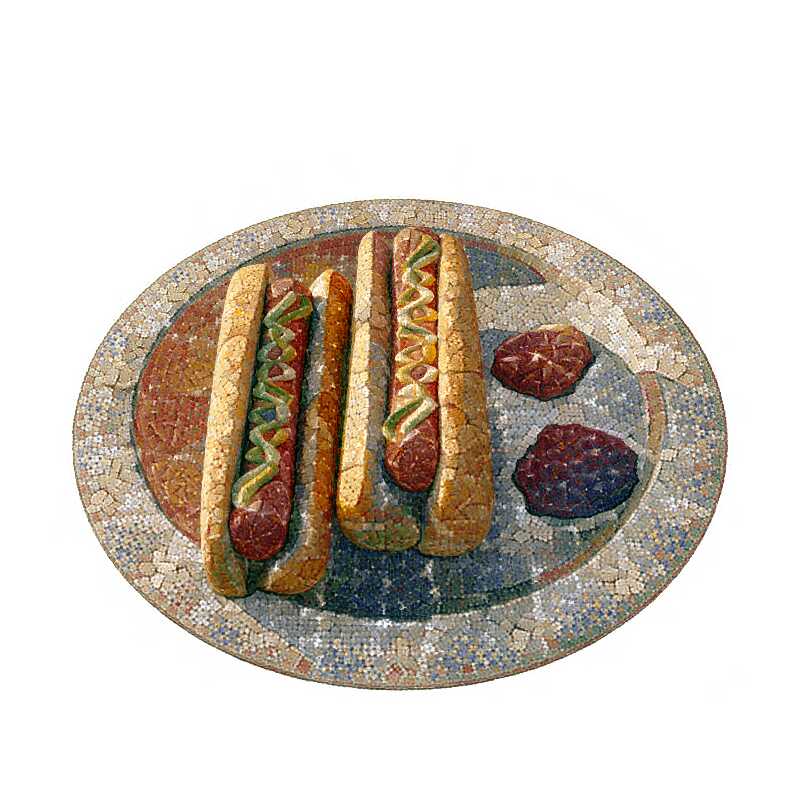} &
  \includegraphics[trim={50 100 50 100},clip, width=0.13\textwidth]{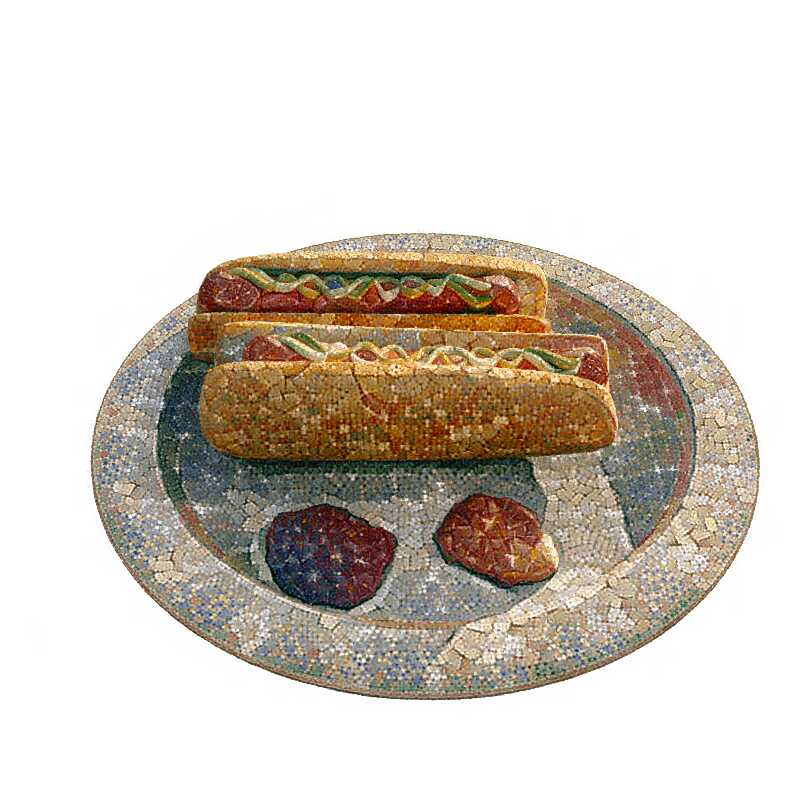} &
  \includegraphics[trim={50 100 50 100},clip, width=0.13\textwidth]{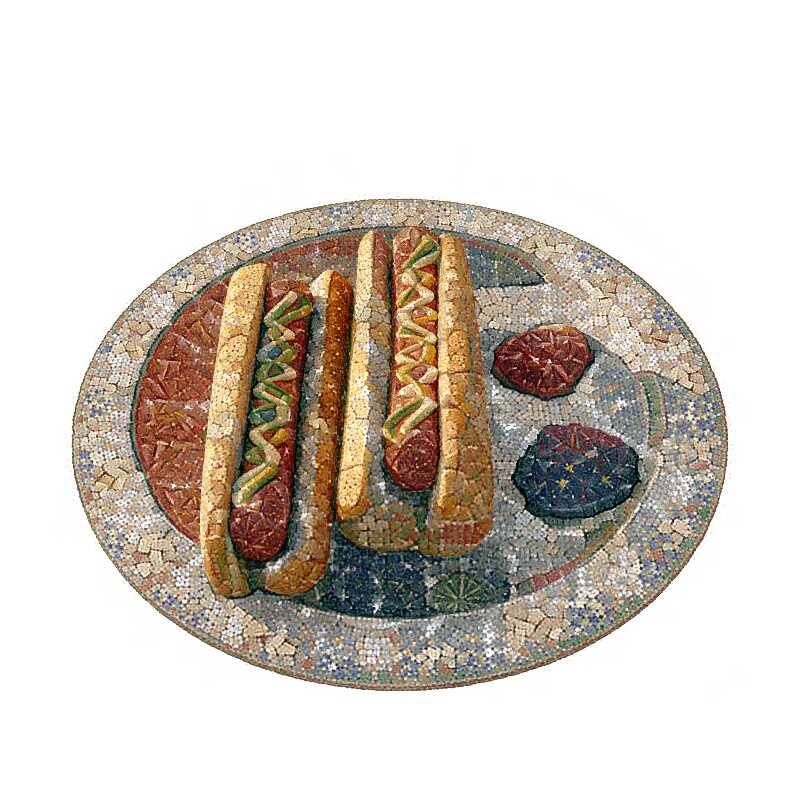} &
  \includegraphics[trim={50 100 50 100},clip, width=0.13\textwidth]{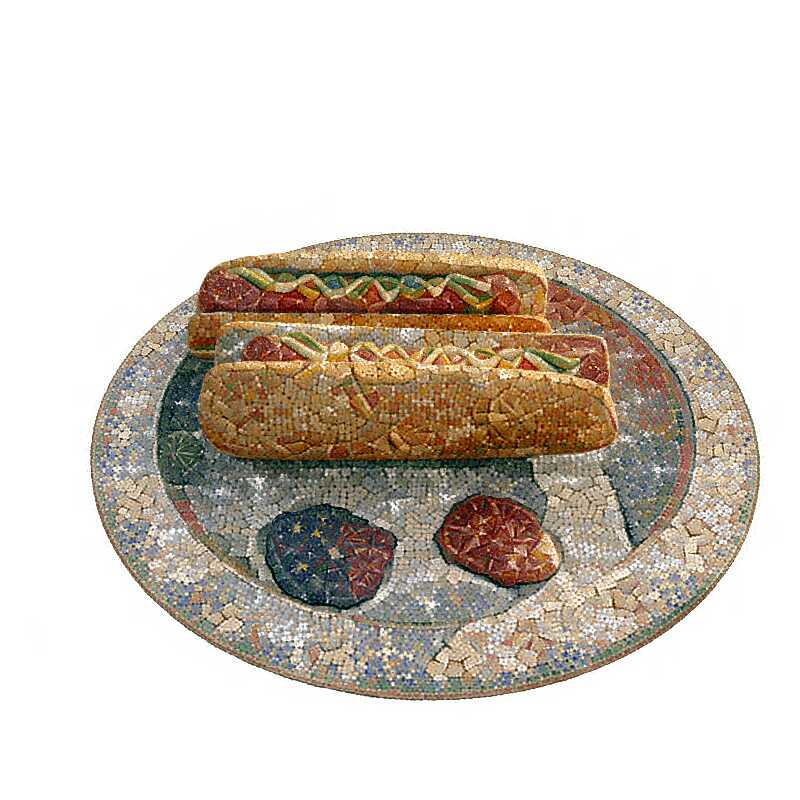} &
  \includegraphics[trim={50 100 50 100},clip, width=0.13\textwidth]{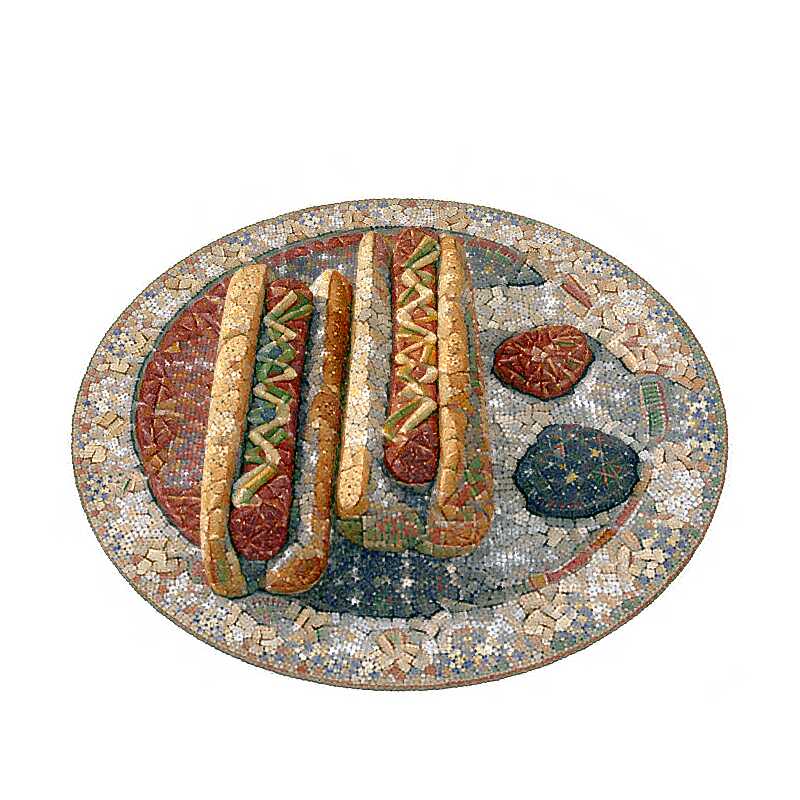} &
  \includegraphics[trim={50 100 50 100},clip, width=0.13\textwidth]{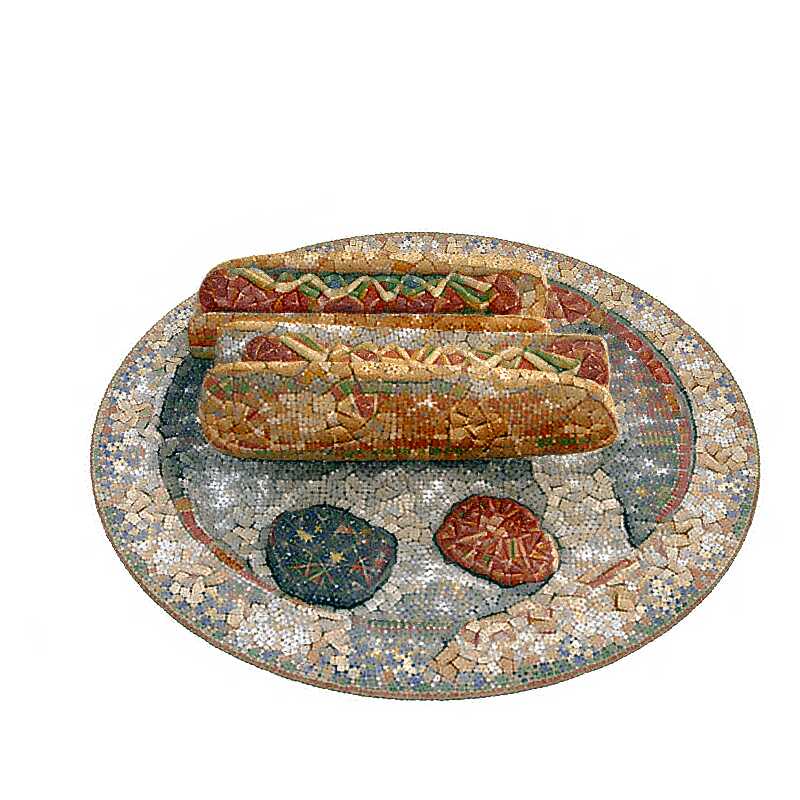} \\
\end{tabular}
\caption{Effect of $\lambda_{d}$ and $\lambda_{p}$ on the performance of \our{} on \textit{lego} and \textit{hotdog} objects from NeRF-Synthetic dataset \cite{mildenhall2020nerf}. Objects are stylized with "Starry Night by Vincent van Gogh" and "Mosaic" prompts.}
\label{fig:app_lp_ld_new}
\end{figure*}

We additionally provide a visual representation of the study  for the text conditioned setting in Figures~\ref{fig:3D_alb_patch} and~\ref{fig:app_lp_ld_new}.

\begin{table}[t!]
\centering
\caption{Ablation study of $\lambda_d$ and $\lambda_p$ parameters under text-conditioned and image-conditioned settings for 3D style transfer. Larger values are better.}
\label{tab:ablation_lambda_conditioned}
\scriptsize
\setlength{\tabcolsep}{2pt}
\resizebox{\columnwidth}{!}{%
    \begin{tabular}{cc|cccc}
        \toprule
            $\lambda_d$ & $\lambda_p$ & \textit{CLIP-S} $\uparrow$ & \textit{CLIP-SIM} $\uparrow$ & \textit{CLIP-F} $\uparrow$ & \textit{CLIP-CONS} $\uparrow$ \\
            \midrule
            \multicolumn{6}{c}{\textit{Text-conditioned}} \\
            \midrule
                 & 540.0  & 18.5091 & 17.7659 & \tbest{98.5833} & \tbest{8.0825} \\
           24.0  & 1080.0 & 19.6185 & 17.8014 & 97.9323 & 4.6924 \\
                 & 2160.0 & \sbest{20.6582} & 17.5859 & 97.2344 & 2.3747 \\
            \midrule
                 & 540.0  & 18.1979 & 17.7288 & \sbest{98.7604} & \sbest{8.8755} \\
            48.0 & 1080.0 & 19.8249 & \sbest{18.2894} & 98.1458 & 5.1358 \\
                 & 2160.0 & \best{\textbf{20.6914}} & 17.8021 & 97.3698 & 2.5132 \\
            \midrule
                & 540.0  & 17.9017 & 17.3130 & \best{\textbf{98.8750}} & \best{\textbf{9.3703}} \\
           96.0 & 1080.0 & 19.6348 & \best{\textbf{18.2946}} & 98.2084 &  4.9963 \\
                 & 2160.0 & \tbest{20.5593} & \tbest{18.0075} & 97.5104 & 2.8079 \\
            \midrule
            \multicolumn{6}{c}{\textit{Image-conditioned}} \\
            \midrule
            & 540.0  & 59.0000 & 9.5239 & \sbest{99.8281} & \tbest{2.6883} \\
            24.0 & 1080.0 & 60.3516 & 12.1074 & \tbest{99.7969} & 1.3852 \\
            & 2160.0 & \tbest{61.5078} & \best{\textbf{13.9268}} & 99.6406 & 0.8419 \\
            \midrule
            & 540.0  & 59.0078 & 9.1992 & \best{\textbf{99.8750}} & \sbest{2.9677} \\
            48.0 & 1080.0 & 60.3438 & 11.9541 & 99.7188 & 1.6991 \\
            & 2160.0 & \sbest{61.5352} & \tbest{13.7754} & 99.6250 & 1.1497 \\
            \midrule
            & 540.0  & 59.1133 & 9.1904 & \tbest{99.7969} & \best{\textbf{3.3152}} \\
            96.0 & 1080.0 & 60.2695 & 11.7402 & 99.7656 & 1.8192 \\
            & 2160.0 & \best{\textbf{61.5586}} & \sbest{13.8740} & 99.6719 & 1.1536 \\
            \bottomrule
    \end{tabular}
}
\end{table}

\begin{table}[t]
    \centering 
    \small
    \caption{Ablation of $\lambda_d$ and $\lambda_p$ for \our{} on a 4D scene
    (image-conditioned, \texttt{coffee\_martini}, $\lambda_c{=}5$,
    $\lambda_t{=}0$). Larger is better.}
    \label{tab:4d_ablation_lambda}
    \begin{tabular}{cc|cc}
        \toprule
        $\lambda_d$ & $\lambda_p$ & \textit{CLIP-S} $\uparrow$ & \textit{CLIP-CONS} $\uparrow$ \\
        \midrule
           & 360 & 70.53 & 0.20 \\
        32 & 480 & 74.47 & \best{\textbf{0.61}} \\
           & 600 & 71.84 & -0.24 \\
        \midrule
           & 360 & 69.83 & \tbest{0.38} \\
        64 & 480 & 75.31 & \sbest{0.54} \\
           & 600 & \tbest{76.29} & -0.08 \\
        \midrule
           & 360 & 67.61 & 0.37 \\
        96 & 480 & \sbest{76.70} & -0.50 \\
           & 600 & \best{\textbf{76.72}} & 0.36 \\
        \bottomrule
    \end{tabular}
\end{table}

\begin{table*}[t]
    \centering
    \caption{Loss-component ablation for \our{} on a 4D scene
    (image-conditioned, \texttt{coffee\_martini}). Each row modifies a single term of our full objective. Larger is better.}
    \label{tab:4d_component_ablation}
    \begin{tabular}{l|cccc}
        \toprule
        Variant & \textit{CLIP-S} $\uparrow$ & \textit{CLIP-SIM} $\uparrow$
                & \textit{CLIP-F} $\uparrow$ & \textit{CLIP-CONS} $\uparrow$ \\
        \midrule
        Default config            & 58.48 & 4.84  & 100.00 & -0.54 \\
        \our{} (full)             & 75.31 & 19.14 & 99.55  & 0.54 \\
        \midrule
        $\;-$ Patch ($\lambda_p{=}0$)        & 58.51 & 1.79  & 99.78  & -0.23 \\
        $\;-$ Directional ($\lambda_d{=}0$)  & 69.61 & 13.04 & 99.54  & 0.07 \\
        $\;-$ Content ($\lambda_c{=}0$)      & 82.55 & 22.73 & 99.94  & -0.20 \\
        $\;+$ Temporal ($\lambda_t{=}2000$)  & 67.95 & 14.96 & 100.03 & -0.90 \\
        $\;+$ Freeze density                 & 70.69 & 13.68 & 99.93  & -0.22 \\
        \bottomrule
    \end{tabular}
\end{table*}

\begin{table*}[t]
\centering
\caption{Ablation study of $\texttt{patch\_size}$ ($p_{size}$) and $\texttt{patch\_num}$ ($p_{num}$) parameters under text-conditioned and image-conditioned settings for 3D style transfer. Larger values are better.}
\label{tab:ablation_3d_lambda_conditioned}
\setlength{\tabcolsep}{2pt}
\resizebox{\columnwidth}{!}{%
    \begin{tabular}{cc|cccc}
        \toprule
            $p_{num}$ &  $p_{size}$ & \textit{CLIP-S} $\uparrow$ & \textit{CLIP-SIM} $\uparrow$ & \textit{CLIP-F} $\uparrow$ & \textit{CLIP-CONS} $\uparrow$ \\
            \midrule
            \multicolumn{6}{c}{\textit{Text-conditioned}} \\
            \midrule
            & 64  & 15.4505 & 12.8164 & \best{\textbf{99.0364}} & \best{\textbf{10.4113}} \\
            25 & 128 & 19.4954 & 18.0394 & 97.9635 & 4.2952 \\
            & 256 & \tbest{22.0378} & \tbest{19.9733} & 97.5885 & 2.8533 \\
            \midrule
            & 64  & 15.7337 & 13.1517 & \tbest{98.9843} & \sbest{9.9658} \\
            50 & 128 & 19.4922 & 18.0026 & 97.7500 & 4.9838 \\
            & 256 & \sbest{22.3932} & \sbest{20.5579} & 97.4479 & 2.6319 \\
            \midrule
            & 64  & 15.7611 & 12.8854 & \sbest{99.0156} & \tbest{9.7855} \\
            100 & 128 & 19.6823 & 18.0980 & 97.7917 & 4.3928 \\
            & 256 & \best{\textbf{22.5755}} & \best{\textbf{20.5625}} & 97.2292 & 2.4138 \\
            \midrule
            \multicolumn{6}{c}{\textit{Image-conditioned}} \\
            \midrule
            & 64  & 58.6406 & 8.5981 & \sbest{99.9141} & \best{\textbf{2.9962}} \\
            25 & 128 & 60.2812 & 11.7451 & 99.6953 & 1.5619 \\
            & 256 & \tbest{65.4961} & \tbest{17.2158} & 99.6250 & 0.6472 \\
            \midrule
            & 64  & 58.8516 & 8.8506 & \best{\textbf{99.9297}} & \sbest{2.9045} \\
            50 & 128 & 60.4492 & 12.0830 & 99.7344 & 1.6569 \\
            & 256 & \sbest{65.5117} & \sbest{17.2988} & 99.6016 & 0.4042 \\
            \midrule
            & 64  & 58.9219 & 9.1162 & \tbest{99.8906} & \tbest{2.6393} \\
            100 & 128 & 60.3633 & 12.1396 & 99.7500 & 1.3890 \\
            & 256 & \best{\textbf{65.6680}} & \best{\textbf{17.5088}} & 99.6172 & 0.4997 \\
            \bottomrule
    \end{tabular}
}
\end{table*}

\begin{figure*}[ht]
\centering
\begin{tabular}{l l c@{}c c@{}c c@{}c}
\multicolumn{8}{c}{$\lambda_{p}$} \\
& & \multicolumn{2}{c}{360} & \multicolumn{2}{c}{480} & \multicolumn{2}{c}{600} \\
\multirow{3}{*}{coffee} & 32 & 
  \includegraphics[width=0.13\textwidth]{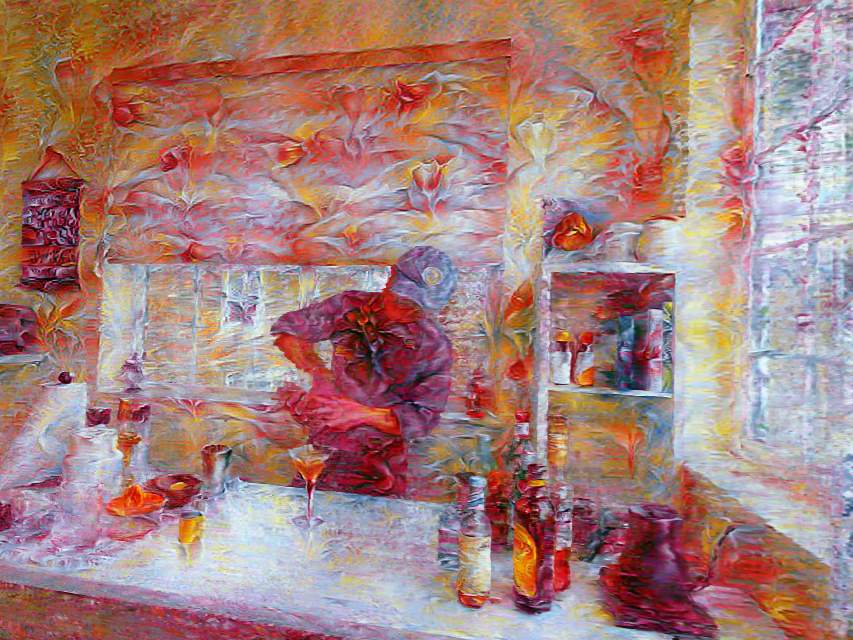} &
  \includegraphics[width=0.13\textwidth]{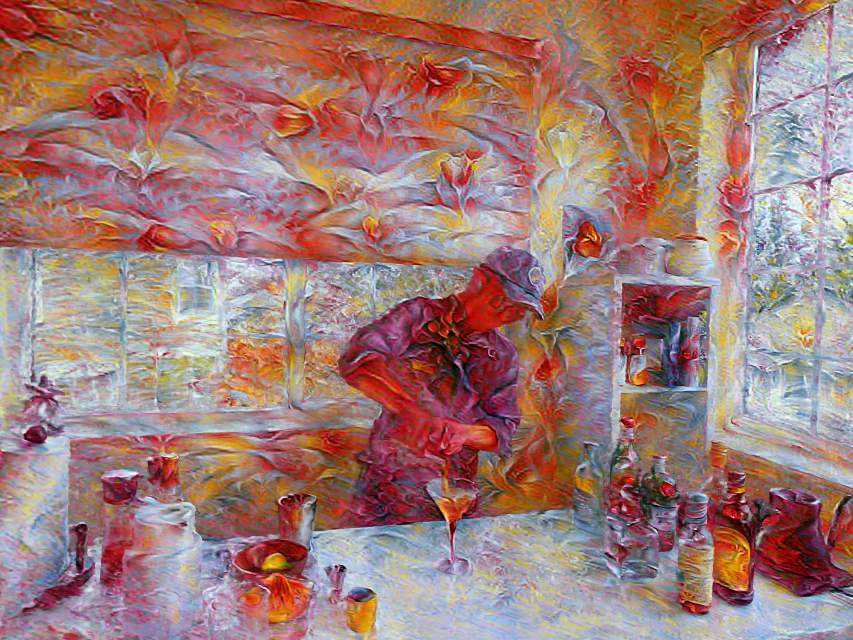} &
  \includegraphics[width=0.13\textwidth]{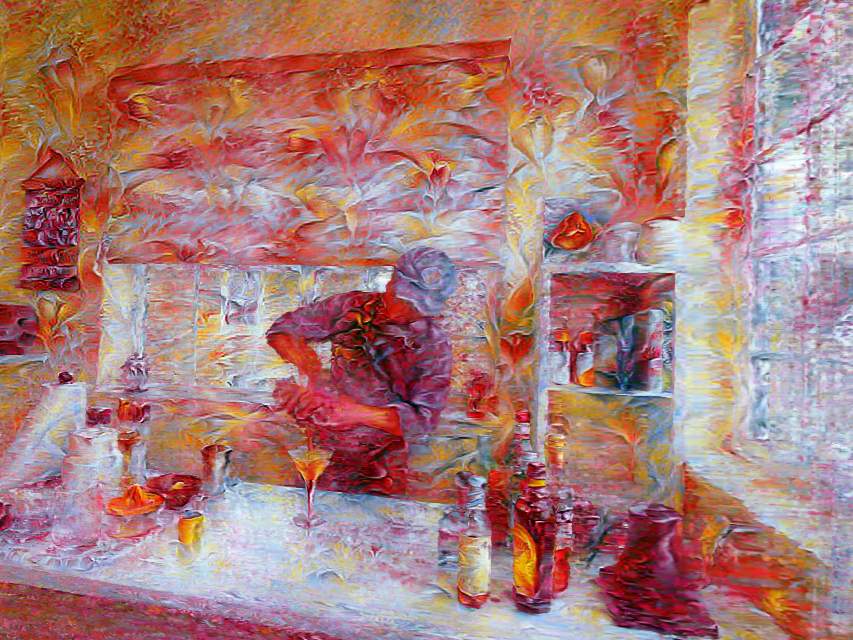} &
  \includegraphics[width=0.13\textwidth]{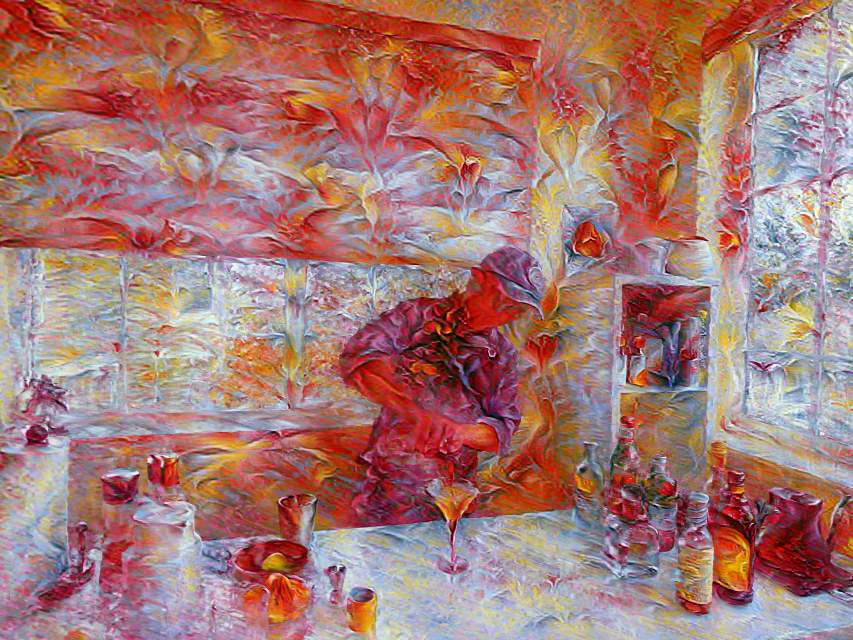} &
  \includegraphics[width=0.13\textwidth]{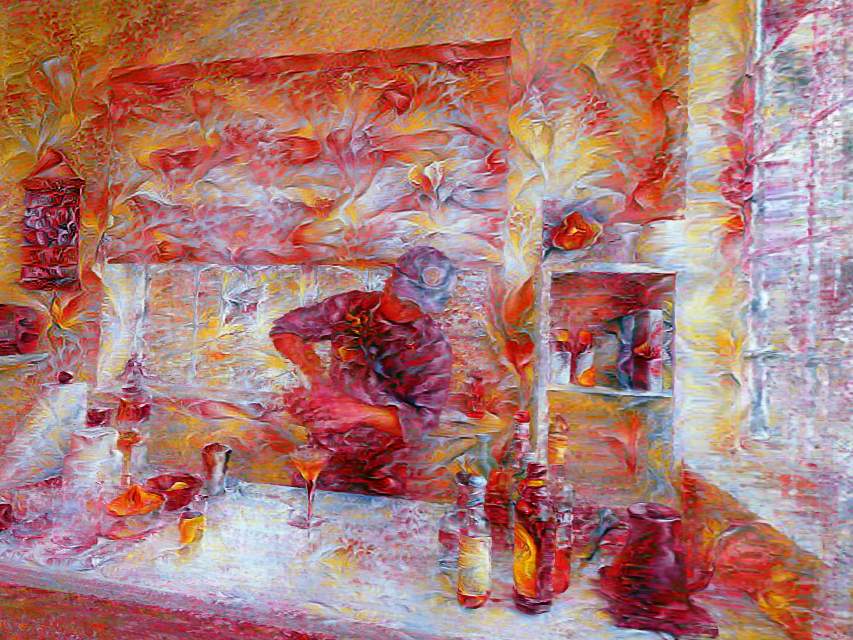} &
  \includegraphics[width=0.13\textwidth]{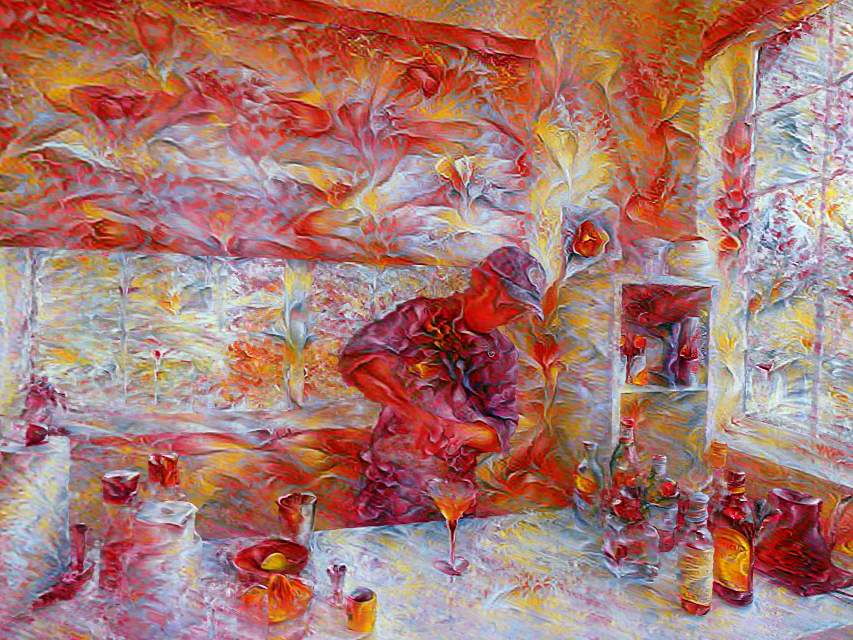} \\
 & 64 & 
  \includegraphics[width=0.13\textwidth]{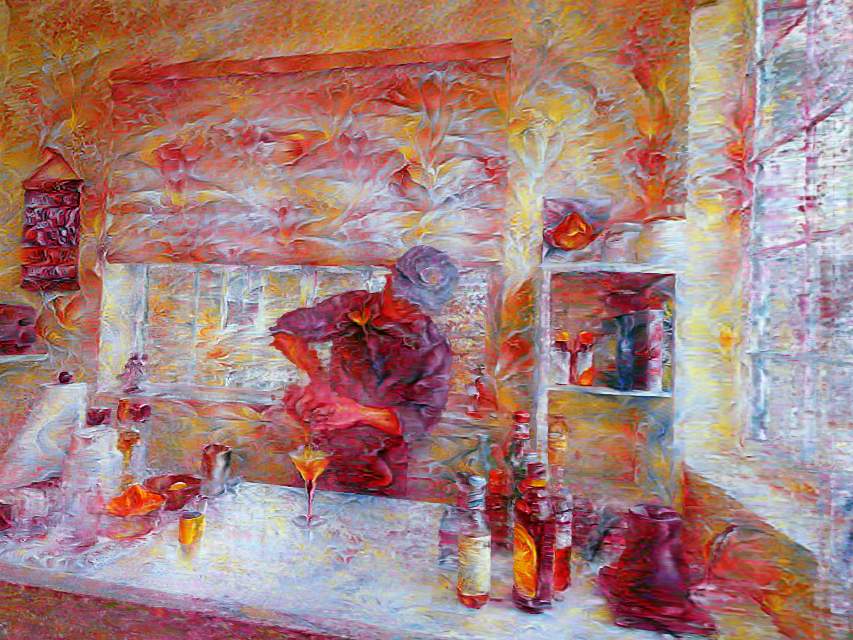} &
  \includegraphics[width=0.13\textwidth]{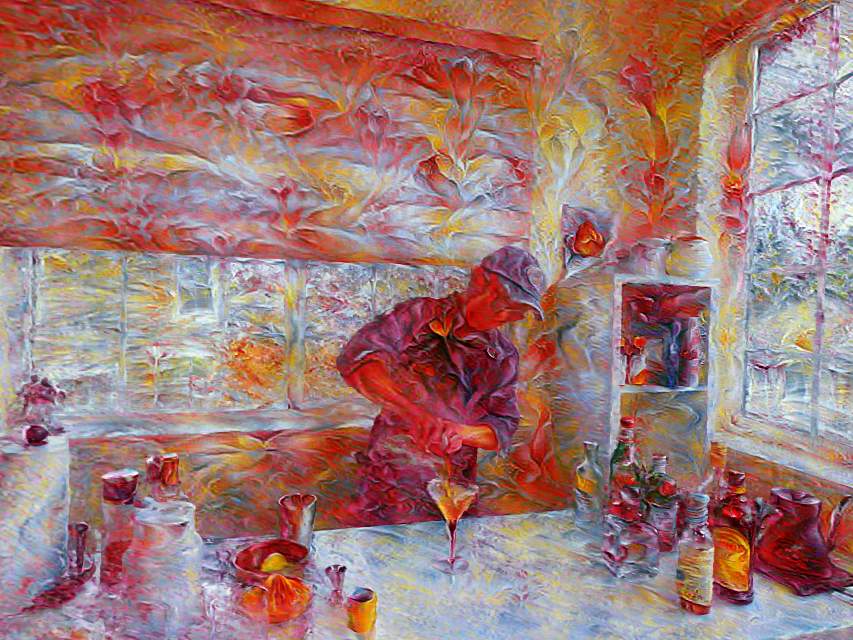} &
  \includegraphics[width=0.13\textwidth]{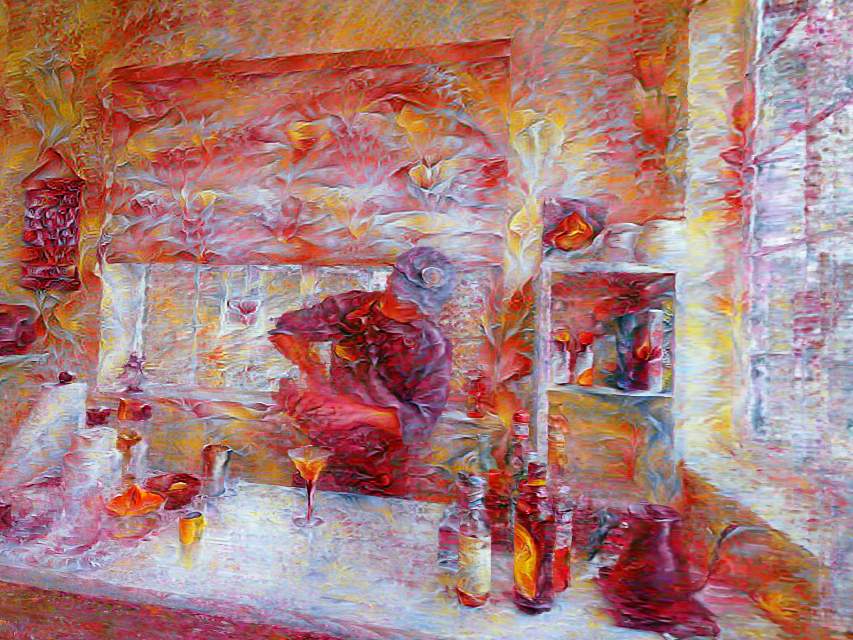} &
  \includegraphics[width=0.13\textwidth]{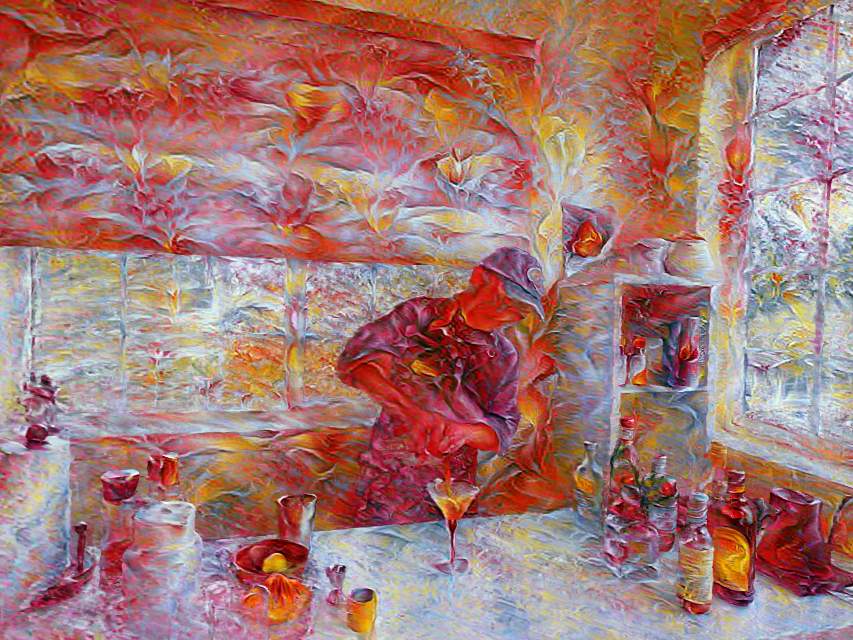} &
  \includegraphics[width=0.13\textwidth]{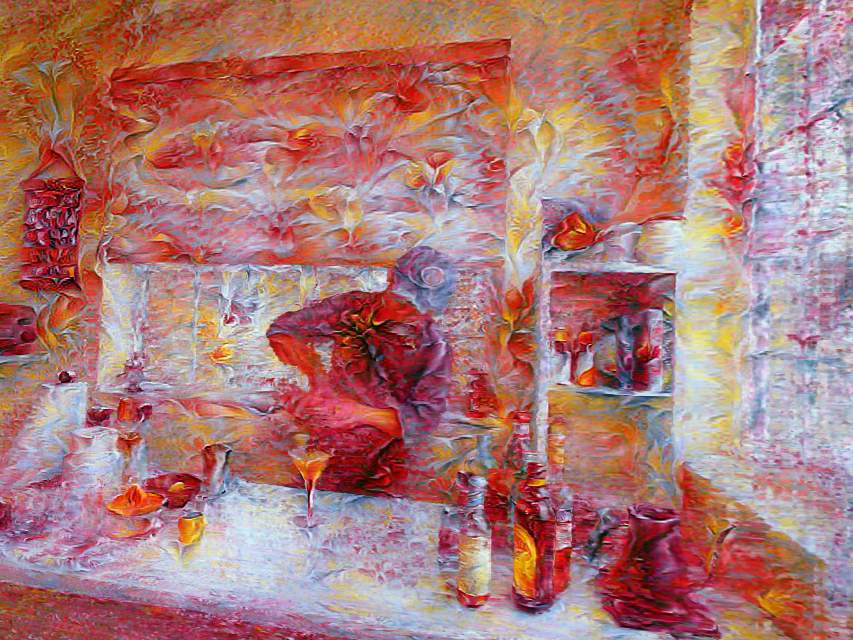} &
  \includegraphics[width=0.13\textwidth]{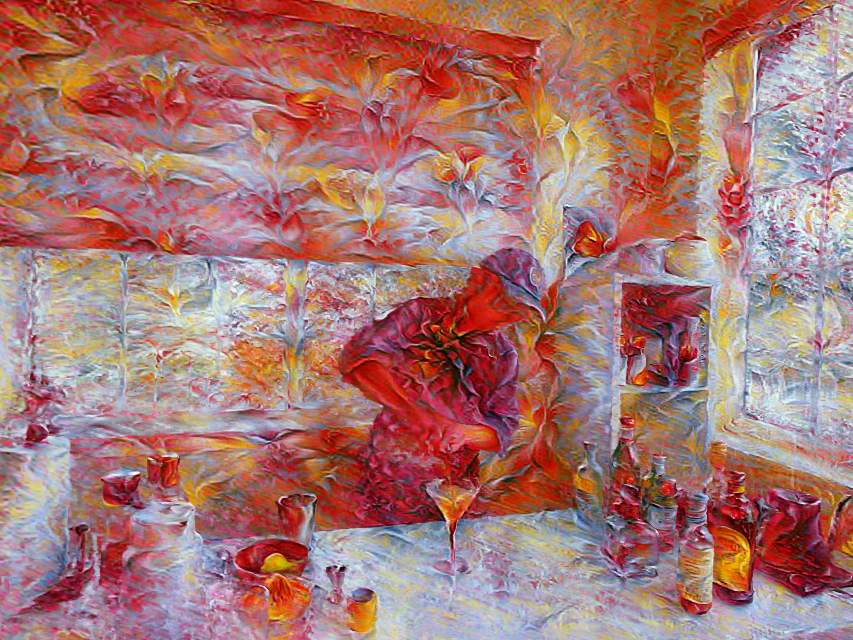} \\
 & 96 & 
  \includegraphics[width=0.13\textwidth]{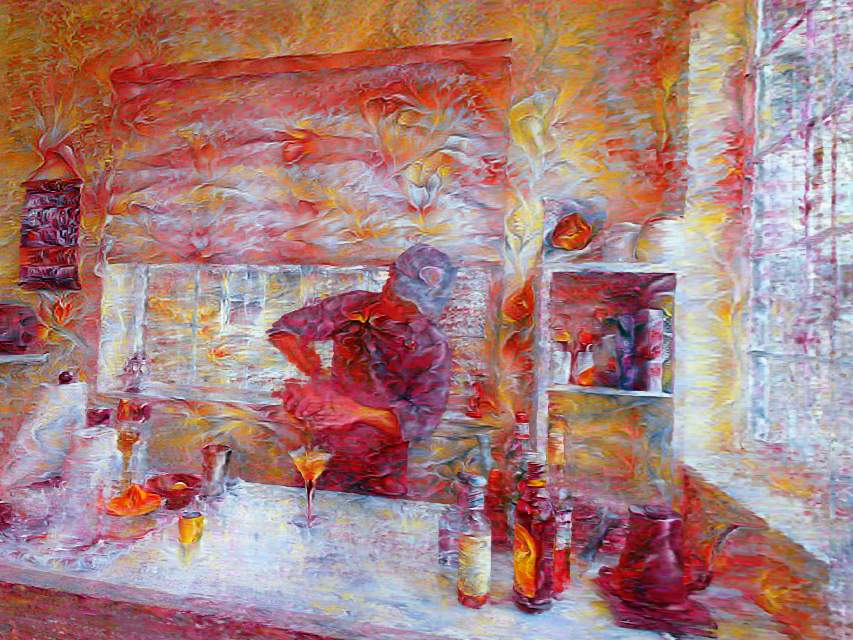} &
  \includegraphics[width=0.13\textwidth]{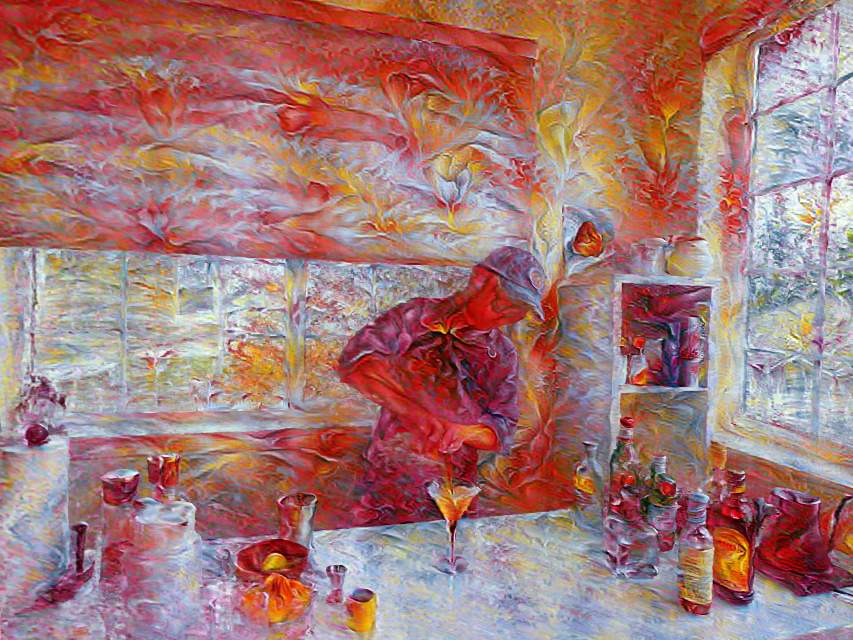} &
  \includegraphics[width=0.13\textwidth]{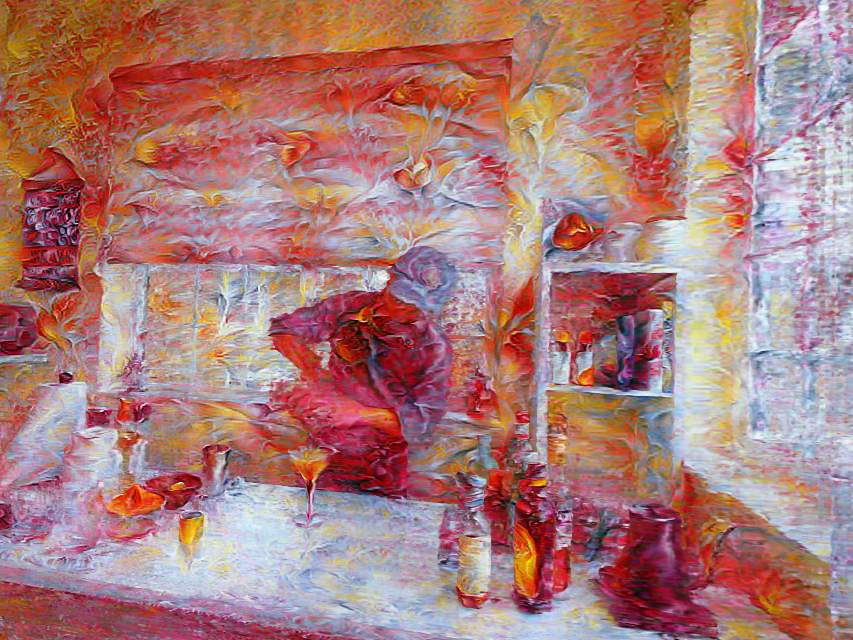} &
  \includegraphics[width=0.13\textwidth]{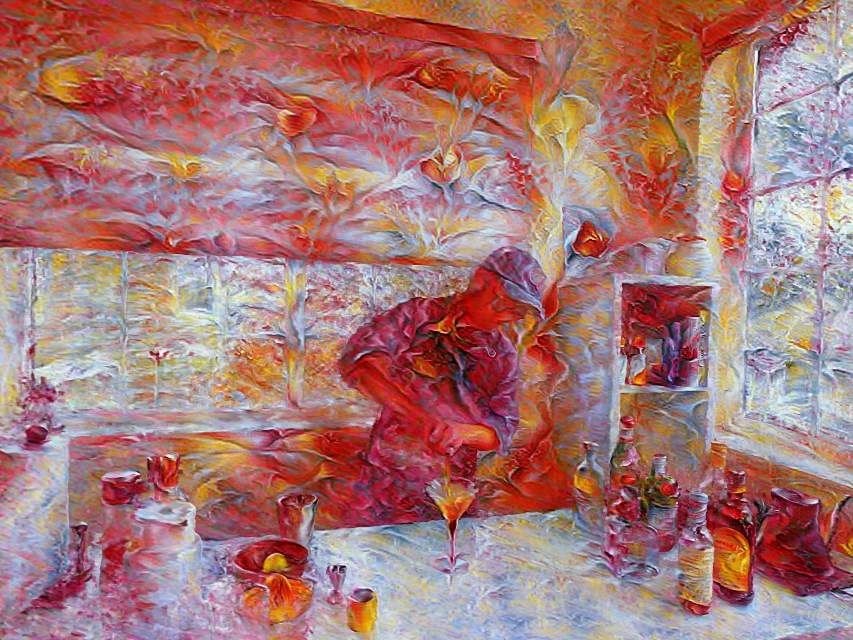} &
  \includegraphics[width=0.13\textwidth]{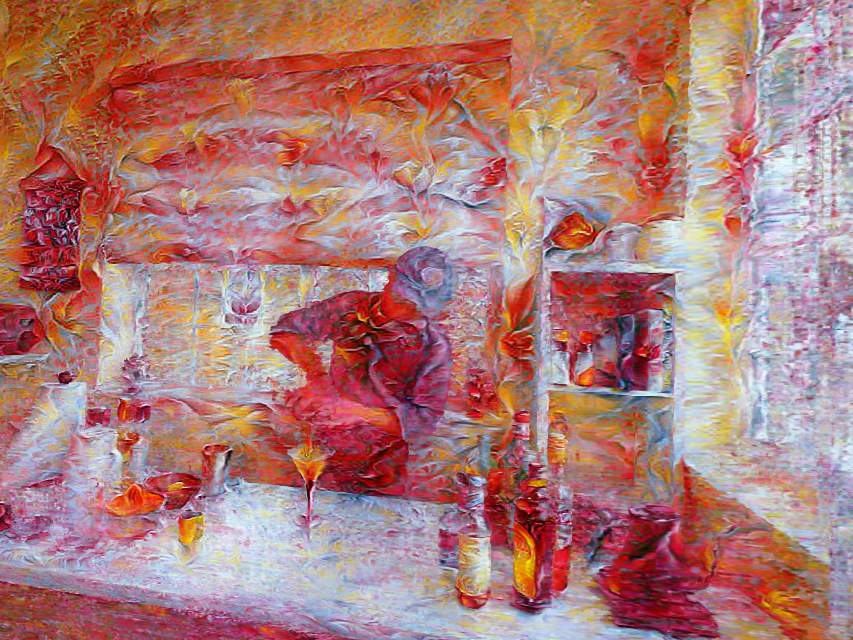} &
  \includegraphics[width=0.13\textwidth]{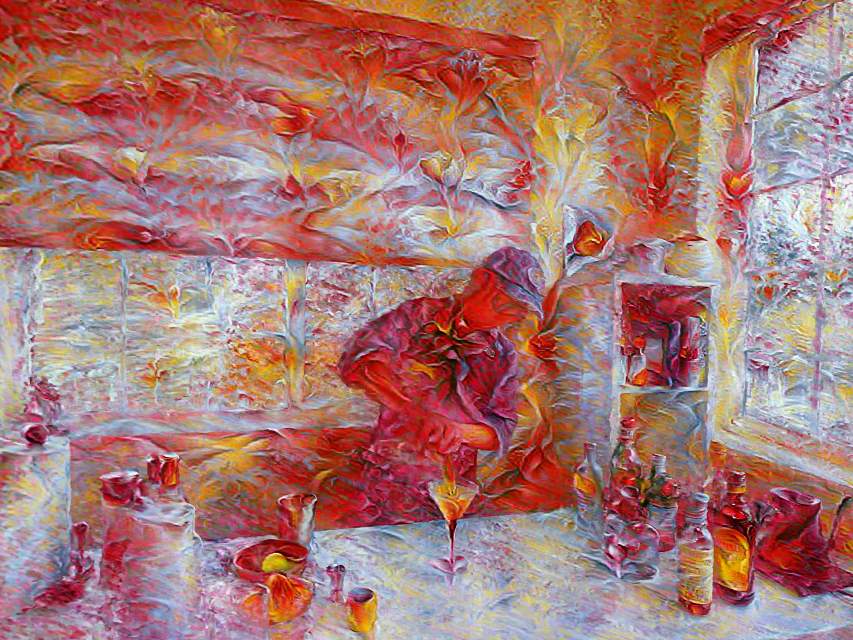} \\
\end{tabular}
\caption{Ablation study on \textit{coffee\_martini} using a '\textit{Red Canna}' Prompt: effect of $\lambda_d$ and $\lambda_p$.  The differences are minimal, but changes in local stylization can be observed}
\label{fig:ablation_coffee}
\end{figure*}

\begin{table*}[p]   %
  \centering

  \renewcommand{\gridonerow}[3]{%
    #3 &
    \includegraphics[width=0.25\linewidth]{#2/grid1_dir#1.0_patch30.0.png} &
    \includegraphics[width=0.25\linewidth]{#2/grid1_dir#1.0_patch90.0.png} &
    \includegraphics[width=0.25\linewidth]{#2/grid1_dir#1.0_patch180.0.png} \\
  }

  \renewcommand{\gridtworow}[3]{%
    #3 &
    \includegraphics[width=0.25\linewidth]{#2/grid2_sz#1_num32.png} &
    \includegraphics[width=0.25\linewidth]{#2/grid2_sz#1_num64.png} &
    \includegraphics[width=0.25\linewidth]{#2/grid2_sz#1_num128.png} \\
  }

  \setlength{\tabcolsep}{1pt}
  \footnotesize               

  \begin{minipage}[b]{0.48\linewidth}
    \centering
    \captionof{table}{ Ablation over $\lambda_{\text{dir}}$ and $\lambda_{\text{path}}$, using a text prompt: \textit{Starry Night by Vincent van Gogh}}
    \label{tab:2d_ablation_text_patch} 
    \begin{tabular}{c c c c}
      \toprule
      $\lambda_{\text{dir}}$ / $\lambda_{\text{path}}$ & 30 & 90 & 180 \\
      \midrule
      \gridonerow{5}{imgs/2D/ablation/collected_images_bridge}{5}
      \gridonerow{10}{imgs/2D/ablation/collected_images_bridge}{10}
      \gridonerow{20}{imgs/2D/ablation/collected_images_bridge}{20}
      \bottomrule
    \end{tabular}
  \end{minipage}%
  \hfill
  \begin{minipage}[b]{0.48\linewidth}
    \centering
    \captionof{table}{Ablation over $\texttt{patch\_size}$ (here denoted as $p_{size}$ and $\texttt{patch\_num}$ (denoted as $p_{num}$), using a text prompt: \textit{Starry Night by Vincent van Gogh}}
    \label{tab:2d_ablation_text_lambda}
    \begin{tabular}{c c c c}
      \toprule
      $p_{size}$ / $p_{num}$ & 32 & 64 & 128 \\
      \midrule
      \gridtworow{64}{imgs/2D/ablation/collected_images_bridge}{64}
      \gridtworow{125}{imgs/2D/ablation/collected_images_bridge}{125}
      \gridtworow{200}{imgs/2D/ablation/collected_images_bridge}{200}
      \bottomrule
    \end{tabular}
  \end{minipage}

  \medskip   %

  \begin{minipage}[b]{0.48\linewidth}
    \centering
    \captionof{table}{Ablation over $\lambda_{\text{dir}}$ and $\lambda_{\text{path}}$, using an image of a mosaic.}
    \label{tab:2d_ablation_photo_patch}
    \begin{tabular}{c c c c}
      \toprule
      $\lambda_{\text{dir}}$ / $\lambda_{\text{path}}$ & 30 & 90 & 180 \\
      \midrule
      \gridonerow{5}{imgs/2D/ablation/photo/collected_images_church}{5}
      \gridonerow{10}{imgs/2D/ablation/photo/collected_images_church}{10}
      \gridonerow{20}{imgs/2D/ablation/photo/collected_images_church}{20}
      \bottomrule
    \end{tabular}
  \end{minipage}%
  \hfill
  \begin{minipage}[b]{0.48\linewidth}
    \centering
    \captionof{table}{Ablation over $\texttt{patch\_size}$ (here denoted as $p_{size}$ and $\texttt{patch\_num}$ (denoted as $p_{num}$), using an image of a mosaic}
    \label{tab:2d_ablation_photo_lambda} 
    \begin{tabular}{c c c c}
      \toprule
      $p_{size}$ / $p_{num}$ & 32 & 64 & 128 \\
      \midrule
      \gridtworow{64}{imgs/2D/ablation/photo/collected_images_church}{64}
      \gridtworow{125}{imgs/2D/ablation/photo/collected_images_church}{125}
      \gridtworow{200}{imgs/2D/ablation/photo/collected_images_church}{200}
      \bottomrule
    \end{tabular}
  \end{minipage}

\end{table*}

\subsubsection{4D Ablation Studies}

We further conduct ablation studies on a 4D scene (\texttt{coffee\_martini}) from N3DV dataset~\cite{Li_2022_CVPR} to analyze the contribution of each loss component and the sensitivity of key hyperparameters.

Table~\ref{tab:4d_component_ablation} evaluates the effect of removing or adding individual loss terms from our full objective. Removing the patch loss ($\lambda_p = 0$) leads to a significant drop in CLIP-SIM, indicating that the patch loss is crucial for capturing local textures. Removing the directional loss ($\lambda_d = 0$) also degrades performance, particularly on CLIP-S and CLIP-SIM, suggesting its importance for style alignment. Interestingly, removing the content loss ($\lambda_c = 0$) yields the highest CLIP-S and CLIP-SIM scores, as the model is free to prioritize stylization over content preservation. Adding the temporal loss ($\lambda_t = 2000$) or freezing volume density prediction during stylization slightly improves consistency but reduces overall style transfer quality. Our full model achieves the best combination of style alignment ($CLIP-S=75.31 $) and temporal consistency ($CLIP-CONS=0.54$),

Table~\ref{tab:4d_ablation_lambda} ablates the directional loss weight $\lambda_d$ and patch loss weight $\lambda_p$ on the same 4D scene, with $\lambda_c = 5$ and $\lambda_t = 0$. An ablation visualization can be seen on Figure \ref{fig:ablation_coffee}.

\

\end{document}